\documentclass[gdplain]{geradwp}

\PassOptionsToPackage{hyphens}{url}

\usepackage[ruled,linesnumbered]{algorithm2e}
\usepackage[french]{babel}
\usepackage{hyperref}

\usepackage{natbib}
\usepackage{subfig}
\usepackage[export]{adjustbox}
\usepackage{bbm}

\GDcoverpagewhitespace{6.8cm}
\graphicspath{{Figures/}} % graphicx pkg setup
\hypersetup{colorlinks,%
citecolor={blue}, % Change for "black" with natbib
urlcolor={blue},
linkcolor={blue},
breaklinks={true}
}

\makeatletter
\ifthenelse{\isundefined{\ALG@name}}{}%
{%
\renewcommand{\ALG@name}{\sffamily\footnotesize Algorithm}
}
\makeatother
\ifthenelse{\isundefined{\SetAlCapNameFnt}}{}%
{% 
\SetAlCapNameFnt{\footnotesize}
\SetAlCapFnt{\sffamily\footnotesize}
}

\GDtitle{Interpolation-Free Deep Learning for Meteorological Downscaling on Unaligned Grids Across Multiple Domains with Application to Wind Power \footnote{This Work has not yet been peer-reviewed and is provided by the contributing Author(s) as a means to ensure timely dissemination of scholarly and technical Work on a noncommercial basis. Copyright and all rights therein are maintained by the Author(s) or by other copyright owners. It is understood that all persons copying this information will adhere to the terms and constraints invoked by each Author's copyright. This Work may not be reposted without explicit permission of the copyright owner. This Work has been submitted to American Meteorological Society - Artificial Intelligence for Earth Systems. Copyright in this Work may be transferred without further notice.}}
\GDmonth{October}{October}
\GDyear{2024}
\GDnumber{XX}
\GDauthorsShort{J.-S. Giroux, S.-P. Breton, J. Carreau}
\GDauthorsCopyright{Giroux, Breton, Carreau}
\begin{document} 

\GDcoverpage

\begin{GDtitlepage}

\begin{GDauthlist}
\GDauthitem{Jean-Sébastien Giroux \ref{affil:eccc}\GDrefsep\ref{affil:usherb}}
\GDauthitem{Simon-Philippe Breton\ref{affil:eccc}}
\GDauthitem{Julie Carreau\ref{affil:polymtl}}
\end{GDauthlist}

\begin{GDaffillist}
\GDaffilitem{affil:eccc}{National Prediction Development, Environment and Climate Change Canada, Dorval, Canada}
\GDaffilitem{affil:usherb}{Department of Computer and Electrical Engineering, Sherbrooke University, Sherbrooke, Canada}
\GDaffilitem{affil:polymtl}{Department of Mathematics and Industrial Engineering, Polytechnique Montréal, Montréal, Canada}
\end{GDaffillist}

\begin{GDemaillist}
\GDemailitem{girj2625@usherbrooke.ca}
\GDemailitem{simon-philippe.breton@ec.gc.ca}
\GDemailitem{julie.carreau@polymtl.ca}
\end{GDemaillist}

\end{GDtitlepage}

%% %%%%%%%%%%%%%%%%%%%%%%%%%%%%%%%%%%%%%%%%%%%%%%%%%%%%%%%
%% %%%%%%%%% Résumés, mots-clés, remerciements %%%%%%%%%%%
%% %%%%%%% Abstract, keywords, acknowledgements %%%%%%%%%%
%% %%%%%%%%%%%%%%%%%%%%%%%%%%%%%%%%%%%%%%%%%%%%%%%%%%%%%%%

\GDabstracts

\begin{GDabstract}{Abstract}
As climate change intensifies, the shift to cleaner energy sources becomes increasingly urgent.
With wind energy production set to accelerate, reliable
wind probabilistic forecasts are essential to ensure its efficient use. However, since numerical weather prediction models are computationally expensive, probabilistic forecasts are produced at resolutions too coarse to capture all mesoscale wind behaviors. Statistical downscaling, typically applied to enchance the resolution of climate model simulations, presents a viable solution with lower computational costs by learning a mapping from low-resolution (LR) variables to high-resolution (HR) meteorological variables. Leveraging deep learning, we evaluate a downscaling model based on a state-of-the-art U-Net architecture, applied to an ensemble member from a coarse-scale probabilistic forecast of wind velocity. The architecture is modified to incorporate (1) a learned grid alignment strategy to resolve LR-HR grid mismatches and (2) a processing module for multi-level atmospheric predictors. To extend the downscaling model's applicability from fixed spatial domains to the entire Canadian region, we assess a transfer learning approach. Our results show that the learned grid alignment strategy performs as well as conventional pre-processing interpolation steps and that LR wind speed at multiple levels is sufficient as a predictor, enabling a more compact architecture.
Additionally, they suggest that extending to new spatial domains using transfer learning is promising, and that downscaled wind velocities demonstrate potential in improving the detection of wind power ramps, a critical phenomenon for wind energy.

\end{GDabstract}

%% %%%%%%%%%%%%%%%%%%%%%%%%%%%%%%%%%%%%%%%%%%%%%%%%%
%% %%%%%%%%%%%%%%%% Article %%%%%%%%%%%%%%%%%%%%%%%%
%% %%%%%%%%%%%%%%%%%%%%%%%%%%%%%%%%%%%%%%%%%%%%%%%%%

\GDarticlestart
\section{Introduction}
\label{sec:intro}
With the acceleration of global warming in the past decades, marked by increasing temperature, shifting weather patterns, and more frequent extreme events, transition to cleaner energy sources is urgent. Adopted in 2015 by 195 countries, the Paris Agreement aims to limit the global temperature rise to well below 2°C above pre-industrial levels through concrete actions~\citep{UN2015}. As one of the signing countries, Canada has the objective of reducing greenhouse gas emissions by close to 40\%\ below the 2005 levels by 2030~\citep{CanadaClimatePlan2021}. One key solution to achieving this goal is transitioning the energy production sector to cleaner sources like wind energy, which has become increasingly attractive due to rapid technological advancements and declining costs. Various studies have investigated how Canada's energy system could be modified to meet the Paris Agreement's objectives, as well as the target of net-zero emissions by 2050, while considering different scenarios and constraints~\citep{CER2023,PolyMTL2024}. Although numbers may differ, one thing is certain: wind energy production is planned to significantly increase in the next decades in Canada.

Reliable wind forecasts shall therefore become more than ever necessary to ensure an efficient use and integration of this intermittent resource. To this end, operation decisions should increasingly depend on probabilistic forecasts as they provide an envelope of possible scenarios that helps dealing with uncertainties~\citep{IEA_Wind2019}. Such forecasts are traditionally obtained from numerical weather prediction models that can produce ensemble members with multiple perturbed initial and boundary conditions~\citep{ECCCREPS}. Presently, however, probabilistic forecasts are generated at horizontal resolutions that are too coarse to capture all mesoscale effects affecting wind behaviour, such as land and sea breezes~\citep{Crosman2010}, or local wind patterns influenced by topography, like mountain-valley circulations~\citep{Wagner2014}. As generating probabilistic forecasts at higher resolutions is computationally challenging, it is of great interest to investigate other ways to obtain such forecasts.

An increasingly popular approach is statistical downscaling, which can produce high-resolution meteorological variables at a small fraction of the computational cost required by running numerical weather prediction models at a corresponding resolution.
It operates by establishing statistical relationships between low-resolution variables (predictors) and high-resolution ones (predictands)~\citep{sun1,ipcc_sres}. Once calibrated, these relationships are applied to estimate predictands from new predictors (e.g., obtained over different time periods or a different ensemble member). 
Although typically applied to climate model simulations, statistical downscaling is also well-suited for post-processing probabilistic forecasts, enhancing the resolution of each member in a low-resolution ensemble.

Numerous approaches have been proposed to model statistical downscaling relationships.
Earlier studies implemented linear methods to downscale wind data~\citep{Davy1, Kirchmeier}. 
Many recent non-linear downscaling methods leverage the latest advances in deep learning. In particular, convolutional neural networks (CNNs) serve as the foundation for several deep learning architectures such as generative adversarial networks (GANs)~\citep{LeCun1, Goodfellow2}.
GANs have been utilized in numerous studies for wind downscaling~\citep{Singh, Stengel, Manepalli, Miralles, Annau}, along with various other types of CNNs \citep{Zhang, Dujardin}, including some that employ a U-Net architecture \citep{Hohlein, Sekiyama, Dupuy}. Additionally, techniques other than CNNs-based, such as Extreme Gradient Boosting \citep{Lianfa, Hu}, have been explored.

In the current study, we selected the U-Net architecture as the basis of our downscaling models. 
Designed originally for biomedical image segmentation,
it combines upsampling and downsampling with convolutional layers and residual
connections to handle coarse and fine resolutions~\citep{Ronneberger}. \citet{Yang} adapted the U-Net for image super-resolution, a task similar to downscaling that aims to enhance the resolution of an image by reconstructing a high-resolution version from a low-resolution one. 
Building on this architecture, U-Net variants were developed for meteorological downscaling, differing in two key aspects from the super-resolution U-Net.
First, predictors additional to the low-resolution version of the predictand, that carry predictive information about the predictand, may be included. Second, the predictors may additionally include high-resolution geophysics variables~\citep{Hohlein, Vandal}.  With the recent update proposed in \citet{Sekiyama}, the U-Net is a widely used architecture for meteorological downscaling.

We focus on two essential factors influencing the design of the U-Net architecture in meteorological downscaling, which, we believe, are often overlooked.
The first factor is related to the inherent mismatch between the predictor and predictand grids, as the physics-based models generating the low-resolution predictors and the high-resolution predictands operate on grids with different spatial resolutions and potentially different orientations. This is usually handled by performing an interpolation as a pre-processing step in order to re-grid the data onto a common grid~\citep{Sekiyama, Dupuy, Annau}. 
As this grid is of higher resolution, the volume of predictor data is inflated without introducing any new information. This could create a computing bottleneck in memory usage, limiting the use of more relevant predictor data. The second factor concerns the selection of predictors to include in the downscaling model, which varies across studies and rarely includes multi-level atmospheric variables~\citep{Hohlein, Sekiyama, Dupuy}. Since predictor selection impacts both accuracy and memory usage, we believe it is crucial to evaluate the predictive power of the chosen predictors.

Since the ultimate goal is to obtain high-resolution probabilistic wind forecasts for all of Canada, another key consideration is the applicability of downscaling models to either large or multiple spatial domains.
Directly incorporating a domain as large as Canada into a downscaling model would demand substantial high-performance computing skills and resources~\citep{chollet2015}. In this regard, most studies have typically focused on relatively small spatial domains. To address larger domains, the area of interest can be partitioned into several smaller sub-domains~\citep{Dupuy,Annau,Sekiyama,Hohlein,Manepalli}.
However, partitioning all of Canada could render the downscaling model intractable due to memory constraints.
An alternative approach, drawing from transfer learning, would be to develop an initial downscaling model across multiple domains and then specialize it for any other domain of interest, such as areas where wind farms are located. This approach allows for leveraging prior knowledge to improve learning efficiency and performance on the new domain. While transfer learning has been utilized in various fields~\citep{Pan2010ASO}, to the best of our knowledge, it has only been applied in a downscaling context for a different purpose~\citep{Miralles} and not to expand applicability to new spatial domains. 

In this work, we build upon the U-Net architecture recently proposed by \citet{Sekiyama} for wind downscaling. We introduce memory-efficient modifications to this architecture to address the previously mentioned grid mismatch by eliminating the pre-processing step of re-gridding and allowing the downscaling model to learn the appropriate interpolation directly. The downscaling model based on this modified architecture is compared with standard re-gridding approaches. We also incorporate in the U-Net architecture the processing of different types of predictors, including multi-level atmospheric ones, leveraging both previous research and expert insights from ECCC~\citep{alzubaidi2023survey}. Comparisons are carried out to assess the importance of some of the chosen predictors. %\spb{Est-ce vraiment ce qu'on a fait?}
Expansion to new spatial domains is explored by first studying the sensitivity of the downscaling model to the choices of domains used to build the initial model, and second, by assessing its performance after specialization on a new domain using transfer learning.
Although the analyses are based on a single member of a probabilistic forecast, they serve as a proof of concept. In addition, the ability of the downscaled forecasts to reveal mesoscale effects responsible for wind power ramp events is explored.

\section{Meteorological data}
\label{sec:Model_data}

\subsection{Meteorological forecast systems}
\label{subsec:Met_model_desc}
The data used as predictands in the downscaling model are taken from a near-surface mesoscale atmospheric analysis (\textit{mesoanalysis}) produced every hour, 24 hours a day, at ECCC. This mesoanalysis is produced on a grid across Canada by combining the most recent available observations (assimilated from surface stations with a 15~min delay) with the latest 2.5~km resolution forecasts produced by the ECCC's High Resolution Deterministic Prediction System (HRDPS) valid at the analysis time~\citep{gmd-2024-55,ECCCHRDPS}. It aims at providing the best possible representation of the weather conditions at the near-surface level at a given time.

Forecasts produced by ECCC's Regional Ensemble Prediction System (REPS) over a 10~km resolution grid covering North America \citep{ECCCREPS} are used as low resolution predictors in the downscaling model. This system is run four times a day at 00Z, 06Z, 12Z and 18Z, outputting forecasts hourly with a leadtime of 78 hours. Its control member was used as a case study for probabilistic forecasts, keeping hours 4 to 9 of each run as a compromise for the forecasts not to be influenced by model spin-up and to be as accurate as possible, see e.g. \citet{Hohlein}. In doing so, hourly data are available 24 hours a day to match with the hourly mesoanalysis. 

Data used from both forecast systems cover a period of 29 months (12/02/2021 - 04/30/2024). As regular updates are implemented, this period is chosen to ensure that the data considered have been produced by the same version of both systems~\citep{ECCCupgrade}.
It is important to mention that the grids of both forecast systems are unaligned and that their pixels (grid cells) are not collocated (Figure \ref{fig:FigGridsDom1}).

\subsection{Configurations of spatial domains of interest}
\label{subsec:Dom_interest}
\begin{figure}
    \centering
    \subfloat[Predictor (red pixels) and predictand (black pixels) grids, showing the grid misalignment and the extent of both grids for domain \#1. Note that the eight outermost rows and columns of pixels (four on each side) of the predictor grid are used for padding.]{\includegraphics[height = 4.6cm]{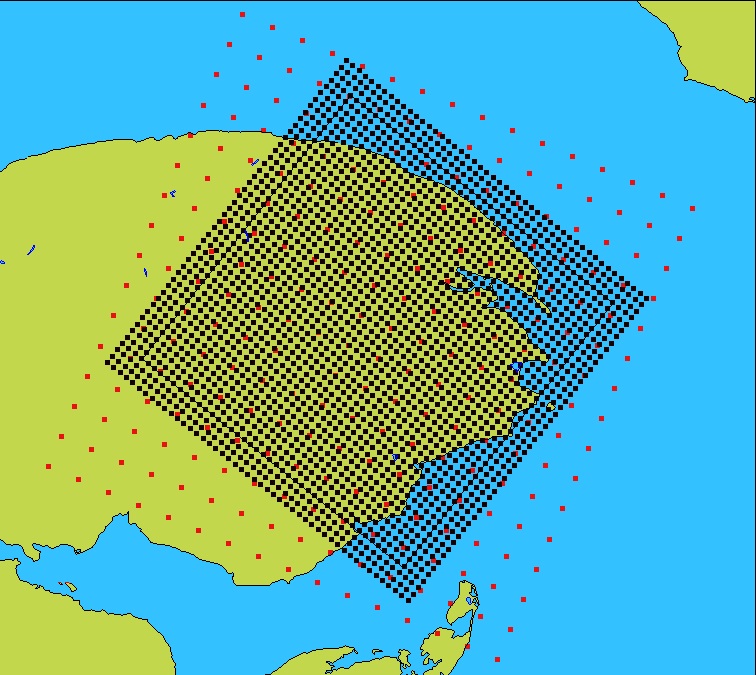}
     \label{fig:FigGridsDom1}}
     \hspace{2mm}
          \subfloat[Configuration A: Five predictand domains located close to each other in Eastern Québec and New Brunswick.]{\includegraphics[height = 4.6cm]{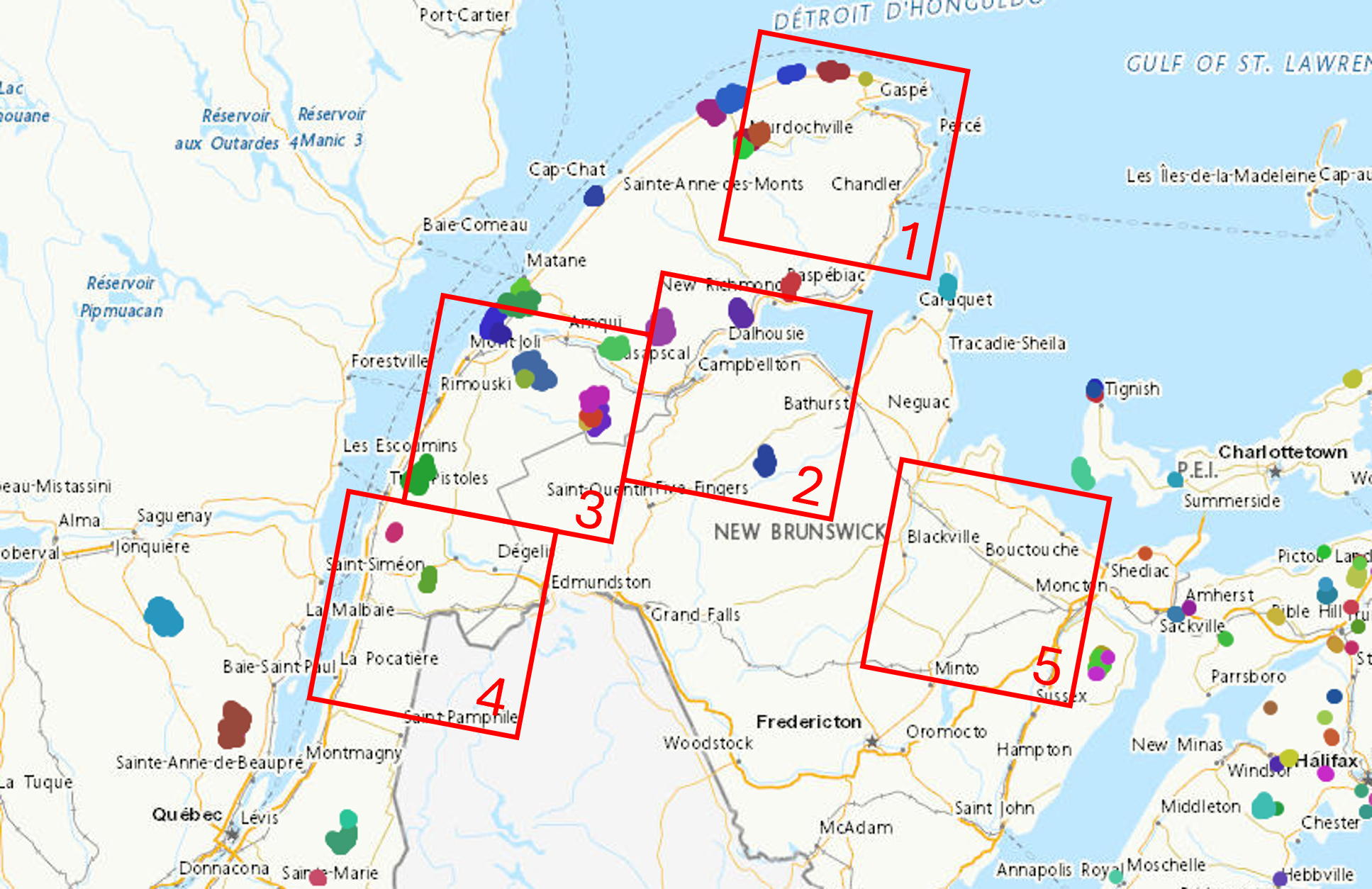}\label{fig:Dom_gaspe}}\\

     \subfloat[ Configuration B: Five predictand domains across Canada.]{\includegraphics[width = 0.48\textwidth]{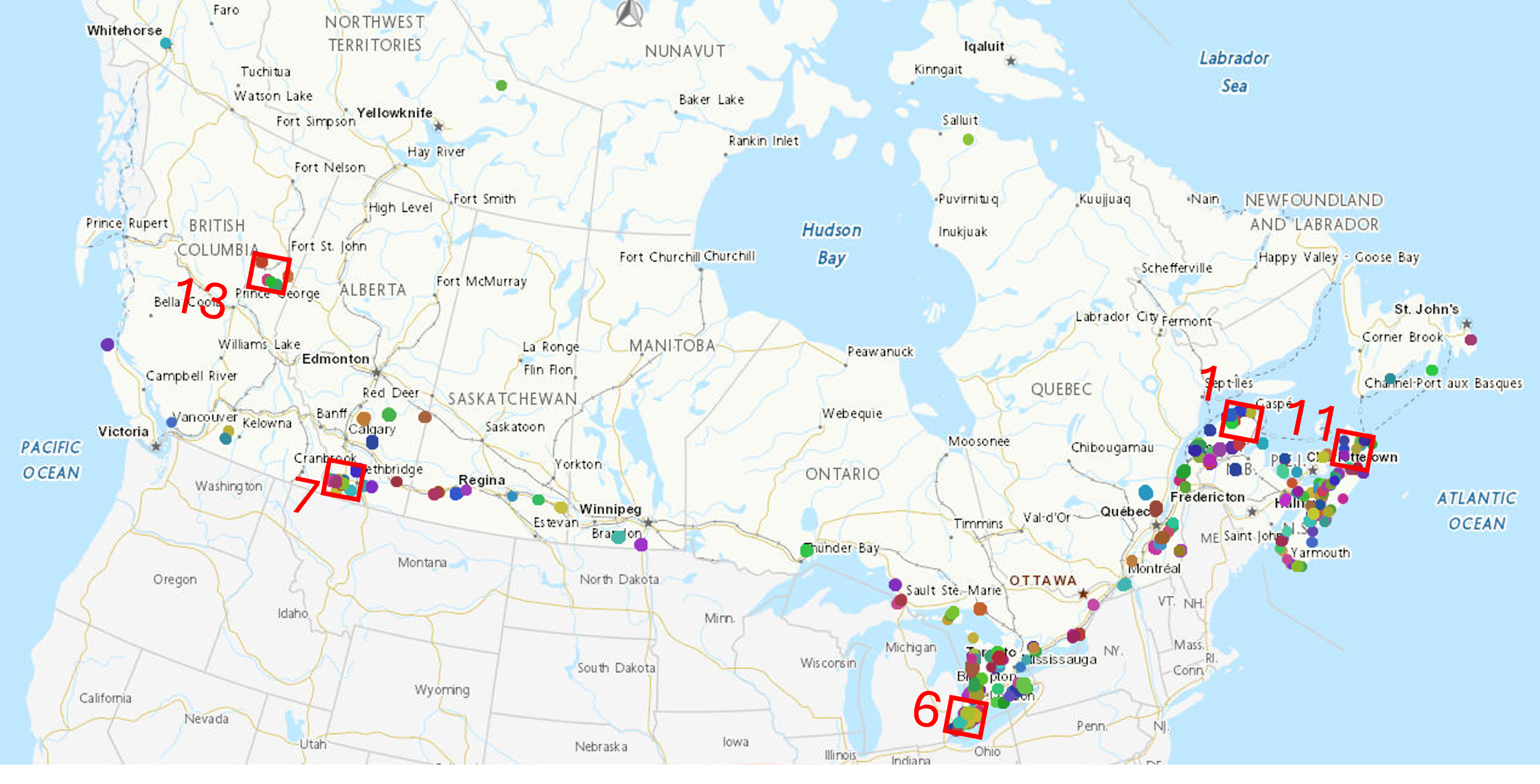}\label{fig:Dom_cad}}
     \hspace{2mm}
     \subfloat[Configuration C: Ten predictand domains across Canada.]{\includegraphics[width = 0.48\textwidth]{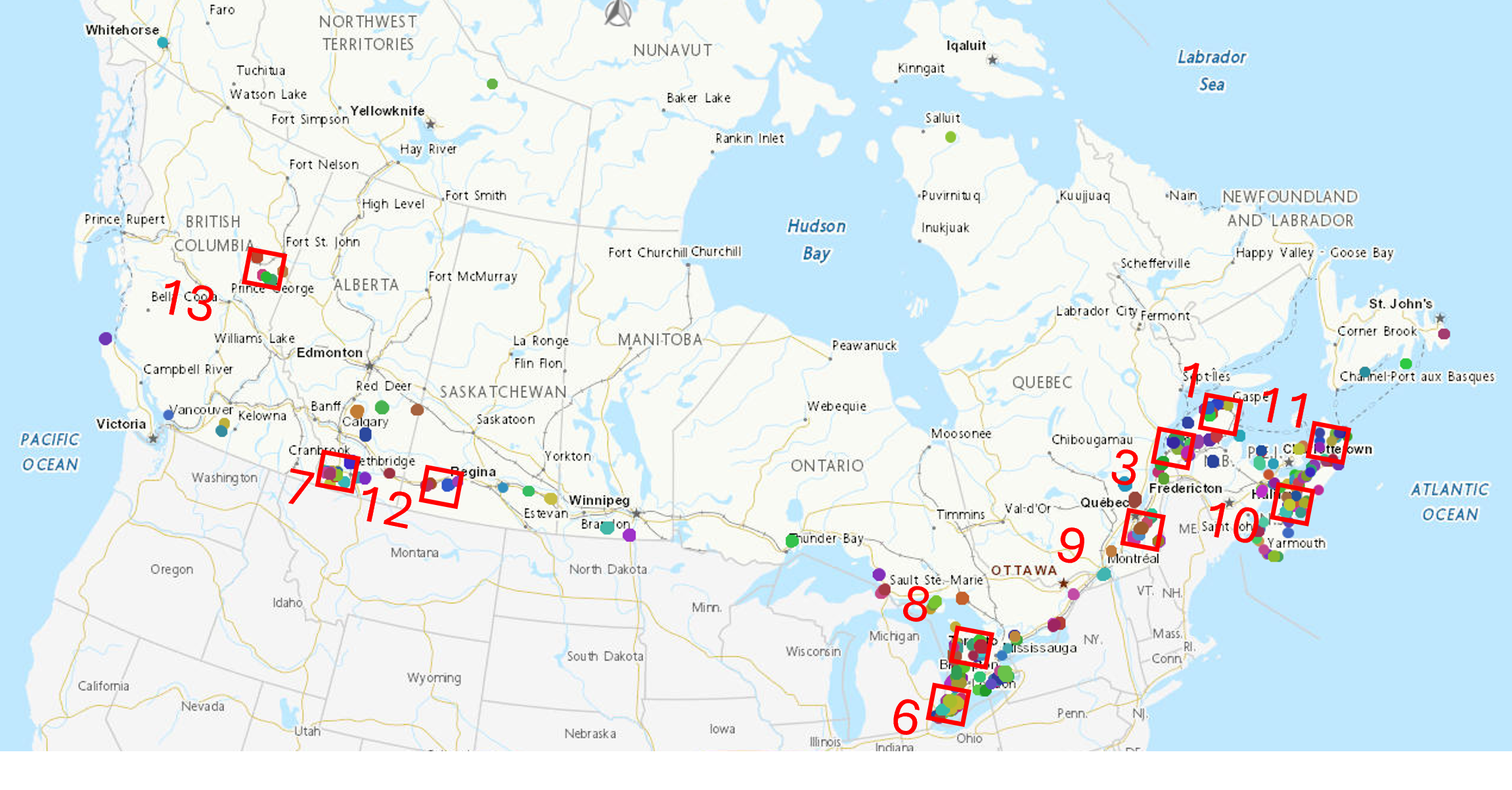}\label{fig:Dom_10cad}}
     
     \caption{Predictor and predictand grid shapes illustrated on one of the spatial domain, and the three configurations of domains used in this work. Each domain is represented by the predictand domain and marked by a red rectangle of size $120 \times 120$~km\textsuperscript{2} including the padding. In addition, each colored dot indicates a wind farm site.}\label{fig:domains}
\end{figure}

We consider a variety of configurations containing multiple spatial domains where Canadian wind farms are located, using the Canadian Wind Turbine Database \citep{govcan2024} to locate their exact position. For the first configuration, inspired by research where only a single or multiple domains in close proximity were used~\citep{Dupuy,Annau,Sekiyama,Hohlein,Manepalli}, five neighbouring domains (Figure \ref{fig:Dom_gaspe}) were selected to assess the need of physical proximity between domains. Then five scattered domains (Figure \ref{fig:Dom_cad}) are chosen to assess the impact of large physical inter-domain distances, associated to climatological and geophysics differences. Finally, 10 scattered spatial domains, including the five domains from the previous configuration, were selected to investigate whether adding more data (Figure \ref{fig:Dom_10cad}) will lead to improved performance, based on the fact that doing so in deep learning often leads to better results~\citep{Bengio,Nakkiran,Su}. Training the downscaling model over multiple domains is indeed expected to provide better prediction results as was presented by \citet{Manepalli} with closely located domains.  

A given spatial domain for the predictors is a square grid of dimensions $160 \times 160$~km\textsuperscript{2}. For the predictand, the grid's dimensions are $120 \times 120$~km\textsuperscript{2}, which includes a $20 \times 20$~km\textsuperscript{2} padding to prevent boundary effects~\citep{Sekiyama}. The padding is removed from the downscaling model prediction, resulting in an effective predictand domain of $100 \times 100$~km\textsuperscript{2} (Figure \ref{fig:FigGridsDom1}). While other techniques exist, see e.g., \citet{Cailing}, we chose padding because it is more in line with similar downscaling models \citep{Hohlein,Sekiyama}. We also selected a larger predictor domain to allow the downscaling model to include atmospheric effects surrounding the predictor domain \citep{Dupuy}. The larger predictor domain is also necessary when applying the non-interpolation strategy, discussed in \S~\ref{sec:methodo}.\ref{subsec:CNN_architecture}, because the grid misalignment would make the predictor domain to inadequately cover the predictand domain if both domains were of the same size.

%Word count temp
%TC:ignore
%Word count temp
%TC:endignore

\subsection{Predictors and predictand}
\label{subsec:Pred_predict}
The wind velocity $UV$ at 10~m above the surface was used as the predictand, i.e., the downscaled variable. We made this choice rather than simultaneously downscaling the eastward and northward components $U$ and $V$ in order to reduce the complexity of the downscaling model and to minimize directly the error for the wind velocity, which is the main variable of interest for wind energy forecasts.

The set of predictors consists of several low resolution variables readily available from REPS, gradients that are computed for some of these variables, along with high resolution geophysics variables.

$UV$, $U$, $V$, and the temperature $T$ at multiple levels above the ground were first chosen as predictors. Multiple levels were considered for these variables to have a better representation of the dynamics of the atmosphere \citep{ahrens2012}. Surface pressure ($P0$) as well as sea level pressure ($PN$) were also used. While $P0$ is known to carry pertinent information about the wind \citep{Curry}, $PN$ was also used to better inform about horizontal changes in pressure different than those due to altitude variations \citep{ahrens2012}. The boundary layer represents the lower part of the atmosphere that has a constant interaction with the surface of the Earth \citep{Holtslag}. Its height ($H$), also a good indicator of atmospheric stability, was selected~\citep{Stull1988}. We also chose the cosine of the azimuth angle ($CX$) and the snow depth ($SD$) to provide an indication, respectively, of the time of the day and period of the year. Moreover, snow coverage is known to impact winds at low levels \citep{Holtslag}. The wind gust estimate ($WGE$) was also included as a predictor as it provides additional information about atmospheric stability and the behaviour of the wind that might not be captured by the wind velocity predictors alone~\citep{Thorarinsdottir, Kahl}. 

Vertical gradients, i.e., rates of change with respect to height, were computed for the wind components and temperature and used as predictors. We selected for the upper level used in the calculation of the respective gradients the model level closer to the top of the atmospheric boundary layer that was available hourly from the outputs of REPS, while the lower level corresponds to the first prognostic model level. Wind velocity vertical gradients inform about the wind shear in the boundary layer, and also enter in the calculation of the Richardson number, as does the temperature vertical gradient~\citep{AMSGlossary}. The Richardson number is a good indicator of atmospheric stability, that has a strong influence on wind velocities in the boundary layer \citep{Stull1988}. For example, a high level of instability will make the winds near the surface to be influenced more importantly by winds at higher levels.

As geophysics variables like the orography ($ME$), land-sea mask ($MG$) and surface roughness ($Z_0$) are also known to impact the wind velocity~\citep{Holtslag,Guo,Hu}, they were used to help the model learn about the geography of the domains being studied and its effect on the wind. A high-resolution version of those static predictors was readily available from ECCC's HRDPS inputs. They were re-generated on a 2.5km resolution grid oriented the same way as the low-resolution predictor grid (see Figure~\ref{fig:FigGridsDom1}). The predictor and predictand variables are summarized in Table~\ref{tab:predictors}.

%Word count temp
%TC:ignore
\begin{table}[ht]
    \caption{Predictor and predictand variables of the downscaling model. The gradient (grad) for a given variable $var$ is calculated as $\text{grad} var = \frac{\text{var(GZ2)} - \text{var(GZ1)}}{\text{GZ2} - \text{GZ1}}$, where $GZ2$ and $GZ1$ are the geopotential heights of respectively the higher and lower model levels used in the calculation of the gradient. Note that the heights provided below are approximative as they were calculated under standard atmospheric conditions.}
    \label{tab:predictors}
    \centering
    \begin{tabular}{|c|c|c|c|}
        \hline
        Variable & Type  & Height above surface ($m$) & Resolution\\
        \hline
        Wind magnitude ($UV$) & Predictor & 10 - 21 - 63 - 117 & Low \\
        Wind vector ($U$) & Predictor &  10 - 21 - 63 - 117 & Low \\
        Wind vector ($V$)  & Predictor &  10 - 21 - 63 - 117 & Low \\
        Temperature ($T$) & Predictor & 1.5 - 11 - 42 - 91 & Low \\
        Gradient $U$ ($UG$) & Predictor & between 273 and 21 & Low \\
        Gradient $V$ ($VG$) & Predictor & between 273 and 21 & Low \\
        Gradient $T$ ($TG$) & Predictor & between 231 and 11 & Low \\
        Surface pressure ($P0$) & Predictor & surface & Low \\
        Sea level pressure ($PN$) & Predictor & surface & Low \\
        Boundary layer height ($H$) & Predictor & NA & Low \\
        Azimuth of the cosine angle ($CX$) & Predictor & NA & Low \\
        Snow coverage ($SD$) & Predictor & surface  & Low \\
        Wind gust estimate ($WGE$) & Predictor & 10 & Low \\
        Orography ($ME$) & Predictor & NA & High\\
        Land-sea mask ($MG$) & Predictor & NA & High\\
        Surface roughness ($Z_0$) & Predictor & NA & High\\
        Wind magnitude ($UV$) & Predictand & 10 & High \\
        \hline
    \end{tabular}
\end{table}
%Word count temp
%TC:endignore

\section{Downscaling methodology}
\label{sec:methodo}

\subsection{Downscaling models for three grid alignment strategies}\label{subsec:CNN_architecture}

To address the misalignment between the predictors' and predictand's grid, three distinct strategies are considered. The first and second, widely used in the literature and referred to as "interpolation strategies", consist in interpolating the predictors' data onto the predictand grid using either nearest neighbour or bi-linear interpolation as a pre-processing step~\citep{Sekiyama,Dupuy,Annau}. We propose a third strategy, termed the "no-interpolation strategy", where this pre-processing step is skipped, and which consists in learning the interpolation as part of the downscaling model. Each of these three grid alignment strategies yields a particular downscaling model. 

The core of the selected architecture for our downscaling models is inspired from an architecture called DeepRU, a kind of modified U-Net used in similar studies~\citep{Ronneberger,Hohlein,Sekiyama}. The complete architecture consists of three modules, namely the multi-modal predictor processing, the U-Net, and the post U-Net, which are described in \S~\ref{par:Module_I}-\ref{par:Module_III} below. There are two main variants of the multi-modal predictor processing module corresponding to the aforementioned interpolation and no-interpolation strategies.

\subsubsection{Multi-modal predictor processing module}
\label{par:Module_I}
This module includes steps that precede the U-Net itself and is illustrated in Figures \ref{fig:pre_interpol} and \ref{fig:pre_nointerpol} for, respectively, the interpolation and the no-interpolation strategies. Based on in-house expert knowledge, suitable predictor processing was determined. It aims to separately extract information on multi-level predictors and account for atmospheric stability using vertical gradients. Such processing could help the downscaling model learn features specific to each predictor. Indeed, research on multi-modal learning shows that separately processing different predictors before combining them in the U-Net can enhance feature extraction and improve performance~\citep{Baltrusaitis}. Single-level low-resolution predictors are processed together. The operation sequence taken by the predictors is as follows: they pass through a block consisting of a 2D convolution, followed by a batch normalization and the application of a LeakyReLU activation function. The predictors pass through two consecutive blocks before being concatenated. Batch normalization is used to accelerate the training process, allowing the use of higher learning rates and reducing the sensitivity to initialization \citep{Ioffe}. The LeakyReLU activation function was used because it precludes the vanishing gradient problem, and was used in similar architectures \citep{Maas2013RectifierNI,Hohlein,Sekiyama}. In this module, the high-resolution geophysics predictors go through two consecutive blocks using 2D stride convolution layers, allowing to concatenate low-resolution predictors with high-resolution ones. The main difference between the interpolation and no-interpolation strategies is that, for the latter, the result of the concatenation passes through two additional blocks of 2D stride convolution layers, while for the former, it passes through two blocks of no-stride convolution layers before being combined with the geophysics predictors block result. This is done because the input resolution for both strategies is not the same (64 versus 16 pixels). After module I, both strategies follow the same steps, with identical filter numbers and dimensions, to allow for fair comparison.

%Word count temp
%TC:ignore
\begin{figure}[ht!]
    \centering
    \includegraphics[scale=1.00]{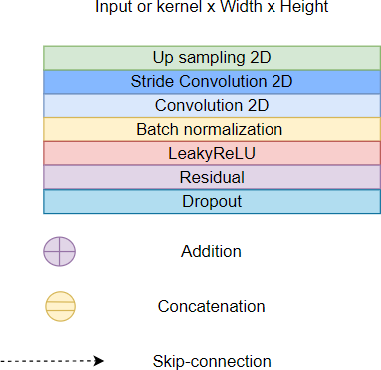}
    \caption{Legend for each kind of layer and operation used in the three modules that form the downscaling models' architecture.}
    \label{fig:legend}
\end{figure}

\begin{figure}[ht!]
    \centering
    \includegraphics[scale=0.80]{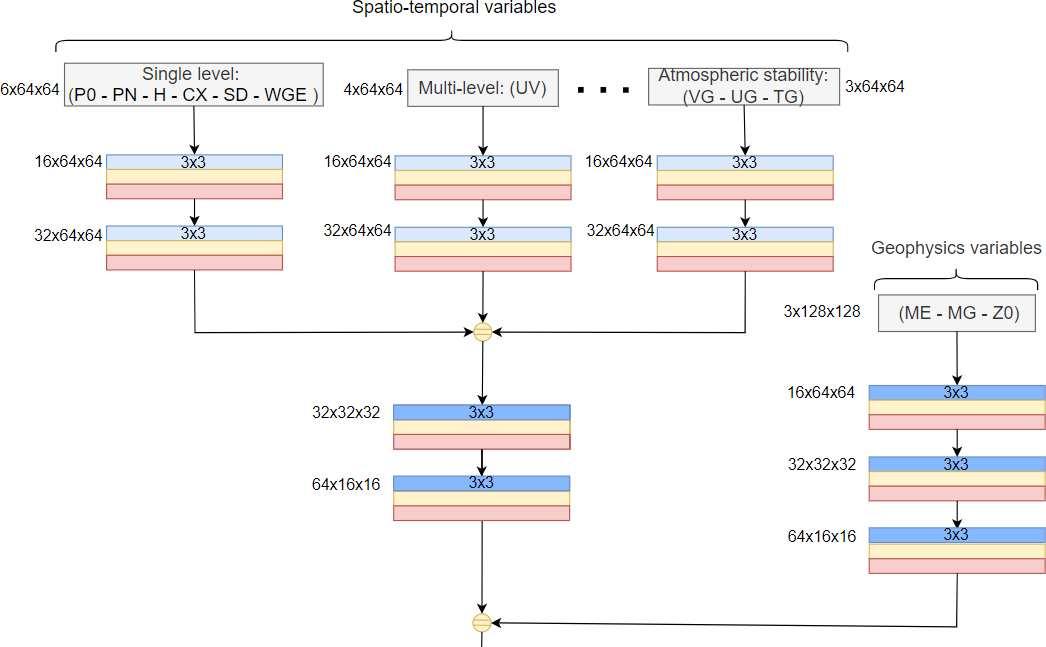}
    \caption{Architecture of the multi-modal predictor processing module for the interpolation strategy. Some predictors are not illustrated and are replaced by $...$. The missing predictors are multi-level $U$,  multi-level $V$, multi-level $T$, see Fig.~\ref{fig:legend} for the legend.} 
    \label{fig:pre_interpol}
\end{figure}

\begin{figure}[ht!]
    \centering
    \includegraphics[scale=0.80]{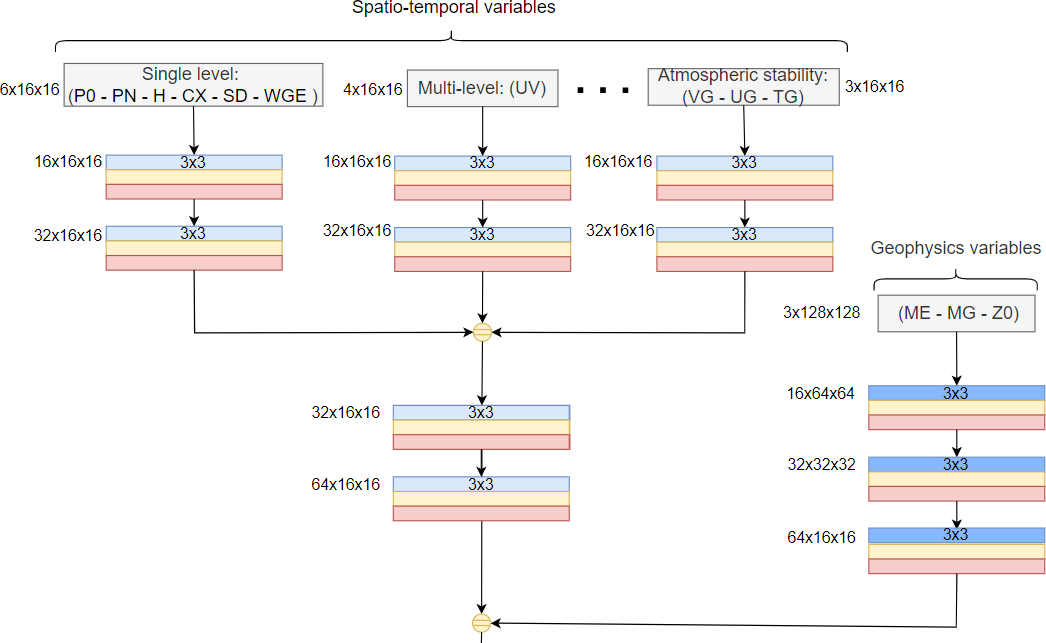}
    \caption{Architecture of the multi-modal predictor processing module for the no-interpolation strategy. Some predictors are not illustrated and are replaced by $...$. The missing predictors are multi-level $U$,  multi-level $V$, multi-level $T$, see Fig.~\ref{fig:legend} for the legend.}
    \label{fig:pre_nointerpol}
\end{figure}
%Word count temp
%TC:endignore

\subsubsection{U-Net module}
\label{par:Module_II}
\begin{figure}[ht!]
    \centering
    \includegraphics[scale=0.60]{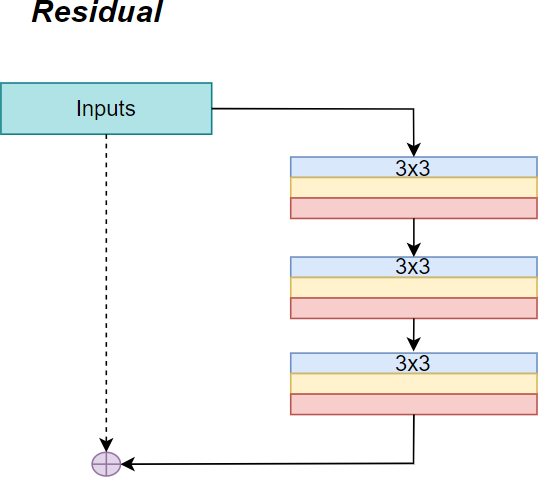}
    \caption{Residual block architecture. See Fig.~\ref{fig:legend} for the legend.}
    \label{fig:residual}
\end{figure}

Figure \ref{fig:u_net} illustrates the U-Net architecture, composed of an encoder, a bottleneck and a decoder. We made a minor adjustment, as this was done in \citet{Laxman}, to encourage the U-Net to enhance high-resolution details in the final stage of the architecture, improving the overall quality of the output. The adjustment consists in inserting an up-sample block at the end of the U-Net, preceded and followed among other layers by convolution layers used for refinement, instead of up-sampling before entering the U-Net architecture. In each block of our U-Net architecture, we added residual blocks, see Figure \ref{fig:residual}, which were proposed to enable a deeper architecture and to address the vanishing gradient and degradation problems~\citep{Kaiming}. It has been shown that, generally, deep architectures outperform shallow ones \citep{Szegedy,Simonyan}, and this is why we chose to reduce the size of the output from the previous block one dimension at a time. Dropout layers have been inserted in each U-Net block to prevent over-fitting and to improve the generalization of the model \citep{Nitish}.

%Word count temp
%TC:ignore
%Word count temp
%TC:endignore

\subsubsection{Post U-Net module} \label{par:Module_III}
\begin{figure}[ht!]
    \centering
    \includegraphics[scale=1]{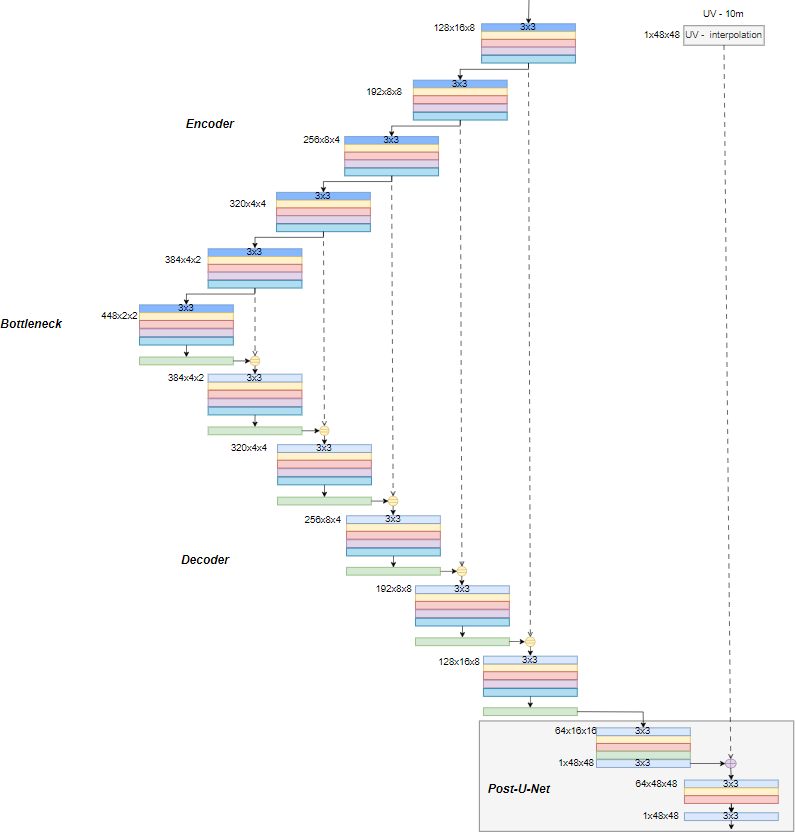}
    \caption{U-Net and post U-Net (in the gray rectangular area) modules. See Fig.~\ref{fig:legend} for the legend.}
    \label{fig:u_net}
\end{figure}

The output of the last block of the U-Net goes through the Post U-Net module (see the gray rectangular area in Figure \ref{fig:u_net}). The first step is to increase the resolution of the output to match that of the predictand. Once completed, the skip connection—created by interpolating the low-resolution $UV$ predictor to the predictand grid using nearest neighbours—is added to the output. The role of this skip connection is to provide an initial guess to the downscaling problem on the predictand grid, onto which high-resolution features predicted by the U-Net are added. 

%Word count temp
%TC:ignore
%Word count temp
%TC:endignore

\subsection{Deep learning setup for downscaling: training, optimization, and transfer learning}
\label{sec:train_proce}

\subsubsection{Python programming environment}
\label{subsec:exp_setup}
TensorFlow \citep{tensorflow2015_whitepaper} with Keras version 2.14.0 \citep{chollet2015} was used to create, train, tune and evaluate the downscaling models. The optimization was performed using the ADAM optimizer \citep{Diederik}. NumPy version 1.26.4 was used for data processing and loading, at is is known to efficiently handle array operations, accelerating the processing of the data \citep{harris2020array}. To improve performance and optimize the resources, we used callbacks implementing early-stopping with 15 epochs of patience, Reduce on Plateau with a patience of 10 epochs, a reduction factor for the learning rate of 0.10, as well as a model checkpoint to save the best model. Callbacks were monitored with the mean squared error (MSE) and were implemented with the Keras library \citep{chollet2015}.

The experiments were conducted on a computer cluster at ECCC equipped with NVIDIA A100 GPUs, each with 40 GB of memory. The system had 110GB of RAM and an Ice Lake processor with 24 cores.

When using the interpolation strategies, which require more memory than the no-interpolation strategy, it was necessary to optimize the data pipeline, because training the model on one GPU was otherwise not possible with the datasets used. Indeed, the interpolation strategies, when performed on the GPU, led the system's memory to be exceeded. To overcome the challenge we used Dataset API from TensorFlow which allows to process data in a way that can efficiently feed the training process. We also used the from$\_$generator method which enables to only load on the GPU the current batch being processed by the downscaling model \citep{tensorflow2023}. 

\subsubsection{Training, validation and test sets}
\label{subsec:data_splitting}
The available data, predictand and predictors variables over the 29 month period, is split into training, validation (for hyperparameter tuning, see \S~\ref{sec:methodo}.\ref{sec:train_proce}.\ref{subsec:hp_tuning}), and test (for performance evaluation, see \S~\ref{sec:Results_Disc}) sets. As meteorological events are continuous phenomena that can be observed across multiple sequential hourly data samples, these three sets are formed sequentially to avoid data leakage \citep{Yagis, Tampu}. Indeed, a model trained with data that includes samples from time steps after those used for testing would have access to future information to predict past events, which would not be the case when using the model in prediction mode.
The training, validation and test sets are composed respectively of data from 12/02/2021-08/31/2023, 09/01/2023-12/31/2023 and 01/01/2024-04/30/2024. Details on these three sets are provided in Table \ref{tab:datadist}.

%Word count temp
%TC:ignore
\begin{table}[ht]
    \caption{Sample size (1 sample = 1 hour) and memory used for storage in each strategy (interpolation vs no-interpolation, see \S~\ref{sec:methodo}.\ref{subsec:CNN_architecture}) for each configuration  (that corresponds to choices of domains, see \S~\ref{sec:Model_data}.\ref{subsec:Dom_interest}) split into training, validation, and test set.
    }
    \label{tab:datadist}
    \centering
    \begin{tabular}{|c|c|c|c|}
        \hline
        Configuration &  A  &  B & C \\
        \hline
        Training (sample size) & 76 120 & 76 120 & 152 240 \\
        Validation (sample size) & 14 615 & 14 615 & 29 320 \\
        Testing (sample size)  & 14 495 & 14 495 & 28 990 \\
        Memory used Bi-linear interpolation (GB) & 45.85 & NA & NA \\
        Memory used nearest neighbour interpolation (GB) & 45.85 & NA & NA \\
        Memory used no interpolation (GB) & 8.45 & 8.45 & 16.90 \\
        \hline
    \end{tabular}
\end{table}
%Word count temp
%TC:endignore

The data are randomly shuffled within the training set, which allows for the generation of batches (a subset of the training set used to perform a learning step) that are more representative of the whole dataset, by obtaining a varied sample of the training set. This would not be the case without shuffling, as batches corresponding to specific temporal periods of the training set would then be obtained \citep{Chollet2017}. It was noticed (results not shown) that the downscaling model performs better when data is shuffled. A fixed random seed was used to ensure that the data shuffle was consistent across different experiments, facilitating a fair comparison between various hyper-parameter configurations \citep{Chollet2017}.

\subsubsection{Standardization procedure}
\label{subsec:data_standard}
The data need to be standardized to help stabilize the downscaling model, accelerate the convergence rate, ensure similar contributions for features from each predictor, and enhance overall model performances \citep{goodfellow2016,Chollet2017}. The standardization is computed by first subtracting a mean value and then dividing by a standard deviation. Details on the standardization procedure are as follows.

The training data for a given domain configuration (A, B, or C, see \S~\ref{sec:Model_data}.\ref{subsec:Dom_interest}) is used to compute the average mean and standard deviation for every predictor. The mean and standard deviation are computed globally for each predictor separately by calculating them first on each sample and by then computing the average over the whole training set. These values are used to standardize the data of each predictor in the training, validation and test sets of the given configuration. It is important to only use the information from the training set to standardize the validation and test sets to prevent data leakage which would artificially boost the results obtained by the model \citep{Yagis, Tampu}. The same strategy was used to standardize the predictand data, i.e., the training set for a given configuration is used to compute the average mean and standard deviation to be used in the standardization of the training, validation and test sets. This approach preserves the predictand characteristics and the integrity of the predictand data since the predictors and predictand data are not on the same scale, due to their difference in resolution \citep{goodfellow2016}. In similar works, the mean and standard deviation were computed pixelwise \citep{Hohlein,Sekiyama}. However, with the multiple domains considered here, each pixel is related to a different geographic location, which make this approach not applicable.

\subsubsection{Loss function}
\label{subsec:loss_fct}
We consider a combination of two metrics to create the loss function, as it was done in~\citet{KAVIANI2023102315}. The first one is the MSE, which was selected to gear the downscaling model to predict values as close as possible to the predictand (ground truth), pixel by pixel, by penalizing large errors more significantly \citep{goodfellow2016}. The second one is the structural similarity index measure (SSIM), which is designed to assess the similarity between two images based on luminance, contrast, and structure, providing a measure of the perceptual similarity between two images which aligns better with the human visual perception \citep{Wang}. SSIM takes values in $[-1,1]$ with an interpretation similar to a correlation coefficient (e.g., values near one indicates high similarity). The loss function is given as:

%Word count temp
%TC:ignore
\begin{equation} 
\label{loss_function}
\text{Loss} = \text{MSE} + \lambda \cdot (1 - \text{SSIM}),
\end{equation} 
%Word count temp
%TC:endignore
where $\lambda \in \mathbbm{R}$ is an hyperparameter that needs to be selected. 

\subsubsection{Hyperparameter tuning}
\label{subsec:hp_tuning}
The following hyperparameters were tuned for each downscaling model: the learning rate, the number of epochs, the value of $\lambda$ in the loss function (see equation \eqref{loss_function}), the Dropout value and the LeakyReLU parameter. Fixed values were chosen for other hyperparameters, namely the batch and kernel sizes, set to respectively 32 and 3, as well as for the number of filters in each convolution layer and the number of layers in the U-net module (Figures \ref{fig:pre_interpol} to \ref{fig:u_net} for details). These choices are based on values for similar downscaling models used for comparable applications, e.g. \citet{Hohlein} and \citet{Sekiyama}, and limit the number of hyperparameters to tune. For each downscaling model, we selected an hyperparameter combination that gave the best performance over the validation set for the MSE values. 

Hyperparameter tuning was done with Keras-Tuner version 1.4.5 using a Bayesian search \citep{chollet2015}, which is known to outperform other methods like grid search~\citep{Bergstra}.

\subsubsection{Transfer Learning}
\label{subsec:train_settings}
A downscaling model is called "general" once it has been trained on the multiple domains composing each configuration. A downscaling model is said to be "specific" if it undergoes another round of training to specialize a general model for a specific geographical region, i.e., by using transfer learning. More precisely, the weights of the multi-modal predictor processing module (\S~\ref{sec:methodo}.\ref{subsec:CNN_architecture}.\ref{par:Module_I})  and in the encoder from the U-Net module (\S~\ref{sec:methodo}.\ref{subsec:CNN_architecture}.\ref{par:Module_II}) are frozen while the weights of the decoder from the U-Net module and from the post-U-Net module (\S~\ref{sec:methodo}.\ref{subsec:CNN_architecture}.\ref{par:Module_III}) are fine-tuned, a technique known as inductive transfer learning using parameter-transfer \citep{Tajbakhsh,Pan2010ASO}.
The general model thus aims to capture as much information as possible on wind patterns and interactions under various geographical conditions, while the specific model is dedicated to a specific domain. With this transfer learning strategy, we aim to cover every wind farm or any domain in Canada even if the general model was not trained directly on the corresponding data.

\section{Comparative experiments and wind ramp detection evaluation}
\label{sec:Results_Disc}

Several experiments (\S~\ref{comparison_interpol}-\ref{subsec:feature:exp} below) are carried out to compare and evaluate various implementation choices of the downscaling models described in \S~\ref{sec:methodo}. 
A baseline model, consisting of a simple application of bi-linear interpolation of low-resolution $UV$ onto the predictand grid, is included to provide a minimum performance level to beat. 
In addition, an investigation (\S~\ref{subsub:windramp} below) is conducted in terms of wind ramp detection, which is particularly important for wind farms.

\subsection{Experimental setup}
\label{subsect:perf_metrics}

In each experiment, training and validation sets are used respectively for training and hyperparameter tuning while the performance metrics are reported for the test set, see \S~\ref{sec:methodo}.\ref{sec:train_proce}.\ref{subsec:data_splitting}. The metrics considered are described below.

We first include two metrics which are temporally and spatially aggregated to provide a global performance overview across all the domains of the configurations considered. Specifically, the RMSE (Root-Mean-Square Error) and the SSIM are averaged over each domain and then across all the domains. These metrics are represented with boxplots, lower RMSE values (SSIM values closer to 1) indicating a better performance. The MAE (mean absolute error) was also considered but led to similar conclusions than the RMSE and is not reported here (see Supplemental Material).

The following three metrics, as a complement to those contained in the loss function, are computed for a single domain to illustrate domain-specific properties. 
The selected domain is domain \#1, see Figure~\ref{fig:domains}, which belongs to the three configurations considered. Other domains were analyzed but led to similar conclusions (see Supplemental Material).
To obtain a spatial view of the performance, the MAE is computed pixel by pixel over the domain.

To examine the spatial frequency content in the wind speed fields, a power spectrum density (PSD) graph is proposed. This graph illustrates the energy distribution across different scales, with low frequencies representing large-scale features and high frequencies indicating rapid changes (such as edges and fine textures). To calculate the PSD, a 2D Fourier transform is applied to a given field and its magnitude is squared to obtain the power of each frequency component. An azimuthal average is then performed, more precisely, the power spectrum is averaged over concentric circles centered at the zero-frequency component, to reduce the 2D power spectrum to a 1D profile of power as a function of frequency (in cycles per pixel) \citep{Gonzalez}. This process is repeated for each sample (hour) in the test set, and the final profile is obtained by averaging the resulting 1D power spectra across all samples.

Last, the distribution of wind speed values is evaluated with a probability density function (PDF) graph. By combining all wind speed fields over the test set, a PDF graph is constructed with a kernel density estimator with 55 bins and a bandwidth of 0.70 to obtain a trade-off between bias (i.e., the inability to reproduce features of the underlying PDF) and variance (which is inversely related to the smoothness of the estimated PDF)~\citep{Scott}.

\subsection{Interpolation versus non-interpolation strategy experiment}
\label{comparison_interpol}
As mentioned in \S~\ref{sec:methodo}, there are three downscaling models corresponding to each grid alignment strategy hereafter referred to as bi-linear or nearest neighbour interpolation and no-interpolation strategy for brevity.  
In this experiment, we compare these three strategies along with the baseline model using the data from the domain configuration A (Figure \ref{fig:Dom_gaspe}).

\subsubsection{Results}
\begin{figure}[ht]
    \centering
     \hspace{2mm}
     \subfloat[RMSE.]{\includegraphics[height=4.4cm]{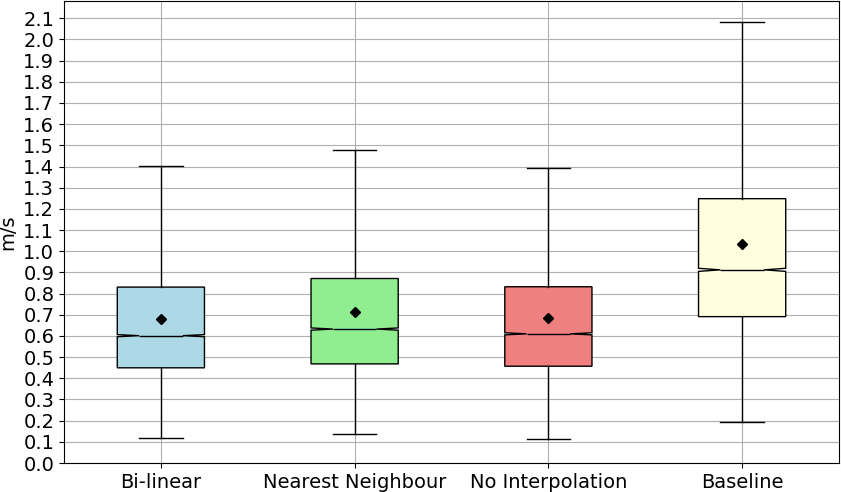}\label{fig:bp_rmse}}
     \hspace{2mm}
     \subfloat[SSIM.]{\includegraphics[height=4.4cm]{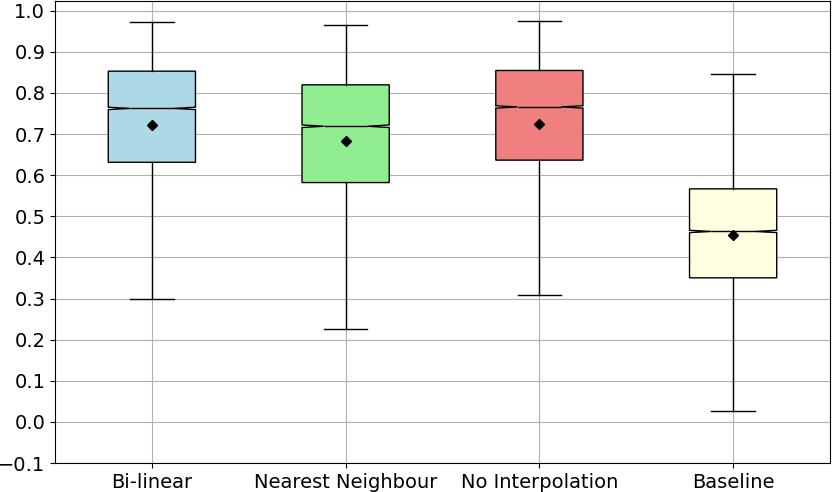}\label{fig:bp_ssim}}
     
     \caption{RMSE and SSIM boxplots for each downscaling model (bi-linear or nearest neighbour interpolation strategy, no-interpolation strategy, see \S~\ref{sec:methodo}.\ref{subsec:CNN_architecture}), and baseline computed on the test set of configuration A in Figure \ref{fig:Dom_gaspe} with the five domains in Eastern Québec and New Brunswick (14 495 hourly samples from January 2024 to April 2024). In each boxplot, the black dot and line represent respectively the mean and the median, the upper and lower box limits indicate the first (Q1) and third (Q3) quartiles and the whiskers depict the highest (lowest) value within the $1.5 \times$ (Q3-Q1) above Q3 (below Q1).}\label{fig:results_I}
\end{figure}

The boxplots in Figure \ref{fig:results_I} display the temporally and spatially aggregated RMSE and SSIM computed on configuration A's test set. 
The three strategies all significantly outperform the baseline, notably decreasing the magnitude and frequencies of extreme values, but otherwise yield similar performance although the interpolation strategy with nearest neighbour performed slightly worse.

%Word count temp
%TC:ignore
%Word count temp
%TC:endignore

\begin{figure}[ht]
    \centering
    \subfloat[Pixel-wise MAE (m/s) on domain \#1 for the different strategies and the baseline. The colorscale is capped at the maximum MAE of the three strategies. The baseline’s MAE exceeds the maximum value of the colorscale in certain locations, causing saturation. The value at the top of each image is the average over the domain.]{\includegraphics[width=0.99\textwidth,valign=c]{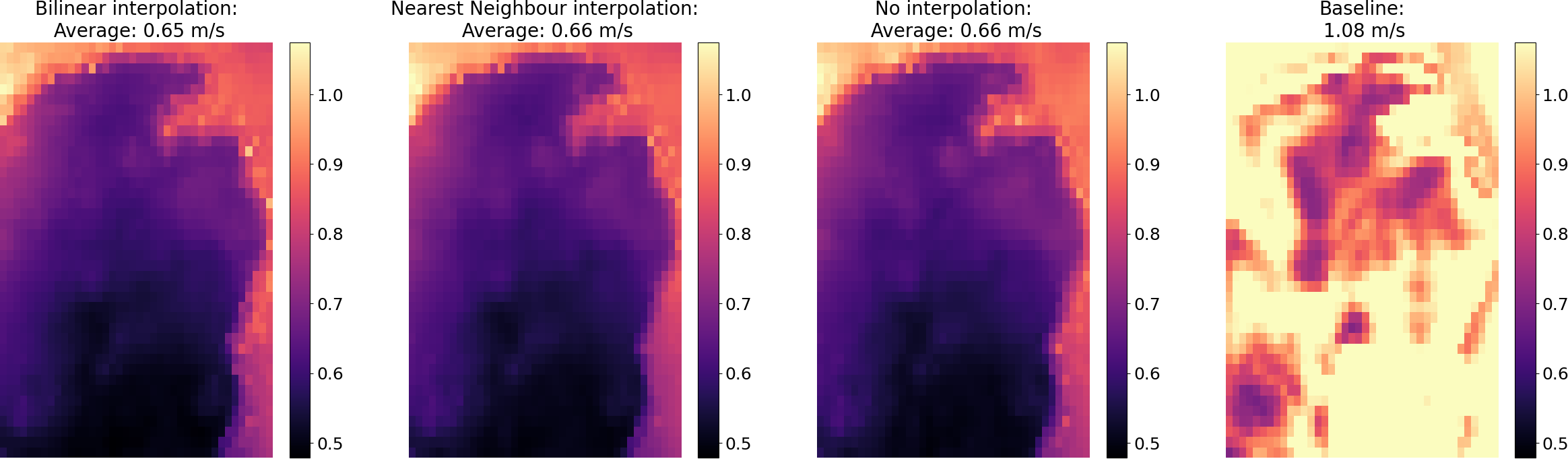}
     \label{fig:error_section_I}} \\
     %\hspace{2mm}
     \subfloat[Average Power Spectrum Density (PSD).]{\includegraphics[width = 0.48\textwidth,valign=c]{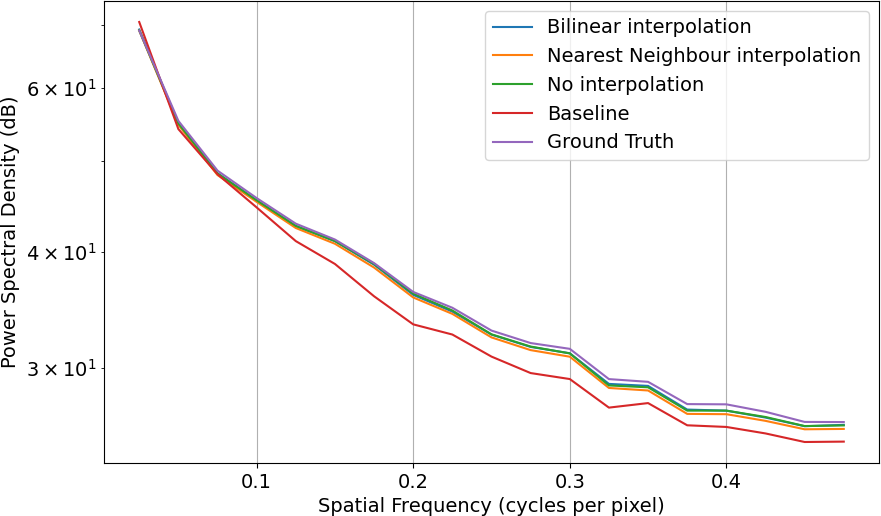}\label{fig:results_psd_I}}
     \hspace{2mm}
     \subfloat[Average Probability Density Function (PDF). Note that the legend is the same as Figure \ref{fig:results_psd_I}.]{\includegraphics[width = 0.48\textwidth,valign=c]{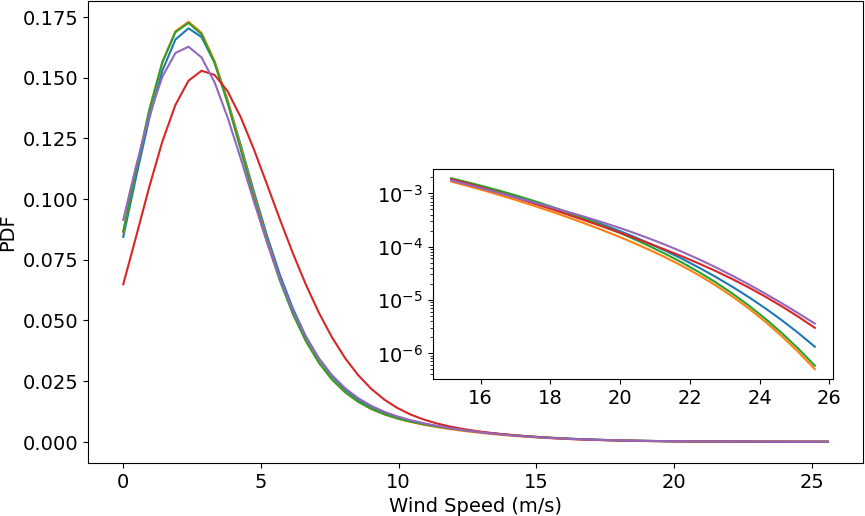}\label{fig:result_pdf_I}}
     
     \caption{MAE, average PSD and average PDF on the test set for domain \#1 of configuration A in Eastern Québec and New Brunswick, see Figure \ref{fig:Dom_gaspe}  (2899 hourly samples from January 2024 to April 2024), for each downscaling model (bi-linear or nearest neighbour interpolation strategy, no-interpolation strategy, see \S~\ref{sec:methodo}.\ref{subsec:CNN_architecture}), and baseline. The ground truth in the PSD and PDF graphs refers to the HR $UV$, i.e., the predictand.}\label{fig:results_domain_I}
\end{figure}

The following analyses concern domain \#1 in Figure \ref{fig:Dom_gaspe}. Figure~\ref{fig:error_section_I} shows the pixel-wise MAE.
All three strategies performed similarly, and they did markedly better than the baseline. The geographical patterns are similar for each strategy with larger values in the water and near the coast. This is expected since wind velocities are also known to be higher in these areas~\citep{CanadianWindAtlas}. 
The PSD graph, Figure \ref{fig:results_psd_I}, shows that the three downscaling strategies also surpassed the baseline in their ability to reproduce the range of spatial frequencies.
Indeed, their curves are closer and more similar to the predictand than the baseline's is. 
The bi-linear and no interpolation strategies yielded almost the same curve, slightly more accurate than the nearest neighbour interpolation's. In addition, the reproduction of higher frequencies is slightly more challenging for all three strategies. 

Last, the three downscaling strategies largely outperformed the baseline in reproducing the distribution of wind speed values, as shown in the PDF graph of Figure~\ref{fig:result_pdf_I}. However, they slightly over-estimated the PDF curve for central values ($[2,5]$ m/s) which is compensated by a small under-estimation of more extreme values ($ <2$ and $>17$ m/s). This is not surprising as the downscaling models are not specifically designed to handle these less frequent wind velocities, that are actually not of primary interest for the application of wind ramp detection, see \S~\ref{subsub:windramp} below. Moreover, it should be pointed out that our loss function is not designed to specifically optimize the velocity distribution. Nonetheless, the downscaling models allow an overall improved reproduction of the PDF curve compared to the baseline's.

%Word count temp
%TC:ignore
%Word count temp
%TC:endignore

\subsubsection{Discussion}
In the super-resolution community, some authors have acknowledged the limitations of using interpolation as a pre-processing step~\citep{Yang}. Indeed, interpolation smooths the details, is time-consuming and might lead to inaccuracies. In contrast, in recent work on wind downscaling, the chosen strategy involves this pre-processing step, i.e., the predictors are interpolated beforehand~\citep{Hohlein, Sekiyama, Dupuy}. The results provided by this experiment show that it is possible to obtain similar even better performance by letting the downscaling model learn the grid alignement operation, i.e., to skip the interpolation pre-processing step. A striking advantage of this no-interpolation strategy is the 5.4-fold decrease in memory allocation (Table~\ref{tab:datadist}). This means that, with the same allocated memory, the no-interpolation strategy could leverage up to 5 times more domains, increasing the number from 5, such as in configuration A used in this experiment, to 25.  
In our viewpoint, the no-interpolation strategy outperforms overall the two interpolation strategies by being more memory-efficient and providing equivalent (with respect to bi-linear) or improved (with respect to nearest neighbour) performance. 
Therefore, in what follows, the experiments are carried out solely for the no-interpolation strategy, hereafter referred to as the downscaling model or U-Net for brevity.

\subsection{Domain configuration experiment}
\label{evaluation_dataset}
In this experiment, we compare the three domain configurations described in \S~\ref{sec:Model_data}.\ref{subsec:Dom_interest} using the downscaling model that relies on the no-interpolation strategy. The aim is to assess the effect of scattered (configuration B) versus nearby (configuration A) domains and of doubling the number of domains (configuration B versus C). 

\subsubsection{Results}
\begin{figure}[ht]
    \centering
     \hspace{2mm}
     \subfloat[RMSE.]{\includegraphics[height=4.3cm]{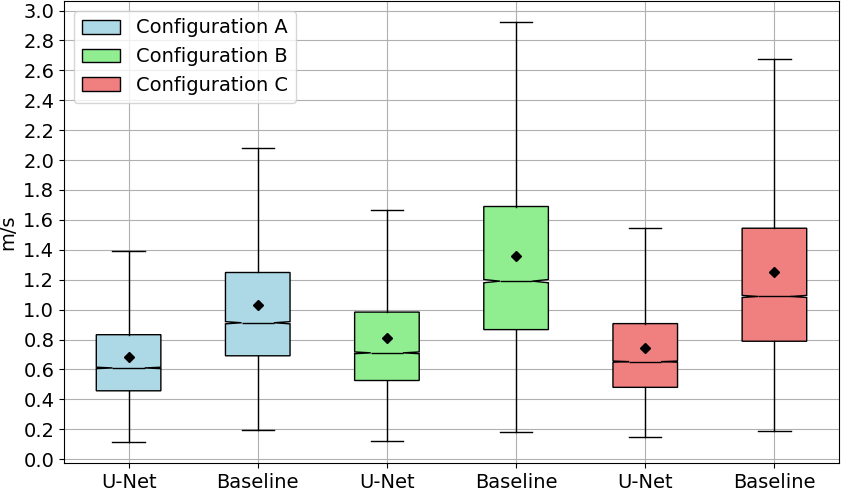}\label{fig:bp_rmse_II}}
     \hspace{1mm}
     \subfloat[SSIM.]{\includegraphics[height=4.3cm]{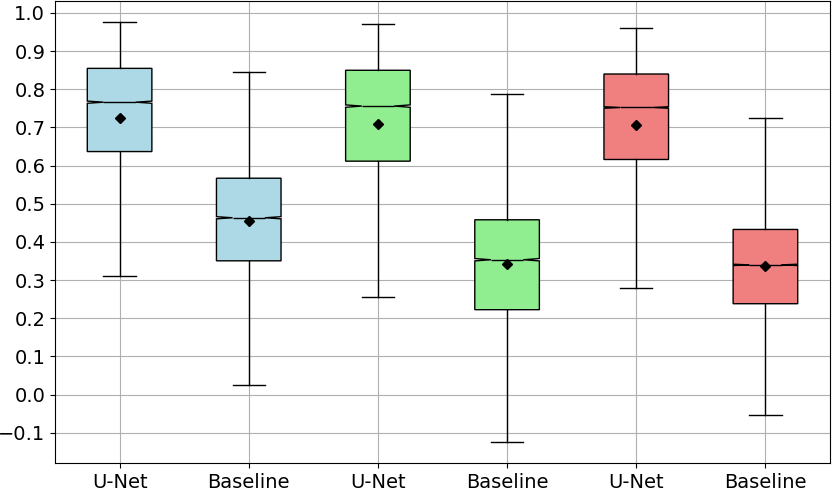}\label{fig:bp_ssim_II}}
     
     \caption{RMSE and SSIM boxplots for the downscaling model (U-Net) predictions on the test set, which consists of hourly samples from January 2024 to April 2024, i.e. respectively 14 495 for configuration A in Eastern Québec and New Brunswick, and 14 495 and 28 990 for configurations B and C across Canada (Figure \ref{fig:domains}). RMSE and SSIM boxplots for the baseline model (bi-linear interpolation of the low resolution $UV$ onto the predictand grid) are presented alongside. In each boxplot, the black dot and line represent respectively the mean and the median, the upper and lower box limits indicate the first (Q1) and third (Q3) quartiles and the whiskers depict the highest (lowest) value within the $1.5 \times$ (Q3-Q1) above Q3 (below Q1).}\label{fig:results_II}
\end{figure}

Figure \ref{fig:results_II} presents the boxplots of the spatiotemporally aggregated RMSE and SSIM on the test set of each configuration.
Since the test set varies for each configuration due to the differences in spatial domains, the comparison between configurations consists in assessing the gain in performance of the downscaling model relative to the baseline model. It can be seen that the downscaling model outperforms the baseline model across all configurations and metrics. The largest gain in performance is observed for configuration B, especially in terms of RMSE. Certain domains, particularly those with complex wind patterns due to intricate topographies, present a greater challenge for the baseline model compared to flat regions. This type of domain accounts for a greater part of configuration B, and the downscaling model is shown to offer significant improvements in such cases. 

%Word count temp
%TC:ignore
%Word count temp
%TC:endignore

\begin{figure}[ht]
    \centering
    \subfloat[Pixel-wise MAE (m/s) on domain \#1 for the downscaling model (U-Net) trained with each configuration along with the baseline's. The colorscale is capped at the maximum MAE of the downscaling model for each configuration. The baseline's MAE  exceeds the maximum value of the colorscale in certain locations, causing saturation. The value at the top of each image is the average over the domain.]{\includegraphics[width=0.99\textwidth,valign=c]{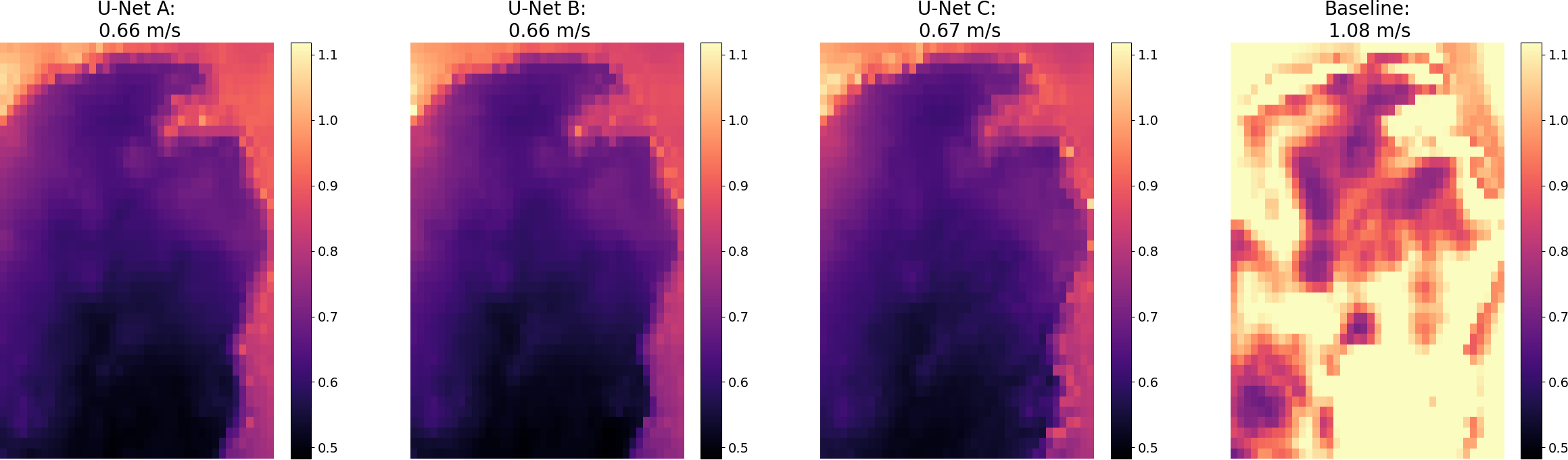}
     \label{fig:section_II_average_error}}
     \\
     \subfloat[Average PSD.]{\includegraphics[width=0.48\textwidth,valign=c]{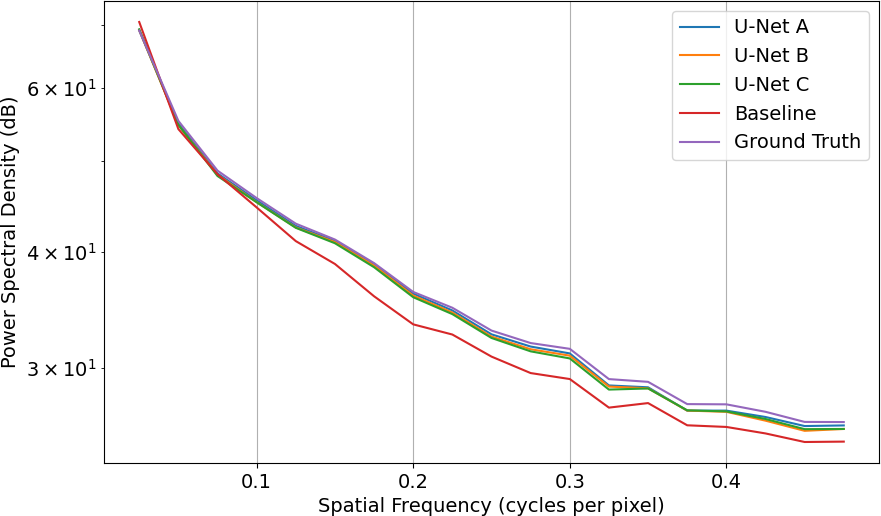}\label{fig:section_II_psd}}
     \hspace{2mm}
     \subfloat[Average PDF. Note that the legend is the same as Figure \ref{fig:section_II_psd}.]{\includegraphics[width=0.48\textwidth,valign=c]{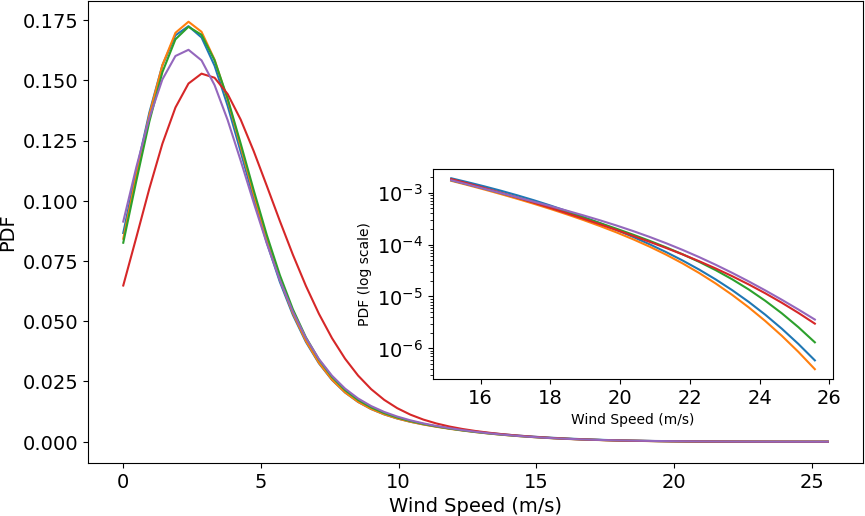}\label{fig:section_II_pdf}}
     \caption{MAE, average PSD and average PDF  on the test set over domain \#1 (2 899 hourly samples from January 2024 to April 2024), for the downscaling model (U-Net)  trained with configuration A (five domains in Eastern Québec and New Brunswick), configuration B (five domains across Canada) or configuration C (10 domains across Canada), see Figure \ref{fig:domains}, along with the baseline (bi-linear interpolation of the low-resolution $UV$ onto the predictand grid). The ground truth in the PSD and PDF graphs refers to the HR $UV$, i.e., the predictand.}\label{fig:results_domain_II}
\end{figure}

In Figure~\ref{fig:results_domain_II}, we analyze the three domain-specific metrics over domain \#1 (Figure \ref{fig:domains}).
The pixel-wise MAE, Figure~\ref{fig:section_II_average_error}, remains very similar across all configurations with comparable geographical error patterns. As in \S~\ref{sec:Results_Disc}.\ref{comparison_interpol}, larger errors are found over the water and along the coasts. The performance of the downscaling model in terms of PSD (Figure~\ref{fig:section_II_psd}) and PDF (Figure~\ref{fig:section_II_pdf}) curves is also similar across all configurations. There are marked improvements with respect to the baseline model in reproducing all frequencies of the predictand's PSD curve, particularly those larger than 0.1 cycle/pixel, as well as the central values of wind speeds in the PDF curve. These analyses for domain \#1, which is included in all three configurations, suggest that the performance of the no-interpolation downscaling model is largely unaffected by the configuration's composition. 

%Word count temp
%TC:ignore
%Word count temp
%TC:endignore

\subsubsection{Discussion}
The results above indicate that the proximity of domains, whether nearby or scattered, does not affect performance. Moreover, increasing the number of domains in the configuration does not reduce the performance. However, the performance of the downscaling model seems to greatly decrease when evaluated on a domain that is not included in the configuration. As an illustration, we evaluated the downscaling model on the test set using each configuration on domain \#10  (Figure \ref{fig:Dom_10cad}) which is only contained in configuration C. As expected (results not shown), the model using configuration C significantly outperformed those using configurations A or B. 
Using configurations with multiple domains paves the way for a highly generalized downscaling model capable of operating across diverse regions, potentially covering all of Canada. Including numerous, scattered and varied domains, by introducing greater variability to the data, is expected to lead to a more generalized and robust downscaling model.

As there would be hardware and optimization limitations, in practice, to including as many domains as necessary to cover all of Canada in a generalized model, we introduce the following idea. A general downscaling model is developed using a configuration with numerous, diverse, and distinct domains. The rationale is to enable the model to learn as many wind patterns as possible. A model specialized for a specific domain—whether or not it is part of the training configuration of the generalized model—is then developed using transfer learning. The specific model should benefit from the general model's training on a wide variety of domains.

\subsection{Transfer learning experiment}
\label{TransferLearning_exp}

Since this is how it would be used in order to perform downscaling Canada-wide, the aim of this experiment is to evaluate the benefit of transfer learning on a domain that is not part of the training configuration.
We propose to use configuration C to create the general dowscaling model, while also developing a specific model for domain \#2 which it is not included in the selected configuration. The specific model is compared against the general model and a model trained only on domain \#2 (denoted as "zero" in the results below, as it is trained from scratch). 

\subsubsection{Results}
\begin{figure}[ht]
    \centering
     \hspace{2mm}
     \subfloat[RMSE.]{\includegraphics[height=4.4cm]{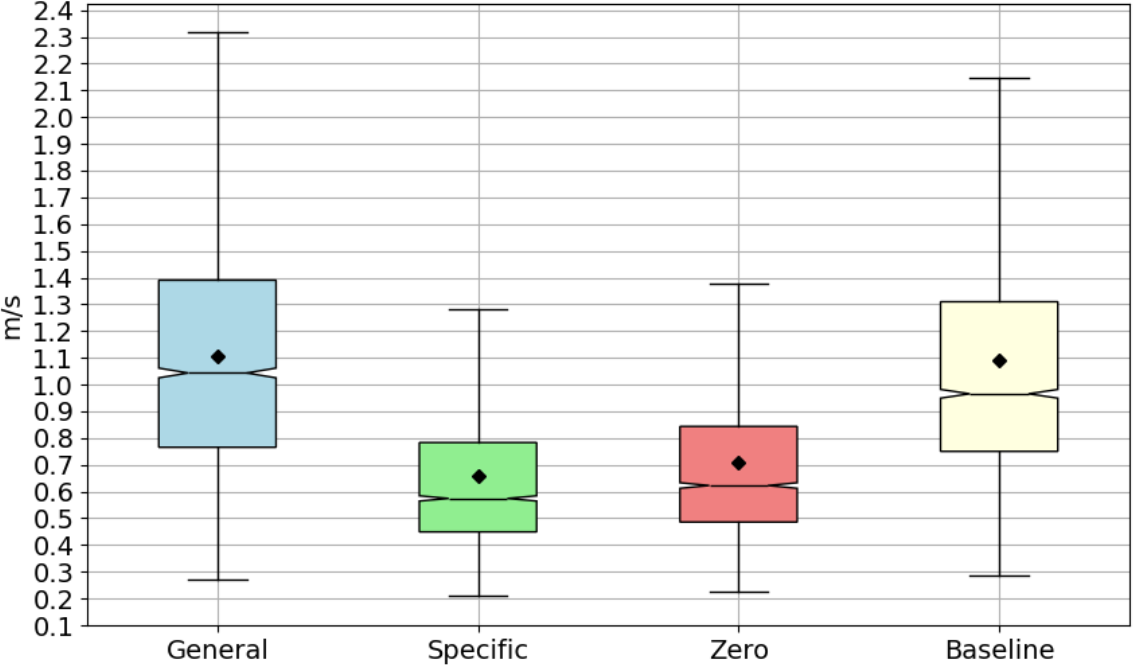}\label{fig:bp_rmse_III}}
     \hspace{2mm}
     \subfloat[SSIM.]{\includegraphics[height=4.4cm]{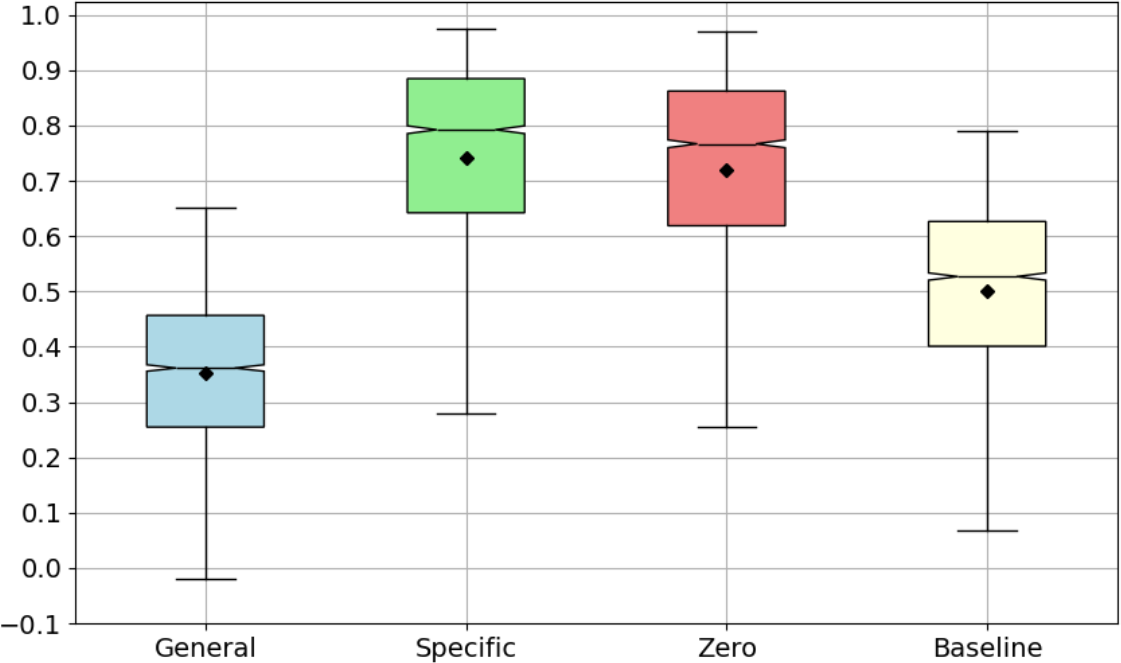}\label{fig:bp_ssim_III}}
     
     \caption{RMSE and SSIM metrics box plots for the model predictions on the test set for domain \#2, Figure \ref{fig:domains} (2 899 hourly samples from January 2024 to April 2024), for specific, generic, and zero models, and for baseline obtained from interpolating the predictor grid on the predictand grid using bi-linear interpolation. In each graph, the black dot and lines represent respectively the mean and the median, the upper and lower box limits indicate the first (Q1) and third (Q3) quartiles and the whiskers depict the highest (lowest) value within the $1.5 \times$ (Q3-Q1) above Q3 (below Q1).}\label{fig:results_III}
\end{figure}

Figure \ref{fig:results_III} illustrates that
the best models in terms of both RMSE and SSIM are the specific and zero models. 
The general model's performance is poor, matching the baseline in terms of RMSE and performing worse in terms of SSIM. It is worth noting that results (not shown) suggest a slight improvement when applying transfer learning to a domain included in the general model's training configuration.

%Word count temp
%TC:ignore
%Word count temp
%TC:endignore

Figure \ref{fig:section_III_average_error} shows that the specific model has the lowest MAE. The zero model produced slightly larger MAE values on the whole domain compared to the specific model, while the general model and the baseline have the largest MAE values across the domain.
Figure \ref{fig:section_III_psd} shows that the PSD curve generated by the specific model closely matches that of the predictand, with only minor deviations at high frequencies, and performs better than the zero model in this range. It also indicates that all models, except the general model, outperform the baseline, as the general model exhibited the lowest SSIM scores.
Figure \ref{fig:section_III_pdf} indicates that the wind speed distribution produced by the general, specific and zero models is quite similar, and much closer to the predictand's than the baselines', except at high wind velocities for the general model.

%Word count temp
%TC:ignore
\begin{figure}[ht]
    \centering
    \subfloat[Pixel-wise MAE (m/s) for the different models and the baseline. The colorscale is capped at the max MAE of the Specific and Zero models. The baseline’s MAE exceeds the maximum value of the colorscale in certain locations, causing saturation. The value at the top of each image is the average over the domain.]{\includegraphics[width=0.99\textwidth,valign=c]{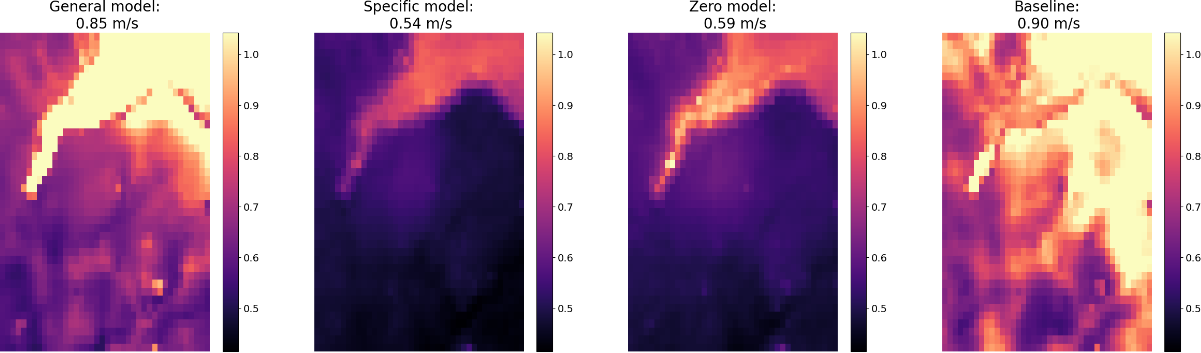}
     \label{fig:section_III_average_error}}
     \\
     \subfloat[Average PSD.]{\includegraphics[width=0.48\textwidth,valign=c]{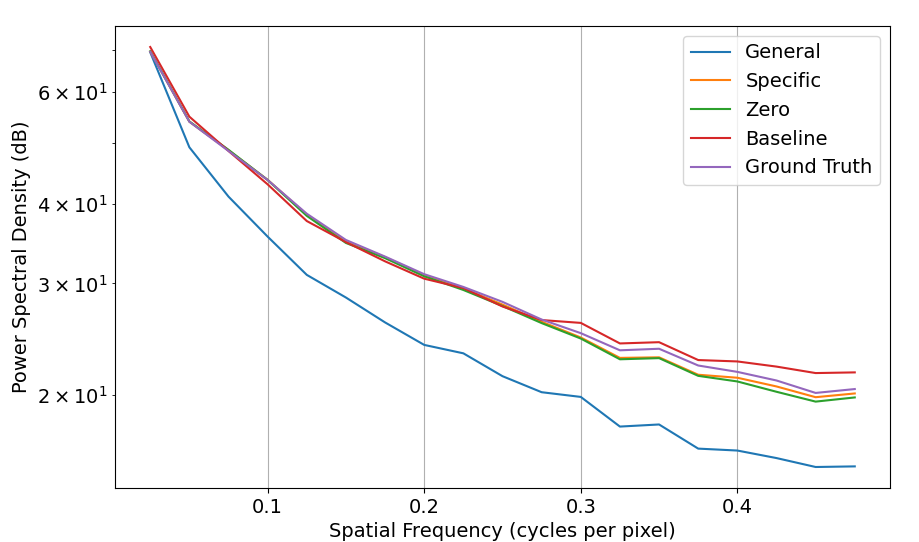}\label{fig:section_III_psd}}
     \hspace{2mm}
     \subfloat[Average PDF. Note that the legend is the same as Figure \ref{fig:section_III_psd}.]{\includegraphics[width=0.48\textwidth,valign=c]{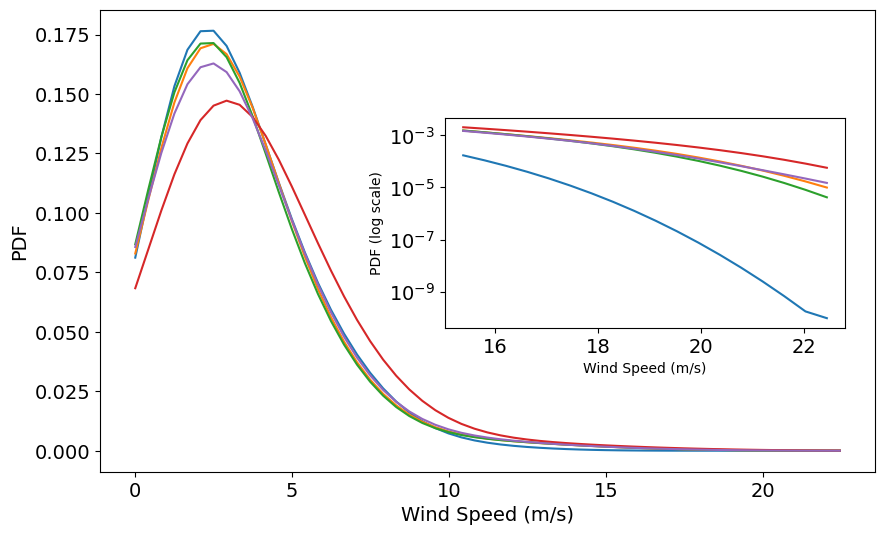}\label{fig:section_III_pdf}}
     
     \caption{MAE, average PSD and average PDF graphs for the model predictions on the test set for domain 2, Figure \ref{fig:domains} (2 899 hourly samples from January 2024 to April 2024) for specific, general and zero models, and for baseline obtained from interpolating the predictor grid on the predictand grid using bi-linear interpolation. The ground truth in the PSD and PDF graphs refers to the HR $UV$, i.e., the predictand.}\label{fig:results_domain_III}
\end{figure}
%Word count temp
%TC:endignore

\subsubsection{Discussion}
The fact that the specific model not only matched but slightly outperformed the zero model on these metrics highlights the effectiveness of the proposed transfer learning approach. It allows for strong performance on a domain without requiring its inclusion in the general model’s training dataset or retraining from scratch. Given that the specialized step of the transfer learning technique converges rapidly—reaching optimal performance in just 10 epochs compared to 50 for training from scratch—it demonstrates significant potential for downscaling across Canada. A general model trained on only a subset of the country can be fine-tuned through transfer learning on multiple specific domains, capturing unique regional characteristics and refining the knowledge within the general model. We anticipate even greater improvements as more domains are incorporated into the general model’s training, paving the way for the development of a comprehensive foundation model.

\subsection{Predictor selection experiment}
\label{subsec:feature:exp}
In this experiment, we compare two alternative predictor combinations with our initial set (\S~\ref{sec:Model_data}). The first set includes only $UV$ at multiple levels to evaluate if this information is sufficient. Conversely, the second set excludes from our initial set all the predictors associated with $UV$, $U$ and $V$.
Two new downscaling models were created using each alternative set of predictors and trained on the data from configuration C. In these two new models, the skip connection in the U-Net architecture (\S~\ref{sec:methodo}.\ref{subsec:CNN_architecture}.\ref{par:Module_II})), which is a nearest-neighbour interpolation of the low-resolution $UV$ variable at 10~m on the predictand grid, was kept.

\subsubsection{Results}
\begin{figure}[ht]
    \centering
     \hspace{2mm}
     \subfloat[RMSE.]{\includegraphics[height=4.4cm]{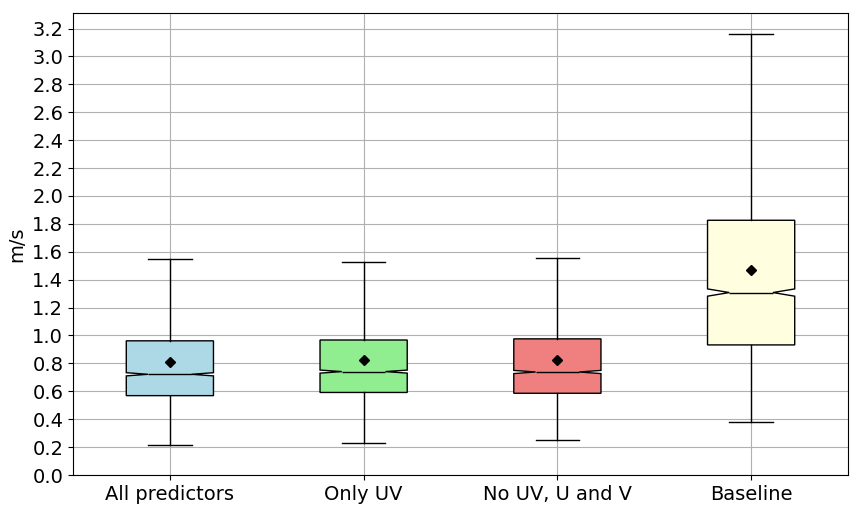}\label{fig:bp_rmse_IIII}}
     \hspace{2mm}
     \subfloat[SSIM.]{\includegraphics[height=4.4cm]{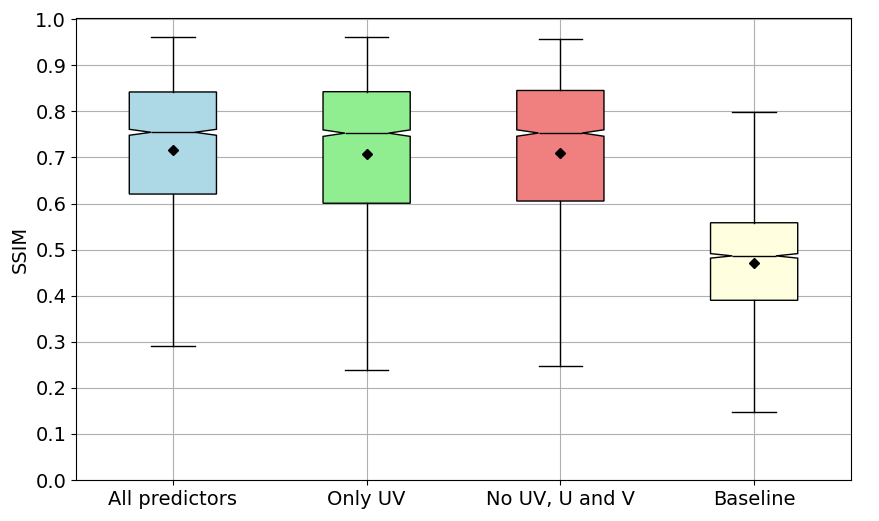}\label{fig:bp_ssim_IIII}}
     
     \caption{RMSE and SSIM metrics box plots for the model predictions on the test set for configuration C, Figure \ref{fig:domains} (28 990 hourly samples from January 2024 to April 2024), for no interpolation with original set of predictors, no interpolation with only $UV$ predictor at multiple levels, and no interpolation with every predictor except $UV$, $U$ and $V$ at multiple levels, and for baseline obtained from interpolating the $UV$ predictor grid at 10m on the predictand grid using bi-linear interpolation. In each graph, the black dot and lines represent respectively the mean and the median, the upper and lower box limits indicate the first (Q1) and third (Q3) quartiles and the whiskers depict the highest (lowest) value within the $1.5 \times$ (Q3-Q1) above Q3 (below Q1).}\label{fig:results_IIII}
\end{figure}

In Figure \ref{fig:results_IIII}, we see that the three downscaling models based on the different predictor combinations share almost the same performance on both metrics. The comparable performance of the associated models suggests that using only $UV$ at multiple levels in the predictor set, or excluding it from the initial set, provides similar predictive information to that contained in the initial set.

%Word count temp
%TC:ignore
%Word count temp
%TC:endignore

\begin{figure}[ht]
    \centering
    \includegraphics[width=0.99\textwidth]{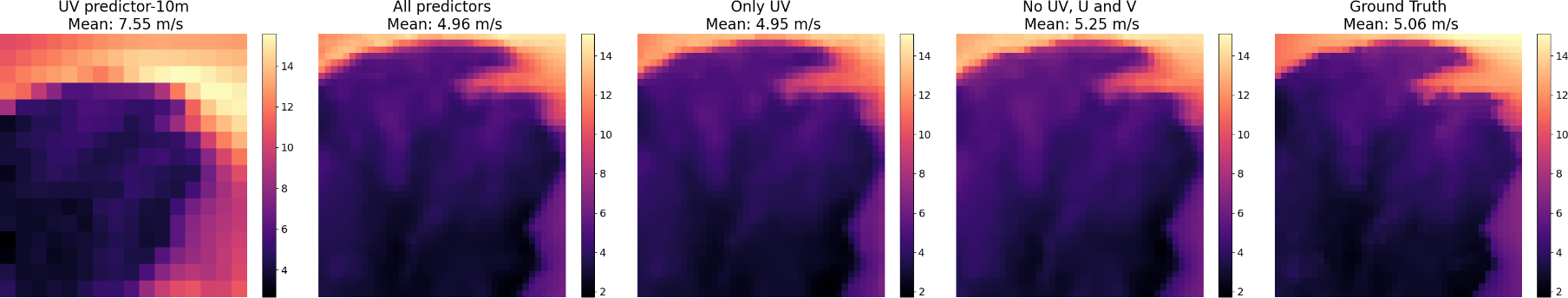}
    \caption{Test sample on 2024-01-05 at 18Z for configuration C on domain \#1 (Figure \ref{fig:domains}). From left to right: $UV$ predictor on the low-resolution grid, downscaling model predictions with the initial set of predictors, with only the $UV$ predictor at multiple levels, and with every predictor except $UV$, $U$ and $V$ at multiple levels, and finally the predictand (ground truth).
    Note that the mean value in m/s is displayed at the top of each image to facilitate comparison for the reader.}\label{fig:prediction}
\end{figure}

In contrast to the other experiments, we look here at the prediction for a specific hour over domain \#1, instead of the MAE, see Figure~\ref{fig:prediction}. This sampled hour is representative of the fact that the downscaling models based on the different combinations of predictors performed similarly. Indeed, after thorough examination of all the test set samples, we did not observe any instances where the larger set of predictors brought significant improvements, contrary to our initial expectations. Furthermore, Figure~\ref{fig:prediction}
 illustrates the learned grid alignment process of the downscaling model using the no-interpolation strategy (\S~\ref{sec:methodo}.\ref{subsec:CNN_architecture}).

%Word count temp
%TC:ignore
%Word count temp
%TC:endignore

\subsubsection{Discussion}
This finding is significant for future research, as it suggests that incorporating predictors beyond wind variables does not substantially improve model performance, making it more data-efficient to downscale wind using only $UV$ at multiple levels. Similar results were reported in related studies, where high-resolution topography was found to reduce training time without impacting model accuracy \citep{Manepalli}, prompting further investigation in this area. Additionally, \citet{Hohlein, Sekiyama} showed that most predictors had minimal influence on the final prediction, with the exception of topography, which was not observed in our results.

\subsection{Wind ramp detection exploration} \label{subsub:windramp}
In wind farms, an important phenomenon to detect is wind power ramps defined as sharp increases or decreases in wind power occurring over a short period of time. The criteria used to define a ramp are variable and in our case defined as a 70\% change in nominal power within a period of two hours. Some ramps are caused by large-scale weather phenomena, while others result from local events requiring high-resolution forecasts to be captured \citep{olson2019improving}. 
In this final section, we explore the potential of our downscaled wind velocities to capture small-scale phenomena that are not detectable in the low-resolution data.

\subsubsection{Results}
\begin{figure}[ht]
    \centering
    \subfloat[Velocity time series, six-day period within the test set in February 2024.]{\includegraphics[width=0.42\textwidth,valign=c]{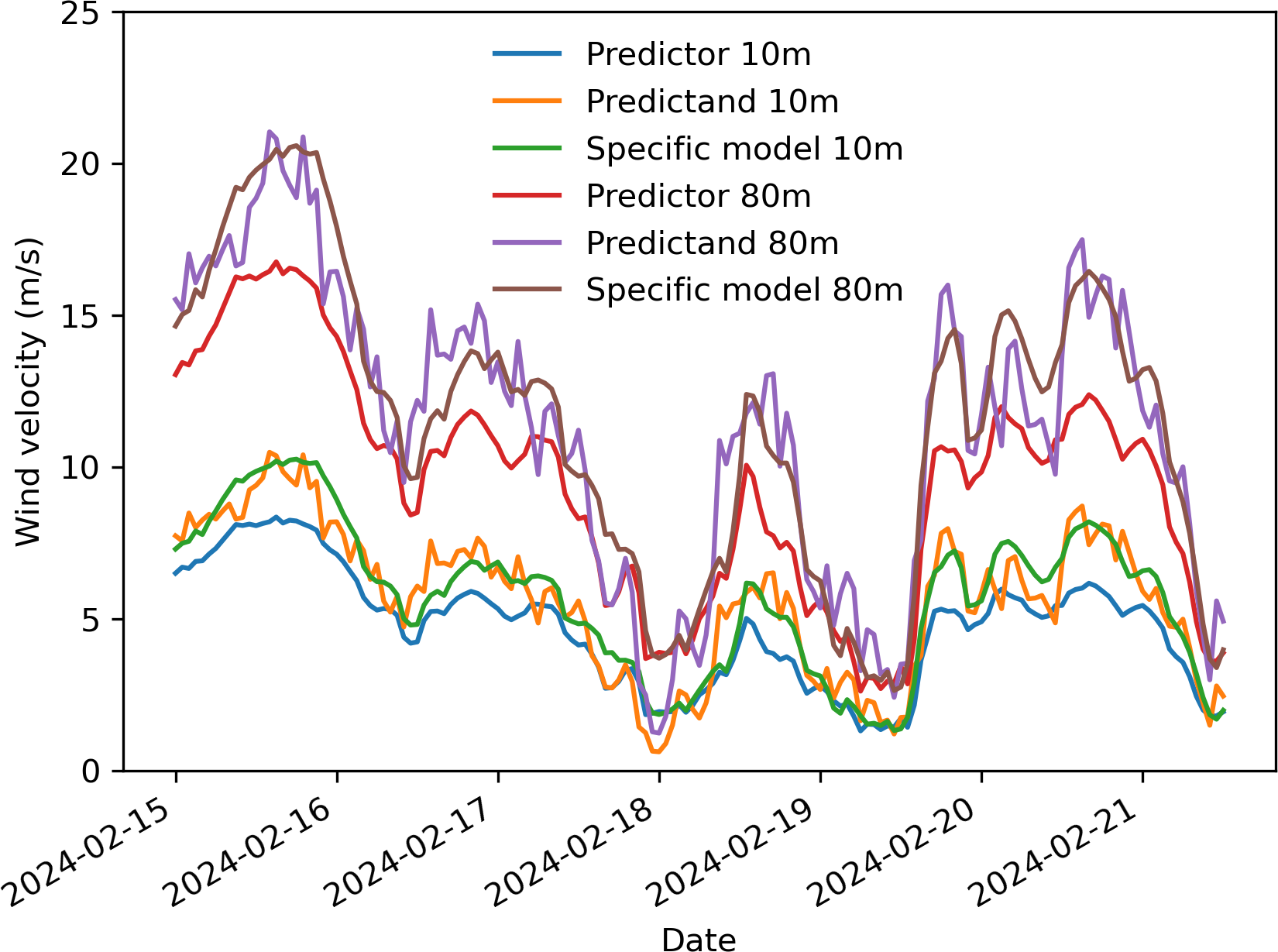}\label{fig:Ramp_TS1}}
     \hspace{2mm}
     \subfloat[Normalized power as a function of time showing an up-ramp centered on February 19th, 2024 at 16Z, detected by the predictand and the prediction from the specific model.]{\includegraphics[width=0.42\textwidth,valign=c]{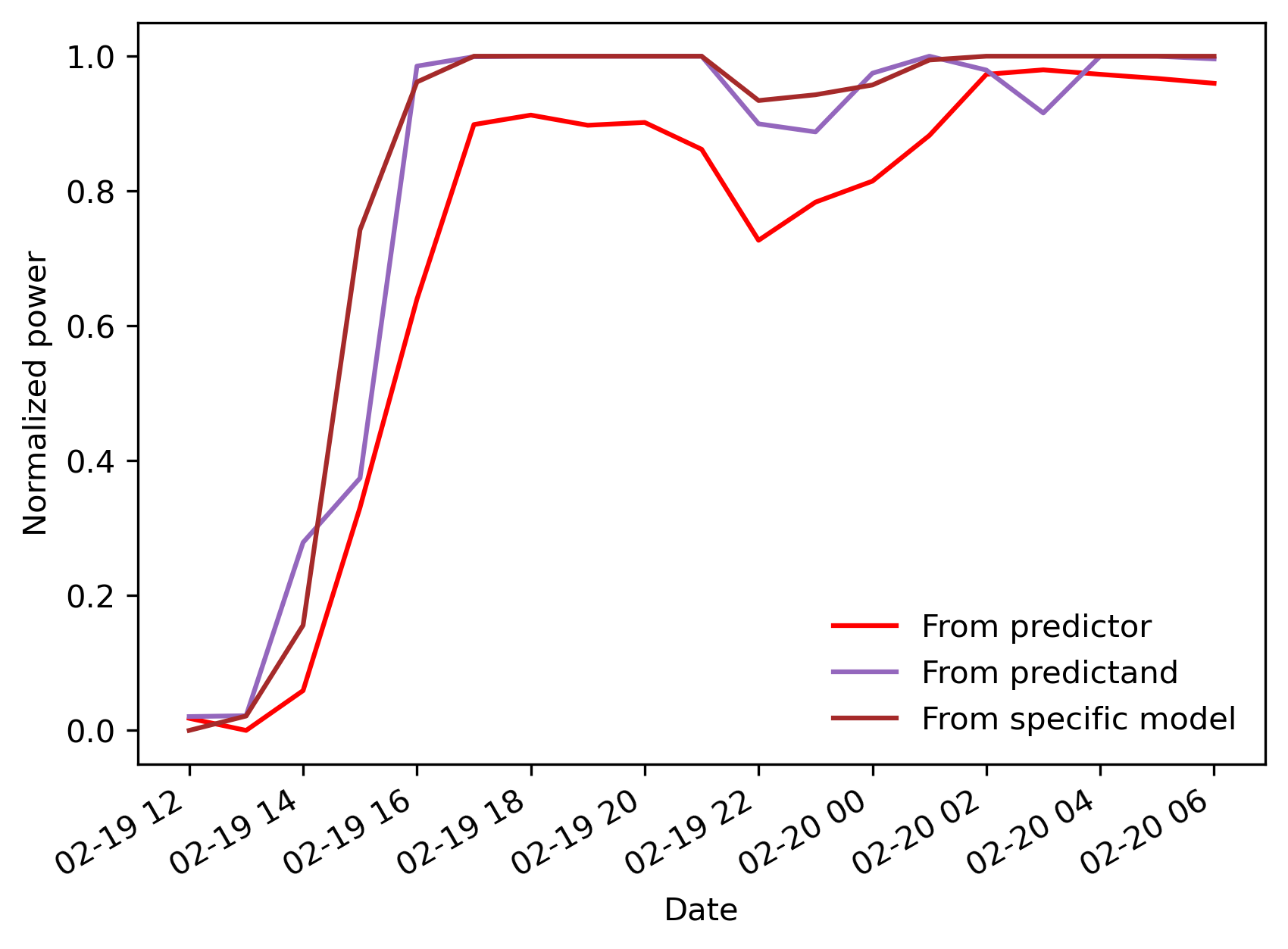}\label{fig:Ramp_Power1}}
     \hspace{2mm}
     \subfloat[Velocity time series, five-day period within the test set in March 2024. Note that the legend
is the same as Figure \ref{fig:Ramp_TS1}.]
{\includegraphics[width=0.42\textwidth,valign=c]{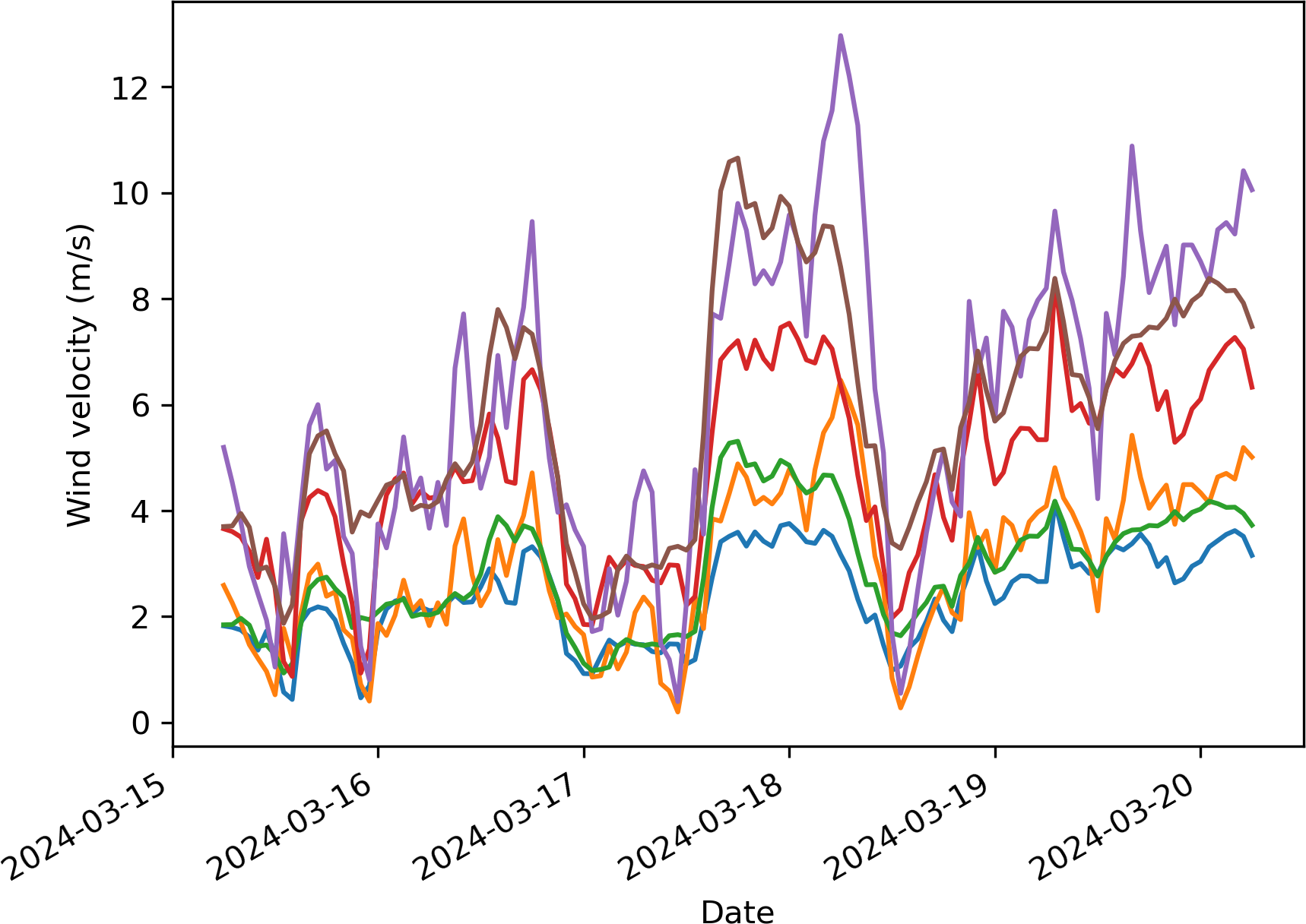}\label{fig:Ramp_TS2}}
 \hspace{2mm}
     \subfloat[Normalized power as a function of time showing a down ramp centered on March 18th, 2024 at 10Z, detected by the predictand but not by the predictor nor by the prediction from the specific model. Note that the legend
is the same as Figure \ref{fig:Ramp_Power1}.]{\includegraphics[width=0.42\textwidth,valign=c]{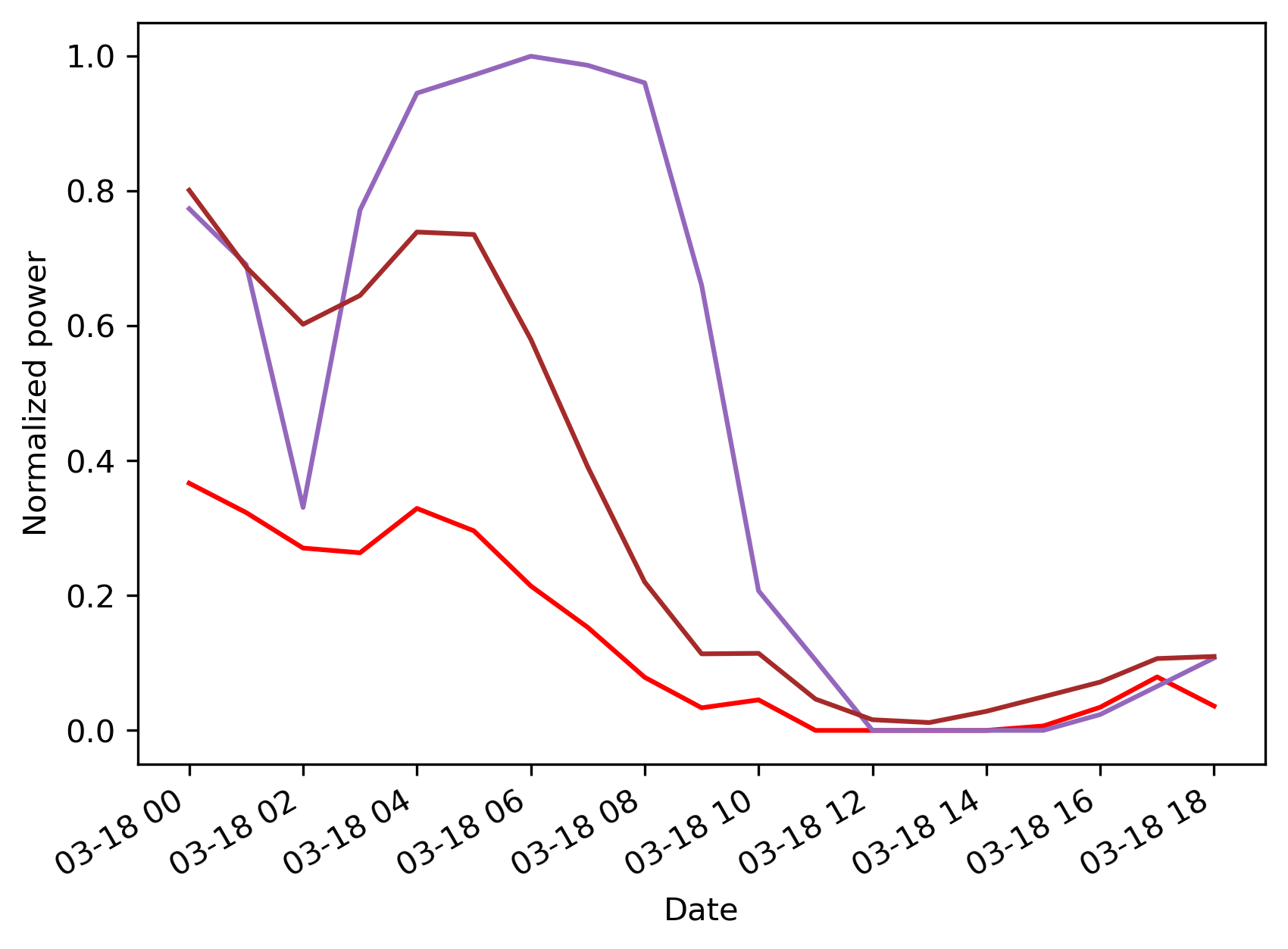}\label{fig:Ramp_Power2}}
     \caption{Velocity time series (left column) and normalized power as a function of time obtained from applying a generic power curve on part of the velocity time series at 80m (right column) for two different periods (upper and lower row) extracted at the (48.99,-64.45) lat-long position near the shore within domain \#1 (see Figure \ref{fig:Dom_gaspe}).  The predictand, the wind velocity predictor and the prediction of the specific model at 10m height were interpolated at the (48.99,-64.45) lat-long position using nearest neighbour and were extrapolated at a 80m height using the logarithmic profile law under neutral stability conditions.}\label{fig:results_IIIII}
\end{figure}

Figure \ref{fig:Ramp_TS1} shows a wind velocity times series extracted at a specific coordinate in domain \#1 from the predictand, the wind $UV$ predictor, and the prediction from the specific model, i.e., the general downscaling model after specialization on domain \#1 (\S~\ref{sec:Results_Disc}.\ref{TransferLearning_exp}). In addition, these velocities were extrapolated at 80m using the logarithmic profile law under neutral stability conditions \citep{ECMWF_rep2020}. It can be seen that the downscaled wind velocities are overall much closer to the predictand than the predictor,  potentially allowing some ramps missed by the predictor to be captured in the downscaled signal. To verify this, we used a generic power curve \citep{PowerCurveSite} to translate the wind velocity at a hub height of 80m to a normalized power, and applied an algorithm to detect wind power ramps from the resulting time series \citep{Bianco2016}. Figure \ref{fig:Ramp_Power1} shows an example of an up-ramp detected by both the predictand and the specific model’s prediction, but missed by the predictor.

Figure \ref{fig:Ramp_TS2} shows a time series of wind velocities for which a down-ramp observed in the predictand (centered at 10Z on March 18th, as shown in figure  \ref{fig:Ramp_Power2}) was not captured by either the predictor or the specific model’s prediction. Although the velocity values are still seen to be overall improved by the downscaling operation, it seems that for some events where the wind predictor is originally far from the predictand, as seen around March 18th at 10Z, the downscaling model is not able to recover all the small-scale phenomena. This may be due to the occurrence of specific weather events that the model needs more exposure to during training in order to predict accurately.
%Word count temp
%TC:ignore
%Word count temp
%TC:endignore

\subsubsection{Discussion}
These results suggest that our downscaling method, by bringing wind velocities closer to the predictand, has the potential to capture power ramps that were not initially detected at a lower resolution. This is a promising first step in assessing whether the resolution enhancement from our model is sufficient to capture various small-scale phenomena. Further investigation is needed, including analyzing weather patterns on the downscaled maps that correspond to specific small-scale phenomena, such as land and sea breezes or local events driving power ramps. Notably, because our downscaling model was optimized not only to minimize pixel-level errors but also to maximize image similarity, it is likely to be effective in reproducing wind patterns across the map.

\clearpage
\section{Conclusion and future work}
This study, in which a U-Net architecture was built for downscaling wind forecasts from 10km to 2.5km on unaligned grids on various domains across Canada,  presented convincing evidence of the following five key findings:
\begin{itemize}
\item Our new memory-efficient architecture that deals with unaligned grids by letting the interpolation of the predictor data over the predictand grid to be learned by the network outperformed traditional interpolation strategies by using significantly less memory and giving better or similar performances than the latter. We believe this can provide valuable guidelines for future downscaling studies, helping to adapt the chosen architecture to effectively learn the interpolation step and utilize a larger dataset; 
\item Geographically dispersed domains or an increased number of domains yield similar performance on a given domain compared to using closely located domains. This means that a larger, more diverse, dataset can be used for training, helping to build a more generalizable model;
\item Performing transfer learning from a general model trained on numerous, geographically dispersed domains enables the use of specialized models for specific regions, allowing coverage of the entire Canadian territory without requiring excessive training data and ensuring efficient training with strong performance across vast areas;
\item Using only the wind velocity $UV$ at multiple vertical levels as a predictor gives similar performance than using multiple and varied predictors. This suggests that, initially, it may be more efficient to focus on wind predictors for downscaling.
\item Some wind power ramp events not present in the low-resolution forecasts could be observed in the downscaled ones, indicating a potential for capturing meteorological phenomena that cannot be seen at lower resolutions.
\end{itemize}

It is interesting to note that the wind velocity components $U$ or $V$ could also be downscaled independently with their own model, as well as any other meteorological variable. Furthermore, probabilistic forecasts that would be obtained by applying this downscaling method on the members of an ensemble prediction system have the potential to show a greater skill to reproduce small-scale phenomena, which opens great opportunities for improving, among others, wind power forecasts. 

Building on our findings, several promising avenues for future work emerge. Two potential directions are the incorporation of physics-informed neural networks, which could enhance model accuracy by embedding domain-specific physical laws directly into the learning process, and the integration of temporal information into the model training, as through the use of a 3D U-Net architecture, to better capture the dynamic nature of atmospheric processes.

% \acknowledgments
\section*{Acknowledgments}
This project was funded by the PERD program from Natural Resources Canada. The contribution from ECCC colleagues is acknowledged, among others Franco Petrucci, Bertrand Denis, Didier Davignon, Oleksandr Huziy and André Plante. The third author would like to acknowledge funding by the Natural Sciences and Engineering Research Council of Canada (NSERC) and by Fonds de Recherche du Québec Nature et Technologies (FRQNT).

\clearpage
\appendix
\section{Appendix A: Hyper parameters selection}
\begin{table}[ht]
    \caption{Selected hyper-parameters for the general models, from an exhaustive search performed using 200 Bayesian trials  for each strategy on the different configurations, which took four training days. Notes: For each configuration, the columns in the first table section describe the hyper-parameters which are respectively the number of epochs that the callback stopped the training at (every model is set to do 70 epochs if Early Stopping callback is not activated) ($Ep$), the learning rate ($LR$), the LeakyReLU ($\alpha$) parameter, the loss weight ($\lambda$) parameter, and the Dropout ($Drop$) value. In the second section, the columns describe the Training Time ($TT$) necessary to train the network with the given set of hyper-parameters in minutes ($min$), the Prediction Time ($PT$) for the whole test set in seconds ($s$), the inference time ($IT$) for a single input in milliseconds ($ms$), the number of trainable parameters ($TP$) in thousands ($k$) and the memory consumption for the model ($MEM$) in megabytes ($MB$).}
    \label{tab:hpgraph}
    \centering
    \footnotesize
    \begin{tabular}{|l|lllll|lllll|}
        \hline
        Conf. & $Ep$  & $LR$ & $\alpha$ & $\lambda$ & $Drop$ & $TT$ & $PT$ & $IT$ & $TP$ & $MEM$\\
                &         &    &          &           &         & (min)                  &  (s) & (ms) &  (k) & (MB) \\
        \hline
        Conf. A: bi-linear & 26 & 1e-3 & 0.25 & 1.00 & 0.25 & 48 & 5 & 9 & 37 142.738 & 148.57 \\
        Conf. A: Nearest-Neighbour & 28 & 1e-3 & 0.25 & 1.00 & 0.25 & 52 & 5 & 9 & 37 142.738 & 141.66\\
        Conf. A: No-interpolation & 32 & 1e-3 & 0.30 & 0.50 & 0.30 & 50 & 5 & 7 & 37 142.738 & 141.66\\
        \hline\hline
        Conf. B: No-interpolation & 24 & 1e-3 & 0.30 & 1.00 & 0.25 & 41 & 4 & 6 & 37 142.738 & 141.66\\
        \hline\hline
        Conf. C: No-interpolation  & 26 & 1e-3 & 0.25 & 1.00 & 0.35 & 80 & 7 & 6 & 37 142.738 & 141.66 \\
        \hline
    \end{tabular}
\end{table}

\begin{table}[ht]
    \caption{Selected hyper-parameters for specific models where we used the same strategy as the general model, but the results were quicker to obtain because fine-tuning a model that has already been trained on similar data converges towards the optimal solution quicker than a network trained from scratch.}
    \label{tab:hp_tl_graph}
    \centering
        \footnotesize
    \begin{tabular}{|l|lllll|lllll|}
        \hline
        Conf. & $Ep$  & $LR$ & $\alpha$ & $\lambda$ & $Drop$ & $TT$ & $PT$ & $IT$ & $TP$ & $MEM$\\
                &         &    &          &           &         & (min)                  &  (s) & (ms) &  (k) & (MB) \\
        \hline
        Conf. C: Specific domain \#2  & 25 & 1e-5 & 0.30 & 1.0 & 0.20 & 4 & 1 & 6 & 37 142.738 & 141.66 \\
        Conf. C: Zero domain \#2  & 41 & 1e-4 & 0.25 & 1.0 & 0.35 & 9 & 1 & 6 & 37 142.738 & 141.66 \\
        \hline
    \end{tabular}
    \caption*{Notes: Ibid as \ref{tab:hpgraph}.}
\end{table}

\clearpage
\bibliographystyle{authordate1}
\bibliography{references}

\clearpage
\section{Supplemental Material}
\subsection{Interpolation versus non-interpolation strategy experiment}
Here are all the results for domains \#1 to \#5. 
\begin{figure}[ht]
    \centering
     \hspace{2mm}
     \subfloat[RMSE.]{\includegraphics[height=4.6cm,valign=c]{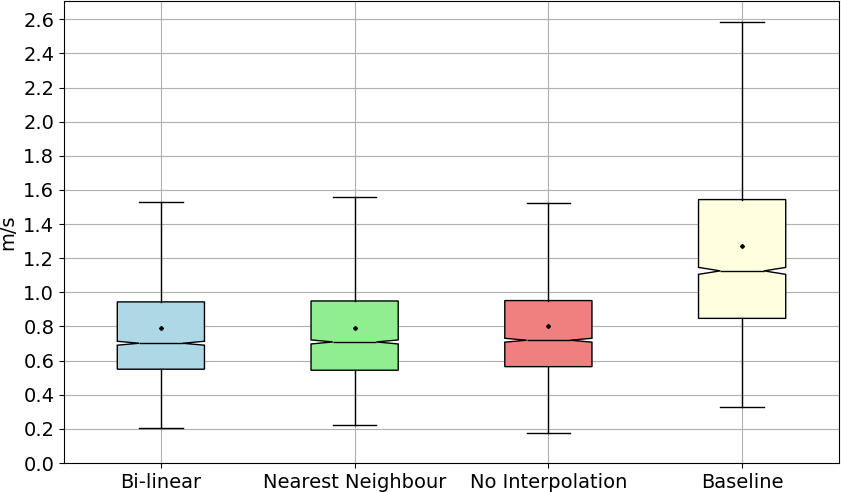}\label{fig:bp_rmse_d1}}
     \hspace{2mm}
     \subfloat[MAE.]{\includegraphics[height=4.6cm,valign=c]{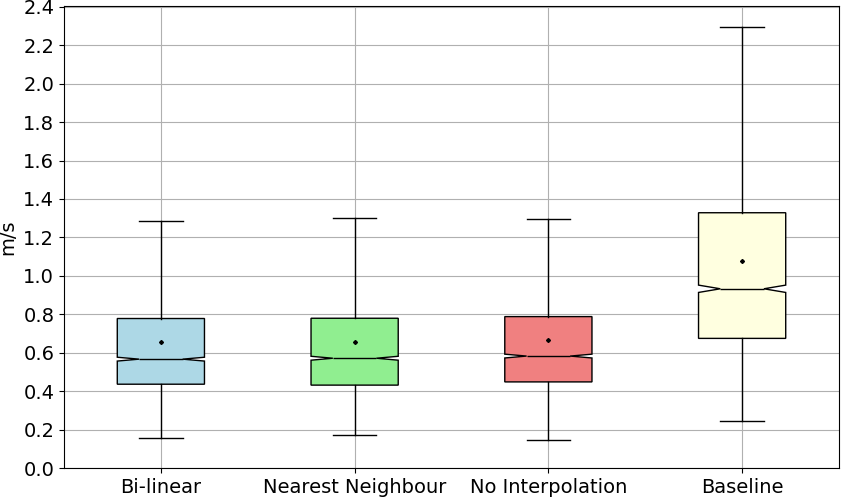}\label{fig:bp_mae_d1}}
     \hspace{2mm}
     \subfloat[SSIM.]{\includegraphics[height=4.6cm,valign=c]{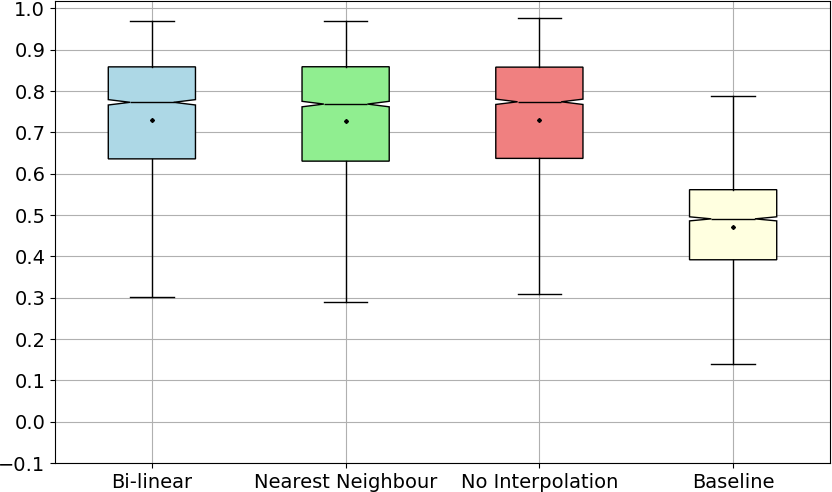}\label{fig:bp_ssim_d1}}
     
     \caption{RMSE, MAE and SSIM boxplots for each downscaling model (bi-linear or nearest neighbour interpolation strategy, no-interpolation strategy), and baseline computed on the test set of configuration A for domain \#1 with the five domains in Eastern Québec and New Brunswick (2 899 hourly samples from January 2024 to April 2024). In each boxplot, the black dot and line represent respectively the mean and the median, the upper and lower box limits indicate the first (Q1) and third (Q3) quartiles and the whiskers depict the highest (lowest) value within the $1.5 \times$ (Q3-Q1) above Q3 (below Q1).}\label{fig:results_I_d1}
\end{figure}

\begin{figure}[ht]
    \centering
    \subfloat[Pixel-wise MAE (m/s) for the different strategies and the baseline. The colorscale is capped at the max MAE of the three strategies, which makes the baseline's values higher than the maximum of the scale to saturate in some locations. The value at the top of each image is the average over the domain.]{\includegraphics[width=0.99\textwidth,valign=c]{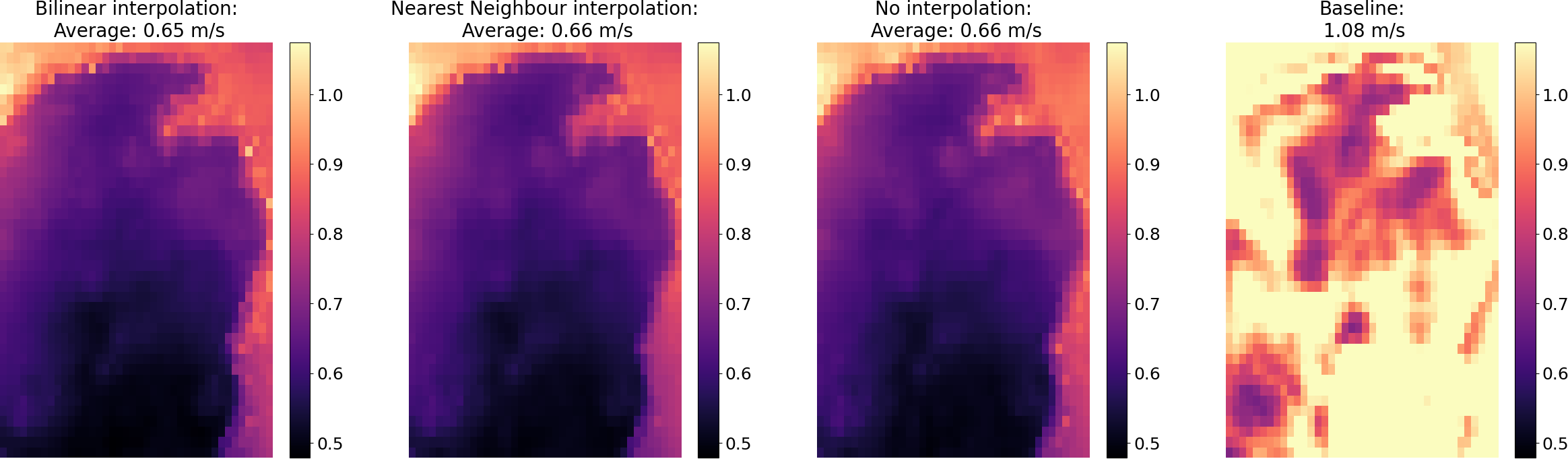}
     \label{fig:error_section_I_d1}} \\
     %\hspace{2mm}
     \subfloat[Average Power Spectrum Density (PSD).]{\includegraphics[width = 0.48\textwidth,valign=c]{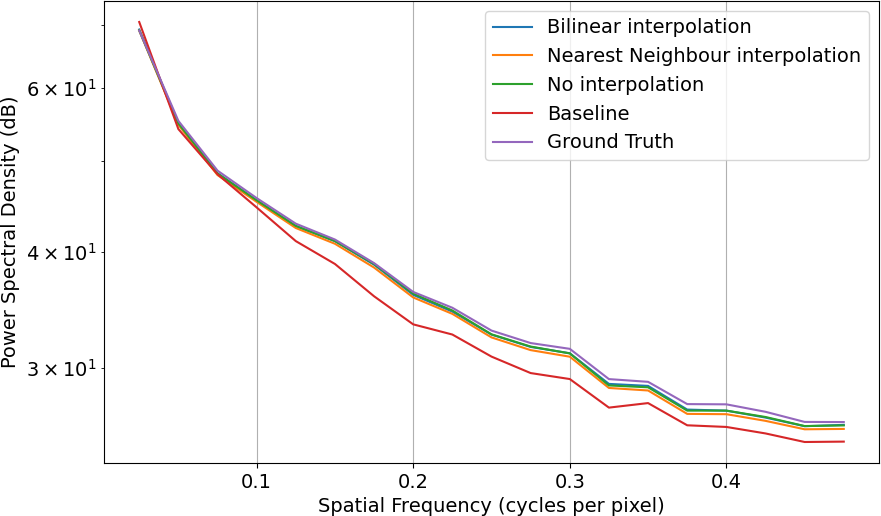}\label{fig:results_psd_I_d1}}
     \hspace{2mm}
     \subfloat[Average Probability Density Function (PDF). Note that the legend is the same as Figure \ref{fig:results_psd_I_d1}.]{\includegraphics[width = 0.48\textwidth,valign=c]{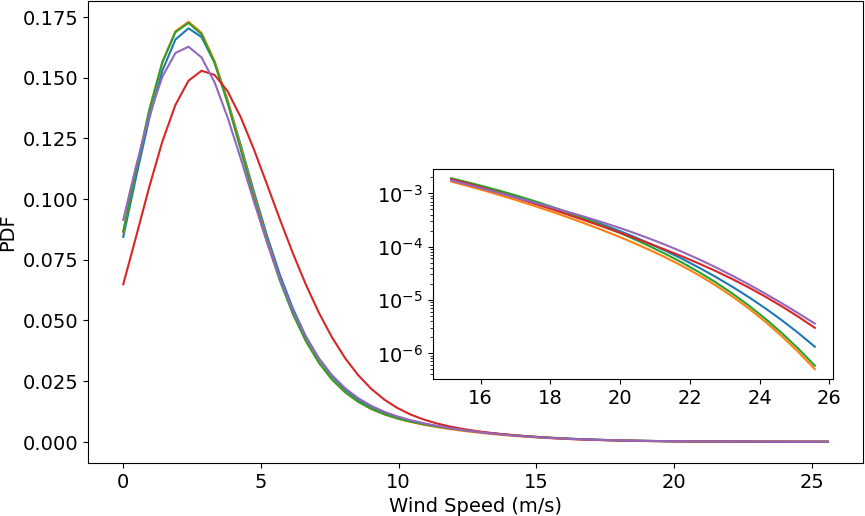}\label{fig:result_pdf_I_d1}}
     
     \caption{MAE, average PSD and average PDF on the test set for domain \#1 of configuration A in Eastern Québec and New Brunswick (2899 hourly samples from January 2024 to April 2024), for each downscaling model (bi-linear or nearest neighbour interpolation strategy, no-interpolation strategy), and baseline. The ground truth in the PSD and PDF graphs refers to the HR $UV$, i.e., the predictand.}\label{fig:results_domain_I_d1}
\end{figure}

\begin{figure}[ht]
    \centering
     \hspace{2mm}
     \subfloat[RMSE.]{\includegraphics[height=4.6cm,valign=c]{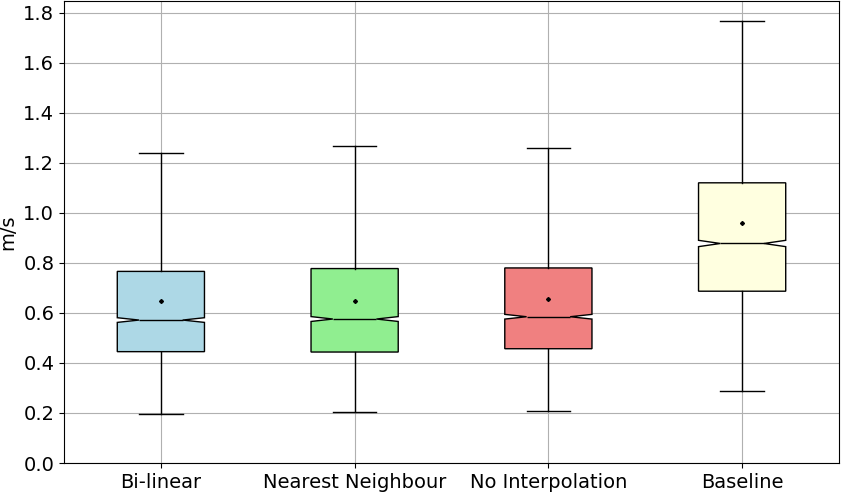}\label{fig:bp_rmse_d2}}
     \hspace{2mm}
     \subfloat[MAE.]{\includegraphics[height=4.6cm,valign=c]{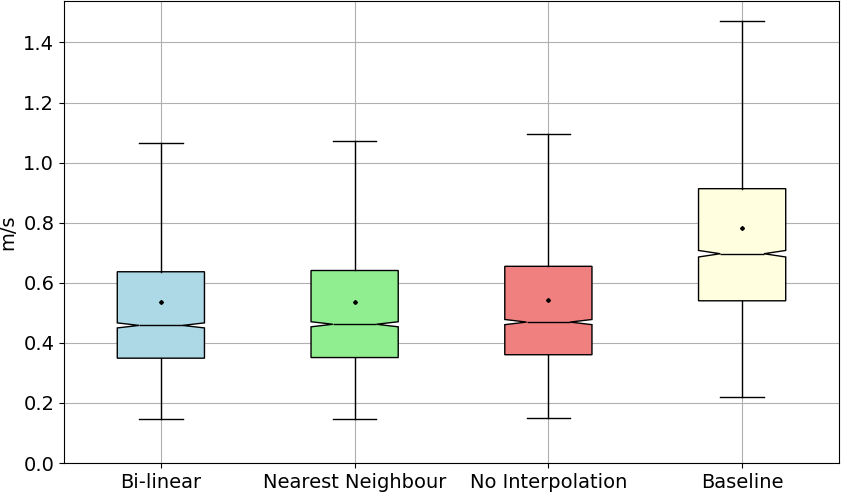}\label{fig:bp_mae_d2}}
     \hspace{2mm}
     \subfloat[SSIM.]{\includegraphics[height=4.6cm,valign=c]{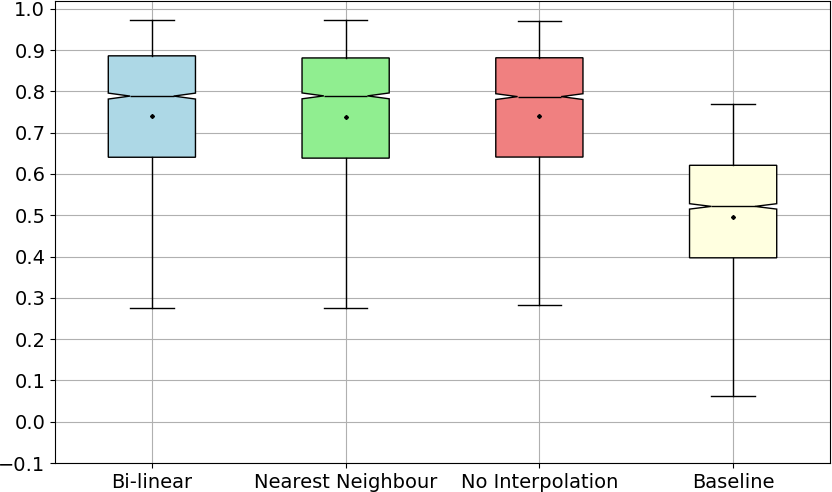}\label{fig:bp_ssim_d2}}
         \caption{RMSE, MAE and SSIM boxplots for each downscaling model (bi-linear or nearest neighbour interpolation strategy, no-interpolation strategy), and baseline computed on the test set of configuration A for domain \#2 with the five domains in Eastern Québec and New Brunswick (2 899 hourly samples from January 2024 to April 2024). In each boxplot, the black dot and line represent respectively the mean and the median, the upper and lower box limits indicate the first (Q1) and third (Q3) quartiles and the whiskers depict the highest (lowest) value within the $1.5 \times$ (Q3-Q1) above Q3 (below Q1).}\label{fig:results_I_d2}
\end{figure}

\begin{figure}[ht]
    \centering
    \subfloat[Pixel-wise MAE (m/s) for the different strategies and the baseline. The colorscale is capped at the max MAE of the three strategies, which makes the baseline's values higher than the maximum of the scale to saturate in some locations. The value at the top of each image is the average over the domain.]{\includegraphics[width=0.99\textwidth,valign=c]{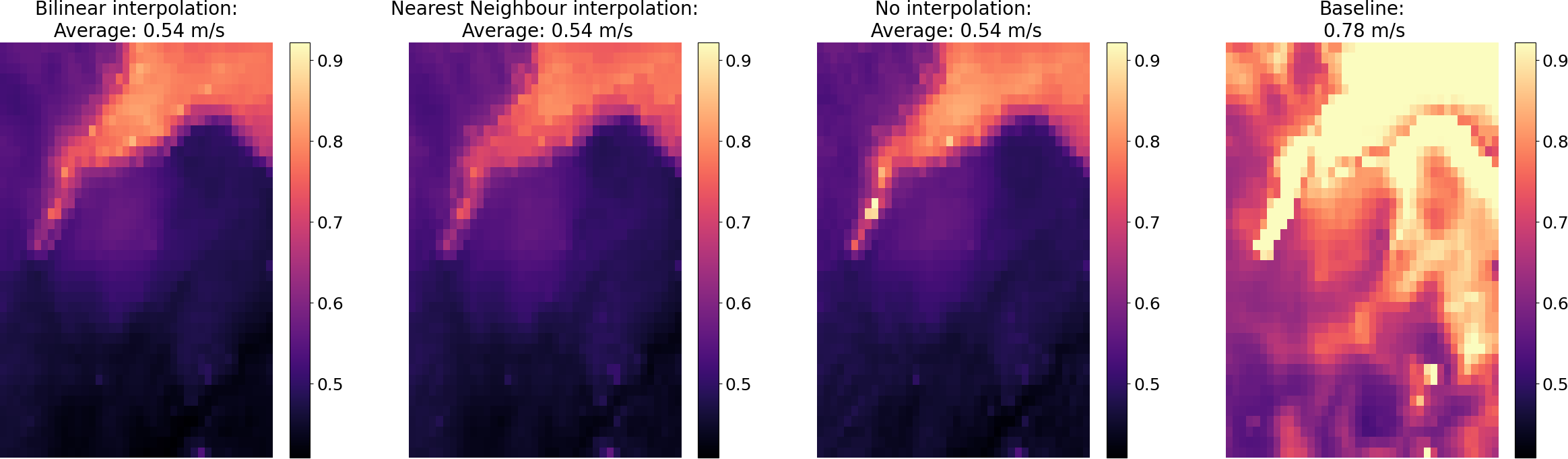}
     \label{fig:error_section_I_d2}} \\
     %\hspace{2mm}
     \subfloat[Average PSD.]{\includegraphics[width = 0.48\textwidth,valign=c]{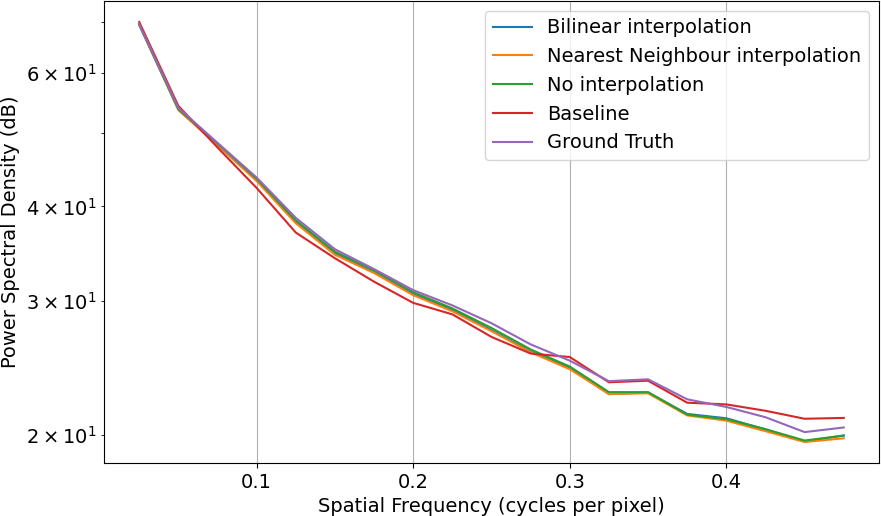}\label{fig:results_psd_I_d2}}
     \hspace{2mm}
     \subfloat[Average PDF. Note that the legend is the same as Figure \ref{fig:results_psd_I_d2}.]{\includegraphics[width = 0.48\textwidth,valign=c]{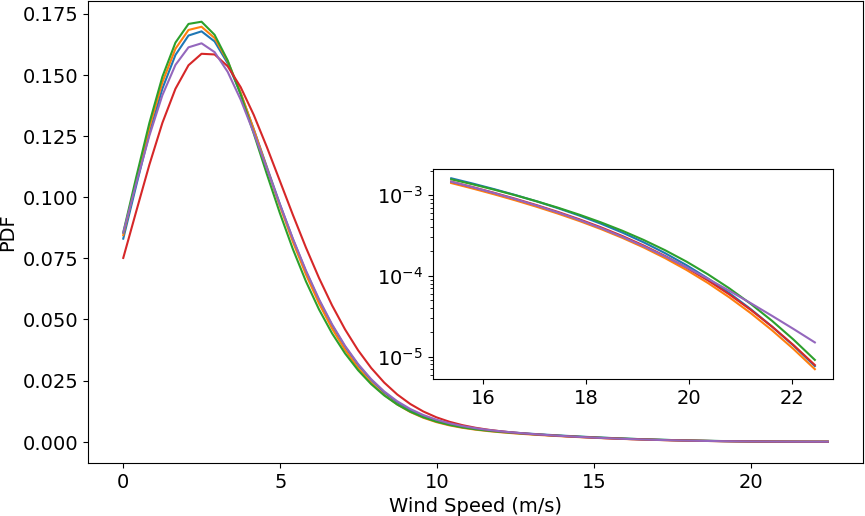}\label{fig:result_pdf_I_d2}}
     
     \caption{MAE, average PSD and average PDF on the test set for domain \#2 of configuration A in Eastern Québec and New Brunswick (2899 hourly samples from January 2024 to April 2024), for each downscaling model (bi-linear or nearest neighbour interpolation strategy, no-interpolation strategy), and baseline. The ground truth is also shown in the PSD and PDF graphs. }\label{fig:results_domain_I_d2}
\end{figure}

\begin{figure}[ht]
    \centering
     \hspace{2mm}
     \subfloat[RMSE.]{\includegraphics[height=4.6cm,valign=c]{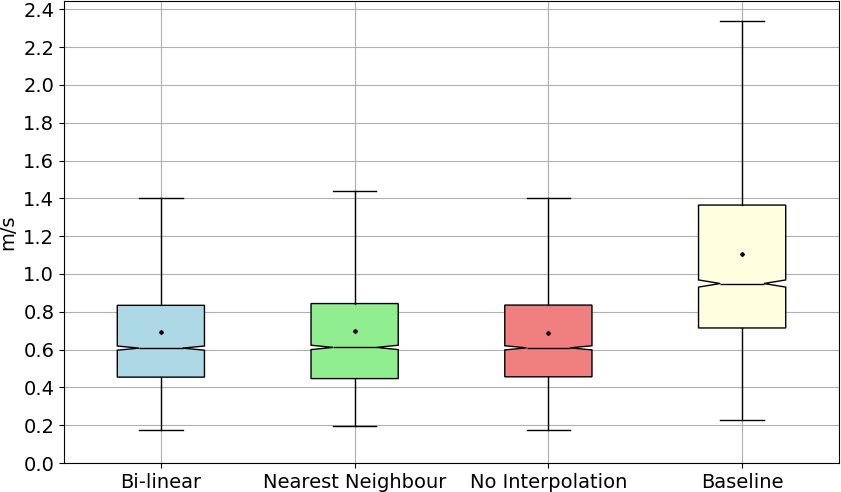}\label{fig:bp_rmse_d3}}
     \hspace{2mm}
     \subfloat[MAE.]{\includegraphics[height=4.6cm,valign=c]{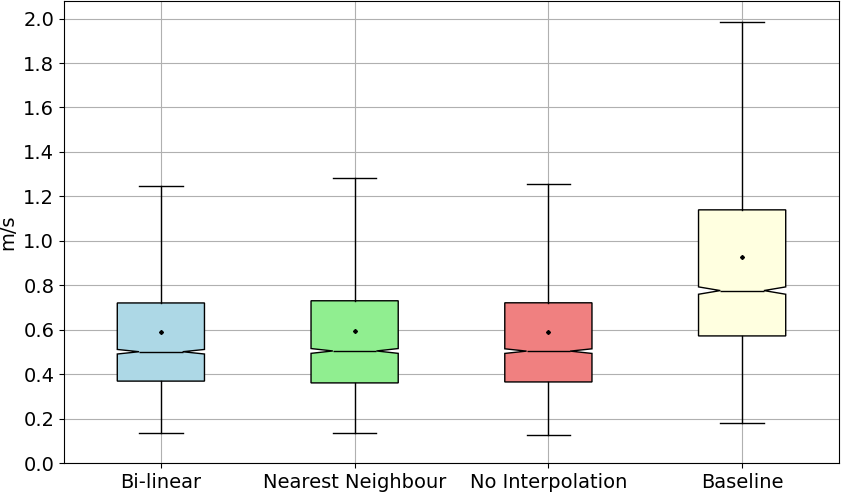}\label{fig:bp_mae_d3}}
     \hspace{2mm}
     \subfloat[SSIM.]{\includegraphics[height=4.6cm,valign=c]{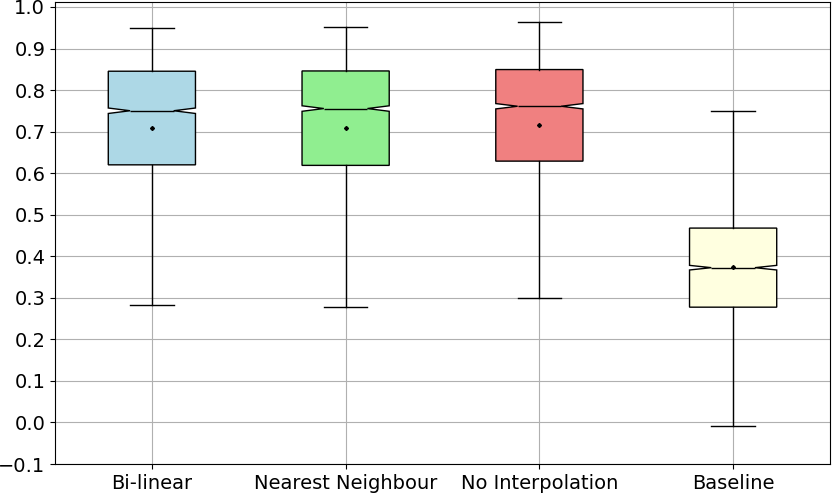}\label{fig:bp_ssim_d3}}
     
     \caption{RMSE, MAE and SSIM boxplots for each downscaling model (bi-linear or nearest neighbour interpolation strategy, no-interpolation strategy), and baseline computed on the test set of configuration A for domain \#3 with the five domains in Eastern Québec and New Brunswick (2 899 hourly samples from January 2024 to April 2024). In each boxplot, the black dot and line represent respectively the mean and the median, the upper and lower box limits indicate the first (Q1) and third (Q3) quartiles and the whiskers depict the highest (lowest) value within the $1.5 \times$ (Q3-Q1) above Q3 (below Q1).}\label{fig:results_I_d3}
\end{figure}

\begin{figure}[ht]
    \centering
    \subfloat[Pixel-wise MAE (m/s) for the different strategies and the baseline. The colorscale is capped at the max MAE of the three strategies, which makes the baseline's values higher than the maximum of the scale to saturate in some locations. The value at the top of each image is the average over the domain.]{\includegraphics[width=0.99\textwidth,valign=c]{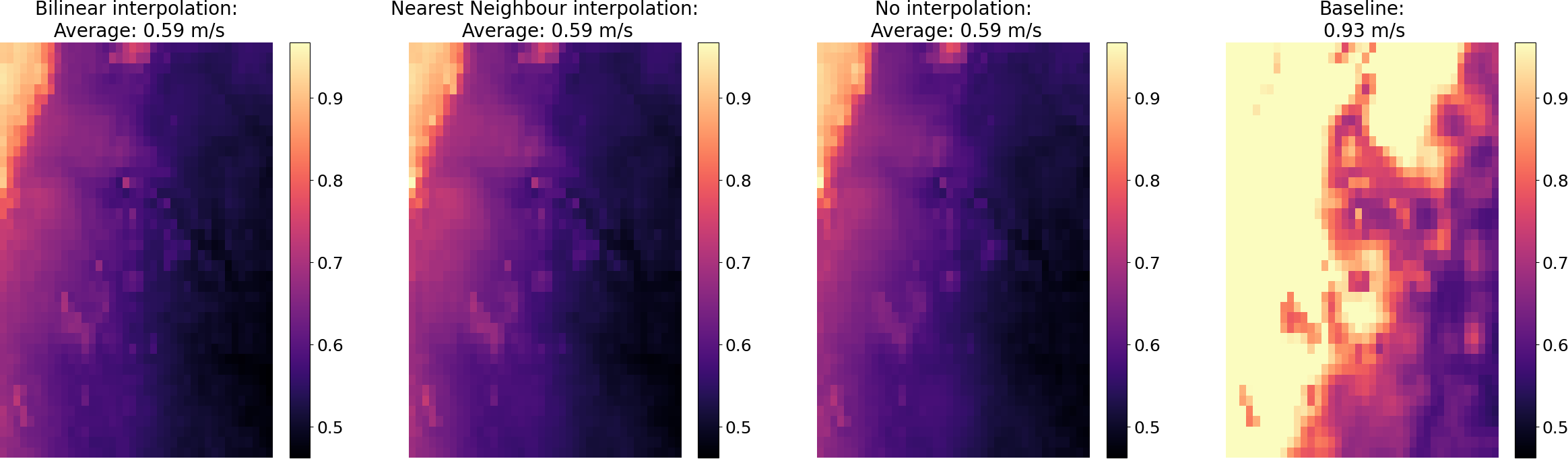}
     \label{fig:error_section_I_d3}} \\
     %\hspace{2mm}
     \subfloat[Average PSD.]{\includegraphics[width = 0.48\textwidth,valign=c]{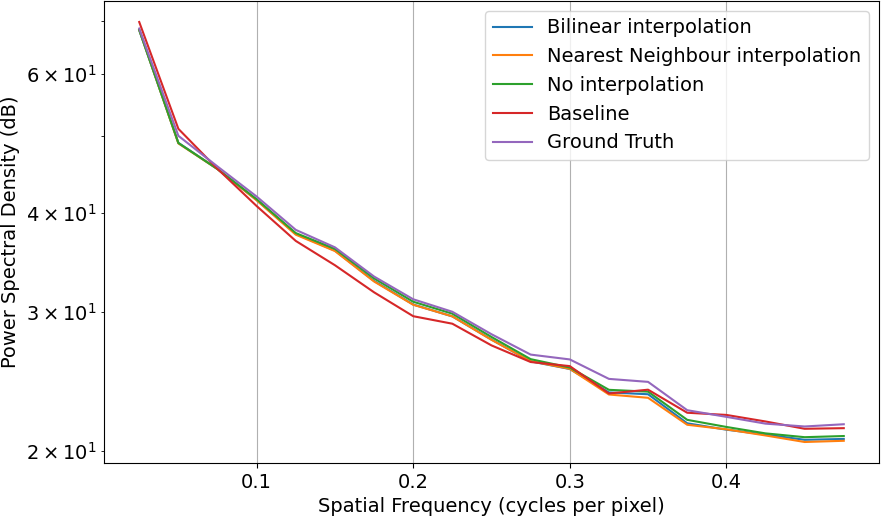}\label{fig:results_psd_I_d3}}
     \hspace{2mm}
     \subfloat[Average PDF. Note that the legend is the same as Figure \ref{fig:results_psd_I_d3}.]{\includegraphics[width = 0.48\textwidth,valign=c]{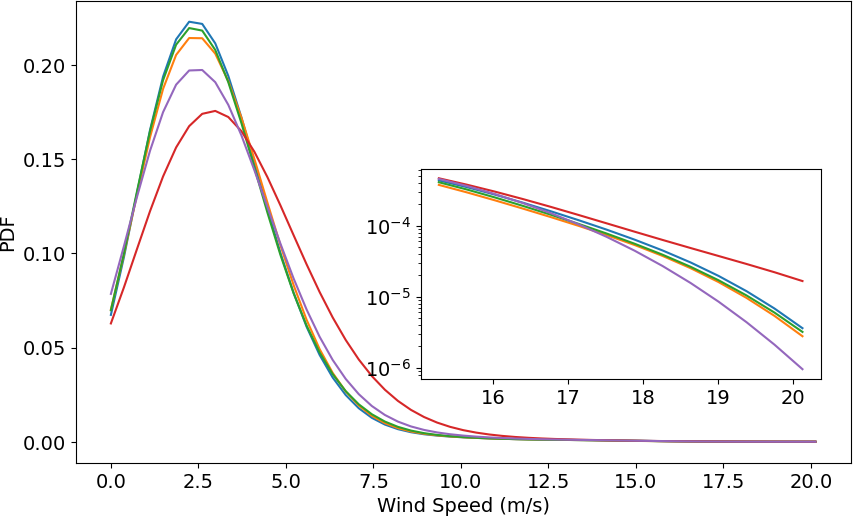}\label{fig:result_pdf_I_d3}}
     
     \caption{MAE, average PSD and average PDF on the test set for domain \#3 of configuration A in Eastern Québec and New Brunswick (2899 hourly samples from January 2024 to April 2024), for each downscaling model (bi-linear or nearest neighbour interpolation strategy, no-interpolation strategy), and baseline. The ground truth in the PSD and PDF graphs refers to the HR $UV$, i.e., the predictand.}\label{fig:results_domain_I_d3}
\end{figure}

\begin{figure}[ht]
    \centering
     \hspace{2mm}
     \subfloat[RMSE.]{\includegraphics[height=4.6cm,valign=c]{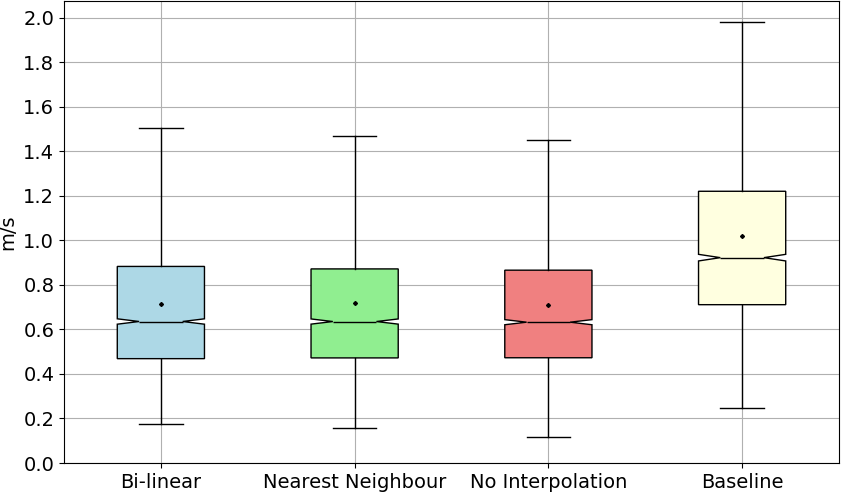}\label{fig:bp_rmse_d4}}
     \hspace{2mm}
     \subfloat[MAE.]{\includegraphics[height=4.6cm,valign=c]{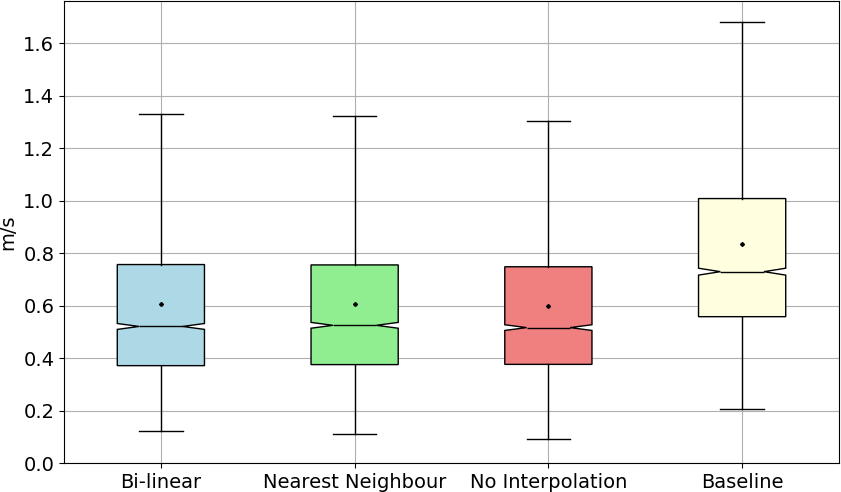}\label{fig:bp_mae_d4}}
     \hspace{2mm}
     \subfloat[SSIM.]{\includegraphics[height=4.6cm,valign=c]{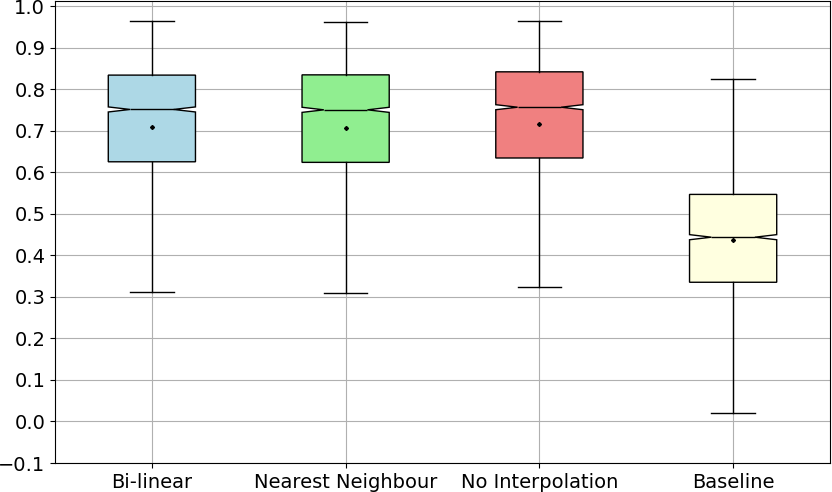}\label{fig:bp_ssim_d4}}
     
     \caption{RMSE, MAE and SSIM boxplots for each downscaling model (bi-linear or nearest neighbour interpolation strategy, no-interpolation strategy), and baseline, computed on the test set of configuration A for domain \#4 with the five domains in Eastern Québec and New Brunswick (2 899 hourly samples from January 2024 to April 2024). In each boxplot, the black dot and line represent respectively the mean and the median, the upper and lower box limits indicate the first (Q1) and third (Q3) quartiles and the whiskers depict the highest (lowest) value within the $1.5 \times$ (Q3-Q1) above Q3 (below Q1).}\label{fig:results_I_d4}
\end{figure}

\begin{figure}[ht]
    \centering
    \subfloat[Pixel-wise MAE (m/s) for the different strategies and the baseline. The colorscale is capped at the max MAE of the three strategies, which makes the baseline's values higher than the maximum of the scale to saturate in some locations. The value at the top of each image is the average over the domain.]{\includegraphics[width=0.99\textwidth,valign=c]{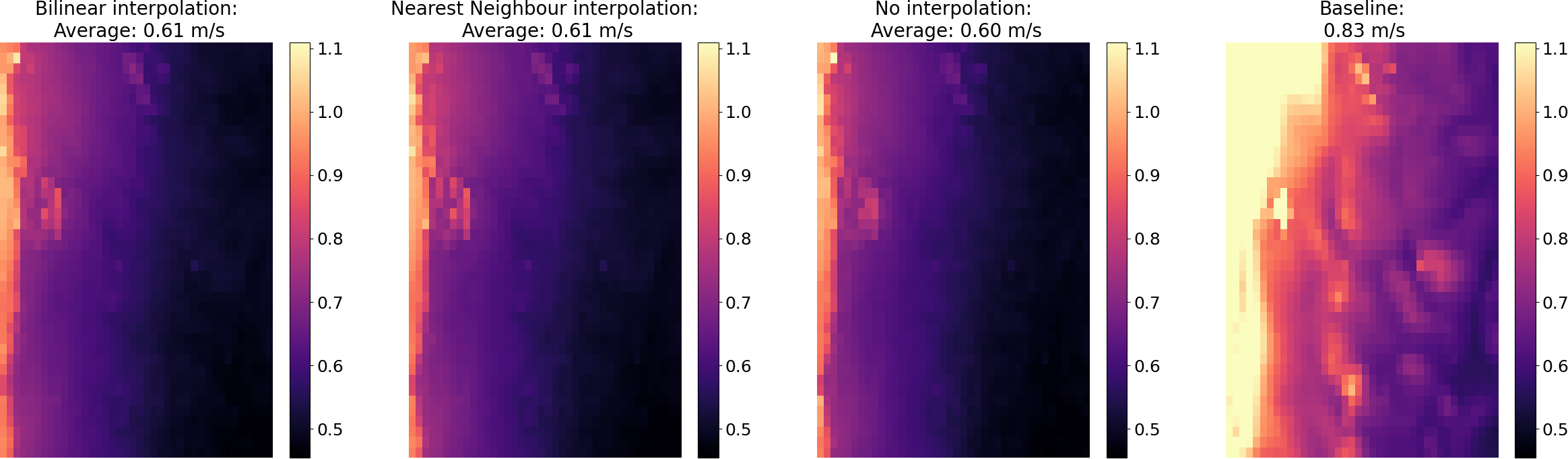}
     \label{fig:error_section_I_d4}} \\
     %\hspace{2mm}
     \subfloat[Average PSD.]{\includegraphics[width = 0.48\textwidth,valign=c]{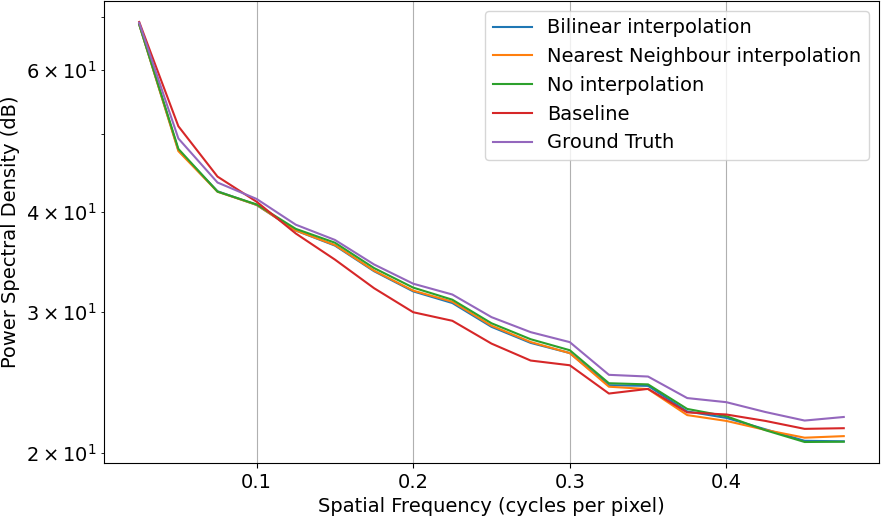}\label{fig:results_psd_I_d4}}
     \hspace{2mm}
     \subfloat[Average PDF. Note that the legend is the same as Figure \ref{fig:results_psd_I_d4}.]{\includegraphics[width = 0.48\textwidth,valign=c]{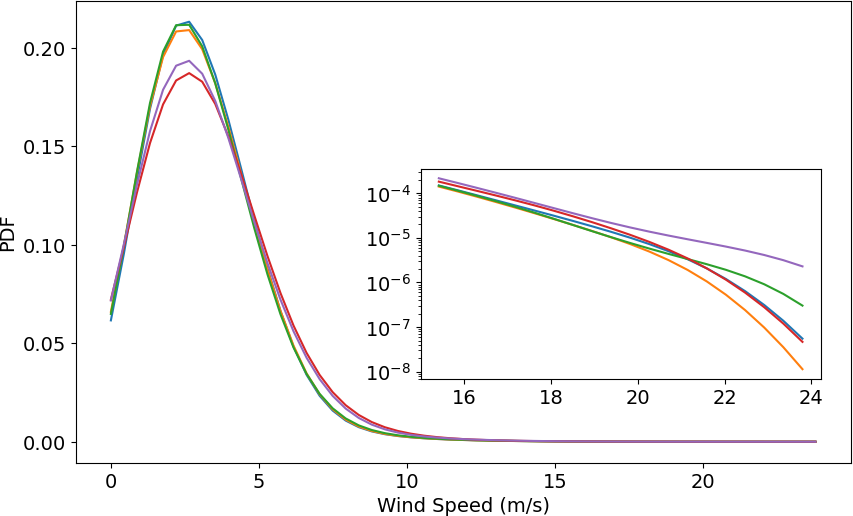}\label{fig:result_pdf_I_d4}}
     
     \caption{MAE, average PSD and average PDF on the test set for domain \#4 of configuration A in Eastern Québec and New Brunswick (2899 hourly samples from January 2024 to April 2024), for each downscaling model (bi-linear or nearest neighbour interpolation strategy, no-interpolation strategy), and baseline. The ground truth in the PSD and PDF graphs refers to the HR $UV$, i.e., the predictand.}\label{fig:results_domain_I_d4}
\end{figure}

\begin{figure}[ht]
    \centering
     \hspace{2mm}
     \subfloat[RMSE.]{\includegraphics[height=4.6cm,valign=c]{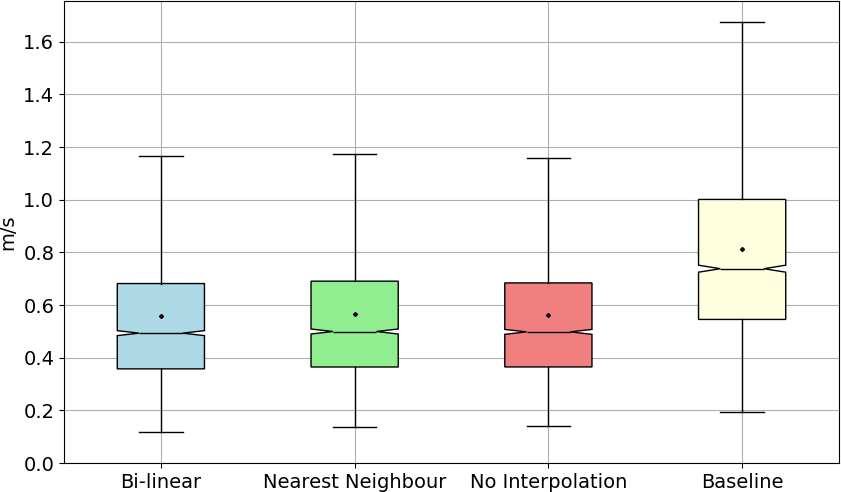}\label{fig:bp_rmse_d5}}
     \hspace{2mm}
     \subfloat[MAE.]{\includegraphics[height=4.6cm,valign=c]{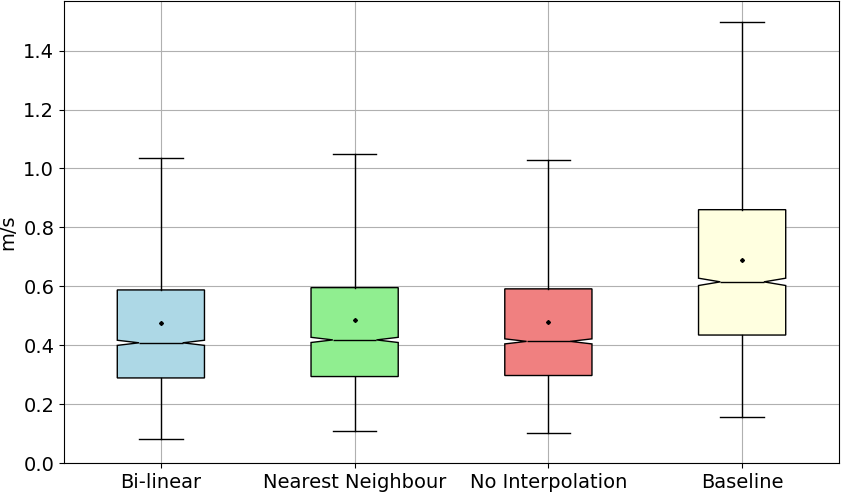}\label{fig:bp_mae_d5}}
     \hspace{2mm}
     \subfloat[SSIM.]{\includegraphics[height=4.6cm,valign=c]{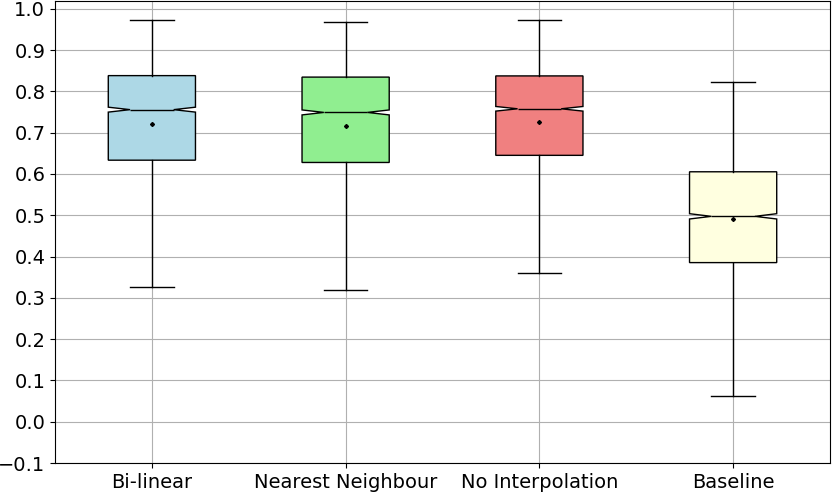}\label{fig:bp_ssim_d5}}
     
     \caption{RMSE, MAE and SSIM boxplots for each downscaling model (bi-linear or nearest neighbour interpolation strategy, no-interpolation strategy), and baseline, computed on the test set of configuration A for domain \#5 with the five domains in Eastern Québec and New Brunswick (2 899 hourly samples from January 2024 to April 2024). In each boxplot, the black dot and line represent respectively the mean and the median, the upper and lower box limits indicate the first (Q1) and third (Q3) quartiles and the whiskers depict the highest (lowest) value within the $1.5 \times$ (Q3-Q1) above Q3 (below Q1).}\label{fig:results_I_d5}
\end{figure}

\begin{figure}[ht]
    \centering
    \subfloat[Pixel-wise MAE (m/s) for the different strategies and the baseline. The colorscale is capped at the max MAE of the three strategies, which makes the baseline's values higher than the maximum of the scale to saturate in some locations. The value at the top of each image is the average over the domain.]{\includegraphics[width=0.99\textwidth,valign=c]{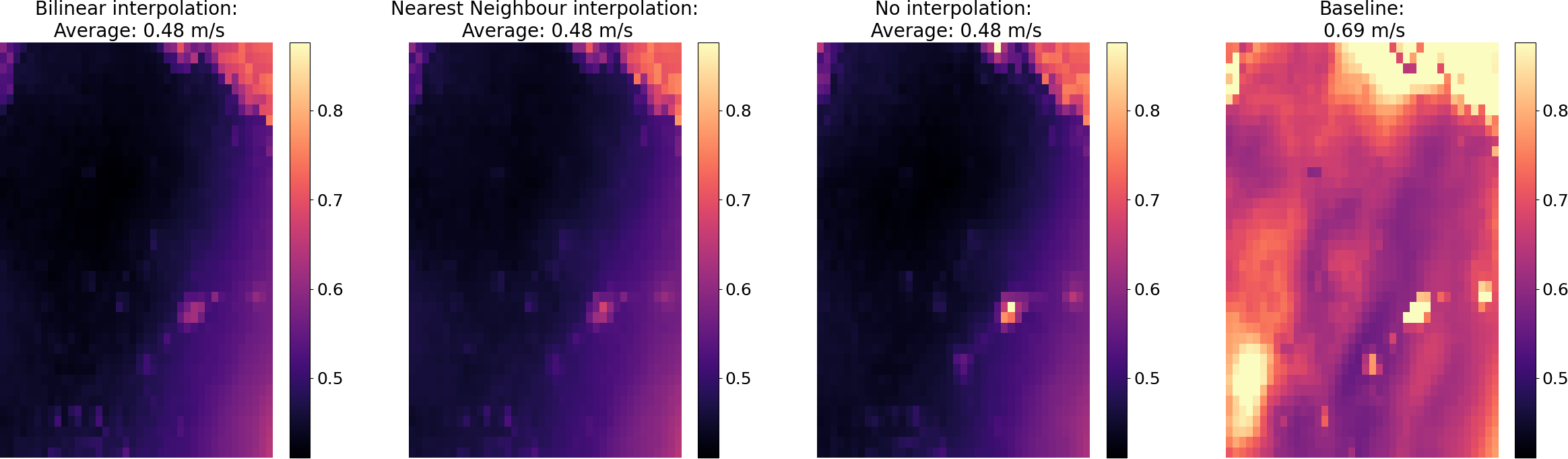}
     \label{fig:error_section_I_d5}} \\
     %\hspace{2mm}
     \subfloat[Average PSD.]{\includegraphics[width = 0.48\textwidth,valign=c]{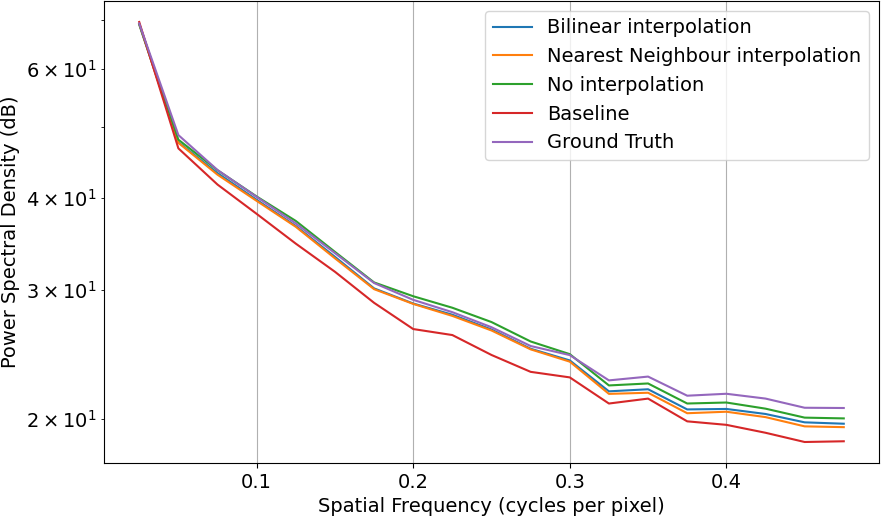}\label{fig:results_psd_I_d5}}
     \hspace{2mm}
     \subfloat[Average PDF. Note that the legend is the same as Figure \ref{fig:results_psd_I_d5}.]{\includegraphics[width = 0.48\textwidth,valign=c]{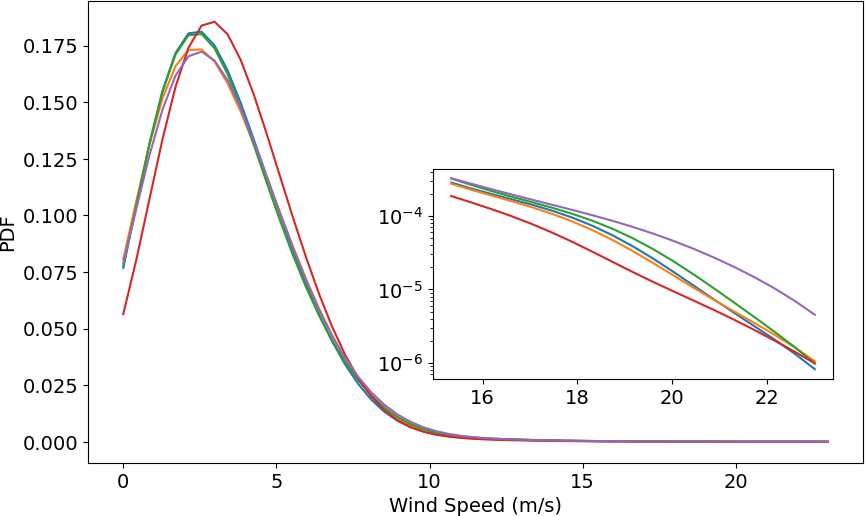}\label{fig:result_pdf_I_d5}}
     
     \caption{MAE, average PSD and average PDF on the test set for domain \#5 of configuration A in Eastern Québec and New Brunswick  (2899 hourly samples from January 2024 to April 2024), for each downscaling model (bi-linear or nearest neighbour interpolation strategy, no-interpolation strategy), and baseline. The ground truth in the PSD and PDF graphs refers to the HR $UV$, i.e., the predictand.}\label{fig:results_domain_I_d5}
\end{figure}

% Domain configuration experiment results
\clearpage
\subsection{Domain configuration experiment}
Here are all the results for domains \#1 to \#13. Configuration A includes domains \#1 to \#5, configuration B includes domains \#1, \#6, \#7, \#11, and \#13 and configuration C includes all configuration B domains in addition to domains \#3, \#8, \#9, \#10, and \#12.

\begin{figure}[ht]
    \centering
     \hspace{2mm}
     \subfloat[RMSE.]{\includegraphics[height=4.6cm,valign=c]{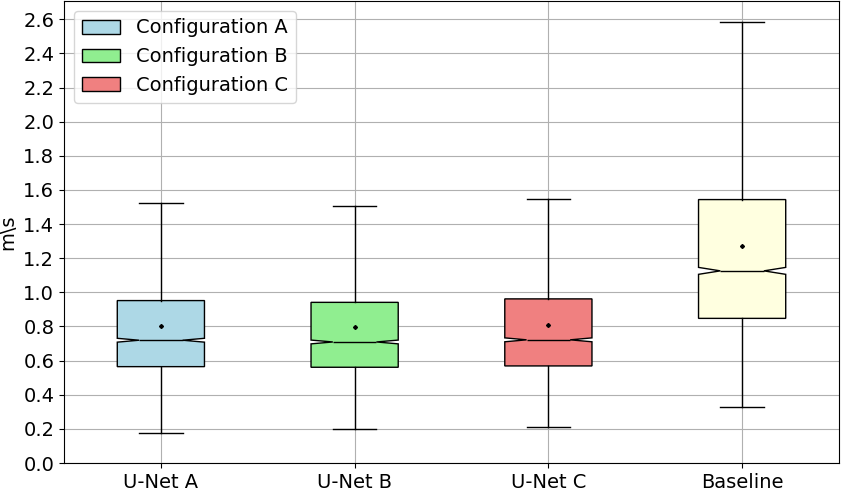}\label{fig:bp_II_rmse_d1}}
     \hspace{2mm}
     \subfloat[MAE.]{\includegraphics[height=4.6cm,valign=c]{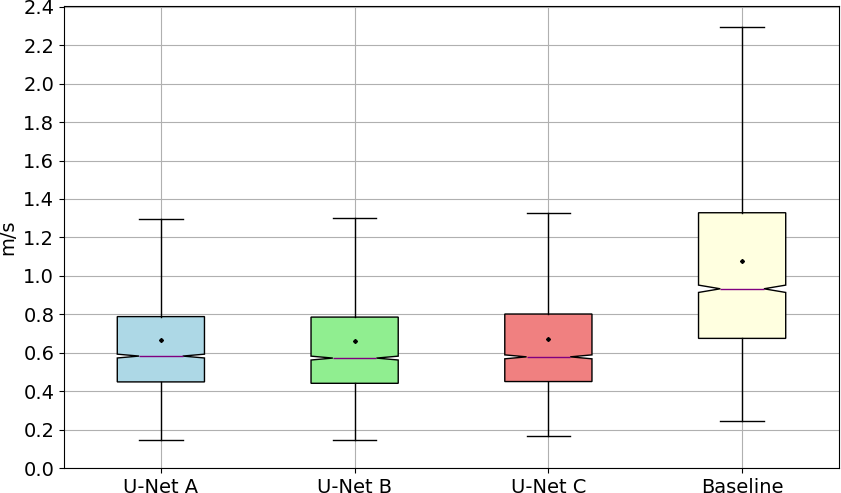}\label{fig:bp_II_mae_d1}}
     \hspace{2mm}
     \subfloat[SSIM.]{\includegraphics[height=4.6cm,valign=c]{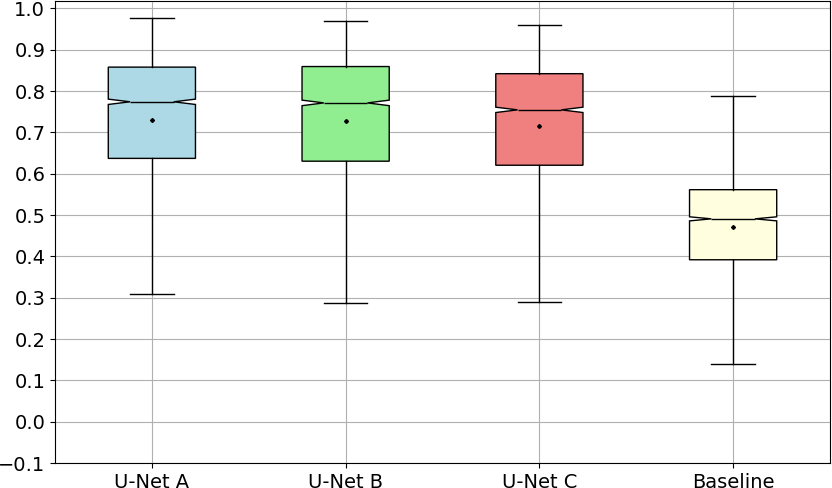}\label{fig:bp__II_ssim_d1}}
     
     \caption{RMSE, MAE and SSIM boxplots for the downscaling model (U-Net) predictions on the test set, which consists of hourly samples from January 2024 to April 2024, i.e. respectively 14 495 for configuration A for domain \#1 in Eastern Québec and New Brunswick, and 14 495 and 28 990 for configurations B and C across Canada. RMSE and SSIM boxplots for the baseline model (bi-linear interpolation of the low resolution $UV$ onto the predictand grid) are presented alongside. In each boxplot, the black dot and line represent respectively the mean and the median, the upper and lower box limits indicate the first (Q1) and third (Q3) quartiles and the whiskers depict the highest (lowest) value within the $1.5 \times$ (Q3-Q1) above Q3 (below Q1).}\label{fig:results_II_d1}
\end{figure}

\begin{figure}[ht]
    \centering
    \subfloat[Pixel-wise MAE (m/s) on domain \#1 for the downscaling model (U-Net) trained with each configuration along with the baseline's. The colorscale is capped at the maximum MAE of the downscaling model for each configuration if the U-Net is trained on that specific domain, else it is ignored from the maximum calculation. The baseline's MAE reaches higher values than the maximum of the colorscale in some locations. The value at the top of each image is the average over the domain.]{\includegraphics[width=0.99\textwidth,valign=c]{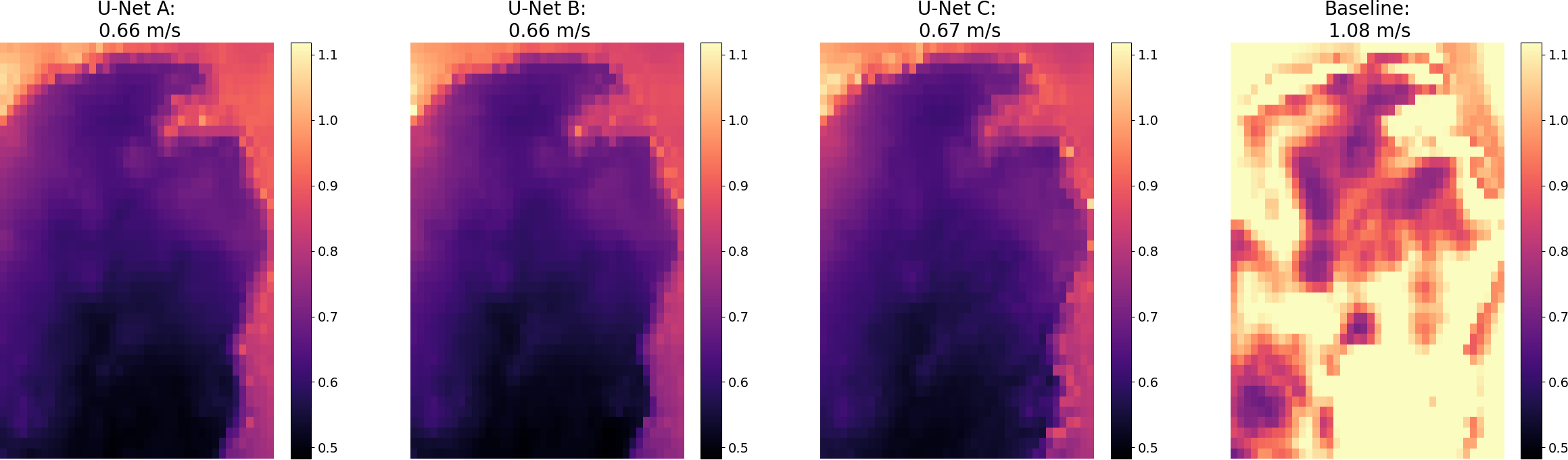}
     \label{fig:section_II_average_error_d1}}
     \\
     \subfloat[Average PSD.]{\includegraphics[width=0.48\textwidth,valign=c]{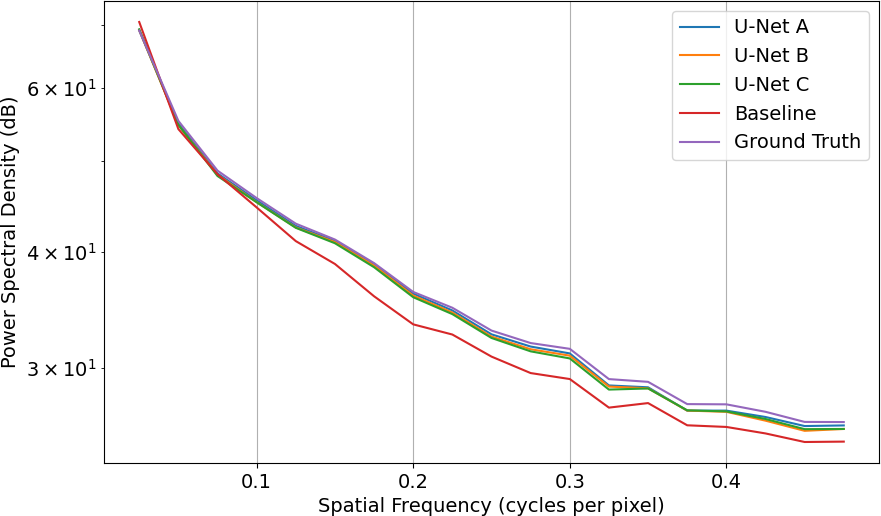}\label{fig:section_II_psd_d1}}
     \hspace{2mm}
     \subfloat[Average PDF. Note that the legend is the same as Figure \ref{fig:section_II_psd_d1}.]{\includegraphics[width=0.48\textwidth,valign=c]{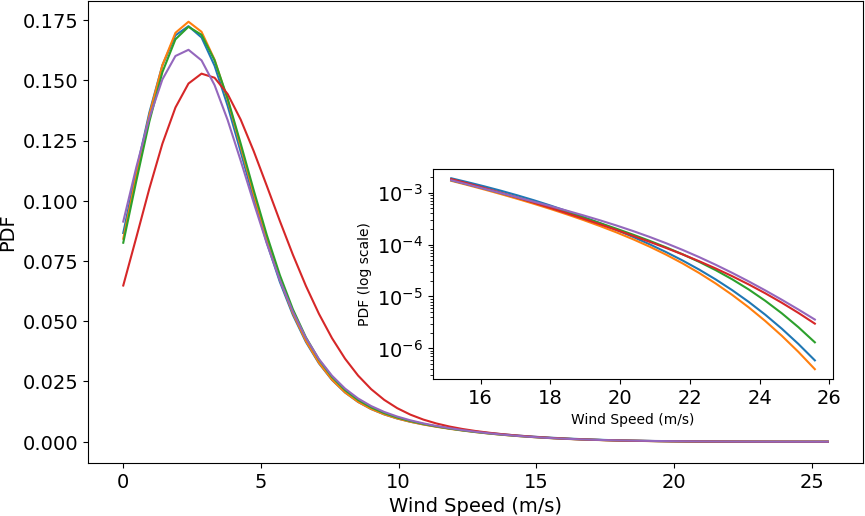}\label{fig:section_II_pdf_d1}}
     \caption{MAE, average PSD and average PDF on the test set over domain \#1 (2 899 hourly samples from January 2024 to April 2024), for the downscaling model (U-Net) trained with configuration A (five domains in Eastern Québec and New Brunswick), configuration B (five domains across Canada) or configuration C (10 domains across Canada), along with the baseline (bi-linear interpolation of the low-resolution $UV$ onto the predictand grid).}\label{fig:results_domain_II_d1}
\end{figure}

\begin{figure}[ht]
    \centering
     \hspace{2mm}
     \subfloat[RMSE.]{\includegraphics[height=4.6cm,valign=c]{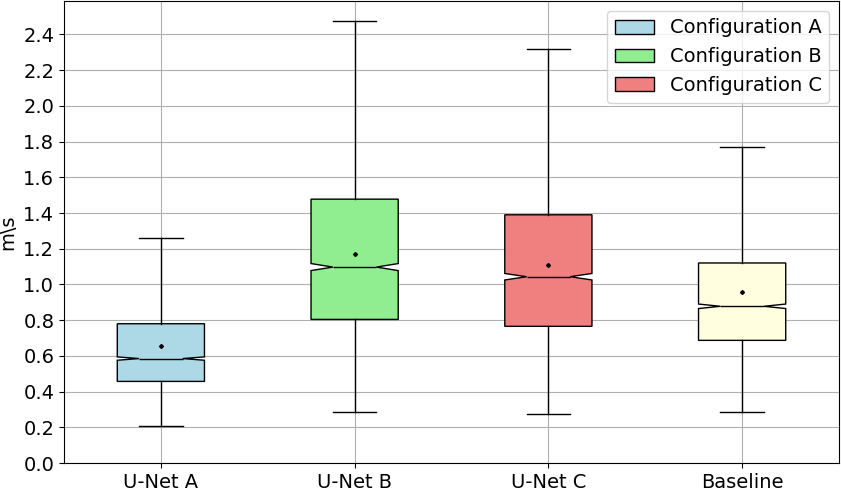}\label{fig:bp_II_rmse_d2}}
     \hspace{2mm}
     \subfloat[MAE.]{\includegraphics[height=4.6cm,valign=c]{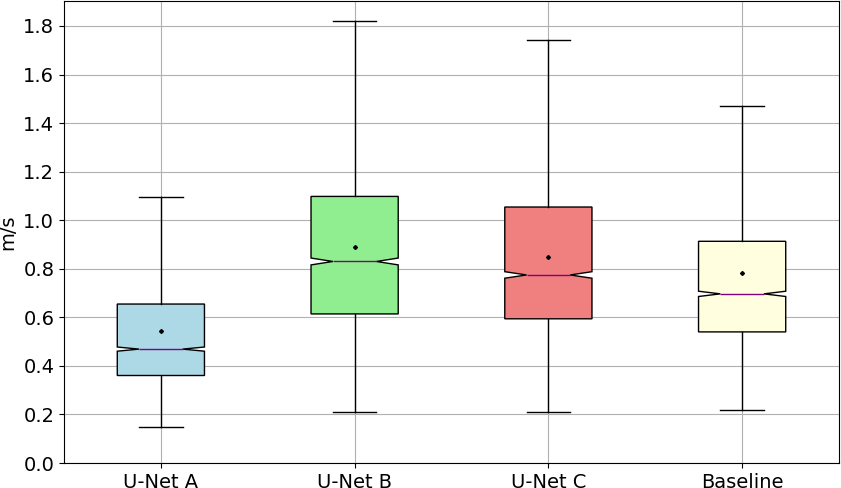}\label{fig:bp_II_mae_d2}}
     \hspace{2mm}
     \subfloat[SSIM.]{\includegraphics[height=4.6cm,valign=c]{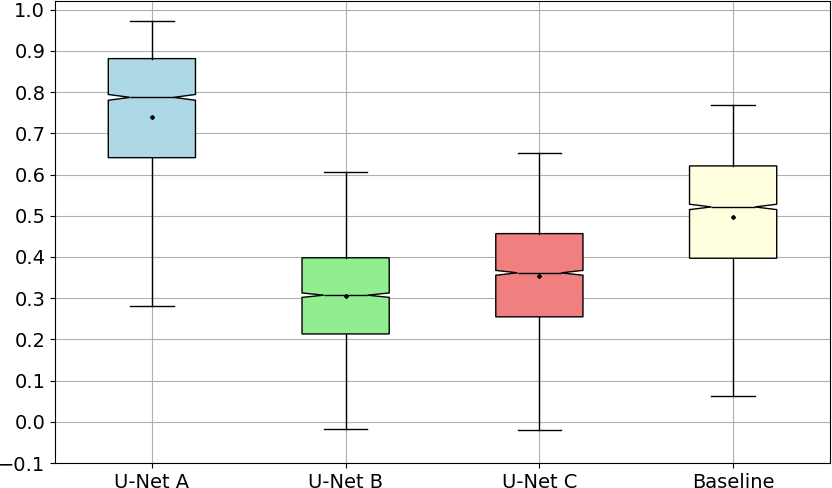}\label{fig:bp__II_ssim_d2}}
     
     \caption{RMSE, MAE and SSIM boxplots for the downscaling model (U-Net) predictions on the test set, which consists of hourly samples from January 2024 to April 2024, i.e. respectively 14 495 for configuration A for domain \#2 in Eastern Québec and New Brunswick, and 14 495 and 28 990 for configurations B and C across Canada. RMSE and SSIM boxplots for the baseline model (bi-linear interpolation of the low resolution $UV$ onto the predictand grid) are presented alongside. In each boxplot, the black dot and line represent respectively the mean and the median, the upper and lower box limits indicate the first (Q1) and third (Q3) quartiles and the whiskers depict the highest (lowest) value within the $1.5 \times$ (Q3-Q1) above Q3  (below Q1).}\label{fig:results_II_d2}
\end{figure}

\begin{figure}[ht]
    \centering
    \subfloat[Pixel-wise MAE (m/s) on domain \#2 for the downscaling model (U-Net) trained with each configuration along with the baseline's. The colorscale is capped at the maximum MAE of the downscaling model for each configuration if the U-Net is trained on that specific domain, else it is ignored from the maximum calculation. The baseline's MAE reaches higher values than the maximum of the colorscale in some locations. The value at the top of each image is the average over the domain.]{\includegraphics[width=0.99\textwidth,valign=c]{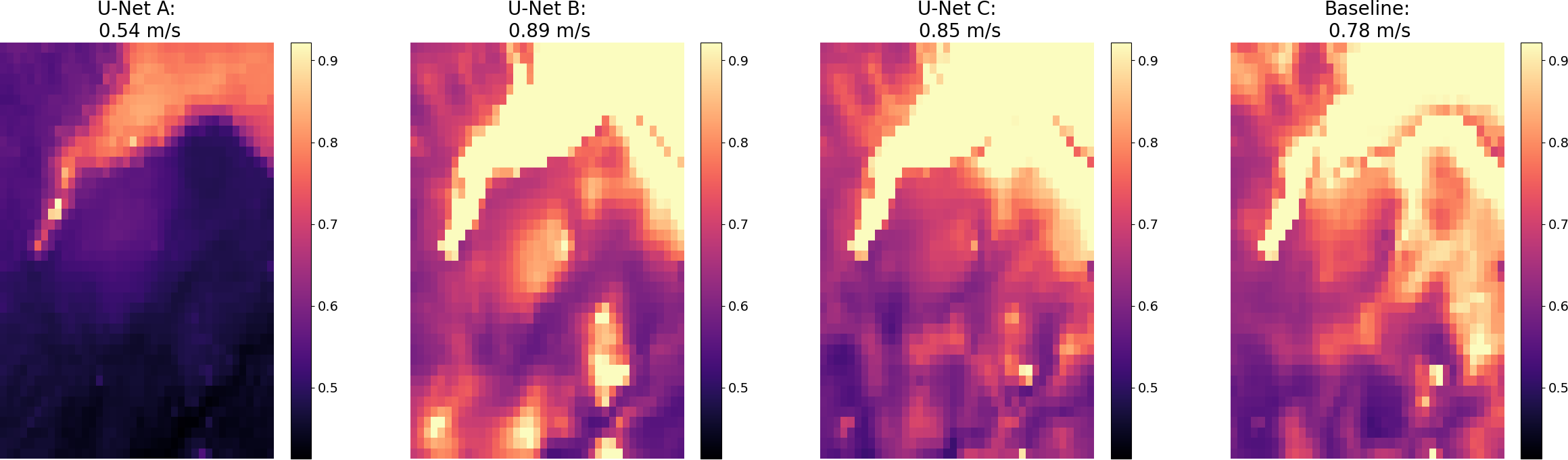}
     \label{fig:section_II_average_error_d2}}
     \\
     \subfloat[Average PSD.]{\includegraphics[width=0.48\textwidth,valign=c]{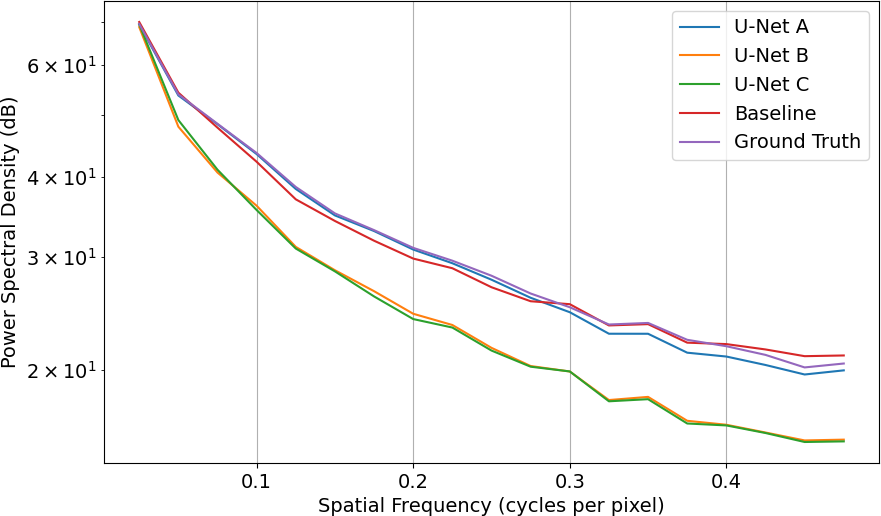}\label{fig:section_II_psd_d2}}
     \hspace{2mm}
     \subfloat[Average PDF. Note that the legend is the same as Figure \ref{fig:section_II_psd_d2}.]{\includegraphics[width=0.48\textwidth,valign=c]{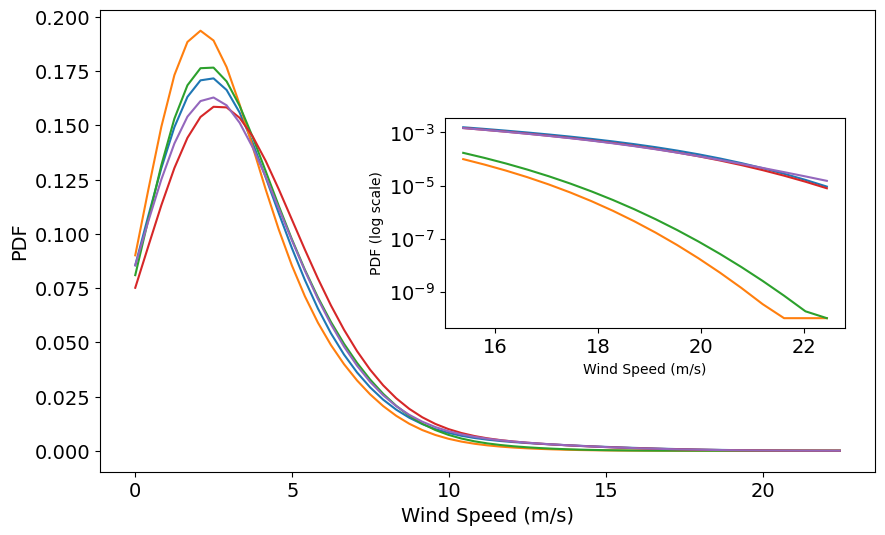}\label{fig:section_II_pdf_d2}}
     \caption{MAE, average PSD and average PDF on the test set over domain \#2 (2 899 hourly samples from January 2024 to April 2024), for the downscaling model (U-Net) trained with configuration A (five domains in Eastern Québec and New Brunswick), configuration B (five domains across Canada) or configuration C (10 domains across Canada), along with the baseline (bi-linear interpolation of the low-resolution $UV$ onto the predictand grid).}\label{fig:results_domain_II_d2}
\end{figure}

\begin{figure}[ht]
    \centering
     \hspace{2mm}
     \subfloat[RMSE.]{\includegraphics[height=4.6cm,valign=c]{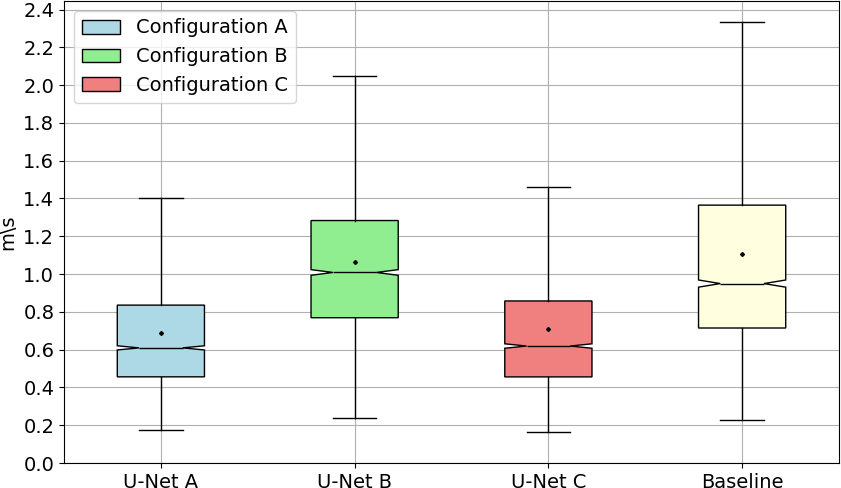}\label{fig:bp_II_rmse_d3}}
     \hspace{2mm}
     \subfloat[MAE.]{\includegraphics[height=4.6cm,valign=c]{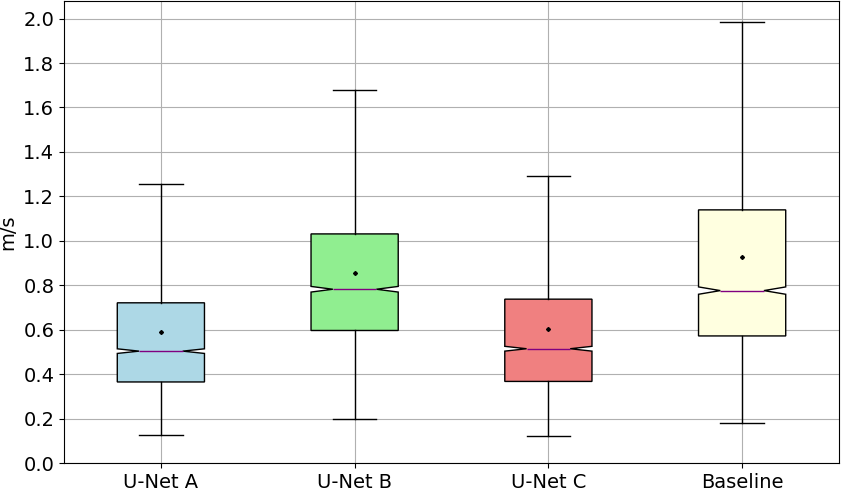}\label{fig:bp_II_mae_d3}}
     \hspace{2mm}
     \subfloat[SSIM.]{\includegraphics[height=4.6cm,valign=c]{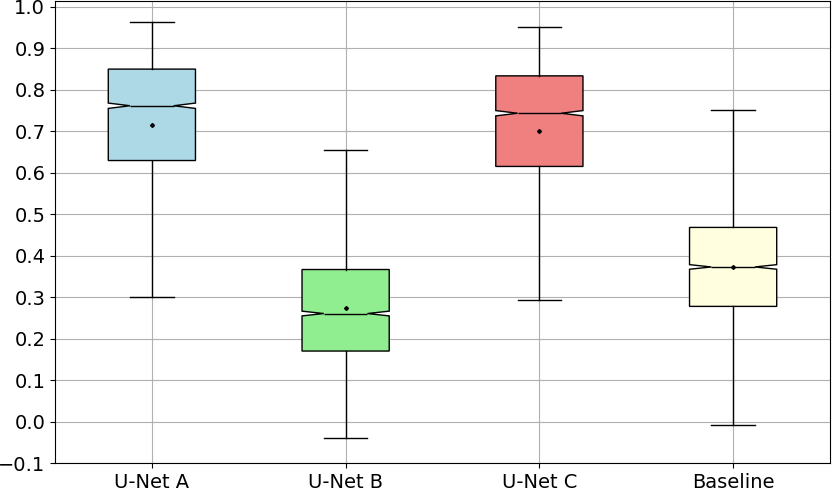}\label{fig:bp__II_ssim_d3}}
     
     \caption{RMSE, MAE and SSIM boxplots for the downscaling model (U-Net) predictions on the test set, which consists of hourly samples from January 2024 to April 2024, i.e. respectively 14 495 for configuration A for domain \#3 in Eastern Québec and New Brunswick, and 14 495 and 28 990 for configurations B and C across Canada. RMSE and SSIM boxplots for the baseline model (bi-linear interpolation of the low resolution $UV$ onto the predictand grid) are presented alongside. In each boxplot, the black dot and line represent respectively the mean and the median, the upper and lower box limits indicate the first (Q1) and third (Q3) quartiles and the whiskers depict the highest (lowest) value within the $1.5 \times$ (Q3-Q1) above Q3  (below Q1).}\label{fig:results_II_d3}
\end{figure}

\begin{figure}[ht]
    \centering
    \subfloat[Pixel-wise MAE (m/s) on domain \#3 for the downscaling model (U-Net) trained with each configuration along with the baseline's. The colorscale is capped at the maximum MAE of the downscaling model for each configuration if the U-Net is trained on that specific domain, else it is ignored from the maximum calculation. The baseline's MAE reaches higher values than the maximum of the colorscale in some locations. The value at the top of each image is the average over the domain.]{\includegraphics[width=0.99\textwidth,valign=c]{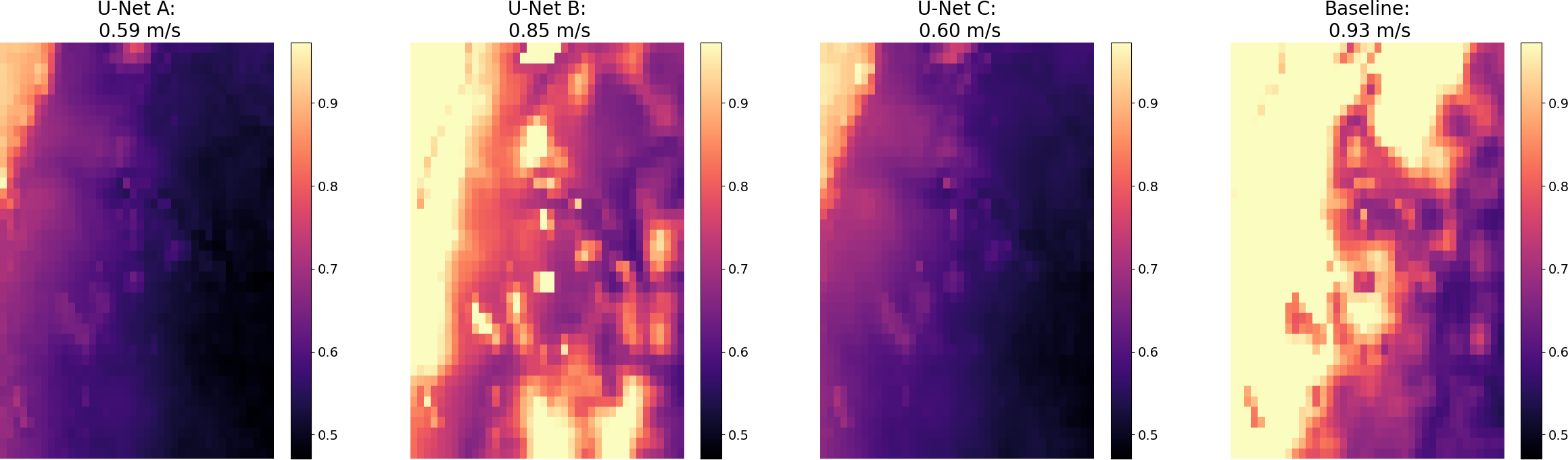}
     \label{fig:section_II_average_error_d3}}
     \\
     \subfloat[Average PSD.]{\includegraphics[width=0.48\textwidth,valign=c]{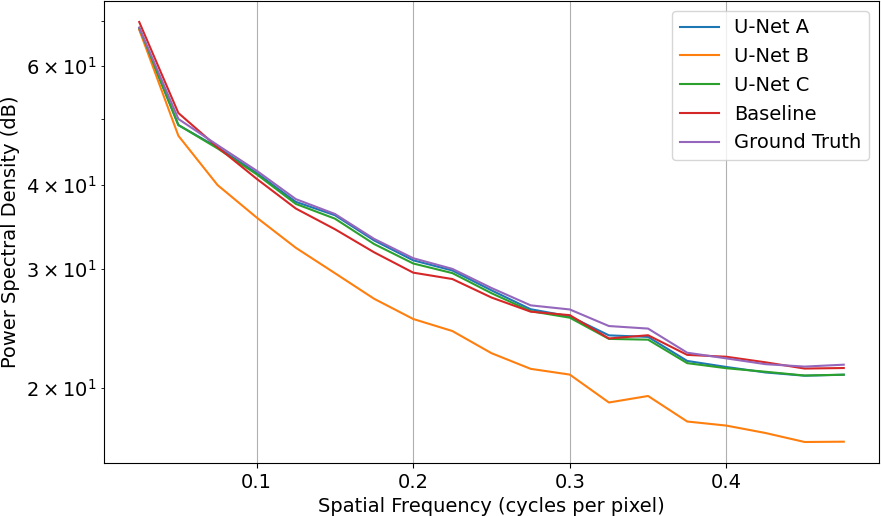}\label{fig:section_II_psd_d3}}
     \hspace{2mm}
     \subfloat[Average PDF. Note that the legend is the same as Figure \ref{fig:section_II_psd_d3}.]{\includegraphics[width=0.48\textwidth,valign=c]{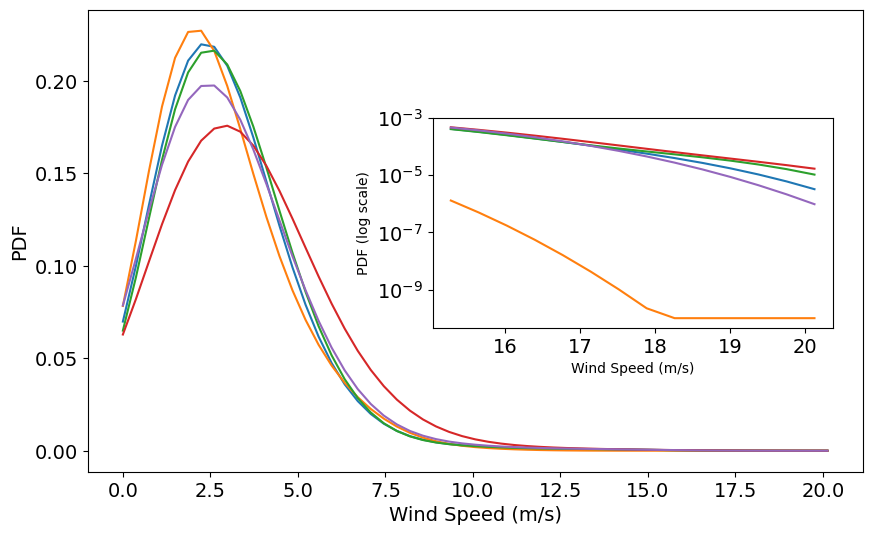}\label{fig:section_II_pdf_d3}}
     \caption{MAE, average PSD and average PDF on the test set over domain \#3 (2 899 hourly samples from January 2024 to April 2024), for the downscaling model (U-Net) trained with configuration A (five domains in Eastern Québec and New Brunswick), configuration B (five domains across Canada) or configuration C (10 domains across Canada), along with the baseline (bi-linear interpolation of the low-resolution $UV$ onto the predictand grid).}\label{fig:results_domain_II_d3}
\end{figure}

\begin{figure}[ht]
    \centering
     \hspace{2mm}
     \subfloat[RMSE.]{\includegraphics[height=4.6cm,valign=c]{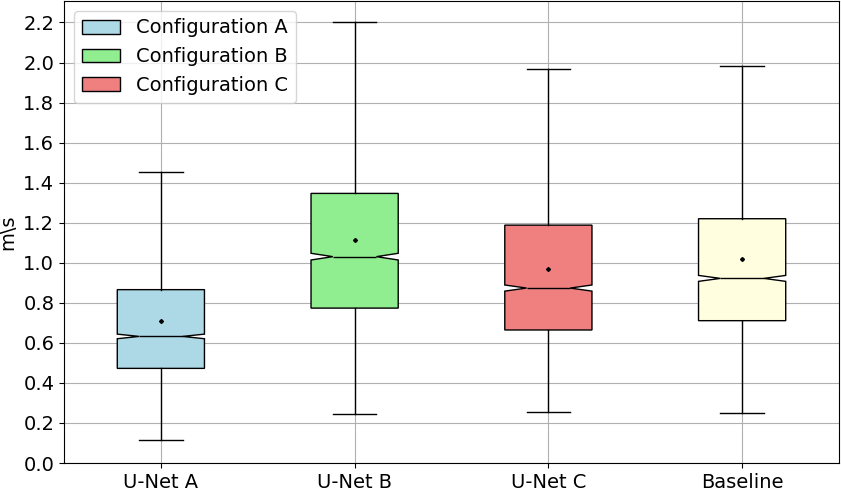}\label{fig:bp_II_rmse_d4}}
     \hspace{2mm}
     \subfloat[MAE.]{\includegraphics[height=4.6cm,valign=c]{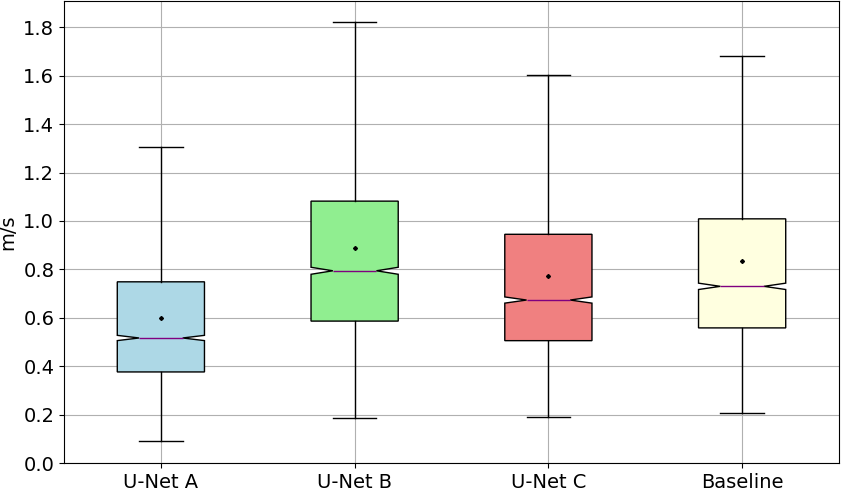}\label{fig:bp_II_mae_d4}}
     \hspace{2mm}
     \subfloat[SSIM.]{\includegraphics[height=4.6cm,valign=c]{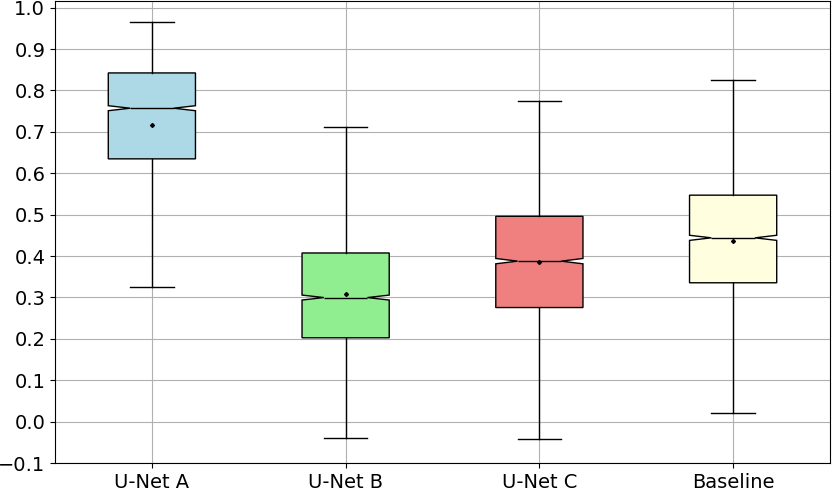}\label{fig:bp__II_ssim_d4}}
     
     \caption{RMSE, MAE and SSIM boxplots for the downscaling model (U-Net) predictions on the test set, which consists of hourly samples from January 2024 to April 2024, i.e. respectively 14 495 for configuration A for domain \#4 in Eastern Québec and New Brunswick, and 14 495 and 28 990 for configurations B and C across Canada. RMSE and SSIM boxplots for the baseline model (bi-linear interpolation of the low resolution $UV$ onto the predictand grid) are presented alongside. In each boxplot, the black dot and line represent respectively the mean and the median, the upper and lower box limits indicate the first (Q1) and third (Q3) quartiles and the whiskers depict the highest (lowest) value within the $1.5 \times$ (Q3-Q1) above Q3  (below Q1).}\label{fig:results_II_d4}
\end{figure}

\begin{figure}[ht]
    \centering
    \subfloat[Pixel-wise MAE (m/s) on domain \#4 for the downscaling model (U-Net) trained with each configuration along with the baseline's. The colorscale is capped at the maximum MAE of the downscaling model for each configuration if the U-Net is trained on that specific domain, else it is ignored from the maximum calculation. The baseline's MAE reaches higher values than the maximum of the colorscale in some locations. The value at the top of each image is the average over the domain.]{\includegraphics[width=0.99\textwidth,valign=c]{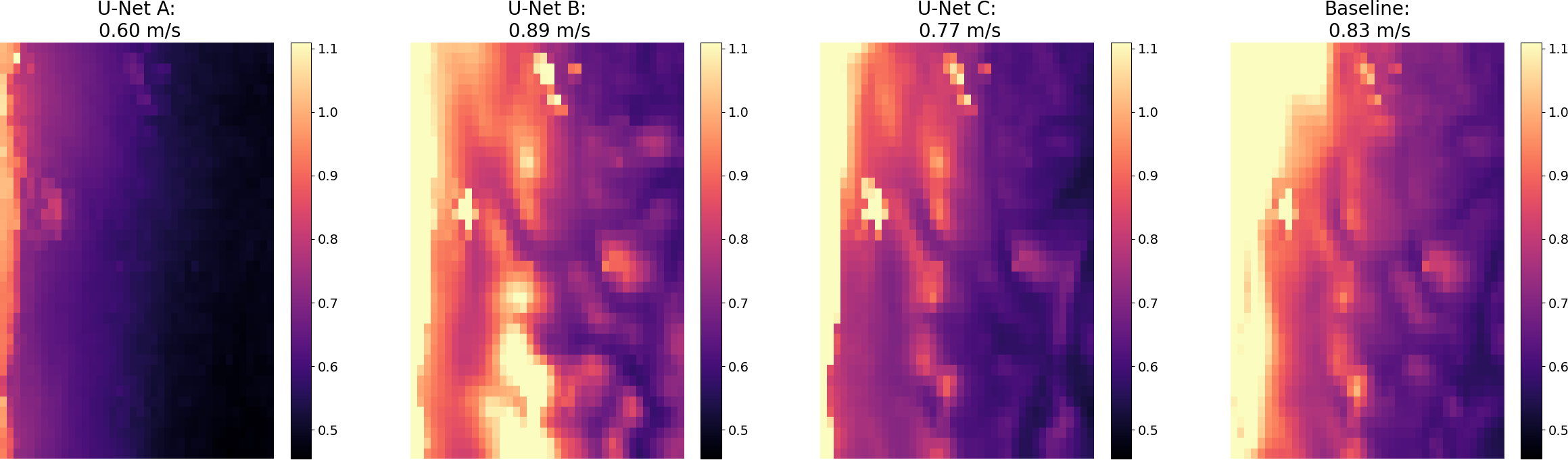}
     \label{fig:section_II_average_error_d4}}
     \\
     \subfloat[Average PSD.]{\includegraphics[width=0.48\textwidth,valign=c]{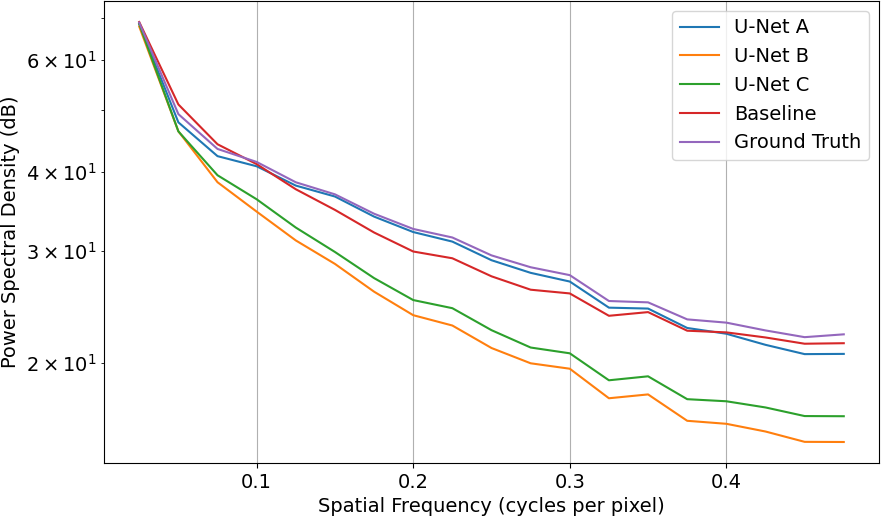}\label{fig:section_II_psd_d4}}
     \hspace{2mm}
     \subfloat[Average PDF. Note that the legend is the same as Figure \ref{fig:section_II_psd_d4}.]{\includegraphics[width=0.48\textwidth,valign=c]{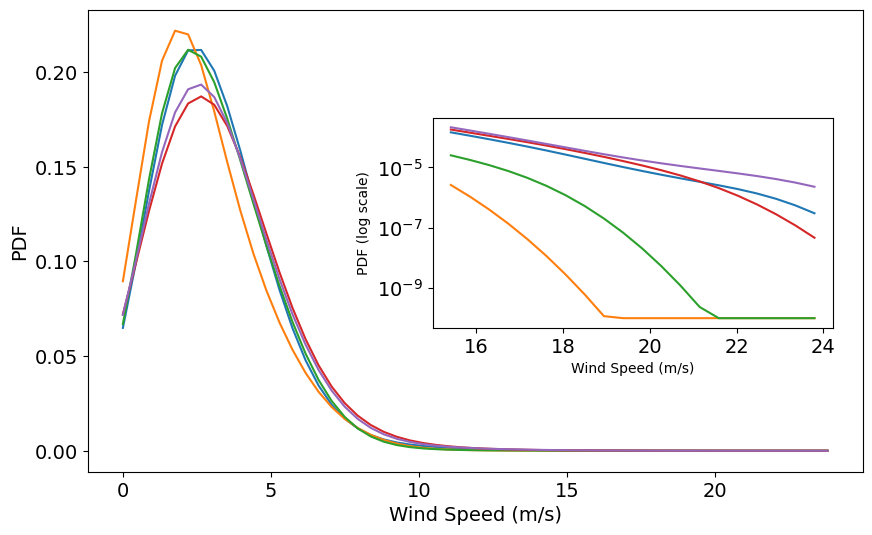}\label{fig:section_II_pdf_d4}}
     \caption{MAE, average PSD and average PDF on the test set over domain \#4 (2 899 hourly samples from January 2024 to April 2024), for the downscaling model (U-Net) trained with configuration A (five domains in Eastern Québec and New Brunswick), configuration B (five domains across Canada) or configuration C (10 domains across Canada), along with the baseline (bi-linear interpolation of the low-resolution $UV$ onto the predictand grid).}\label{fig:results_domain_II_d4}
\end{figure}

\begin{figure}[ht]
    \centering
     \hspace{2mm}
     \subfloat[RMSE.]{\includegraphics[height=4.6cm,valign=c]{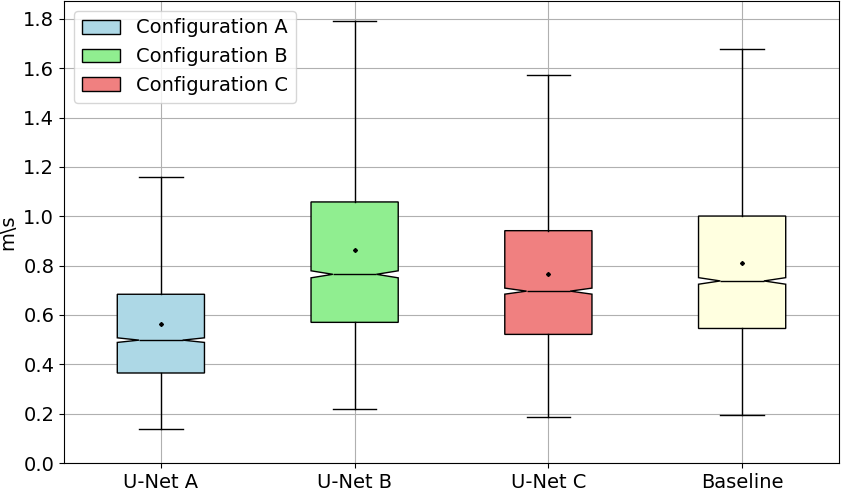}\label{fig:bp_II_rmse_d5}}
     \hspace{2mm}
     \subfloat[MAE.]{\includegraphics[height=4.6cm,valign=c]{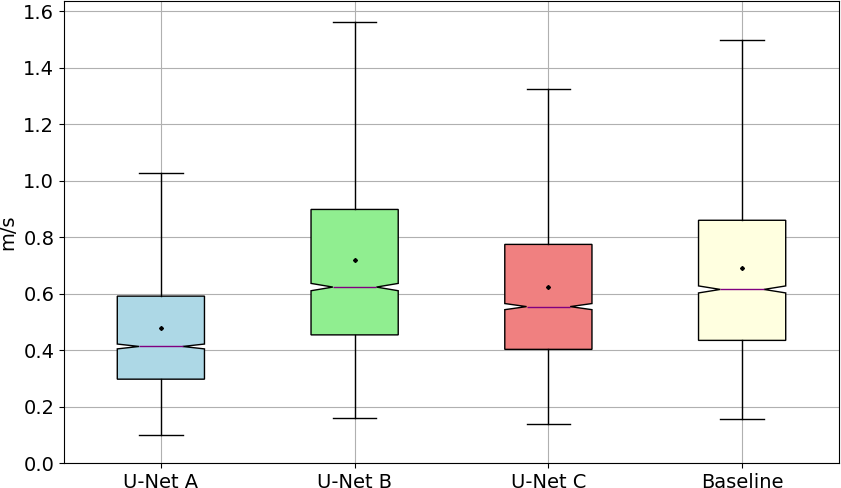}\label{fig:bp_II_mae_d5}}
     \hspace{2mm}
     \subfloat[SSIM.]{\includegraphics[height=4.6cm,valign=c]{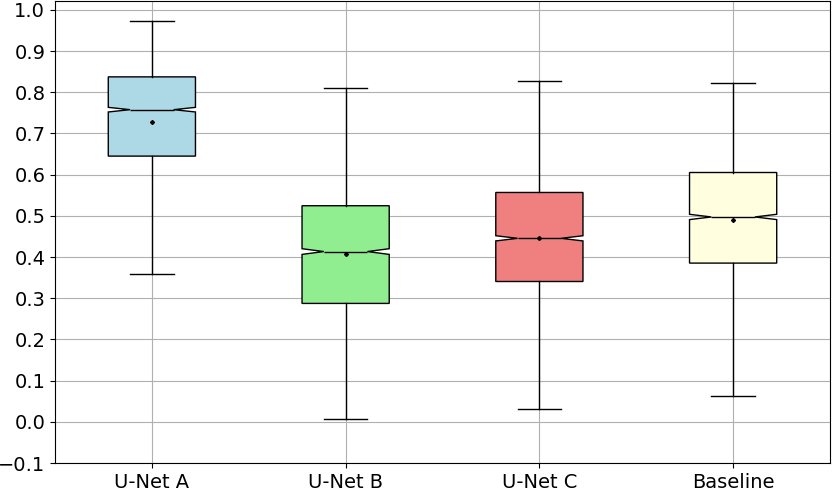}\label{fig:bp__II_ssim_d5}}
     
     \caption{RMSE, MAE and SSIM boxplots for the downscaling model (U-Net) predictions on the test set, which consists of hourly samples from January 2024 to April 2024, i.e. respectively 14 495 for configuration A for domain \#5 in Eastern Québec and New Brunswick, and 14 495 and 28 990 for configurations B and C across Canada. RMSE and SSIM boxplots for the baseline model (bi-linear interpolation of the low resolution $UV$ onto the predictand grid) are presented alongside. In each boxplot, the black dot and line represent respectively the mean and the median, the upper and lower box limits indicate the first (Q1) and third (Q3) quartiles and the whiskers depict the highest (lowest) value within the $1.5 \times$ (Q3-Q1) above Q3  (below Q1).}\label{fig:results_II_d5}
\end{figure}

\begin{figure}[ht]
    \centering
    \subfloat[Pixel-wise MAE (m/s) on domain \#5 for the downscaling model (U-Net) trained with each configuration along with the baseline's. The colorscale is capped at the maximum MAE of the downscaling model for each configuration if the U-Net is trained on that specific domain, else it is ignored from the maximum calculation. The baseline's MAE reaches higher values than the maximum of the colorscale in some locations. The value at the top of each image is the average over the domain.]{\includegraphics[width=0.99\textwidth,valign=c]{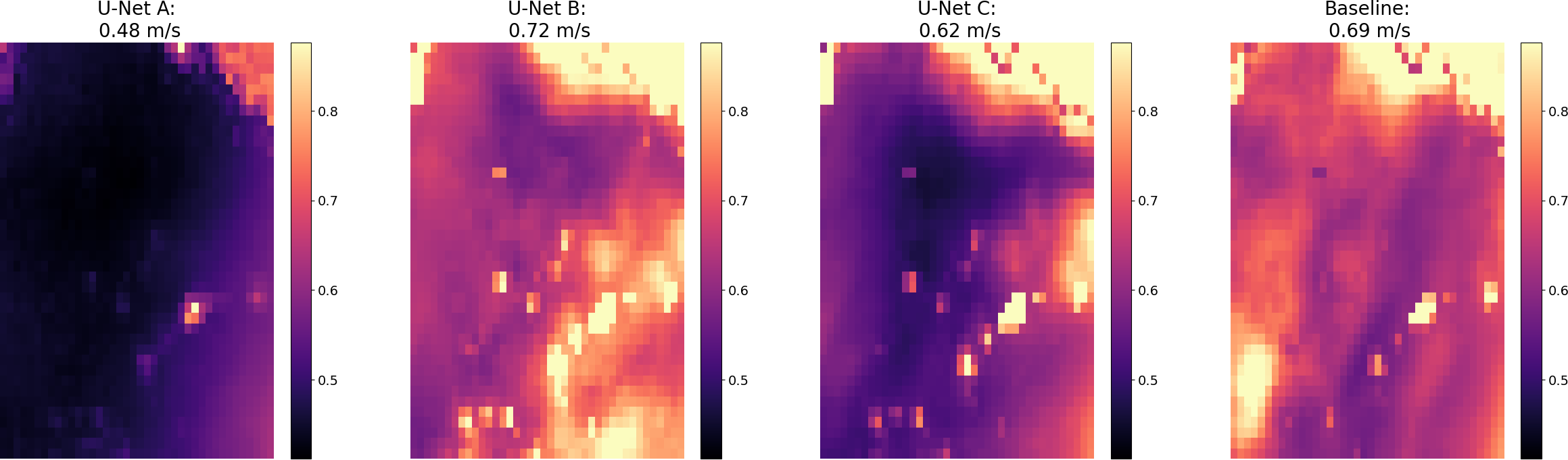}
     \label{fig:section_II_average_error_d5}}
     \\
     \subfloat[Average PSD.]{\includegraphics[width=0.48\textwidth,valign=c]{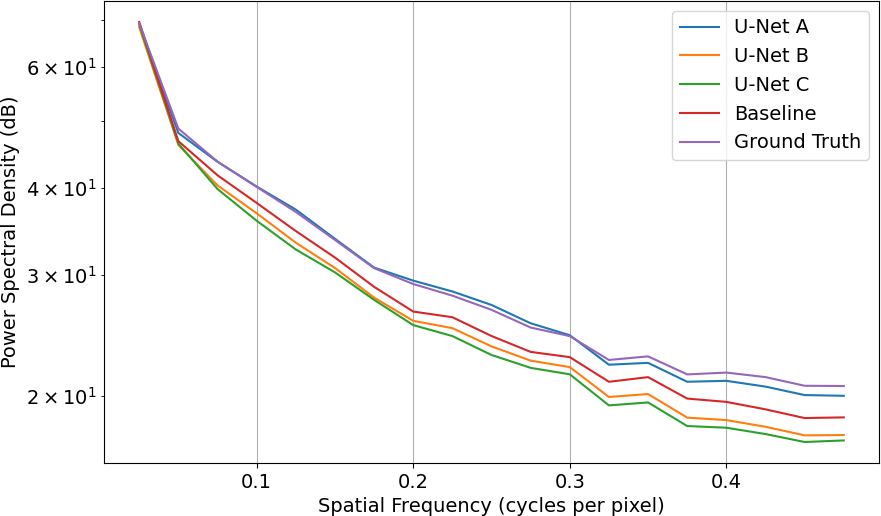}\label{fig:section_II_psd_d5}}
     \hspace{2mm}
     \subfloat[Average PDF. Note that the legend is the same as Figure \ref{fig:section_II_psd_d5}.]{\includegraphics[width=0.48\textwidth,valign=c]{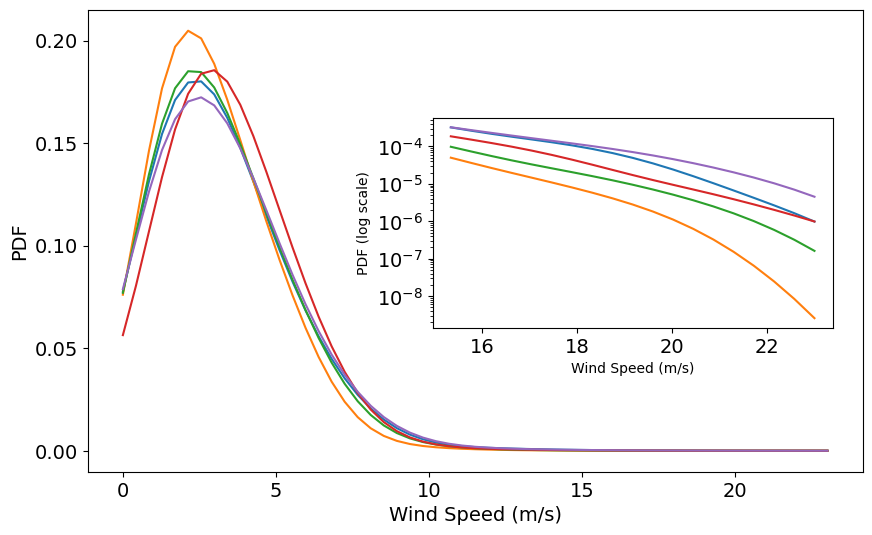}\label{fig:section_II_pdf_d5}}
     \caption{MAE, average PSD and average PDF on the test set over domain \#5 (2 899 hourly samples from January 2024 to April 2024), for the downscaling model (U-Net) trained with configuration A (five domains in Eastern Québec and New Brunswick), configuration B (five domains across Canada) or configuration C (10 domains across Canada), along with the baseline (bi-linear interpolation of the low-resolution $UV$ onto the predictand grid).}\label{fig:results_domain_II_d5}
\end{figure}

\begin{figure}[ht]
    \centering
     \hspace{2mm}
     \subfloat[RMSE.]{\includegraphics[height=4.6cm,valign=c]{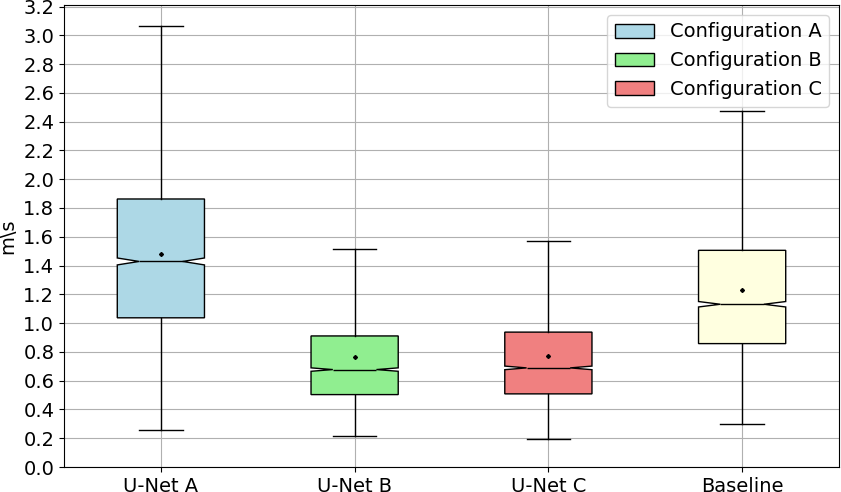}\label{fig:bp_II_rmse_d6}}
     \hspace{2mm}
     \subfloat[MAE.]{\includegraphics[height=4.6cm,valign=c]{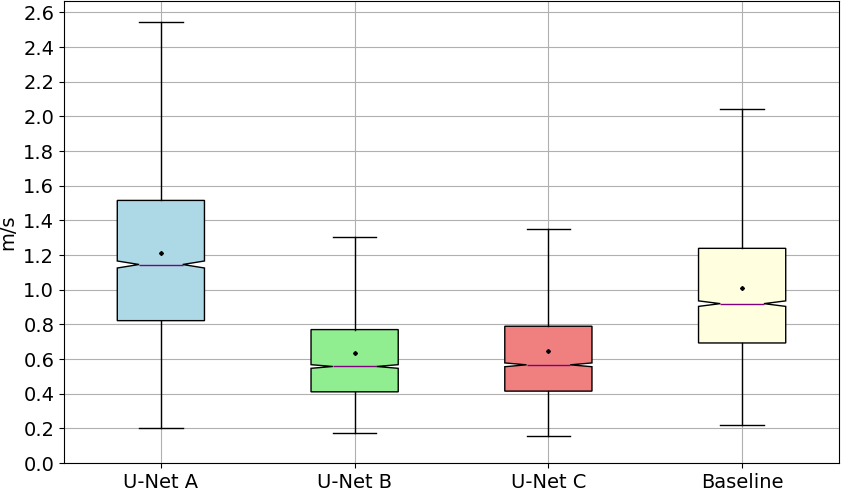}\label{fig:bp_II_mae_d6}}
     \hspace{2mm}
     \subfloat[SSIM.]{\includegraphics[height=4.6cm,valign=c]{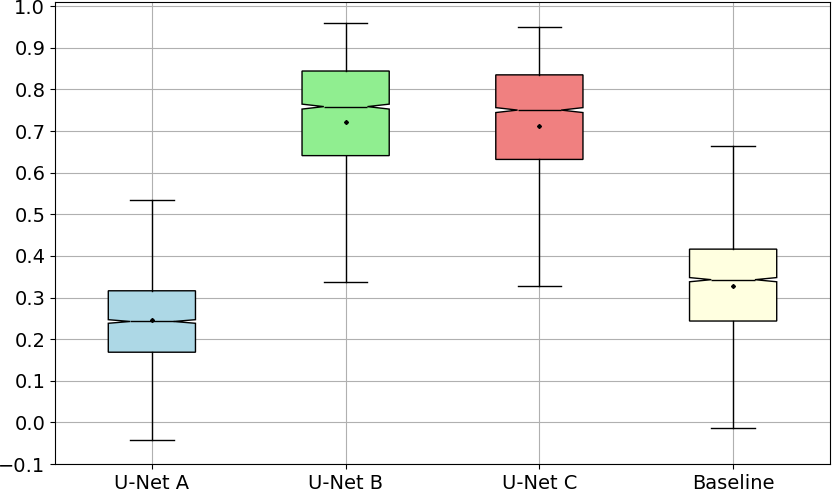}\label{fig:bp__II_ssim_d6}}
     
     \caption{RMSE, MAE and SSIM boxplots for the downscaling model (U-Net) predictions on the test set, which consists of hourly samples from January 2024 to April 2024, i.e. respectively 14 495 for configuration A for domain \#6 in Eastern Québec and New Brunswick, and 14 495 and 28 990 for configurations B and C across Canada. RMSE and SSIM boxplots for the baseline model (bi-linear interpolation of the low resolution $UV$ onto the predictand grid) are presented alongside. In each boxplot, the black dot and line represent respectively the mean and the median, the upper and lower box limits indicate the first (Q1) and third (Q3) quartiles and the whiskers depict the highest (lowest) value within the $1.5 \times$ (Q3-Q1) above Q3  (below Q1).}\label{fig:results_II_d6}
\end{figure}

\begin{figure}[ht]
    \centering
    \subfloat[Pixel-wise MAE (m/s) on domain \#6 for the downscaling model (U-Net) trained with each configuration along with the baseline's. The colorscale is capped at the maximum MAE of the downscaling model for each configuration if the U-Net is trained on that specific domain, else it is ignored from the maximum calculation. The baseline's MAE reaches higher values than the maximum of the colorscale in some locations. The value at the top of each image is the average over the domain.]{\includegraphics[width=0.99\textwidth,valign=c]{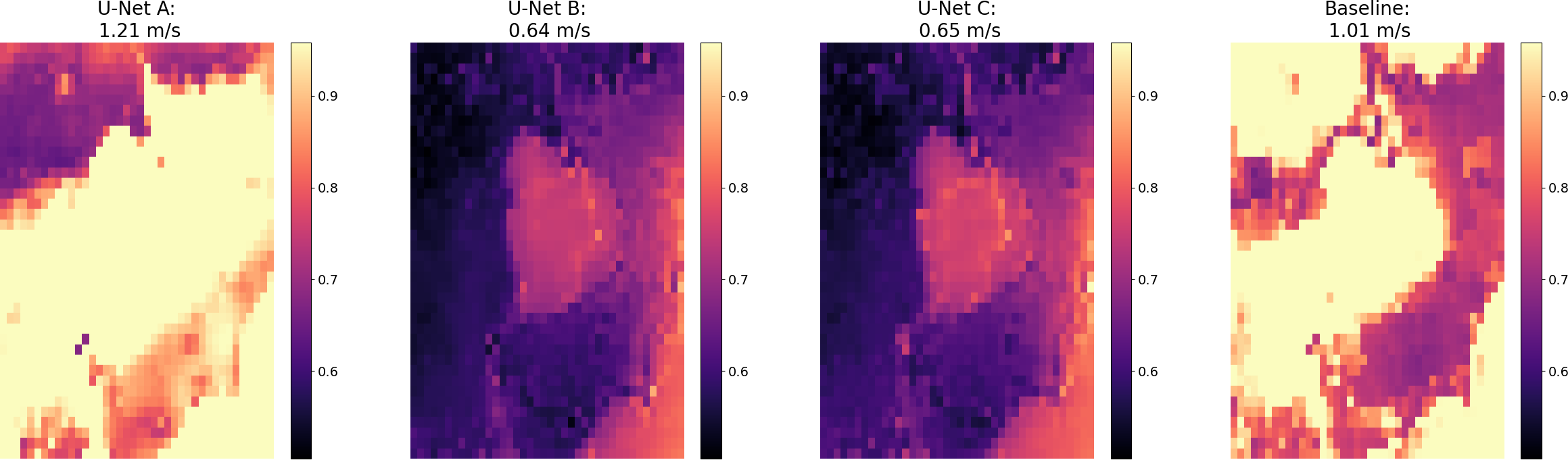}
     \label{fig:section_II_average_error_d6}}
     \\
     \subfloat[Average PSD.]{\includegraphics[width=0.48\textwidth,valign=c]{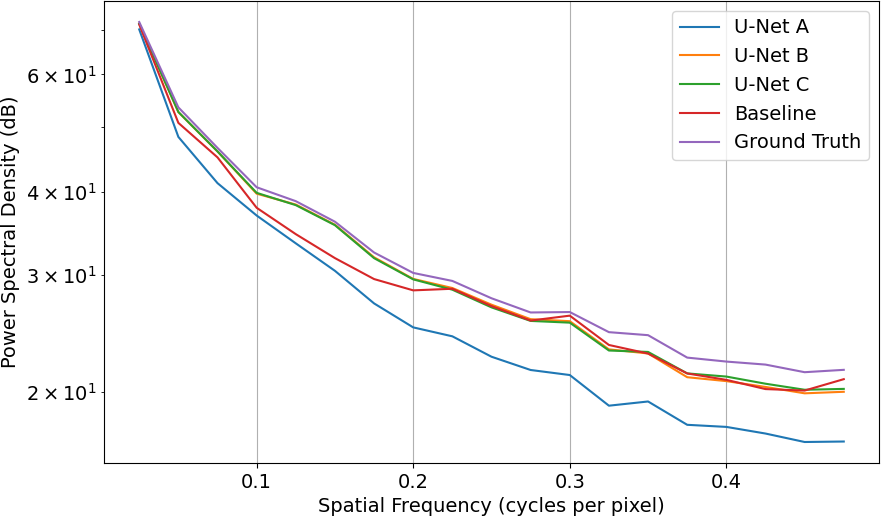}\label{fig:section_II_psd_d6}}
     \hspace{2mm}
     \subfloat[Average PDF. Note that the legend is the same as Figure \ref{fig:section_II_psd_d6}.]{\includegraphics[width=0.48\textwidth,valign=c]{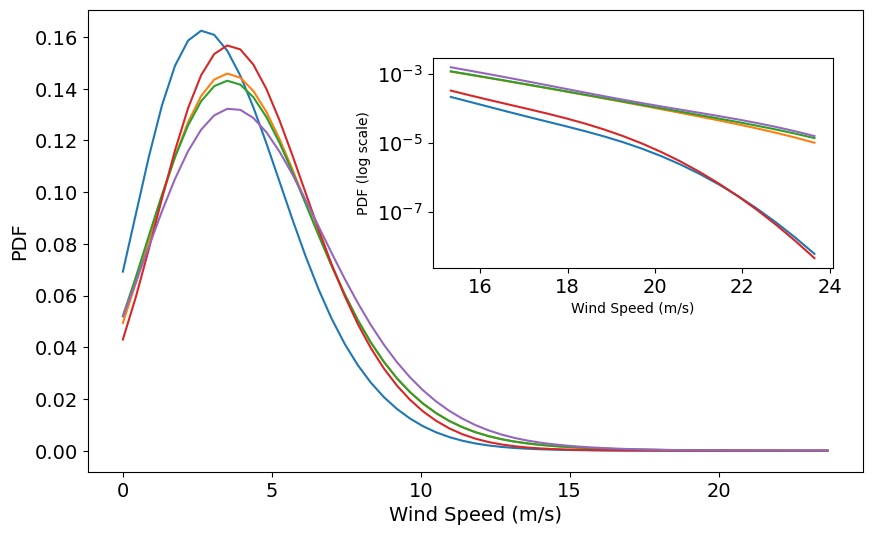}\label{fig:section_II_pdf_d6}}
     \caption{MAE, average PSD and average PDF on the test set over domain \#6 (2 899 hourly samples from January 2024 to April 2024), for the downscaling model (U-Net) trained with configuration A (five domains in Eastern Québec and New Brunswick), configuration B (five domains across Canada) or configuration C (10 domains across Canada), along with the baseline (bi-linear interpolation of the low-resolution $UV$ onto the predictand grid).}\label{fig:results_domain_II_d6}
\end{figure}

\begin{figure}[ht]
    \centering
     \hspace{2mm}
     \subfloat[RMSE.]{\includegraphics[height=4.6cm,valign=c]{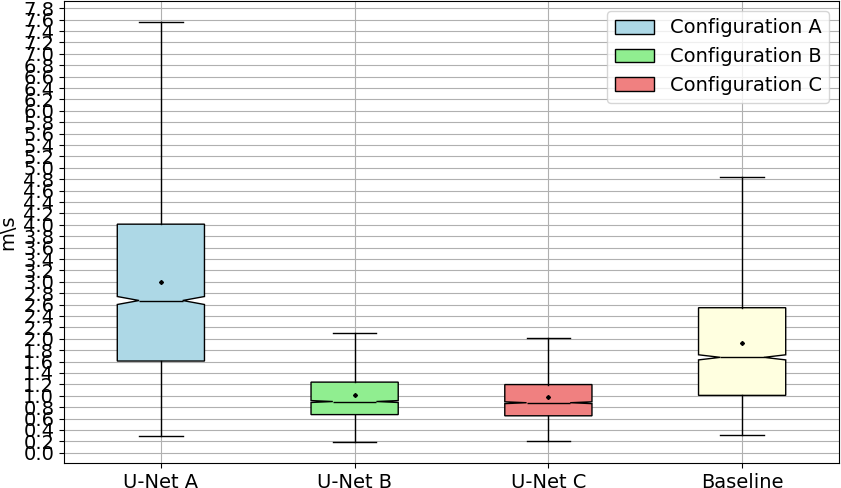}\label{fig:bp_II_rmse_d7}}
     \hspace{2mm}
     \subfloat[MAE.]{\includegraphics[height=4.6cm,valign=c]{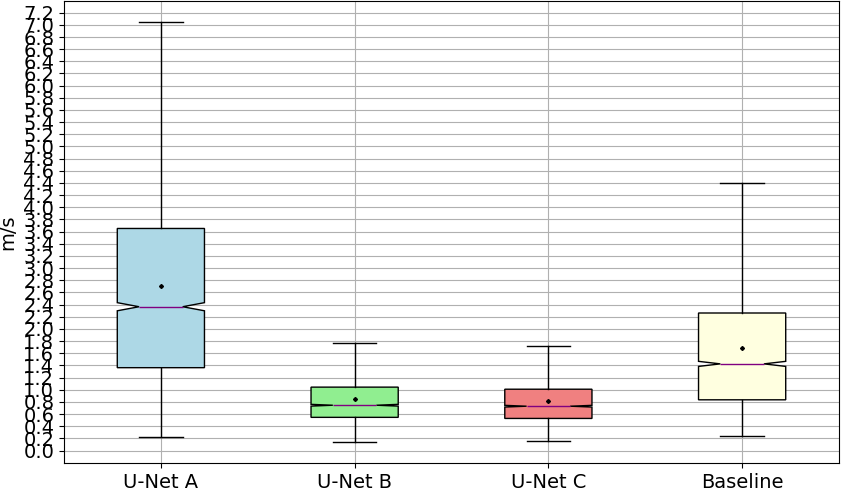}\label{fig:bp_II_mae_d7}}
     \hspace{2mm}
     \subfloat[SSIM.]{\includegraphics[height=4.6cm,valign=c]{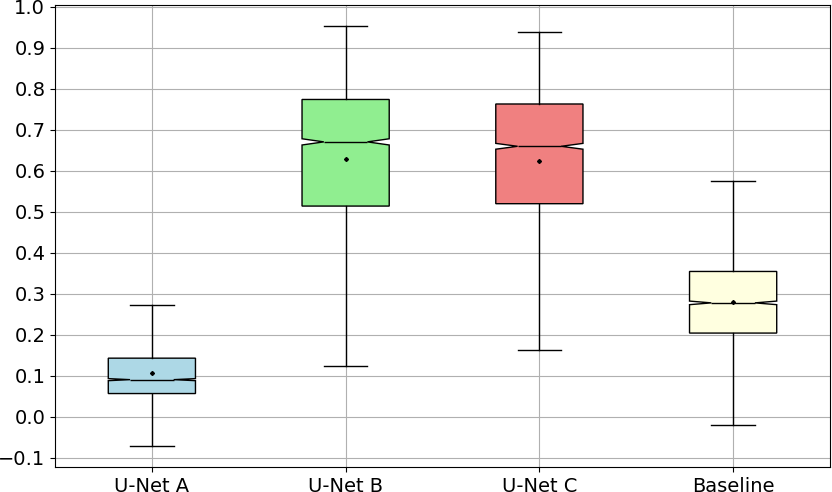}\label{fig:bp__II_ssim_d7}}
     
     \caption{RMSE, MAE and SSIM boxplots for the downscaling model (U-Net) predictions on the test set, which consists of hourly samples from January 2024 to April 2024, i.e. respectively 14 495 for configuration A for domain \#7 in Eastern Québec and New Brunswick, and 14 495 and 28 990 for configurations B and C across Canada. RMSE and SSIM boxplots for the baseline model (bi-linear interpolation of the low resolution $UV$ onto the predictand grid) are presented alongside. In each boxplot, the black dot and line represent respectively the mean and the median, the upper and lower box limits indicate the first (Q1) and third (Q3) quartiles and the whiskers depict the highest (lowest) value within the $1.5 \times$ (Q3-Q1) above Q3  (below Q1).}\label{fig:results_II_d7}
\end{figure}

\begin{figure}[ht]
    \centering
    \subfloat[Pixel-wise MAE (m/s) on domain \#7 for the downscaling model (U-Net) trained with each configuration along with the baseline's. The colorscale is capped at the maximum MAE of the downscaling model for each configuration if the U-Net is trained on that specific domain, else it is ignored from the maximum calculation. The baseline's MAE reaches higher values than the maximum of the colorscale in some locations. The value at the top of each image is the average over the domain.]{\includegraphics[width=0.99\textwidth,valign=c]{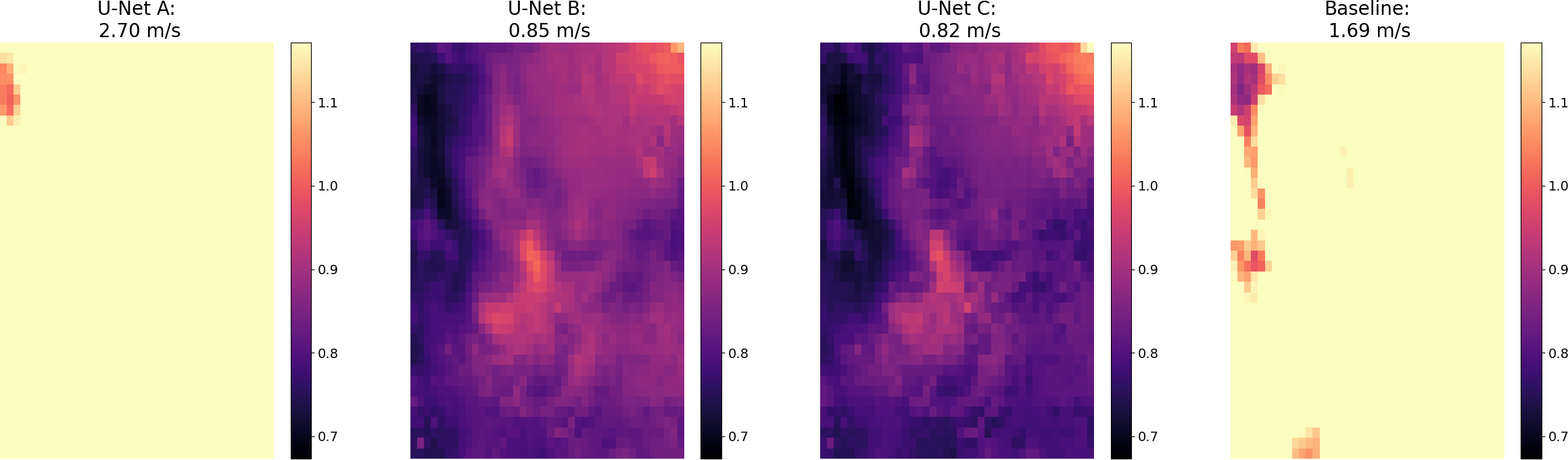}
     \label{fig:section_II_average_error_d7}}
     \\
     \subfloat[Average PSD.]{\includegraphics[width=0.48\textwidth,valign=c]{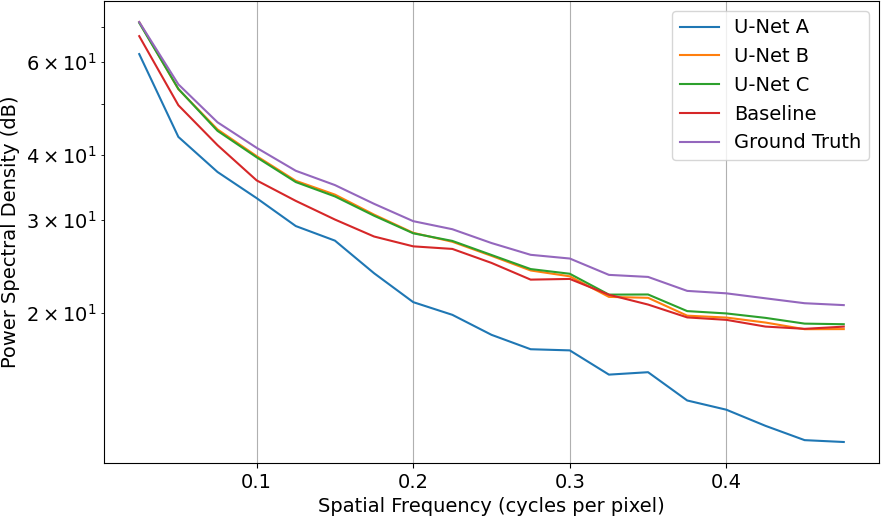}\label{fig:section_II_psd_d7}}
     \hspace{2mm}
     \subfloat[Average PDF. Note that the legend is the same as Figure \ref{fig:section_II_psd_d7}.]{\includegraphics[width=0.48\textwidth,valign=c]{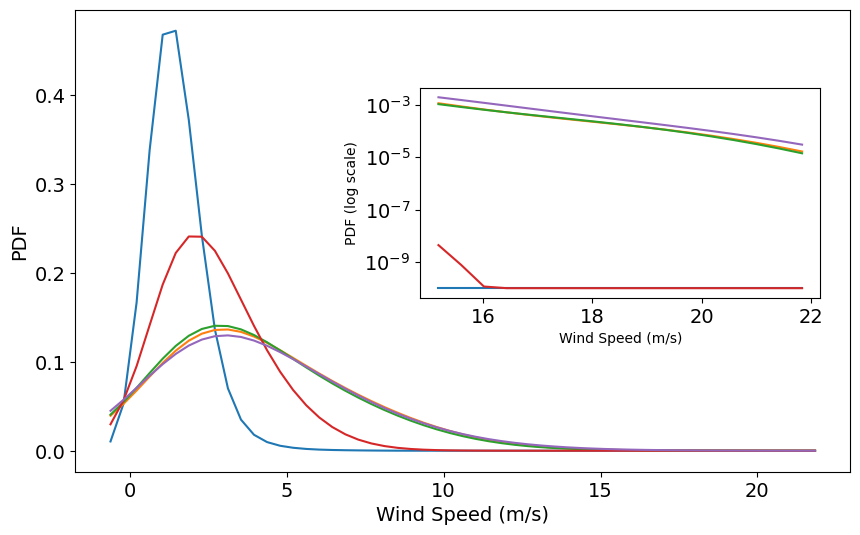}\label{fig:section_II_pdf_d7}}
     \caption{MAE, average PSD and average PDF on the test set over domain \#7 (2 899 hourly samples from January 2024 to April 2024), for the downscaling model (U-Net) trained with configuration A (five domains in Eastern Québec and New Brunswick), configuration B (five domains across Canada) or configuration C (10 domains across Canada), along with the baseline (bi-linear interpolation of the low-resolution $UV$ onto the predictand grid).}\label{fig:results_domain_II_d7}
\end{figure}

\begin{figure}[ht]
    \centering
     \hspace{2mm}
     \subfloat[RMSE.]{\includegraphics[height=4.6cm,valign=c]{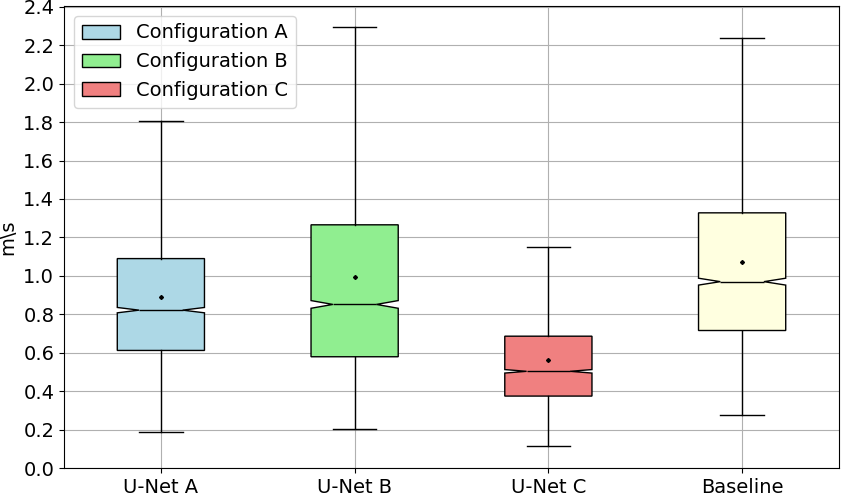}\label{fig:bp_II_rmse_d10}}
     \hspace{2mm}
     \subfloat[MAE.]{\includegraphics[height=4.6cm,valign=c]{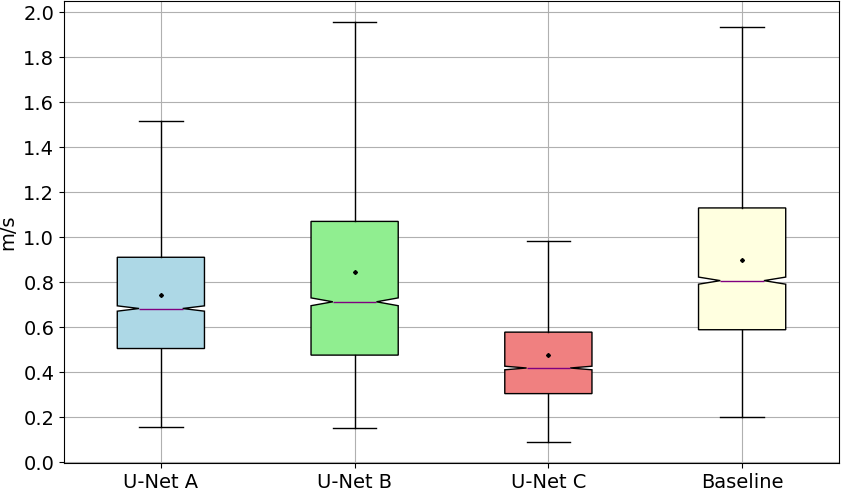}\label{fig:bp_II_mae_d10}}
     \hspace{2mm}
     \subfloat[SSIM.]{\includegraphics[height=4.6cm,valign=c]{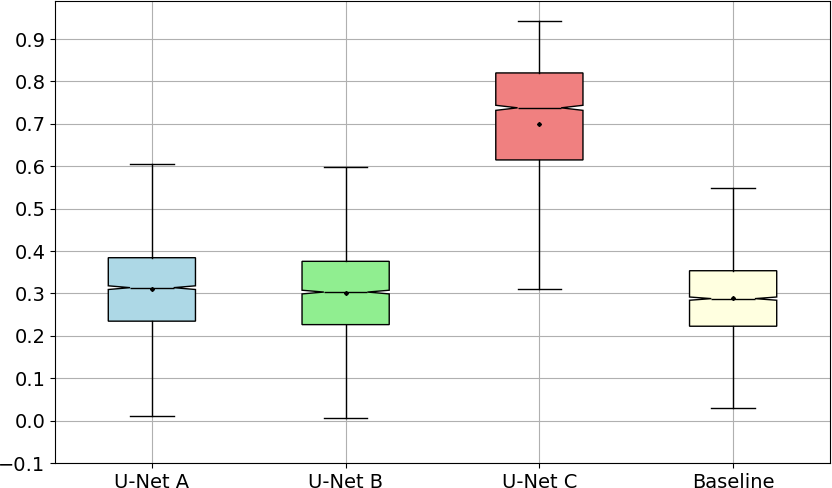}\label{fig:bp__II_ssim_d10}}
     
     \caption{RMSE, MAE and SSIM boxplots for the downscaling model (U-Net) predictions on the test set, which consists of hourly samples from January 2024 to April 2024, i.e. respectively 14 495 for configuration A for domain \#8 in Eastern Québec and New Brunswick, and 14 495 and 28 990 for configurations B and C across Canada. RMSE and SSIM boxplots for the baseline model (bi-linear interpolation of the low resolution $UV$ onto the predictand grid) are presented alongside. In each boxplot, the black dot and line represent respectively the mean and the median, the upper and lower box limits indicate the first (Q1) and third (Q3) quartiles and the whiskers depict the highest (lowest) value within the $1.5 \times$ (Q3-Q1) above Q3  (below Q1).}\label{fig:results_II_d10}
\end{figure}

\begin{figure}[ht]
    \centering
    \subfloat[Pixel-wise MAE (m/s) on domain \#8 for the downscaling model (U-Net) trained with each configuration along with the baseline's. The colorscale is capped at the maximum MAE of the downscaling model for each configuration if the U-Net is trained on that specific domain, else it is ignored from the maximum calculation. The baseline's MAE reaches higher values than the maximum of the colorscale in some locations. The value at the top of each image is the average over the domain.]{\includegraphics[width=0.99\textwidth,valign=c]{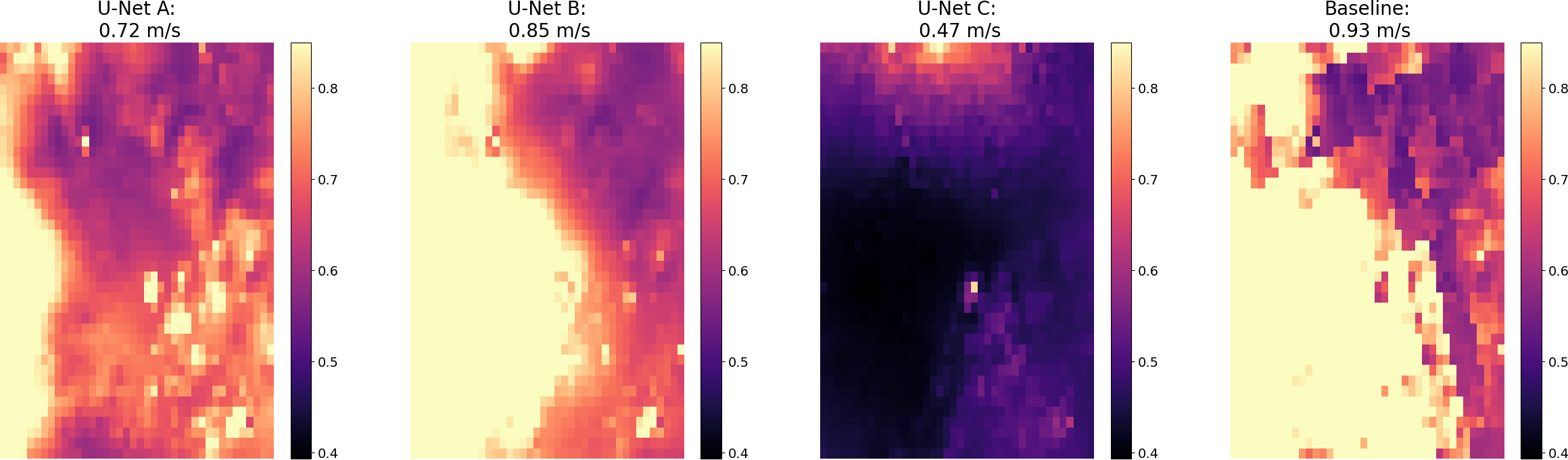}
     \label{fig:section_II_average_error_d10}}
     \\
     \subfloat[Average PSD.]{\includegraphics[width=0.48\textwidth,valign=c]{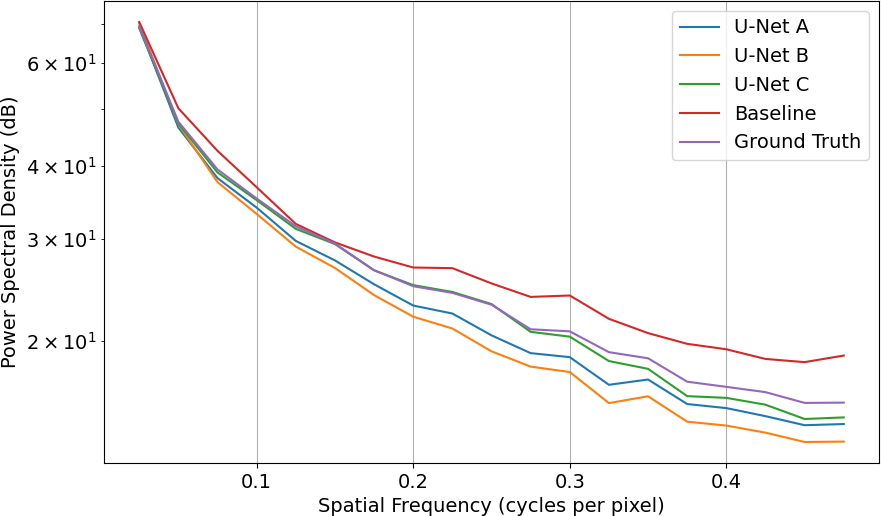}\label{fig:section_II_psd_d10}}
     \hspace{2mm}
     \subfloat[Average PDF. Note that the legend is the same as Figure \ref{fig:section_II_psd_d10}.]{\includegraphics[width=0.48\textwidth,valign=c]{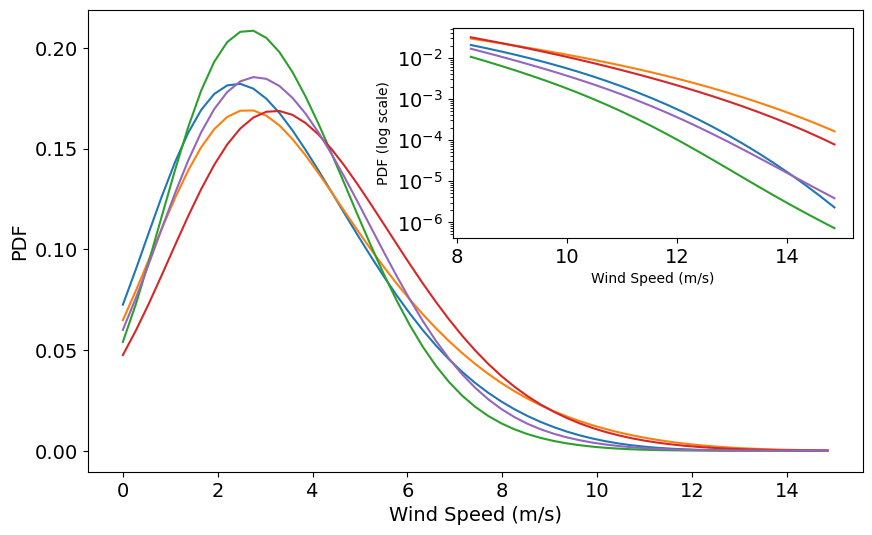}\label{fig:section_II_pdf_d10}}
     \caption{MAE, average PSD and average PDF on the test set over domain \#8 (2 899 hourly samples from January 2024 to April 2024), for the downscaling model (U-Net) trained with configuration A (five domains in Eastern Québec and New Brunswick), configuration B (five domains across Canada) or configuration C (10 domains across Canada), along with the baseline (bi-linear interpolation of the low-resolution $UV$ onto the predictand grid).}\label{fig:results_domain_II_d10}
\end{figure}

\begin{figure}[ht]
    \centering
     \hspace{2mm}
     \subfloat[RMSE.]{\includegraphics[height=4.6cm,valign=c]{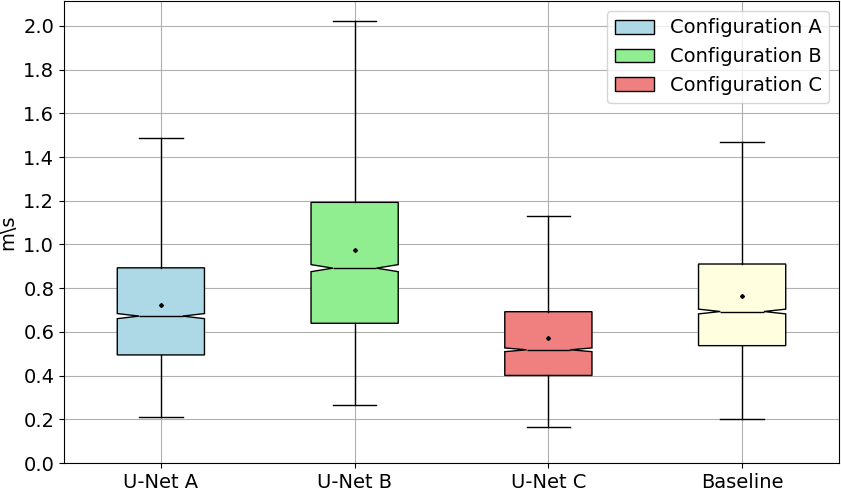}\label{fig:bp_II_rmse_d11}}
     \hspace{2mm}
     \subfloat[MAE.]{\includegraphics[height=4.6cm,valign=c]{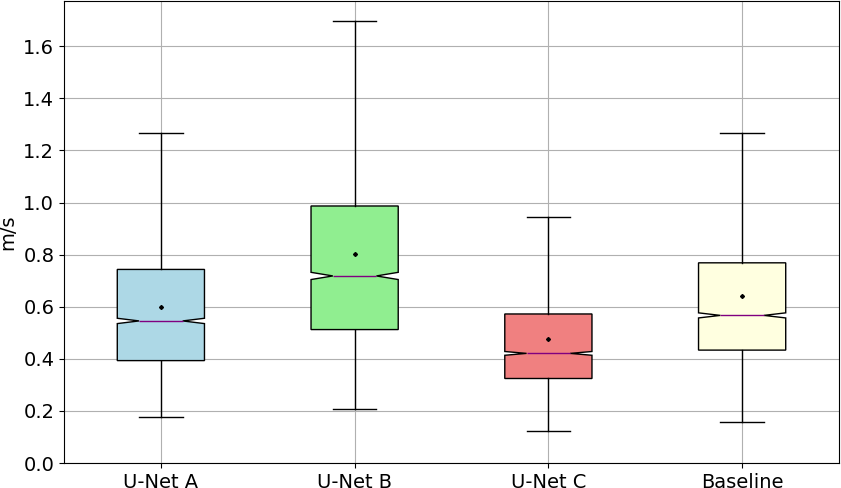}\label{fig:bp_II_mae_d11}}
     \hspace{2mm}
     \subfloat[SSIM.]{\includegraphics[height=4.6cm,valign=c]{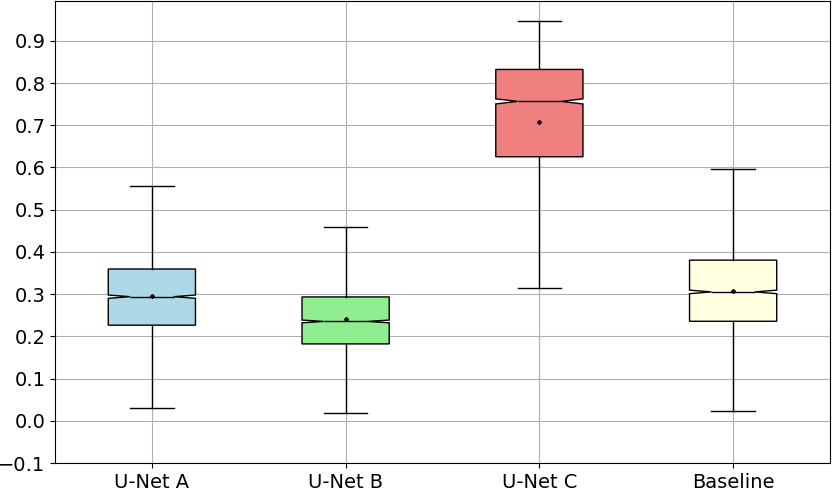}\label{fig:bp__II_ssim_d11}}
     
     \caption{RMSE, MAE and SSIM boxplots for the downscaling model (U-Net) predictions on the test set, which consists of hourly samples from January 2024 to April 2024, i.e. respectively 14 495 for configuration A for domain \#9 in Eastern Québec and New Brunswick, and 14 495 and 28 990 for configurations B and C across Canada. RMSE and SSIM boxplots for the baseline model (bi-linear interpolation of the low resolution $UV$ onto the predictand grid) are presented alongside. In each boxplot, the black dot and line represent respectively the mean and the median, the upper and lower box limits indicate the first (Q1) and third (Q3) quartiles and the whiskers depict the highest (lowest) value within the $1.5 \times$ (Q3-Q1) above Q3  (below Q1).}\label{fig:results_II_d11}
\end{figure}

\begin{figure}[ht]
    \centering
    \subfloat[Pixel-wise MAE (m/s) on domain \#9 for the downscaling model (U-Net) trained with each configuration along with the baseline's. The colorscale is capped at the maximum MAE of the downscaling model for each configuration if the U-Net is trained on that specific domain, else it is ignored from the maximum calculation. The baseline's MAE reaches higher values than the maximum of the colorscale in some locations. The value at the top of each image is the average over the domain.]{\includegraphics[width=0.99\textwidth,valign=c]{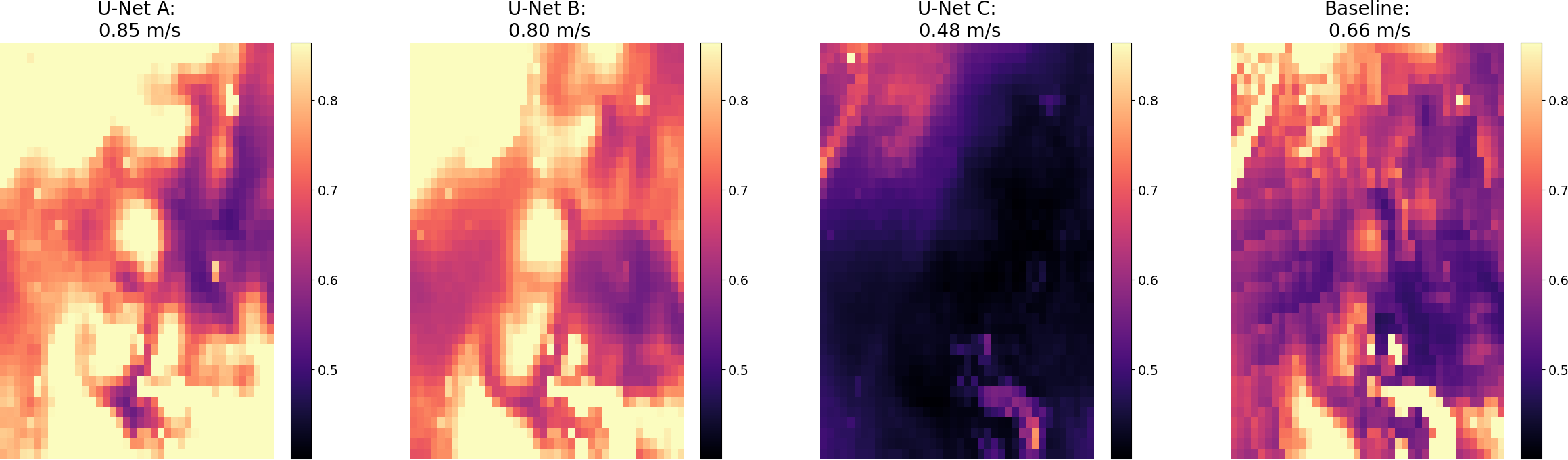}
     \label{fig:section_II_average_error_d11}}
     \\
     \subfloat[Average PSD.]{\includegraphics[width=0.48\textwidth,valign=c]{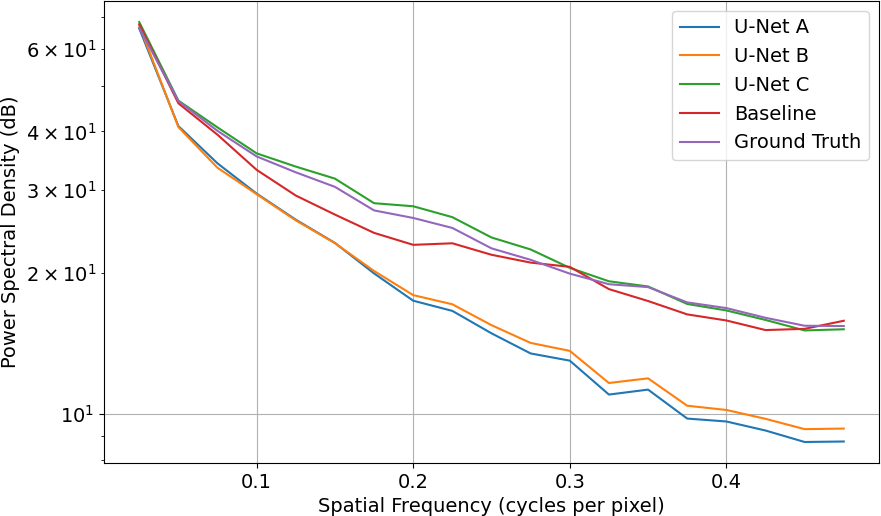}\label{fig:section_II_psd_d11}}
     \hspace{2mm}
     \subfloat[Average PDF. Note that the legend is the same as Figure \ref{fig:section_II_psd_d11}.]{\includegraphics[width=0.48\textwidth,valign=c]{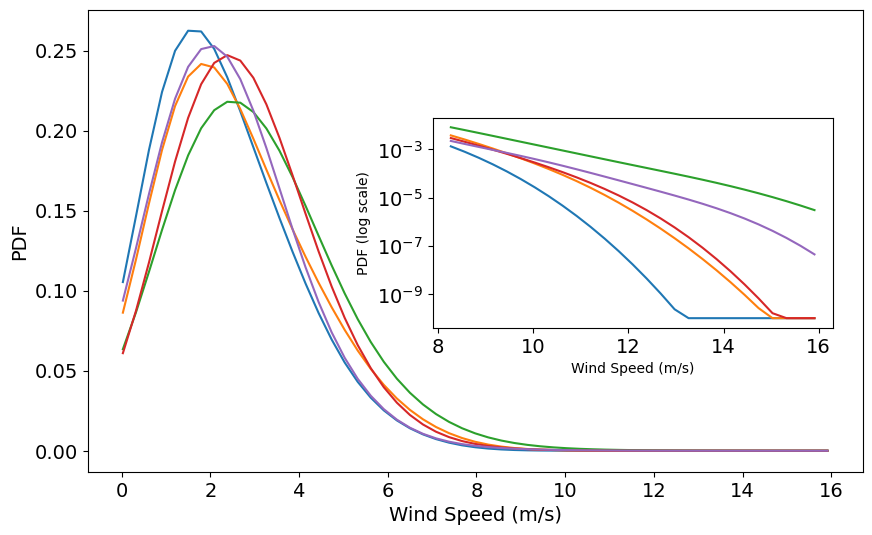}\label{fig:section_II_pdf_d11}}
     \caption{MAE, average PSD and average PDF on the test set over domain \#9 (2 899 hourly samples from January 2024 to April 2024), for the downscaling model (U-Net) trained with configuration A (five domains in Eastern Québec and New Brunswick), configuration B (five domains across Canada) or configuration C (10 domains across Canada), along with the baseline (bi-linear interpolation of the low-resolution $UV$ onto the predictand grid).}\label{fig:results_domain_II_d11}
\end{figure}

\begin{figure}[ht]
    \centering
     \hspace{2mm}
     \subfloat[RMSE.]{\includegraphics[height=4.6cm,valign=c]{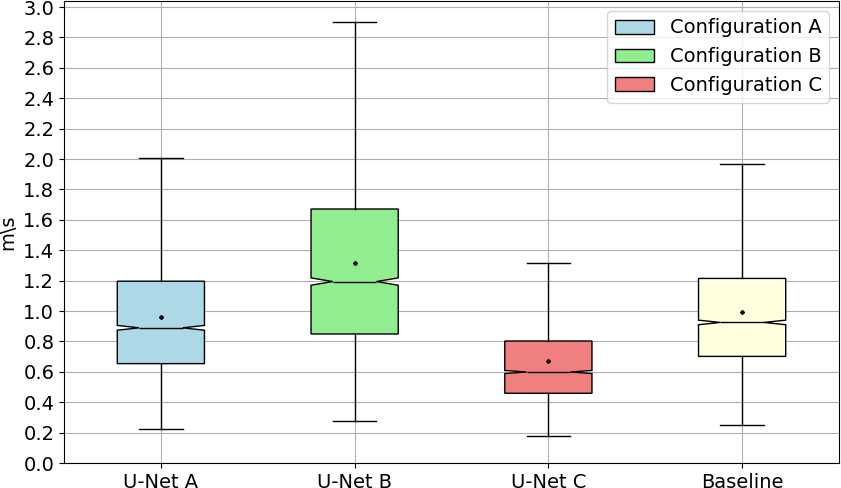}\label{fig:bp_II_rmse_d12}}
     \hspace{2mm}
     \subfloat[MAE.]{\includegraphics[height=4.6cm,valign=c]{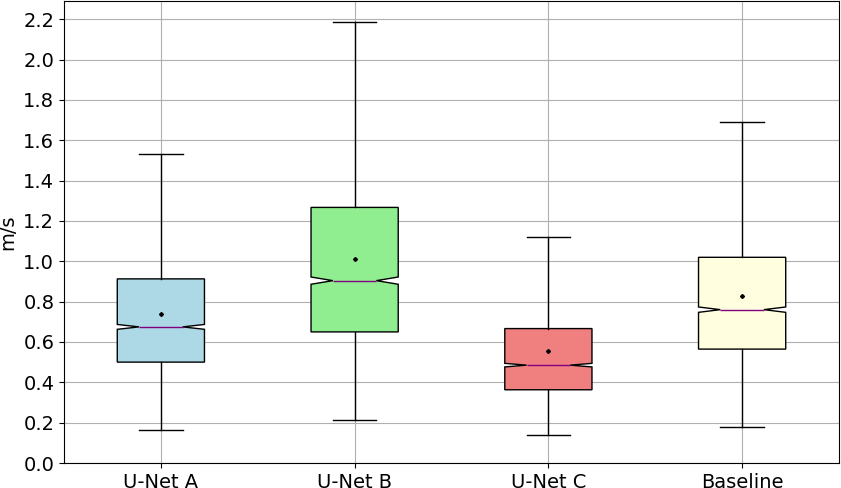}\label{fig:bp_II_mae_d12}}
     \hspace{2mm}
     \subfloat[SSIM.]{\includegraphics[height=4.6cm,valign=c]{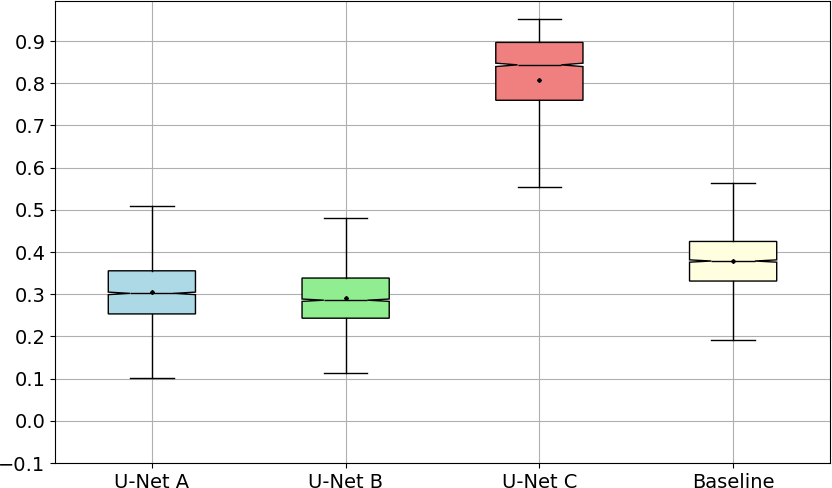}\label{fig:bp__II_ssim_d12}}
     
     \caption{RMSE, MAE and SSIM boxplots for the downscaling model (U-Net) predictions on the test set, which consists of hourly samples from January 2024 to April 2024, i.e. respectively 14 495 for configuration A for domain \#10 in Eastern Québec and New Brunswick, and 14 495 and 28 990 for configurations B and C across Canada. RMSE and SSIM boxplots for the baseline model (bi-linear interpolation of the low resolution $UV$ onto the predictand grid) are presented alongside. In each boxplot, the black dot and line represent respectively the mean and the median, the upper and lower box limits indicate the first (Q1) and third (Q3) quartiles and the whiskers depict the highest (lowest) value within the $1.5 \times$ (Q3-Q1) above Q3  (below Q1).}\label{fig:results_II_d12}
\end{figure}

\begin{figure}[ht]
    \centering
    \subfloat[Pixel-wise MAE (m/s) on domain \#10 for the downscaling model (U-Net) trained with each configuration along with the baseline's. The colorscale is capped at the maximum MAE of the downscaling model for each configuration if the U-Net is trained on that specific domain, else it is ignored from the maximum calculation. The baseline's MAE reaches higher values than the maximum of the colorscale in some locations. The value at the top of each image is the average over the domain.]{\includegraphics[width=0.99\textwidth,valign=c]{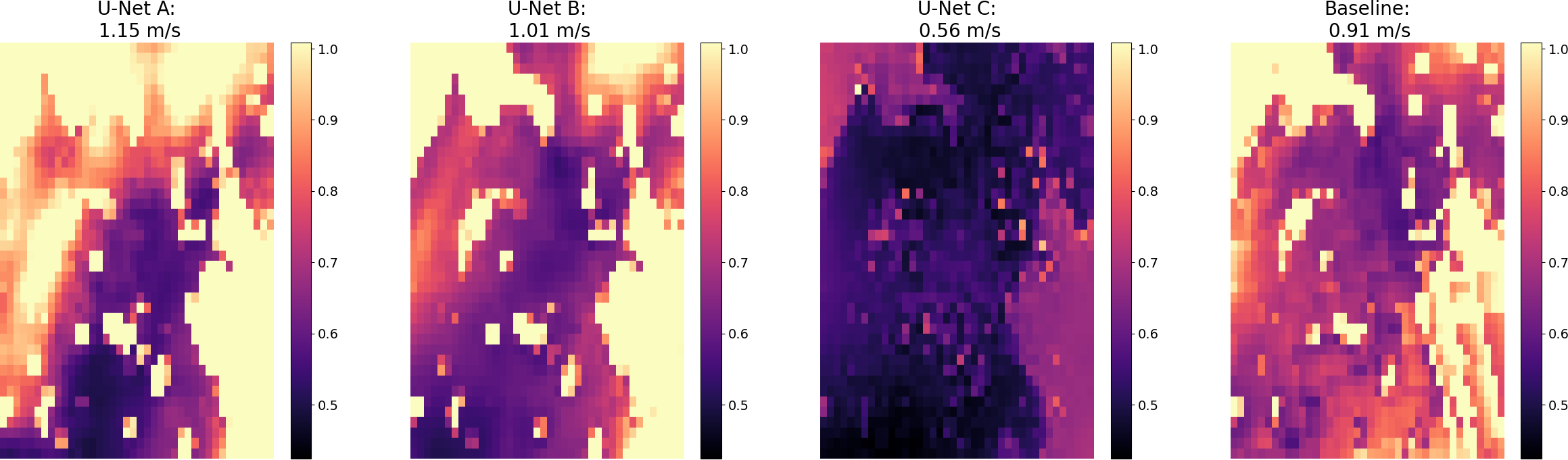}
     \label{fig:section_II_average_error_d12}}
     \\
     \subfloat[Average PSD.]{\includegraphics[width=0.48\textwidth,valign=c]{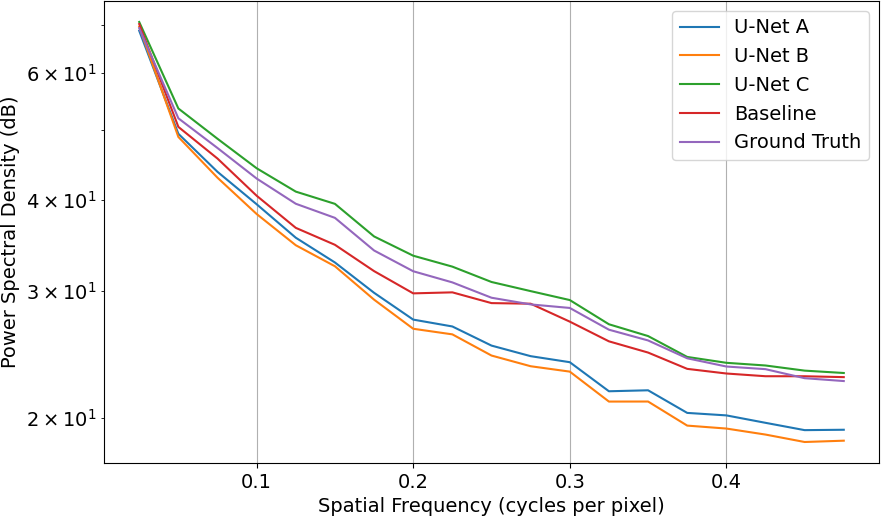}\label{fig:section_II_psd_d12}}
     \hspace{2mm}
     \subfloat[Average PDF. Note that the legend is the same as Figure \ref{fig:section_II_psd_d12}.]{\includegraphics[width=0.48\textwidth,valign=c]{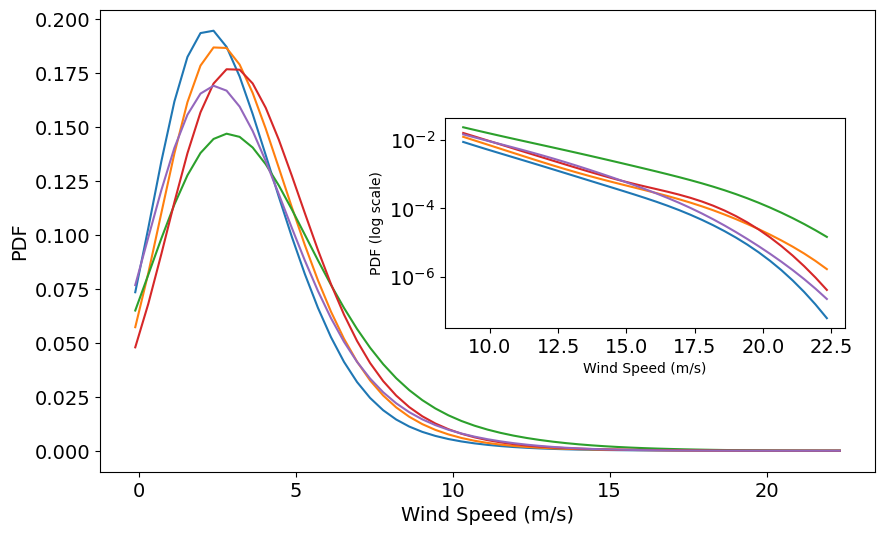}\label{fig:section_II_pdf_d12}}
     \caption{MAE, average PSD and average PDF on the test set over domain \#10 (2 899 hourly samples from January 2024 to April 2024), for the downscaling model (U-Net) trained with configuration A (five domains in Eastern Québec and New Brunswick), configuration B (five domains across Canada) or configuration C (10 domains across Canada), along with the baseline (bi-linear interpolation of the low-resolution $UV$ onto the predictand grid).}\label{fig:results_domain_II_d12}
\end{figure}

\begin{figure}[ht]
    \centering
     \hspace{2mm}
     \subfloat[RMSE.]{\includegraphics[height=4.6cm,valign=c]{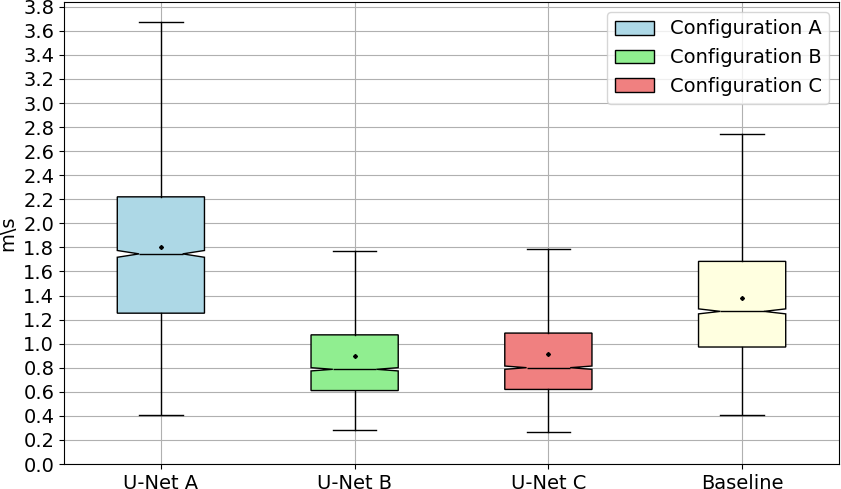}\label{fig:bp_II_rmse_d13}}
     \hspace{2mm}
     \subfloat[MAE.]{\includegraphics[height=4.6cm,valign=c]{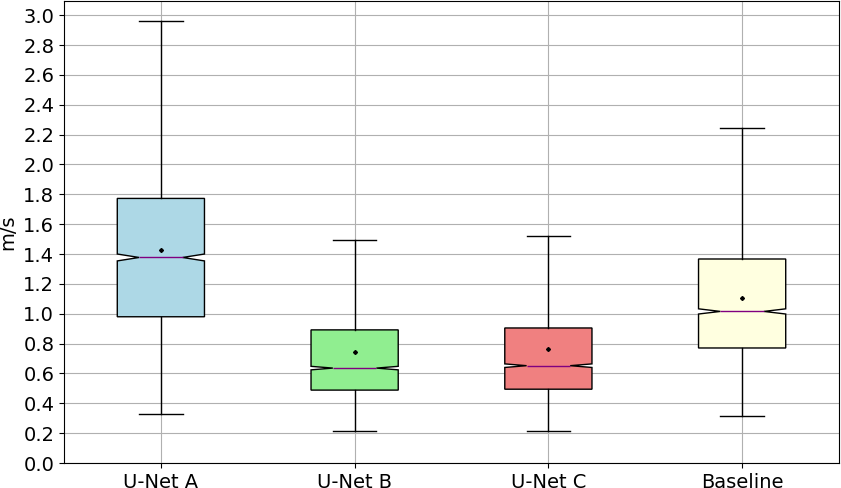}\label{fig:bp_II_mae_d13}}
     \hspace{2mm}
     \subfloat[SSIM.]{\includegraphics[height=4.6cm,valign=c]{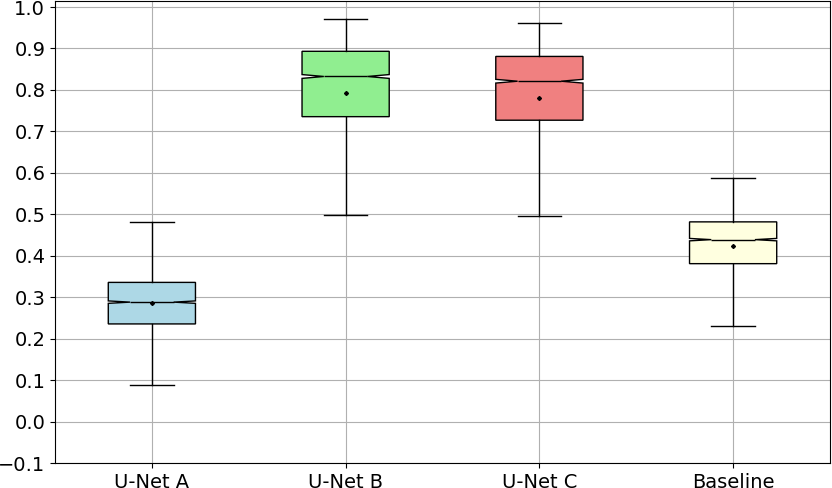}\label{fig:bp__II_ssim_d13}}
     
     \caption{RMSE, MAE and SSIM boxplots for the downscaling model (U-Net) predictions on the test set, which consists of hourly samples from January 2024 to April 2024, i.e. respectively 14 495 for configuration A for domain \#11 in Eastern Québec and New Brunswick, and 14 495 and 28 990 for configurations B and C across Canada. RMSE and SSIM boxplots for the baseline model (bi-linear interpolation of the low resolution $UV$ onto the predictand grid) are presented alongside. In each boxplot, the black dot and line represent respectively the mean and the median, the upper and lower box limits indicate the first (Q1) and third (Q3) quartiles and the whiskers depict the highest (lowest) value within the $1.5 \times$ (Q3-Q1) above Q3  (below Q1).}\label{fig:results_II_d13}
\end{figure}

\begin{figure}[ht]
    \centering
    \subfloat[Pixel-wise MAE (m/s) on domain \#11 for the downscaling model (U-Net) trained with each configuration along with the baseline's. The colorscale is capped at the maximum MAE of the downscaling model for each configuration if the U-Net is trained on that specific domain, else it is ignored from the maximum calculation. The baseline's MAE reaches higher values than the maximum of the colorscale in some locations. The value at the top of each image is the average over the domain.]{\includegraphics[width=0.99\textwidth,valign=c]{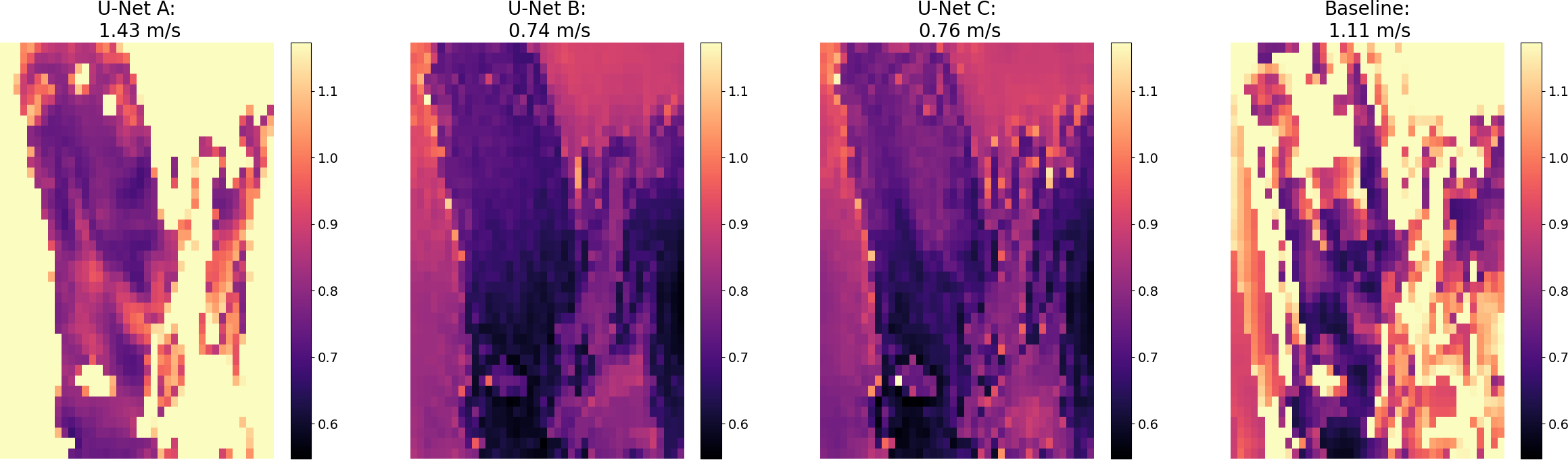}
     \label{fig:section_II_average_error_d13}}
     \\
     \subfloat[Average PSD.]{\includegraphics[width=0.48\textwidth,valign=c]{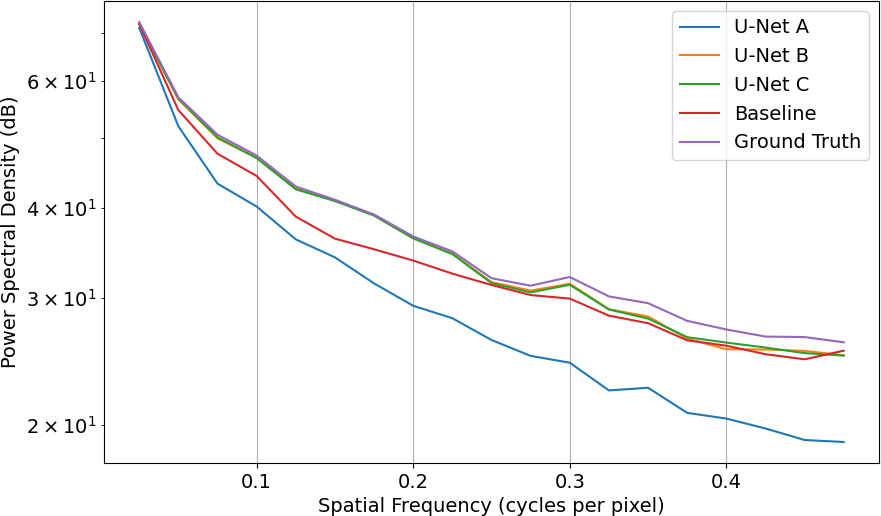}\label{fig:section_II_psd_d13}}
     \hspace{2mm}
     \subfloat[Average PDF. Note that the legend is the same as Figure \ref{fig:section_II_psd_d13}.]{\includegraphics[width=0.48\textwidth,valign=c]{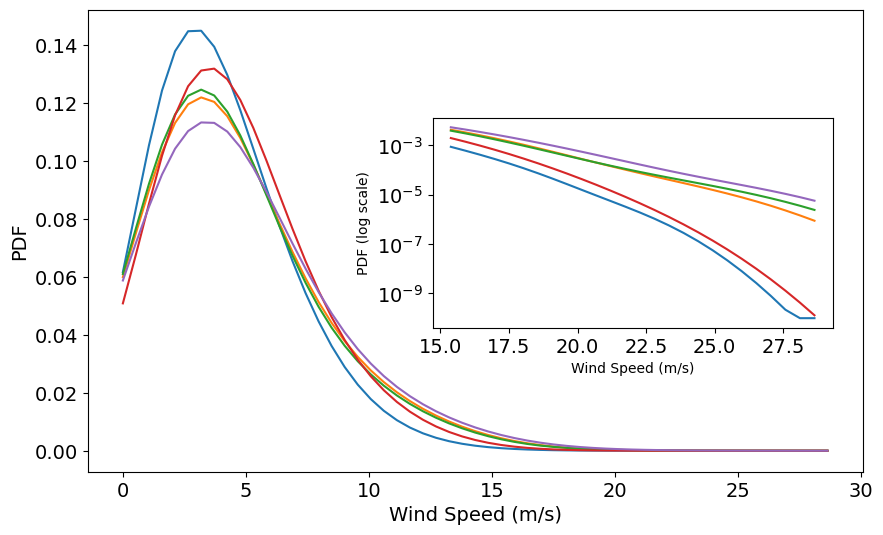}\label{fig:section_II_pdf_d13}}
     \caption{MAE, average PSD and average PDF on the test set over domain \#11 (2 899 hourly samples from January 2024 to April 2024), for the downscaling model (U-Net) trained with configuration A (five domains in Eastern Québec and New Brunswick), configuration B (five domains across Canada) or configuration C (10 domains across Canada), along with the baseline (bi-linear interpolation of the low-resolution $UV$ onto the predictand grid).}\label{fig:results_domain_II_d13}
\end{figure}

\begin{figure}[ht]
    \centering
     \hspace{2mm}
     \subfloat[RMSE.]{\includegraphics[height=4.6cm,valign=c]{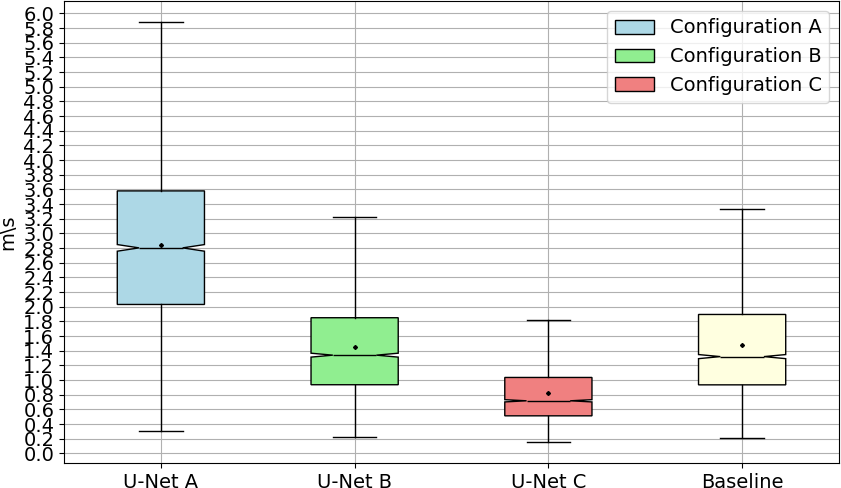}\label{fig:bp_II_rmse_d14}}
     \hspace{2mm}
     \subfloat[MAE.]{\includegraphics[height=4.6cm,valign=c]{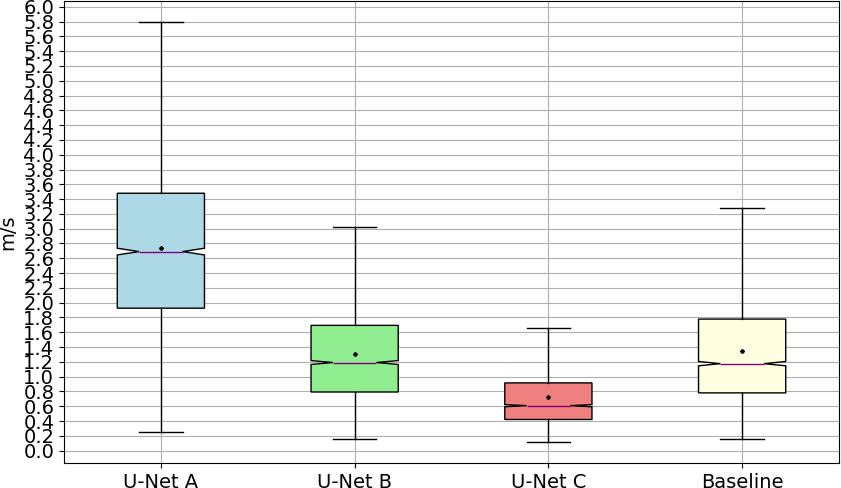}\label{fig:bp_II_mae_d14}}
     \hspace{2mm}
     \subfloat[SSIM.]{\includegraphics[height=4.6cm,valign=c]{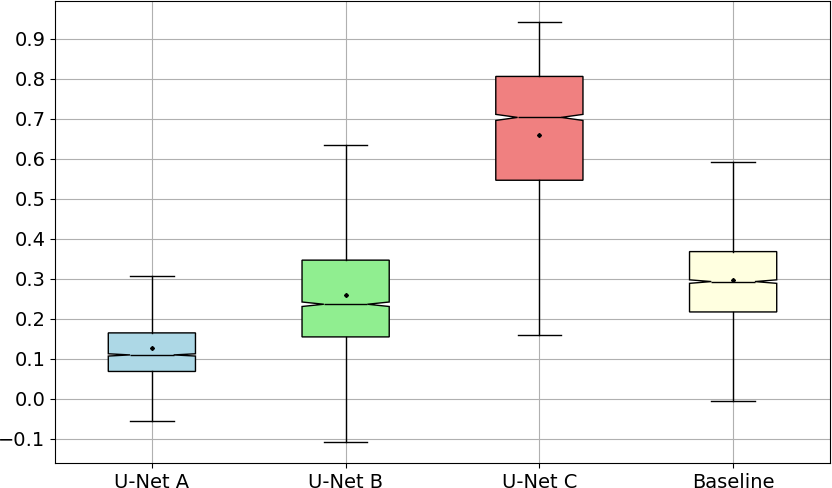}\label{fig:bp__II_ssim_d14}}
     
     \caption{RMSE, MAE and SSIM boxplots for the downscaling model (U-Net) predictions on the test set, which consists of hourly samples from January 2024 to April 2024, i.e. respectively 14 495 for configuration A for domain \#12 in Eastern Québec and New Brunswick, and 14 495 and 28 990 for configurations B and C across Canada. RMSE and SSIM boxplots for the baseline model (bi-linear interpolation of the low resolution $UV$ onto the predictand grid) are presented alongside. In each boxplot, the black dot and line represent respectively the mean and the median, the upper and lower box limits indicate the first (Q1) and third (Q3) quartiles and the whiskers depict the highest (lowest) value within the $1.5 \times$ (Q3-Q1) above Q3  (below Q1).}\label{fig:results_II_d14}
\end{figure}

\begin{figure}[ht]
    \centering
    \subfloat[Pixel-wise MAE (m/s) on domain \#12 for the downscaling model (U-Net) trained with each configuration along with the baseline's. The colorscale is capped at the maximum MAE of the downscaling model for each configuration if the U-Net is trained on that specific domain, else it is ignored from the maximum calculation. The baseline's MAE reaches higher values than the maximum of the colorscale in some locations. The value at the top of each image is the average over the domain.]{\includegraphics[width=0.99\textwidth,valign=c]{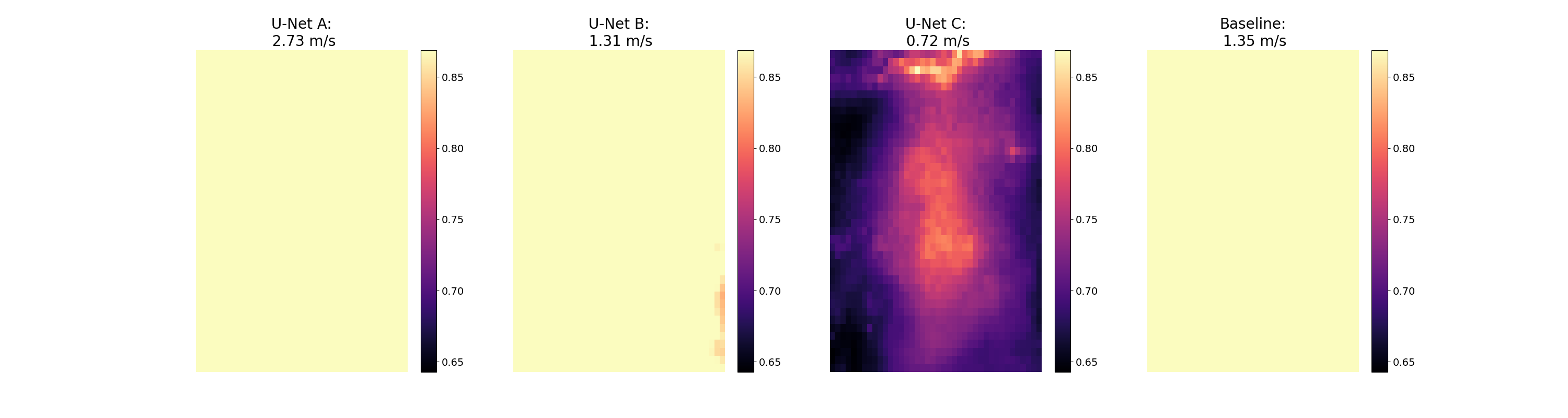}
     \label{fig:section_II_average_error_d14}}
     \\
     \subfloat[Average PSD.]{\includegraphics[width=0.48\textwidth,valign=c]{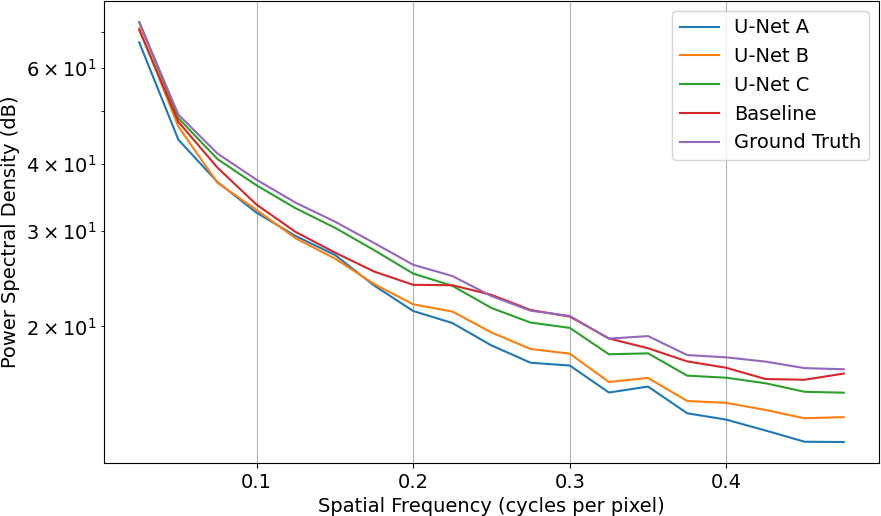}\label{fig:section_II_psd_d14}}
     \hspace{2mm}
     \subfloat[Average PDF. Note that the legend is the same as Figure \ref{fig:section_II_psd_d14}.]{\includegraphics[width=0.48\textwidth,valign=c]{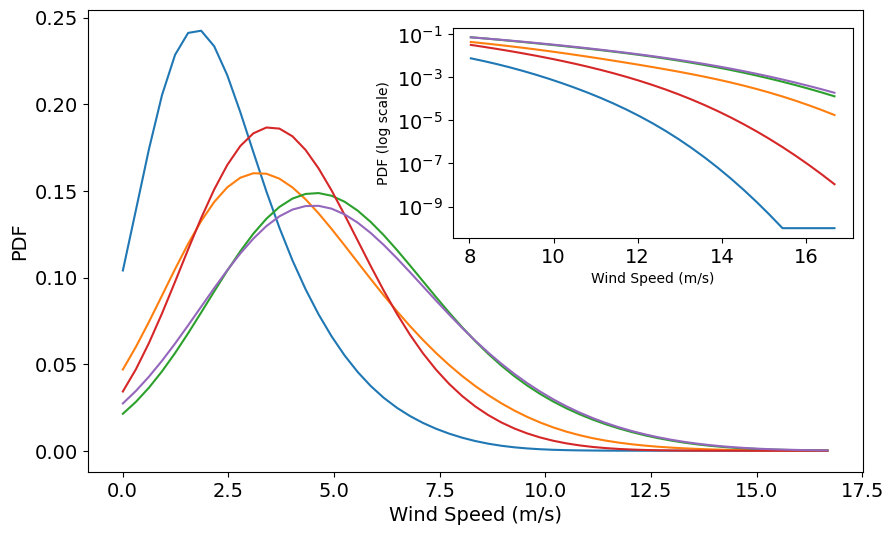}\label{fig:section_II_pdf_d14}}
     \caption{MAE, average PSD and average PDF on the test set over domain \#12 (2 899 hourly samples from January 2024 to April 2024), for the downscaling model (U-Net) trained with configuration A (five domains in Eastern Québec and New Brunswick), configuration B (five domains across Canada) or configuration C (10 domains across Canada), along with the baseline (bi-linear interpolation of the low-resolution $UV$ onto the predictand grid).}\label{fig:results_domain_II_d14}
\end{figure}

\begin{figure}[ht]
    \centering
     \hspace{2mm}
     \subfloat[RMSE.]{\includegraphics[height=4.6cm,valign=c]{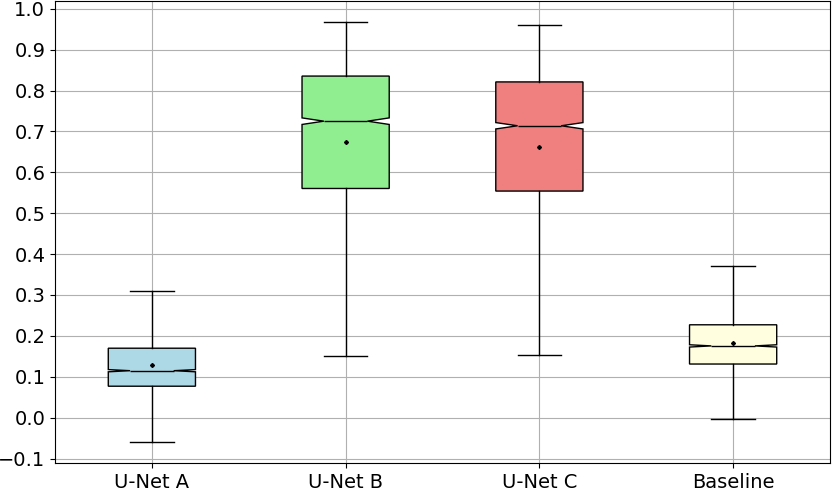}\label{fig:bp_II_rmse_d15}}
     \hspace{2mm}
     \subfloat[MAE.]{\includegraphics[height=4.6cm,valign=c]{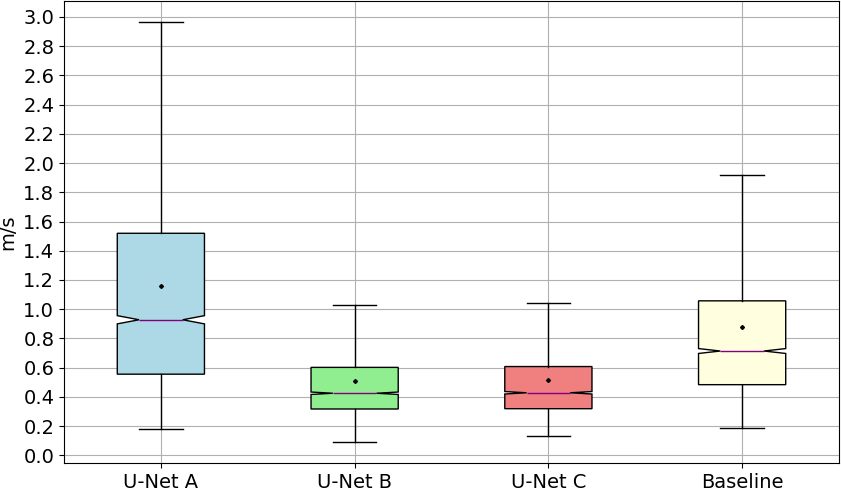}\label{fig:bp_II_mae_d15}}
     \hspace{2mm}
     \subfloat[SSIM.]{\includegraphics[height=4.6cm,valign=c]{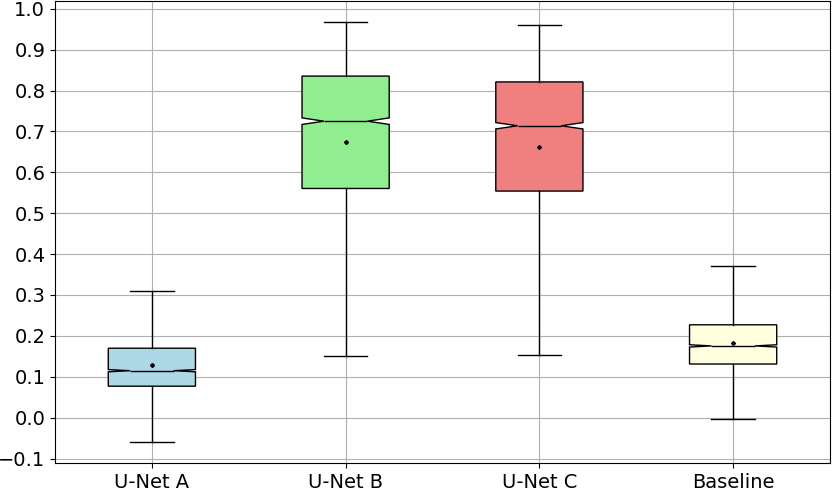}\label{fig:bp__II_ssim_d15}}
     
     \caption{RMSE, MAE and SSIM boxplots for the downscaling model (U-Net) predictions on the test set, which consists of hourly samples from January 2024 to April 2024, i.e. respectively 14 495 for configuration A for domain \#13 in Eastern Québec and New Brunswick, and 14 495 and 28 990 for configurations B and C across Canada. RMSE and SSIM boxplots for the baseline model (bi-linear interpolation of the low resolution $UV$ onto the predictand grid) are presented alongside. In each boxplot, the black dot and line represent respectively the mean and the median, the upper and lower box limits indicate the first (Q1) and third (Q3) quartiles and the whiskers depict the highest (lowest) value within the $1.5 \times$ (Q3-Q1) above Q3  (below Q1).}\label{fig:results_II_d15}
\end{figure}

\begin{figure}[ht]
    \centering
    \subfloat[Pixel-wise MAE (m/s) on domain \#13 for the downscaling model (U-Net) trained with each configuration along with the baseline's. The colorscale is capped at the maximum MAE of the downscaling model for each configuration if the U-Net is trained on that specific domain, else it is ignored from the maximum calculation. The baseline's MAE reaches higher values than the maximum of the colorscale in some locations. The value at the top of each image is the average over the domain.]{\includegraphics[width=0.99\textwidth,valign=c]{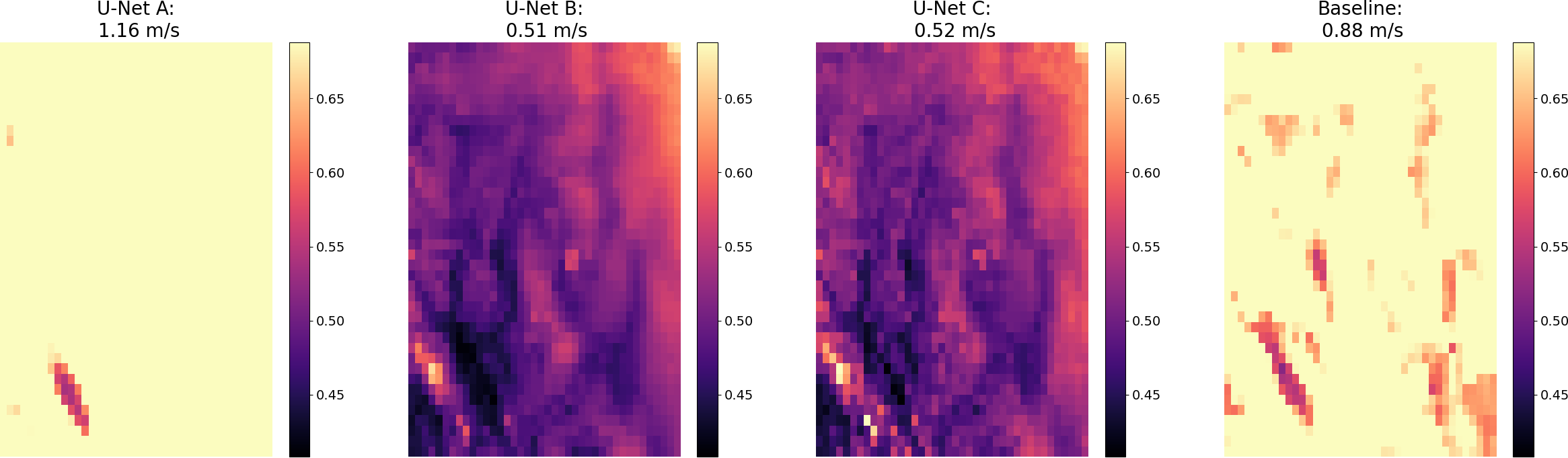}
     \label{fig:section_II_average_error_d15}}
     \\
     \subfloat[Average PSD.]{\includegraphics[width=0.48\textwidth,valign=c]{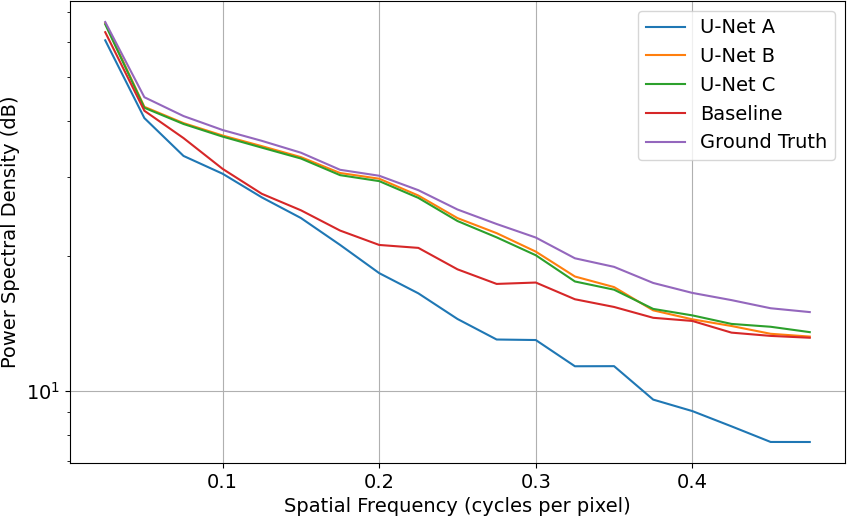}\label{fig:section_II_psd_d15}}
     \hspace{2mm}
     \subfloat[Average PDF. Note that the legend is the same as Figure \ref{fig:section_II_psd_d15}.]{\includegraphics[width=0.48\textwidth,valign=c]{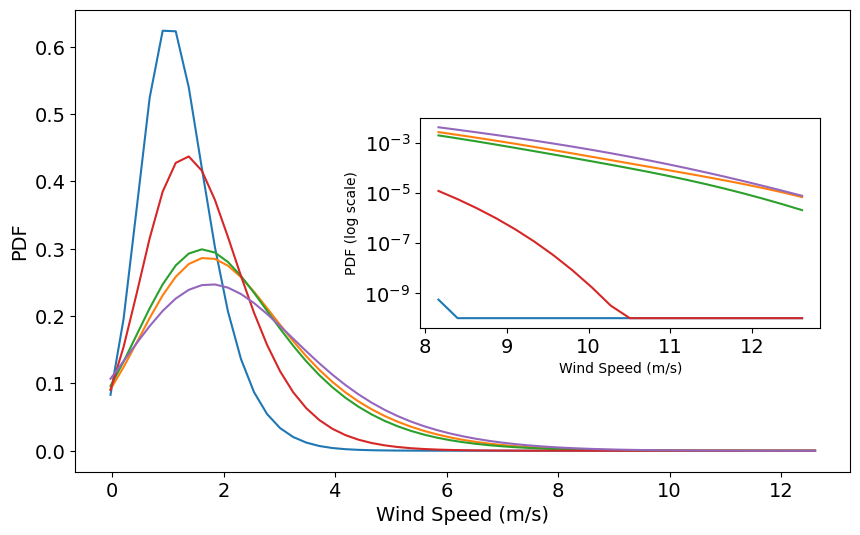}\label{fig:section_II_pdf_d15}}
     \caption{MAE, average PSD and average PDF on the test set over domain \#13 (2 899 hourly samples from January 2024 to April 2024), for the downscaling model (U-Net) trained with configuration A (five domains in Eastern Québec and New Brunswick), configuration B (five domains across Canada) or configuration C (10 domains across Canada), along with the baseline (bi-linear interpolation of the low-resolution $UV$ onto the predictand grid).}\label{fig:results_domain_II_d15}
\end{figure}

% transfer learning results
\clearpage
\subsection{Transfer learning experiment}
Here are all the results for domains \#1 to \#13 where domains \#2, \#4 and \#5 were not in configuration C used for training.

\begin{figure}[ht]
    \centering
     \hspace{2mm}
     \subfloat[RMSE.]{\includegraphics[height=4.6cm,valign=c]{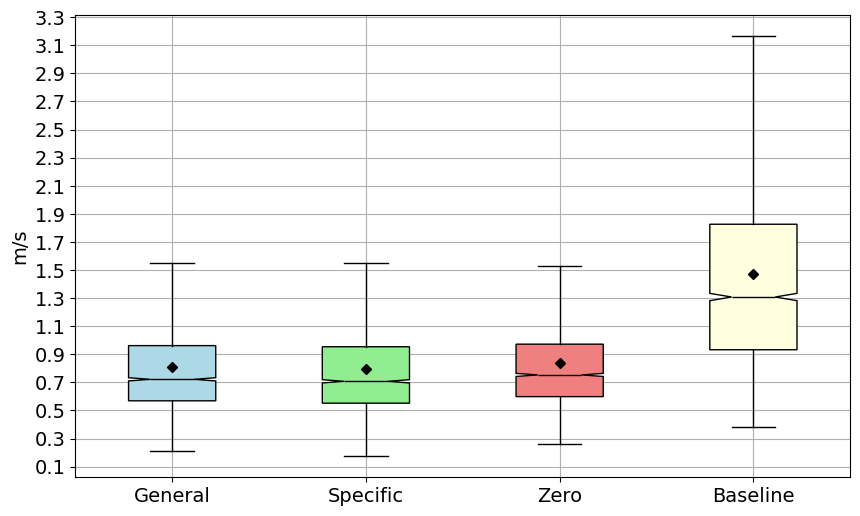}\label{fig:section_III_bp_rmse_d1}}
     \hspace{2mm}
     \subfloat[MAE.]{\includegraphics[height=4.6cm,valign=c]{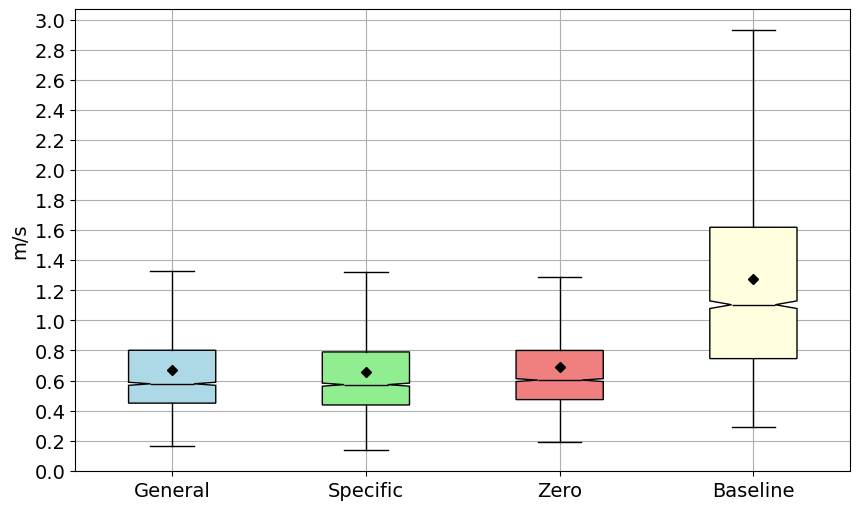}\label{fig:section_III_bp_mae_d1}}
     \hspace{2mm}
     \subfloat[SSIM.]{\includegraphics[height=4.6cm,valign=c]{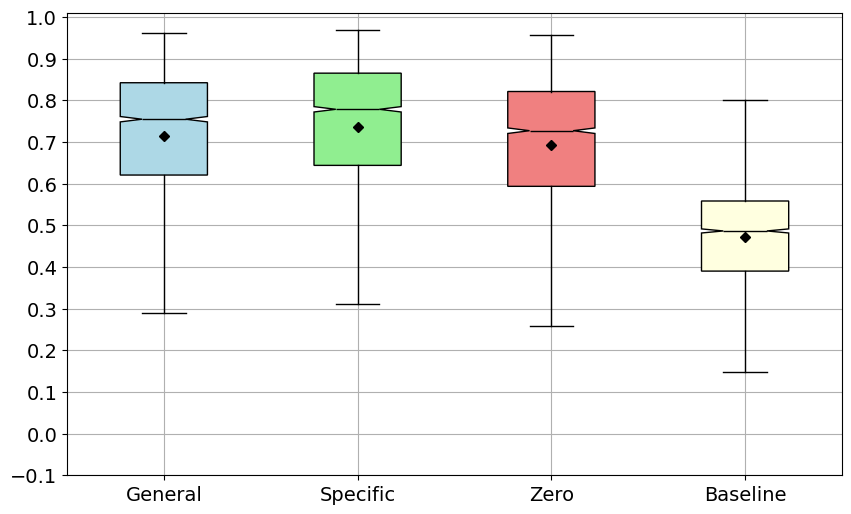}\label{fig:section_III_bp_ssim_d1}}
     
     \caption{RMSE and SSIM metrics box plots for the model predictions on the test set for domain \#1, (2 899 hourly samples from January 2024 to April 2024), for specific, generic, and zero models, and for baseline obtained from interpolating the predictor grid on the predictand grid using bi-linear interpolation. In each graph, the black dot and lines represent respectively the mean and the median, the upper and lower box limits indicate the first (Q1) and third (Q3) quartiles and the whiskers depict the highest (lowest) value within the $1.5 \times$ (Q3-Q1) above Q3 (below Q1).}\label{fig:results_III_d1}
\end{figure}

\begin{figure}[ht]
    \centering
    \subfloat[Pixel-wise MAE (m/s) for the different models and the baseline. The colorscale is capped at the max MAE of the Specific and Zero models. The baseline’s MAE exceeds the maximum value of the colorscale in certain locations, causing saturation. The value at the top of each image is the average over the domain.]{\includegraphics[width=0.99\textwidth,valign=c]{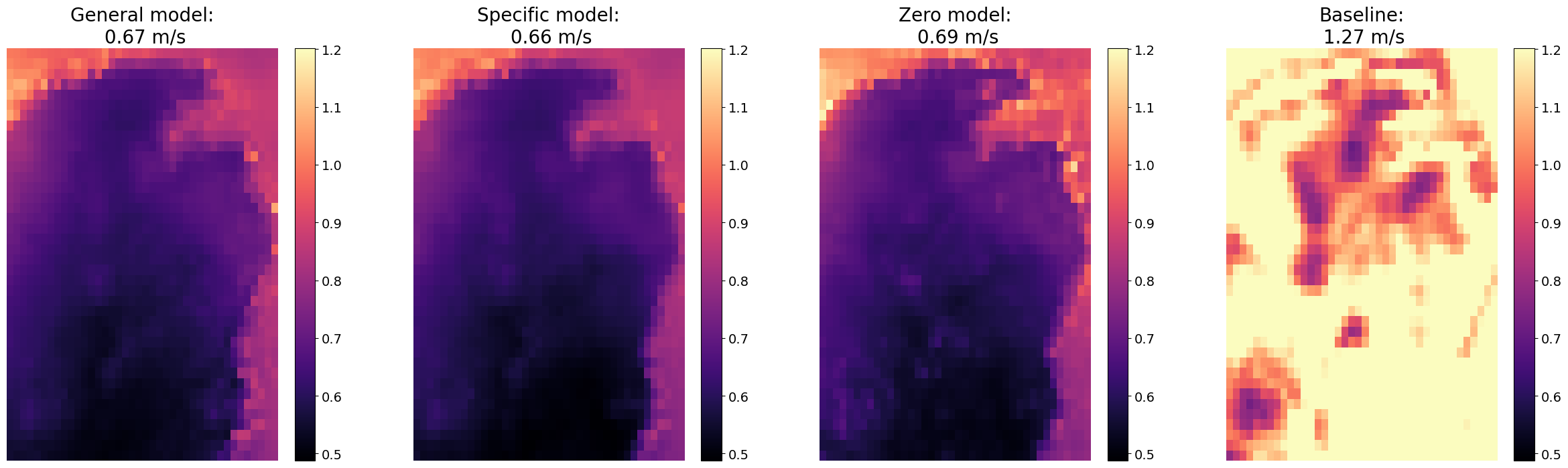}
     \label{fig:section_III_error_d1}}
     \\
     \subfloat[Average PSD.]{\includegraphics[width=0.48\textwidth,valign=c]{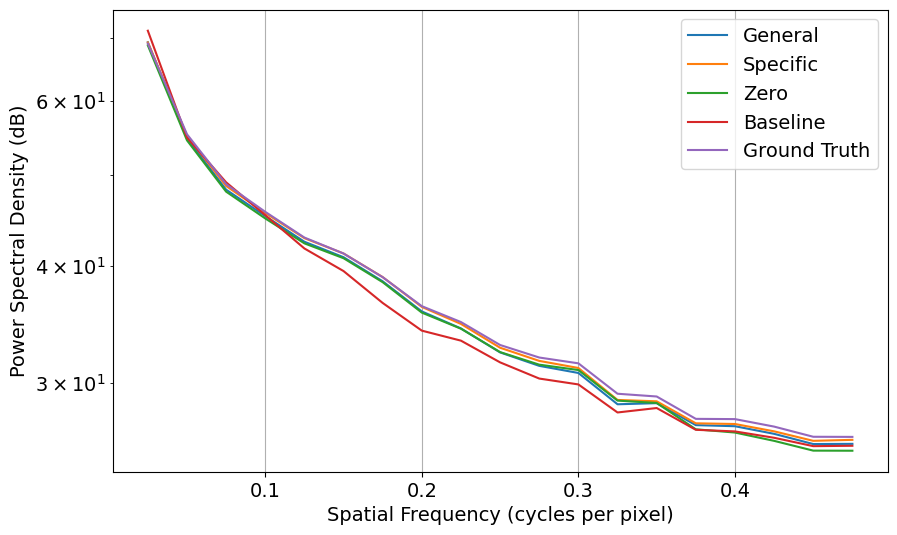}\label{fig:section_III_psd_d1}}
     \hspace{2mm}
     \subfloat[Average PDF. Note that the legend is the same as Figure \ref{fig:section_III_psd_d1}.]{\includegraphics[width=0.48\textwidth,valign=c]{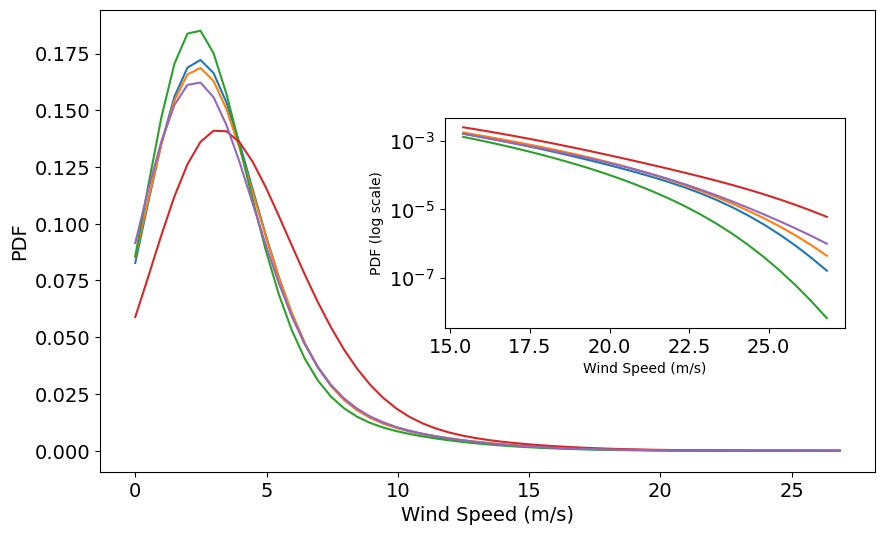}\label{fig:section_III_pdf_d1}}
     
     \caption{MAE, average PSD and average PDF graphs for the model predictions on the test set for domain \#1,(2 899 hourly samples from January 2024 to April 2024) for specific, general and zero models, and for baseline obtained from interpolating the predictor grid on the predictand grid using bi-linear interpolation. The ground truth in the PSD and PDF graphs refers to the HR $UV$, i.e., the predictand.}\label{fig:results_domain_III_d1}
\end{figure}

\begin{figure}[ht]
    \centering
     \hspace{2mm}
     \subfloat[RMSE.]{\includegraphics[height=4.6cm,valign=c]{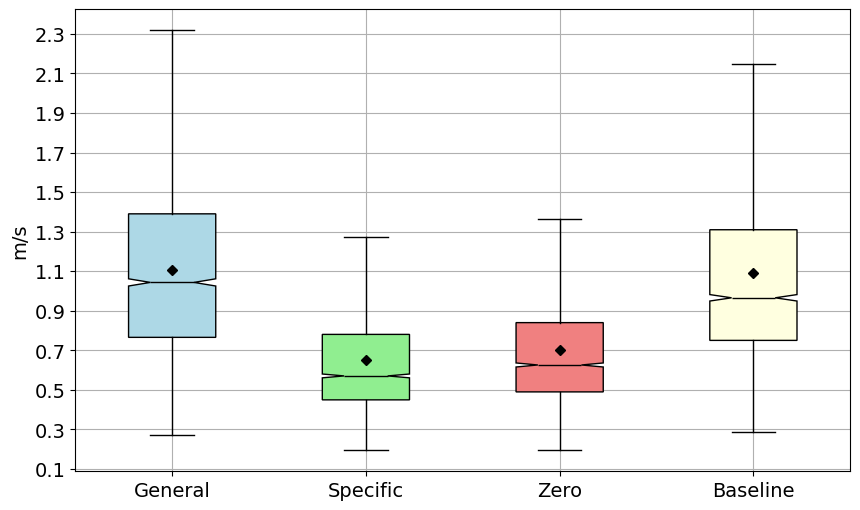}\label{fig:section_III_bp_rmse_d2}}
     \hspace{2mm}
     \subfloat[MAE.]{\includegraphics[height=4.6cm,valign=c]{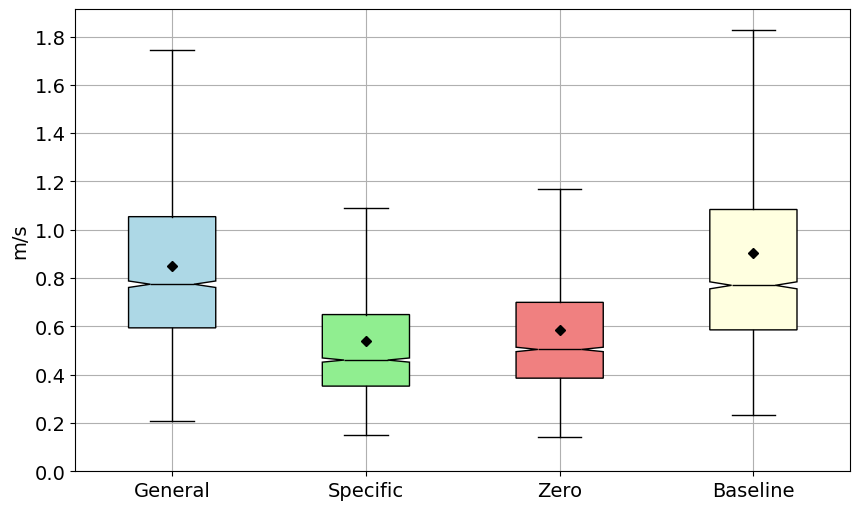}\label{fig:section_III_bp_mae_d2}}
     \hspace{2mm}
     \subfloat[SSIM.]{\includegraphics[height=4.6cm,valign=c]{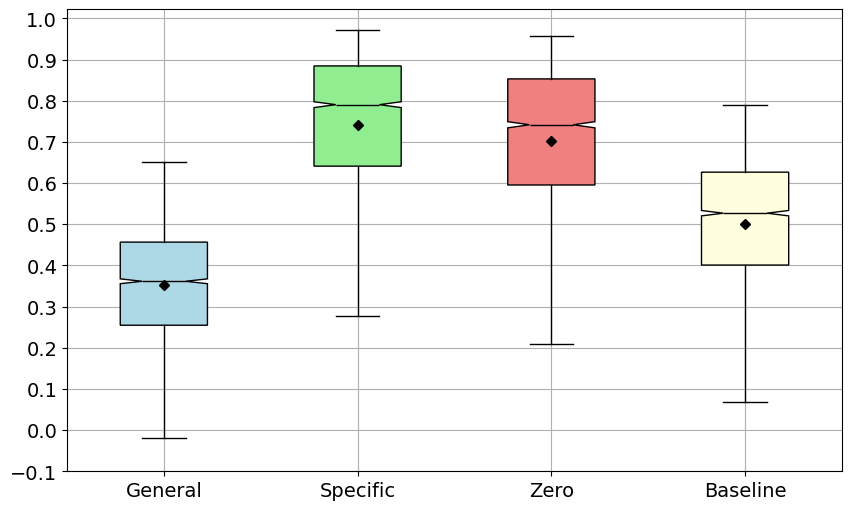}\label{fig:section_III_bp_ssim_d2}}
     
     \caption{RMSE and SSIM metrics box plots for the model predictions on the test set for domain \#2, (2 899 hourly samples from January 2024 to April 2024), for specific, generic, and zero models, and for baseline obtained from interpolating the predictor grid on the predictand grid using bi-linear interpolation. In each graph, the black dot and lines represent respectively the mean and the median, the upper and lower box limits indicate the first (Q1) and third (Q3) quartiles and the whiskers depict the highest (lowest) value within the $1.5 \times$ (Q3-Q1) above Q3 (below Q1).}\label{fig:results_III_d2}
\end{figure}

\begin{figure}[ht]
    \centering
    \subfloat[Pixel-wise MAE (m/s) for the different models and the baseline. The colorscale is capped at the max MAE of the Specific and Zero models. The baseline’s MAE exceeds the maximum value of the colorscale in certain locations, causing saturation. The value at the top of each image is the average over the domain.]{\includegraphics[width=0.99\textwidth,valign=c]{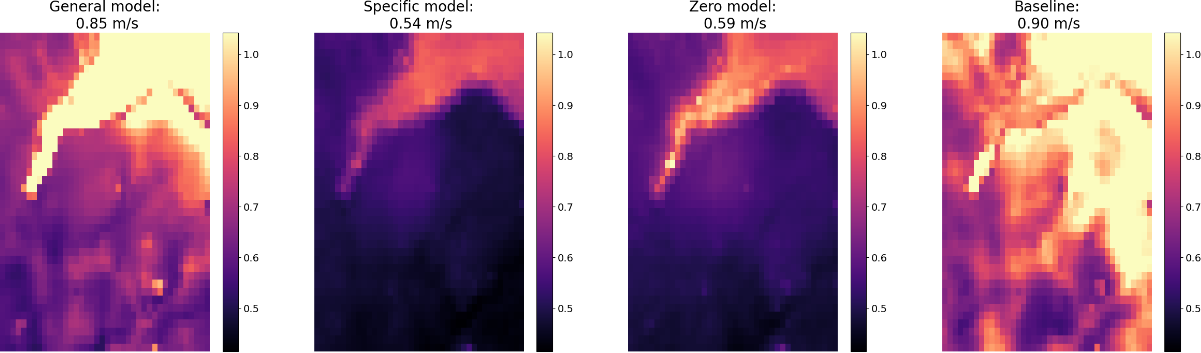}
     \label{fig:section_III_average_error}}
     \\
     \subfloat[Average PSD.]{\includegraphics[width=0.48\textwidth,valign=c]{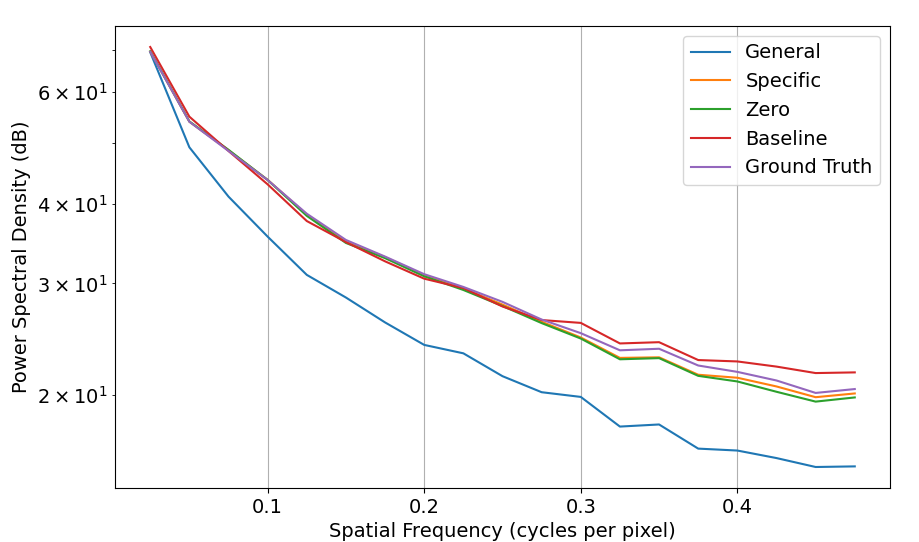}\label{fig:section_III_psd}}
     \hspace{2mm}
     \subfloat[Average PDF. Note that the legend is the same as Figure \ref{fig:section_III_psd}.]{\includegraphics[width=0.48\textwidth,valign=c]{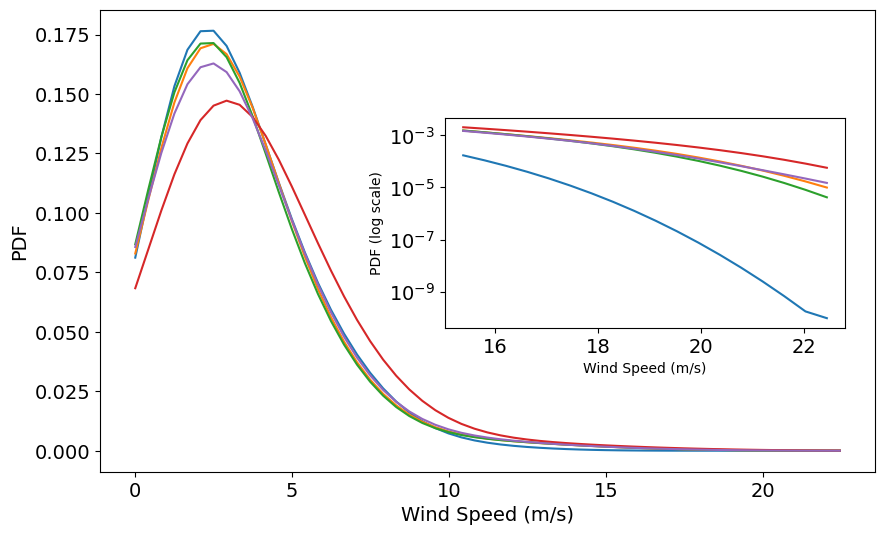}\label{fig:section_III_pdf}}
     
     \caption{MAE, average PSD and average PDF graphs for the model predictions on the test set for domain \#2,(2 899 hourly samples from January 2024 to April 2024) for specific, general and zero models, and for baseline obtained from interpolating the predictor grid on the predictand grid using bi-linear interpolation. The ground truth in the PSD and PDF graphs refers to the HR $UV$, i.e., the predictand.}\label{fig:results_domain_III}
\end{figure}

\begin{figure}[ht]
    \centering
     \hspace{2mm}
     \subfloat[RMSE.]{\includegraphics[height=4.6cm,valign=c]{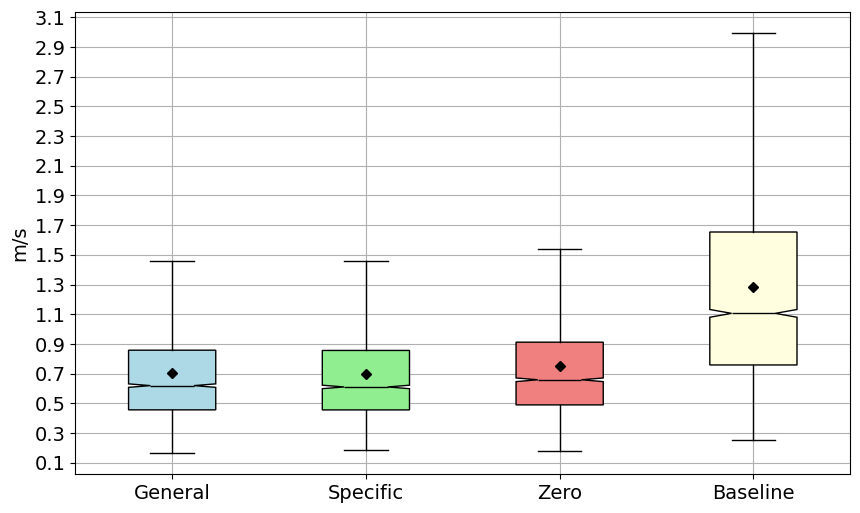}\label{fig:section_III_bp_rmse_d3}}
     \hspace{2mm}
     \subfloat[MAE.]{\includegraphics[height=4.6cm,valign=c]{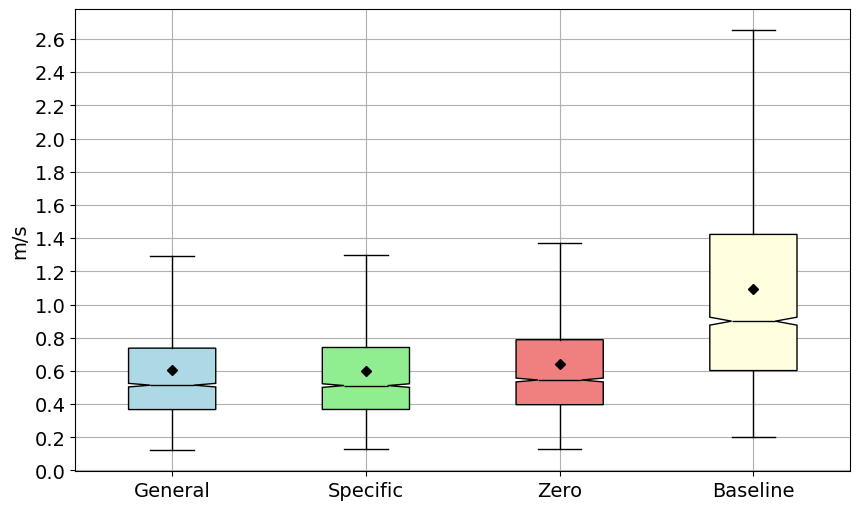}\label{fig:section_III_bp_mae_d3}}
     \hspace{2mm}
     \subfloat[SSIM.]{\includegraphics[height=4.6cm,valign=c]{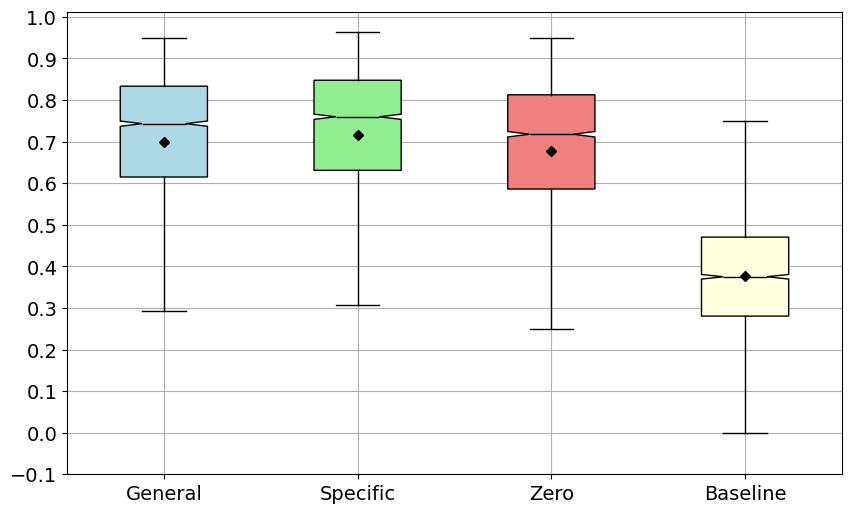}\label{fig:section_III_bp_ssim_d3}}
     
     \caption{RMSE and SSIM metrics box plots for the model predictions on the test set for domain \#3, (2 899 hourly samples from January 2024 to April 2024), for specific, generic, and zero models, and for baseline obtained from interpolating the predictor grid on the predictand grid using bi-linear interpolation. In each graph, the black dot and lines represent respectively the mean and the median, the upper and lower box limits indicate the first (Q1) and third (Q3) quartiles and the whiskers depict the highest (lowest) value within the $1.5 \times$ (Q3-Q1) above Q3 (below Q1).}\label{fig:results_III_d3}
\end{figure}

\begin{figure}[ht]
    \centering
    \subfloat[Pixel-wise MAE (m/s) for the different models and the baseline. The colorscale is capped at the max MAE of the Specific and Zero models. The baseline’s MAE exceeds the maximum value of the colorscale in certain locations, causing saturation. The value at the top of each image is the average over the domain.]{\includegraphics[width=0.99\textwidth,valign=c]{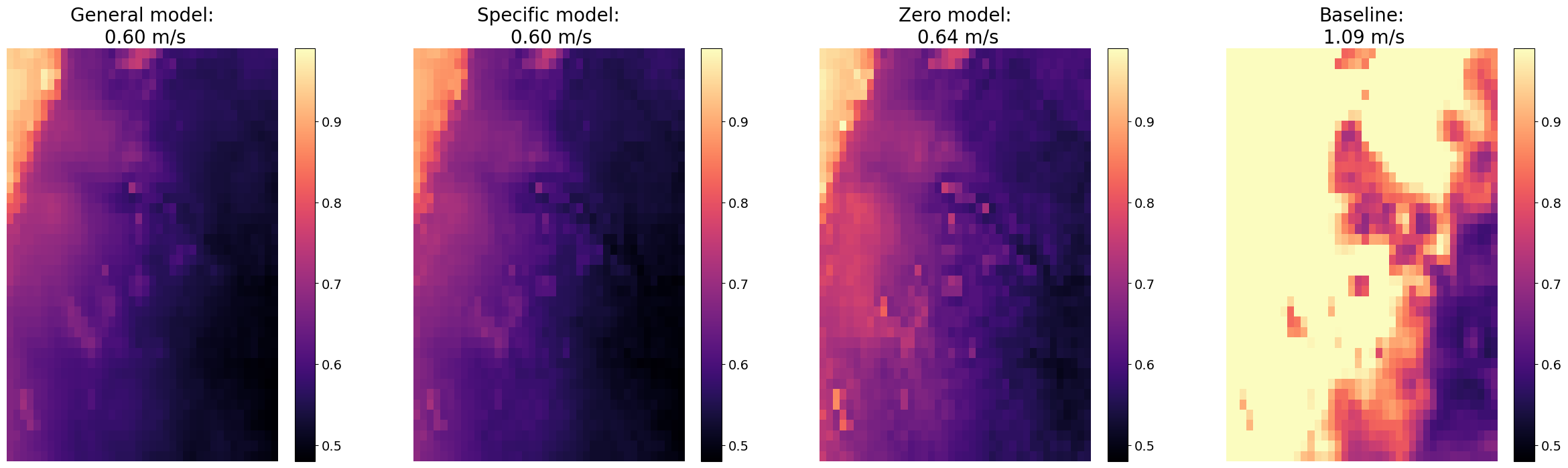}
     \label{fig:section_III_error_d3}}
     \\
     \subfloat[Average PSD.]{\includegraphics[width=0.48\textwidth,valign=c]{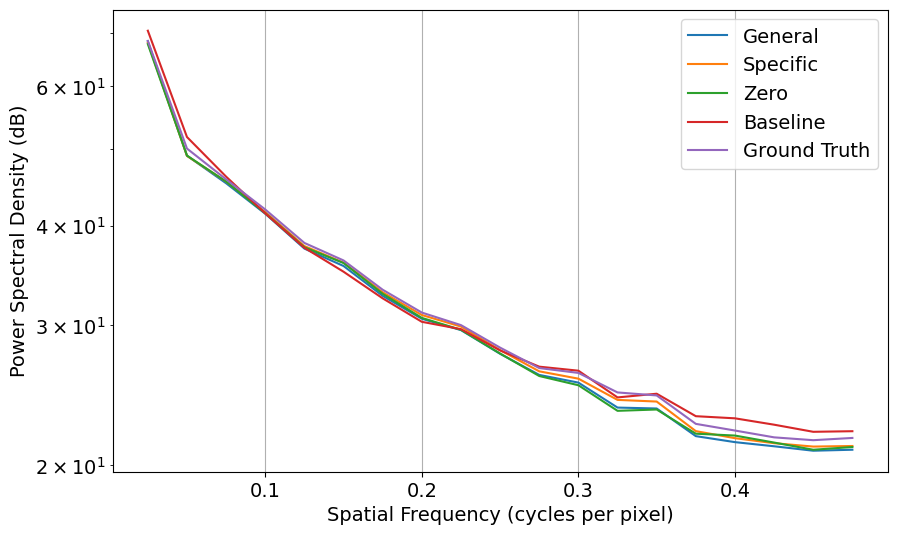}\label{fig:section_III_psd_d3}}
     \hspace{2mm}
     \subfloat[Average PDF. Note that the legend is the same as Figure \ref{fig:section_III_psd_d3}.]{\includegraphics[width=0.48\textwidth,valign=c]{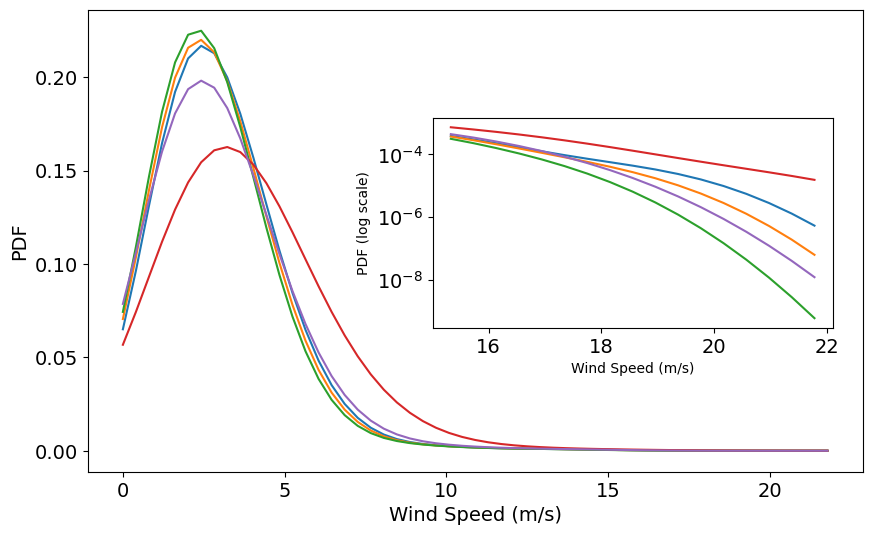}\label{fig:section_III_pdf_d3}}
     
     \caption{MAE, average PSD and average PDF graphs for the model predictions on the test set for domain \#3,(2 899 hourly samples from January 2024 to April 2024) for specific, general and zero models, and for baseline obtained from interpolating the predictor grid on the predictand grid using bi-linear interpolation. The ground truth in the PSD and PDF graphs refers to the HR $UV$, i.e., the predictand.}\label{fig:results_domain_III_d3}
\end{figure}

\begin{figure}[ht]
    \centering
     \hspace{2mm}
     \subfloat[RMSE.]{\includegraphics[height=4.6cm,valign=c]{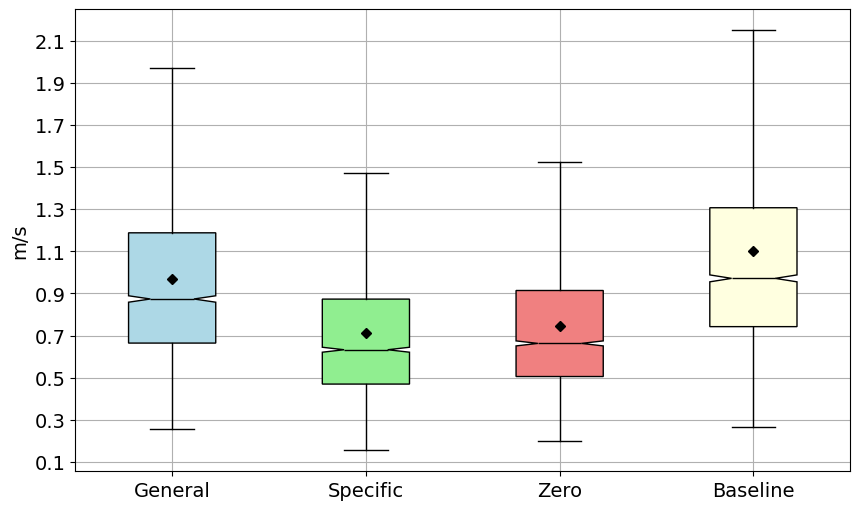}\label{fig:section_III_bp_rmse_d4}}
     \hspace{2mm}
     \subfloat[MAE.]{\includegraphics[height=4.6cm,valign=c]{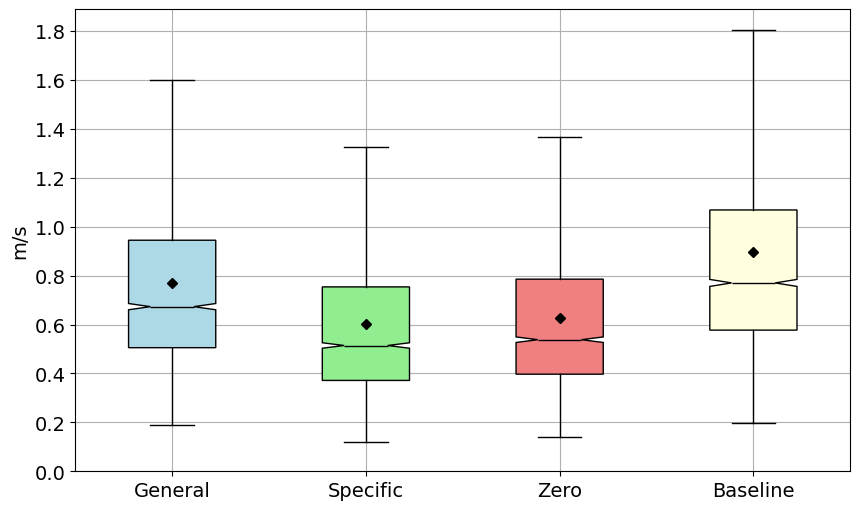}\label{fig:section_III_bp_mae_d4}}
     \hspace{2mm}
     \subfloat[SSIM.]{\includegraphics[height=4.6cm,valign=c]{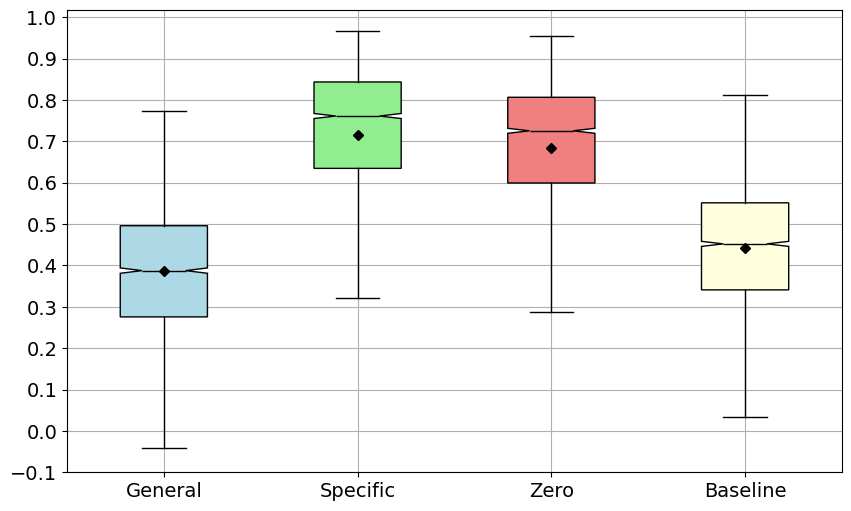}\label{fig:section_III_bp_ssim_d4}}
     
     \caption{RMSE and SSIM metrics box plots for the model predictions on the test set for domain \#4, (2 899 hourly samples from January 2024 to April 2024), for specific, generic, and zero models, and for baseline obtained from interpolating the predictor grid on the predictand grid using bi-linear interpolation. In each graph, the black dot and lines represent respectively the mean and the median, the upper and lower box limits indicate the first (Q1) and third (Q3) quartiles and the whiskers depict the highest (lowest) value within the $1.5 \times$ (Q3-Q1) above Q3 (below Q1).}\label{fig:results_III_d4}
\end{figure}

\begin{figure}[ht]
    \centering
    \subfloat[Pixel-wise MAE (m/s) for the different models and the baseline. The colorscale is capped at the max MAE of the Specific and Zero models. The baseline’s MAE exceeds the maximum value of the colorscale in certain locations, causing saturation. The value at the top of each image is the average over the domain.]{\includegraphics[width=0.99\textwidth,valign=c]{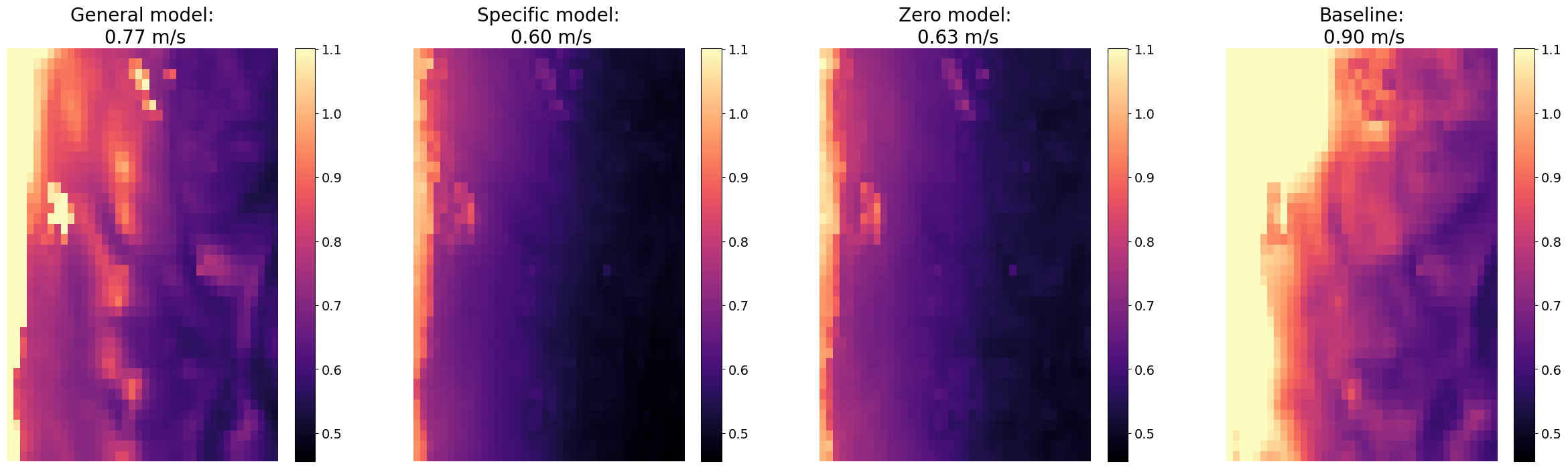}
     \label{fig:section_III_error_d4}}
     \\
     \subfloat[Average PSD.]{\includegraphics[width=0.48\textwidth,valign=c]{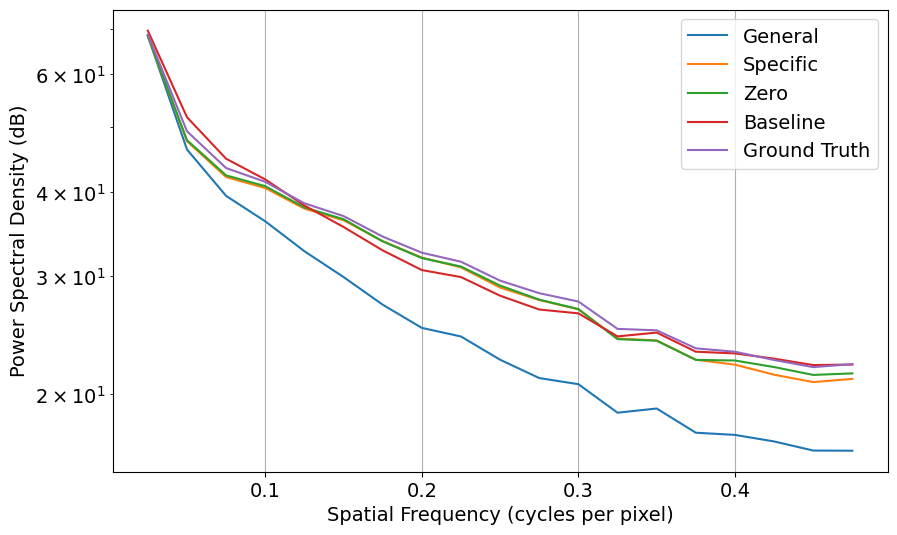}\label{fig:section_III_psd_d4}}
     \hspace{2mm}
     \subfloat[Average PDF. Note that the legend is the same as Figure \ref{fig:section_III_psd_d4}.]{\includegraphics[width=0.48\textwidth,valign=c]{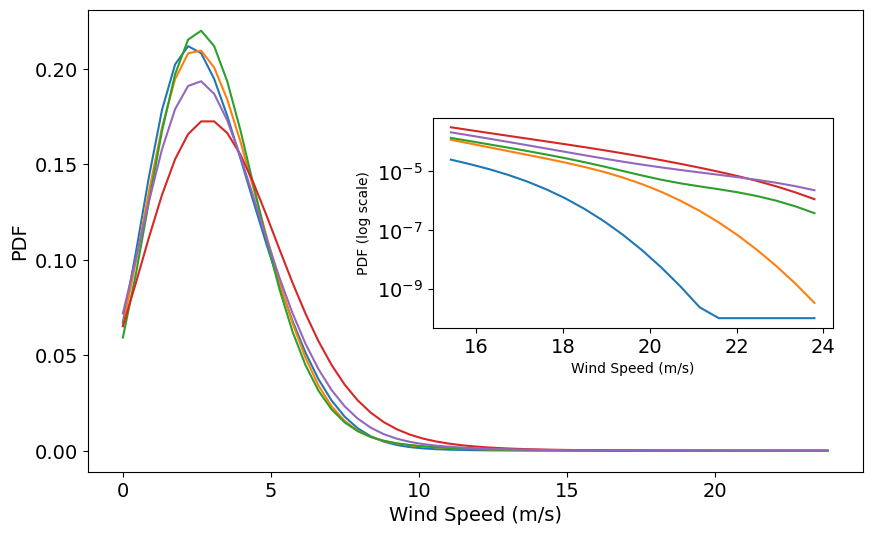}\label{fig:section_III_pdf_d4}}
     
     \caption{MAE, average PSD and average PDF graphs for the model predictions on the test set for domain \#4,(2 899 hourly samples from January 2024 to April 2024) for specific, general and zero models, and for baseline obtained from interpolating the predictor grid on the predictand grid using bi-linear interpolation. The ground truth in the PSD and PDF graphs refers to the HR $UV$, i.e., the predictand.}\label{fig:results_domain_III_d4}
\end{figure}

\begin{figure}[ht]
    \centering
     \hspace{2mm}
     \subfloat[RMSE.]{\includegraphics[height=4.6cm,valign=c]{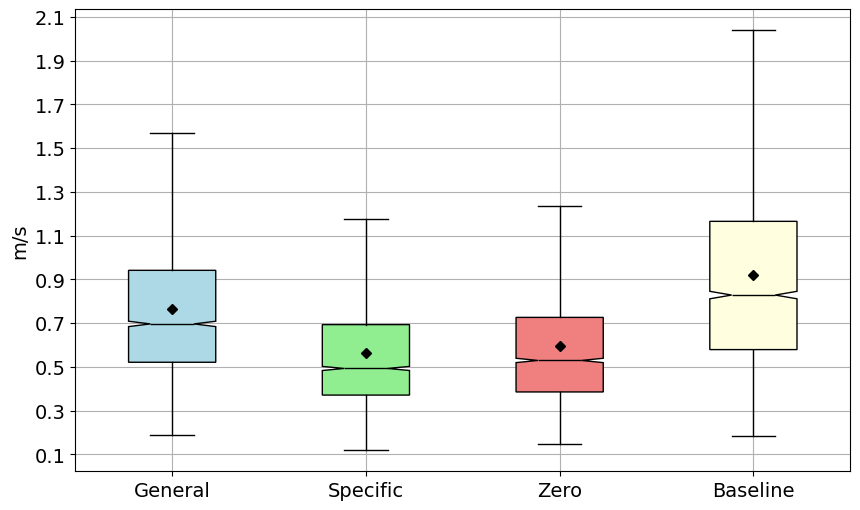}\label{fig:section_III_bp_rmse_d5}}
     \hspace{2mm}
     \subfloat[MAE.]{\includegraphics[height=4.6cm,valign=c]{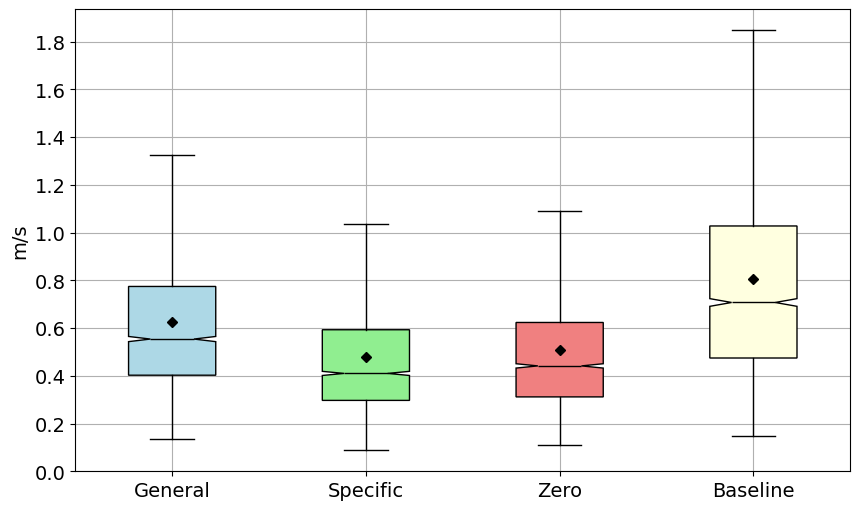}\label{fig:section_III_bp_mae_d5}}
     \hspace{2mm}
     \subfloat[SSIM.]{\includegraphics[height=4.6cm,valign=c]{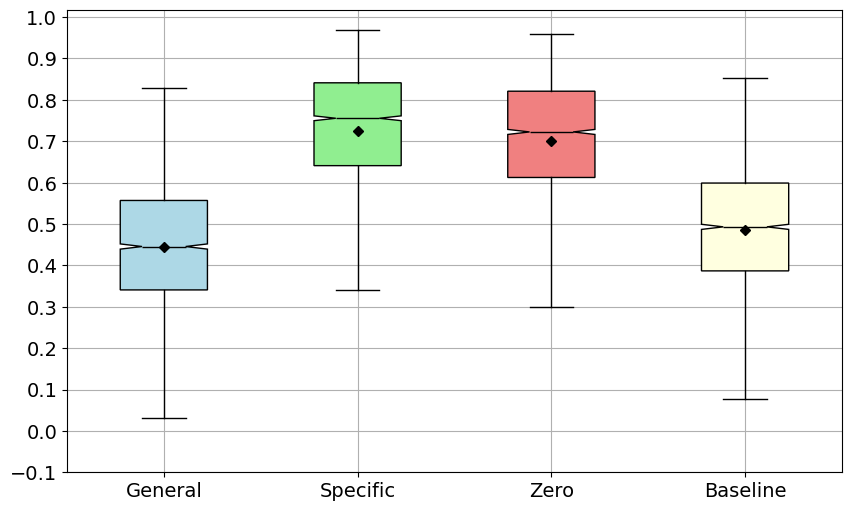}\label{fig:section_III_bp_ssim_d5}}
     
     \caption{RMSE and SSIM metrics box plots for the model predictions on the test set for domain \#5, (2 899 hourly samples from January 2024 to April 2024), for specific, generic, and zero models, and for baseline obtained from interpolating the predictor grid on the predictand grid using bi-linear interpolation. In each graph, the black dot and lines represent respectively the mean and the median, the upper and lower box limits indicate the first (Q1) and third (Q3) quartiles and the whiskers depict the highest (lowest) value within the $1.5 \times$ (Q3-Q1) above Q3 (below Q1).}\label{fig:results_III_d5}
\end{figure}

\begin{figure}[ht]
    \centering
    \subfloat[Pixel-wise MAE (m/s) for the different models and the baseline. The colorscale is capped at the max MAE of the Specific and Zero models. The baseline’s MAE exceeds the maximum value of the colorscale in certain locations, causing saturation. The value at the top of each image is the average over the domain.]{\includegraphics[width=0.99\textwidth,valign=c]{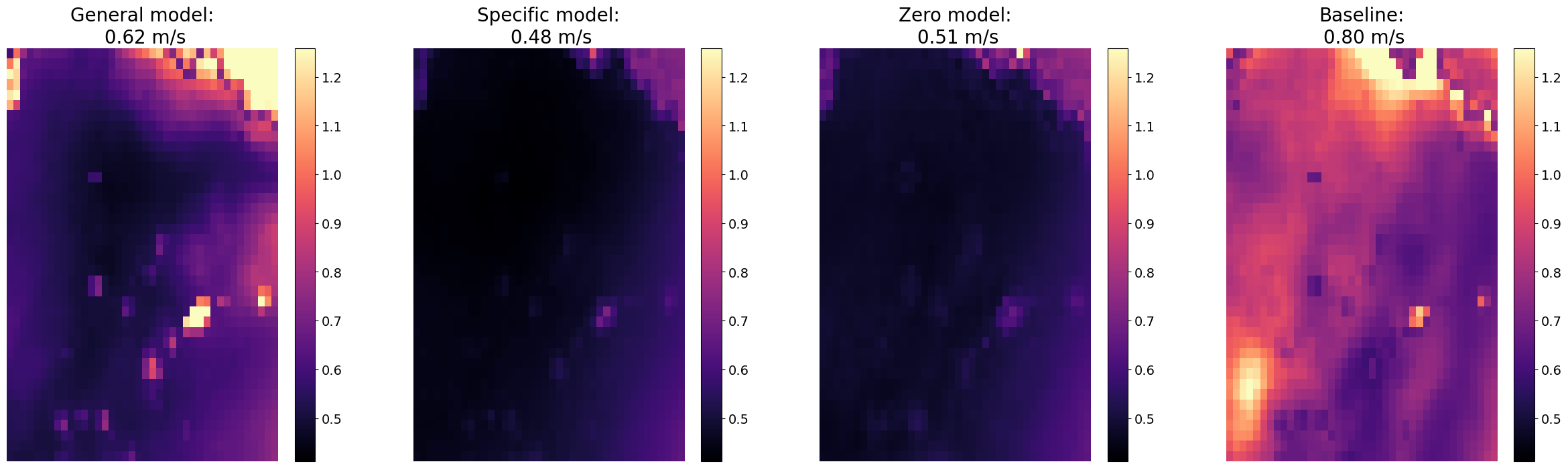}
     \label{fig:section_III_error_d5}}
     \\
     \subfloat[Average PSD.]{\includegraphics[width=0.48\textwidth,valign=c]{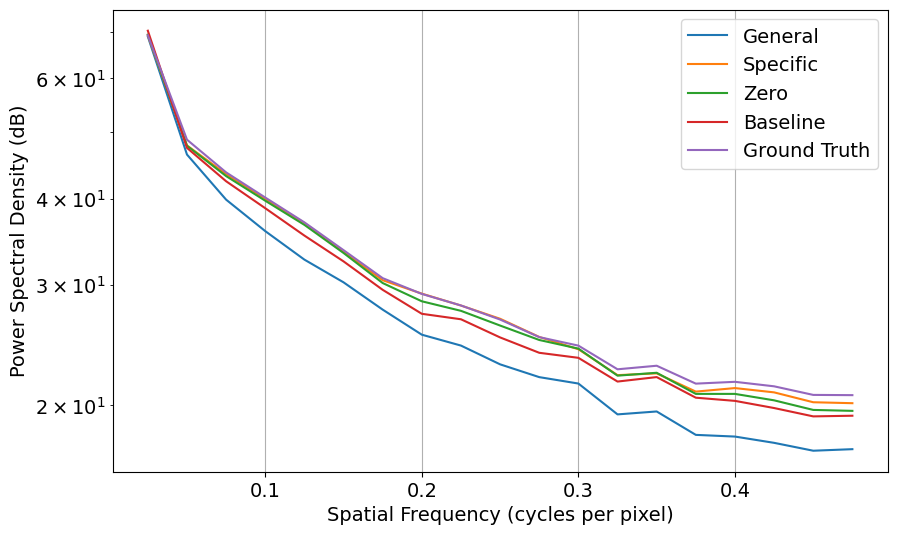}\label{fig:section_III_psd_d5}}
     \hspace{2mm}
     \subfloat[Average PDF. Note that the legend is the same as Figure \ref{fig:section_III_psd_d5}.]{\includegraphics[width=0.48\textwidth,valign=c]{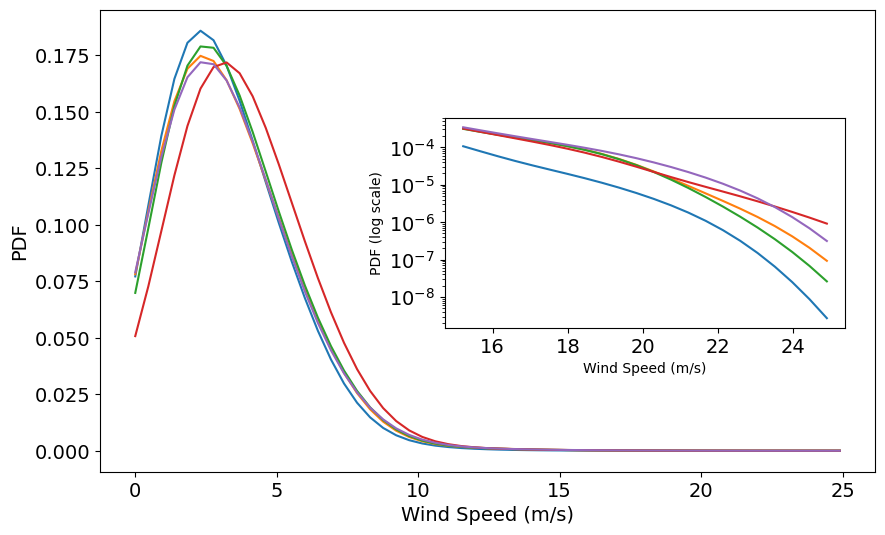}\label{fig:section_III_pdf_d5}}
     
     \caption{MAE, average PSD and average PDF graphs for the model predictions on the test set for domain \#5,(2 899 hourly samples from January 2024 to April 2024) for specific, general and zero models, and for baseline obtained from interpolating the predictor grid on the predictand grid using bi-linear interpolation. The ground truth in the PSD and PDF graphs refers to the HR $UV$, i.e., the predictand.}\label{fig:results_domain_III_d5}
\end{figure}

\begin{figure}[ht]
    \centering
     \hspace{2mm}
     \subfloat[RMSE.]{\includegraphics[height=4.6cm,valign=c]{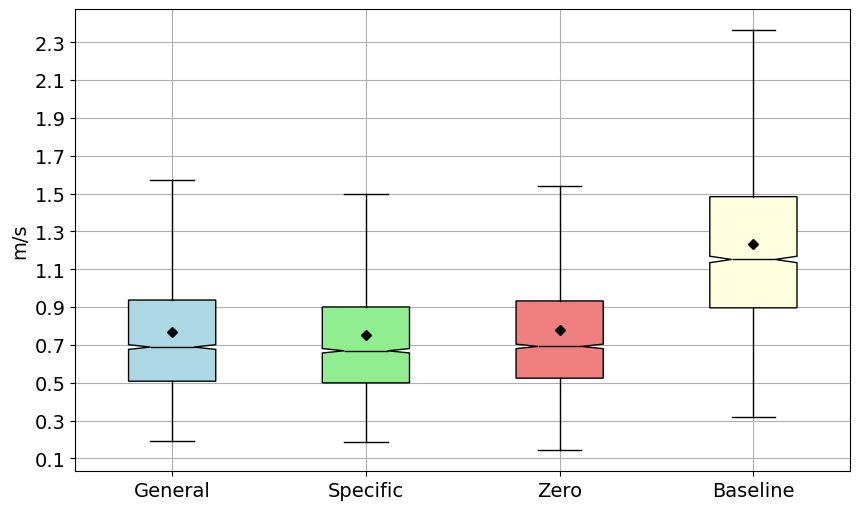}\label{fig:section_III_bp_rmse_d6}}
     \hspace{2mm}
     \subfloat[MAE.]{\includegraphics[height=4.6cm,valign=c]{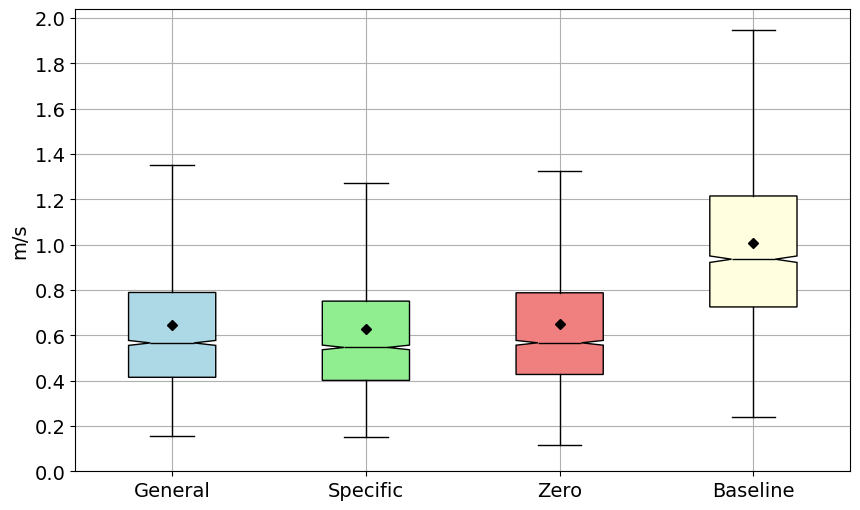}\label{fig:section_III_bp_mae_d6}}
     \hspace{2mm}
     \subfloat[SSIM.]{\includegraphics[height=4.6cm,valign=c]{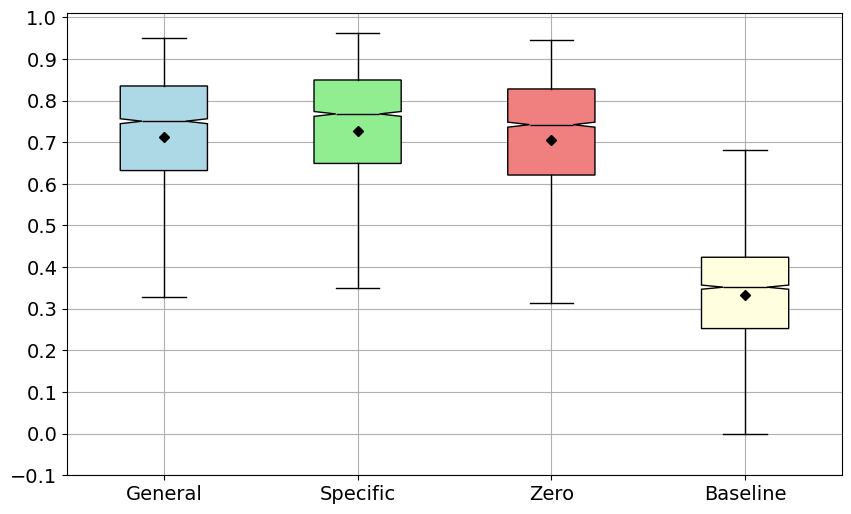}\label{fig:section_III_bp_ssim_d6}}
     
     \caption{RMSE and SSIM metrics box plots for the model predictions on the test set for domain \#6, (2 899 hourly samples from January 2024 to April 2024), for specific, generic, and zero models, and for baseline obtained from interpolating the predictor grid on the predictand grid using bi-linear interpolation. In each graph, the black dot and lines represent respectively the mean and the median, the upper and lower box limits indicate the first (Q1) and third (Q3) quartiles and the whiskers depict the highest (lowest) value within the $1.5 \times$ (Q3-Q1) above Q3 (below Q1).}\label{fig:results_III_d6}
\end{figure}

\begin{figure}[ht]
    \centering
    \subfloat[Pixel-wise MAE (m/s) for the different models and the baseline. The colorscale is capped at the max MAE of the Specific and Zero models. The baseline’s MAE exceeds the maximum value of the colorscale in certain locations, causing saturation. The value at the top of each image is the average over the domain.]{\includegraphics[width=0.99\textwidth,valign=c]{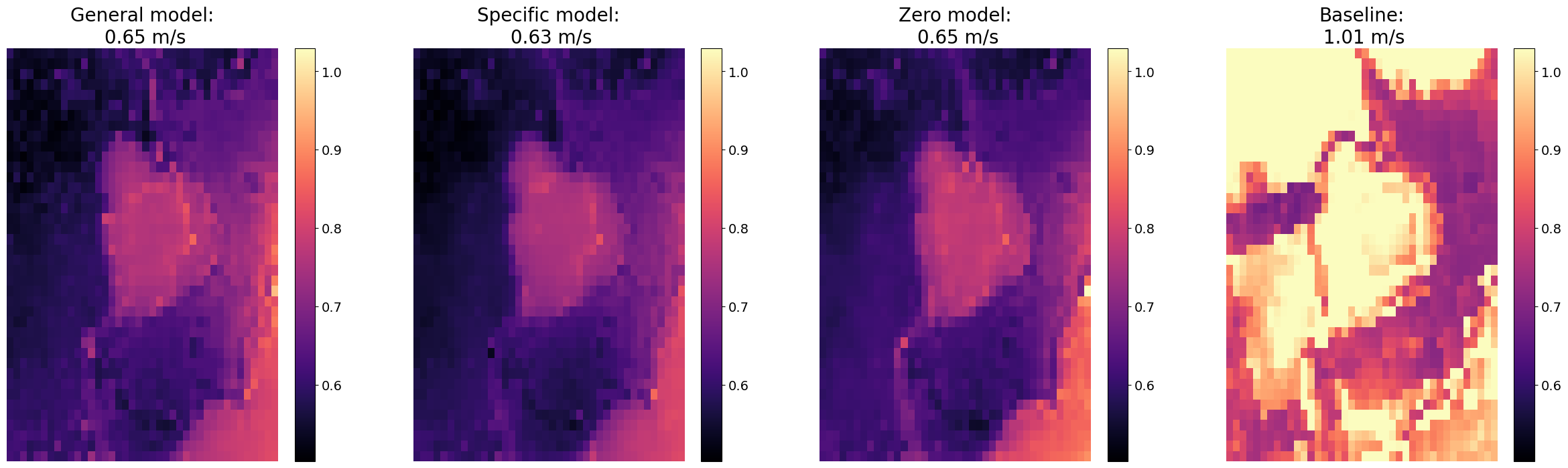}
     \label{fig:section_III_error_d6}}
     \\
     \subfloat[Average PSD.]{\includegraphics[width=0.48\textwidth,valign=c]{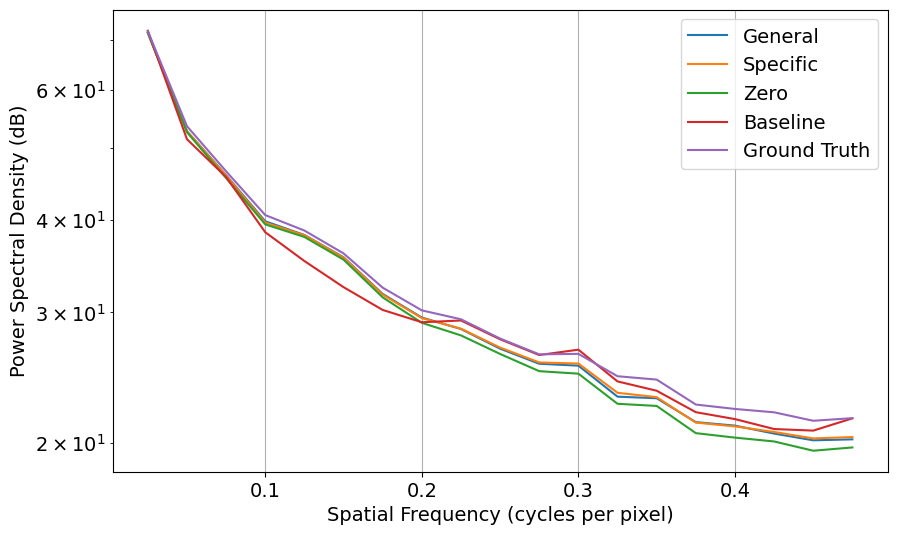}\label{fig:section_III_psd_d6}}
     \hspace{2mm}
     \subfloat[Average PDF. Note that the legend is the same as Figure \ref{fig:section_III_psd_d6}.]{\includegraphics[width=0.48\textwidth,valign=c]{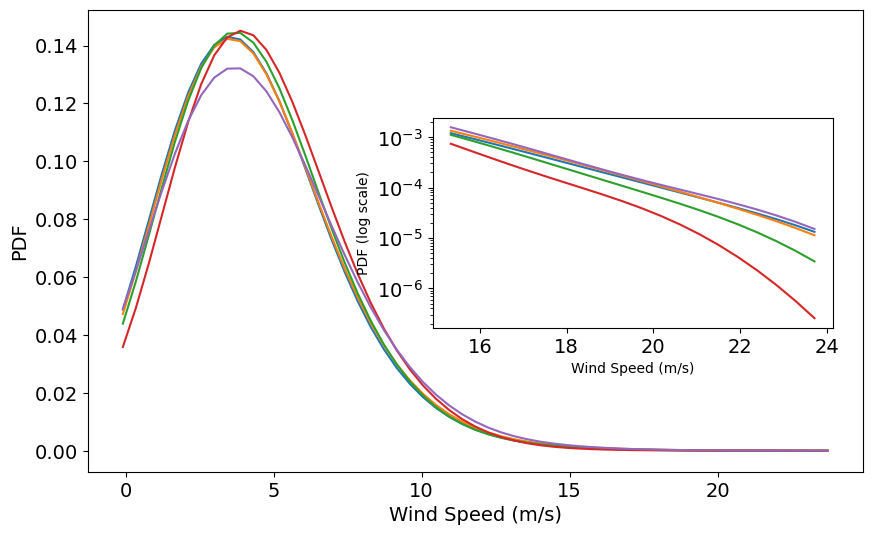}\label{fig:section_III_pdf_d6}}
     
     \caption{MAE, average PSD and average PDF graphs for the model predictions on the test set for domain \#6,(2 899 hourly samples from January 2024 to April 2024) for specific, general and zero models, and for baseline obtained from interpolating the predictor grid on the predictand grid using bi-linear interpolation. The ground truth in the PSD and PDF graphs refers to the HR $UV$, i.e., the predictand.}\label{fig:results_domain_III_d6}
\end{figure}

\begin{figure}[ht]
    \centering
     \hspace{2mm}
     \subfloat[RMSE.]{\includegraphics[height=4.6cm,valign=c]{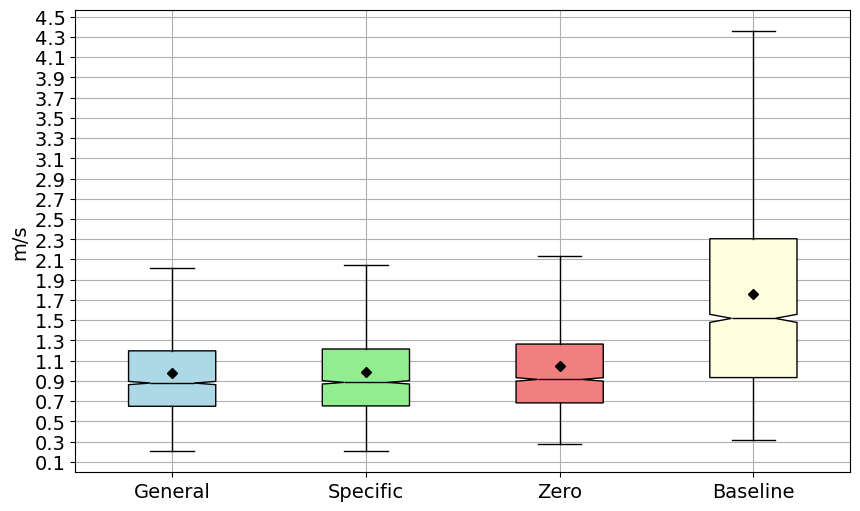}\label{fig:section_III_bp_rmse_d7}}
     \hspace{2mm}
     \subfloat[MAE.]{\includegraphics[height=4.6cm,valign=c]{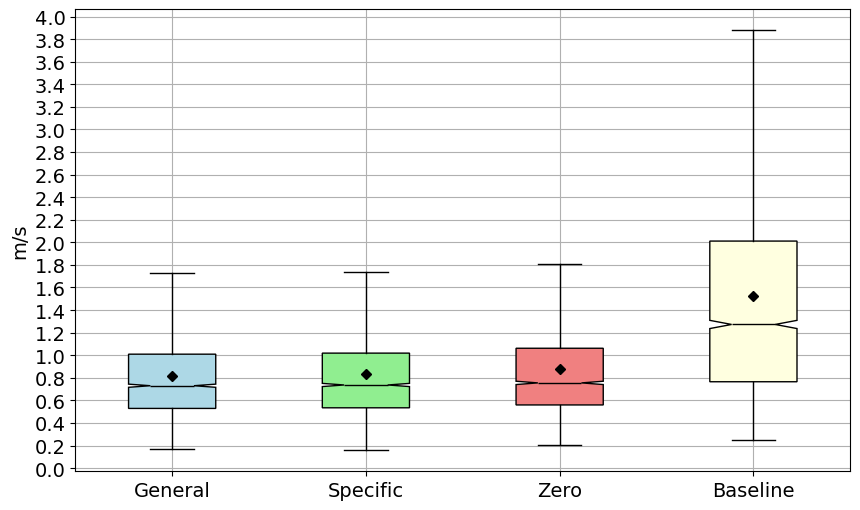}\label{fig:section_III_bp_mae_d7}}
     \hspace{2mm}
     \subfloat[SSIM.]{\includegraphics[height=4.6cm,valign=c]{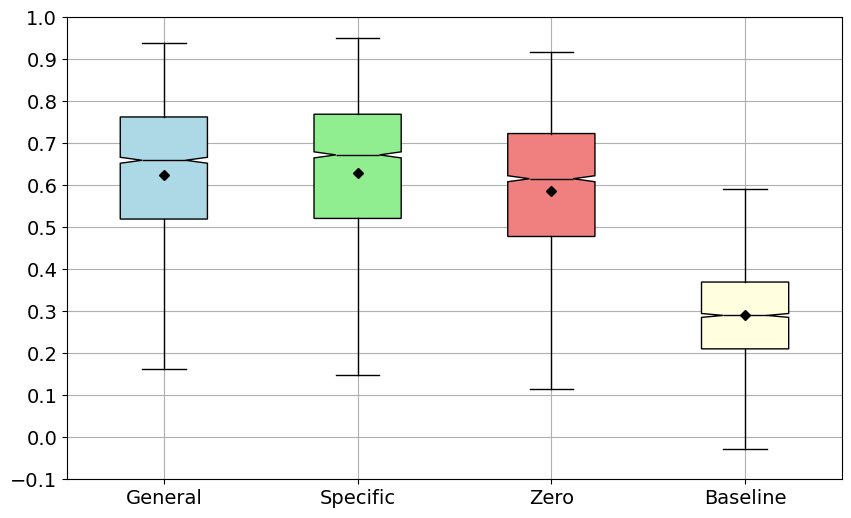}\label{fig:section_III_bp_ssim_d7}}
     
     \caption{RMSE and SSIM metrics box plots for the model predictions on the test set for domain \#7, (2 899 hourly samples from January 2024 to April 2024), for specific, generic, and zero models, and for baseline obtained from interpolating the predictor grid on the predictand grid using bi-linear interpolation. In each graph, the black dot and lines represent respectively the mean and the median, the upper and lower box limits indicate the first (Q1) and third (Q3) quartiles and the whiskers depict the highest (lowest) value within the $1.5 \times$ (Q3-Q1) above Q3 (below Q1).}\label{fig:results_III_d7}
\end{figure}

\begin{figure}[ht]
    \centering
    \subfloat[Pixel-wise MAE (m/s) for the different models and the baseline. The colorscale is capped at the max MAE of the Specific and Zero models. The baseline’s MAE exceeds the maximum value of the colorscale in certain locations, causing saturation. The value at the top of each image is the average over the domain.]{\includegraphics[width=0.99\textwidth,valign=c]{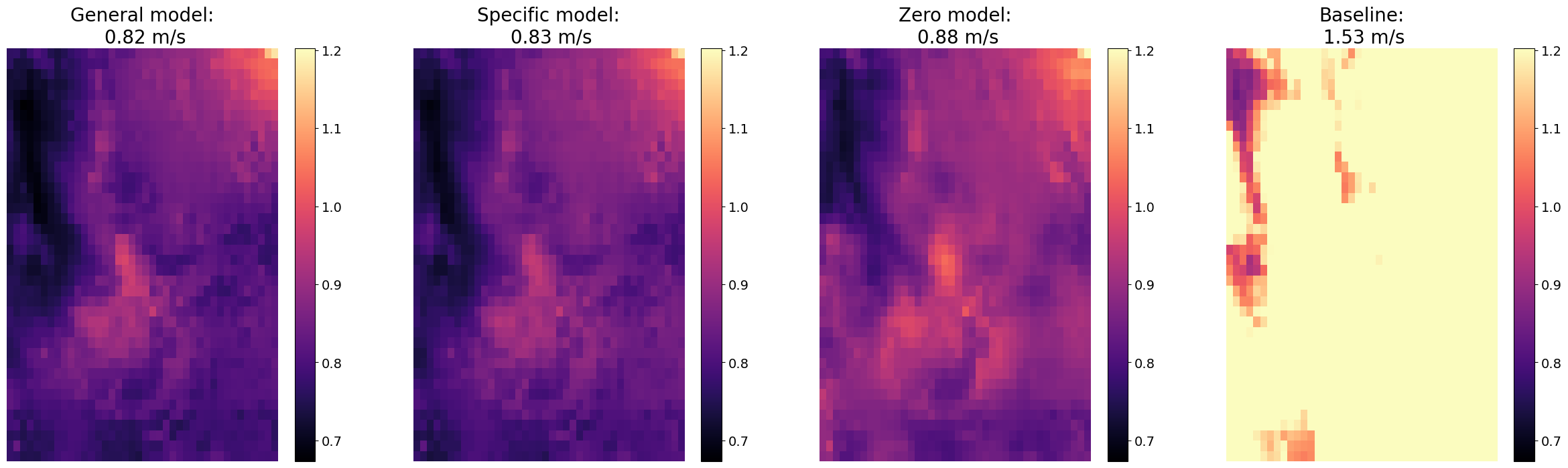}
     \label{fig:section_III_error_d7}}
     \\
     \subfloat[Average PSD.]{\includegraphics[width=0.48\textwidth,valign=c]{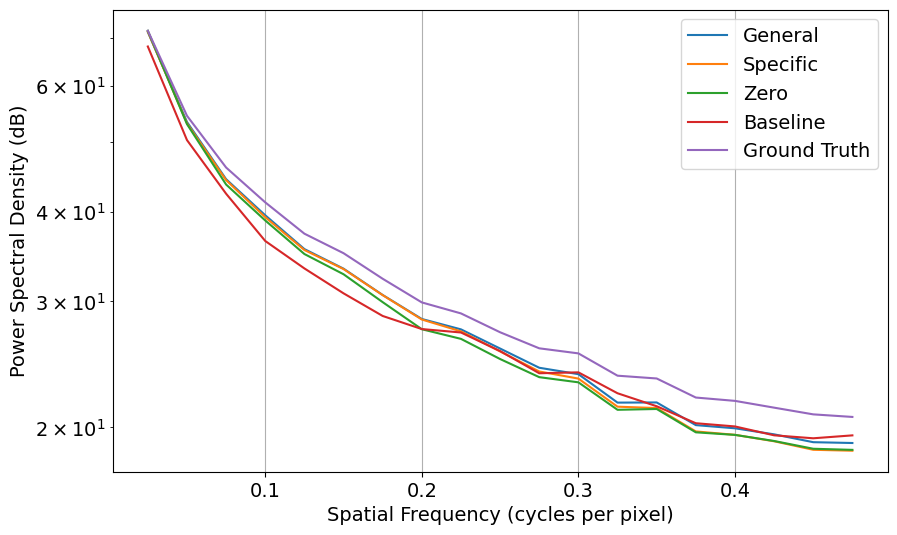}\label{fig:section_III_psd_d7}}
     \hspace{2mm}
     \subfloat[Average PDF. Note that the legend is the same as Figure \ref{fig:section_III_psd_d7}.]{\includegraphics[width=0.48\textwidth,valign=c]{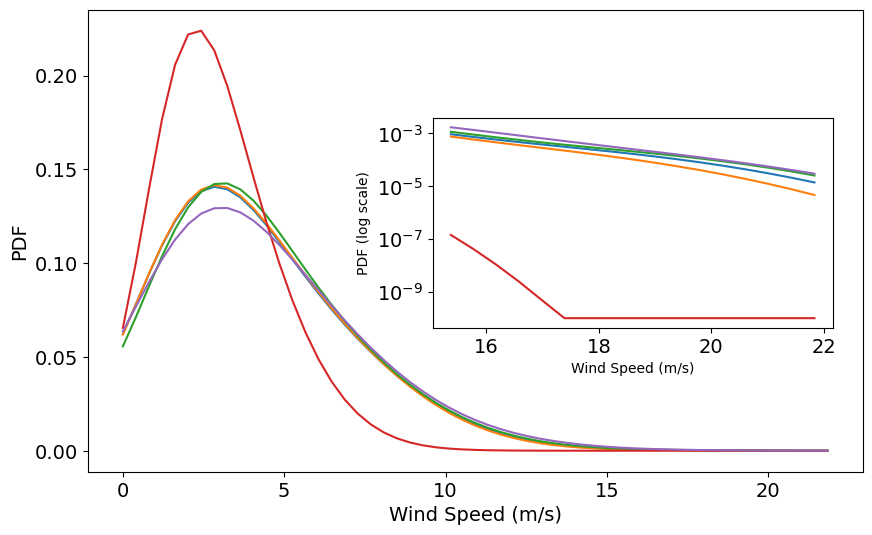}\label{fig:section_III_pdf_d7}}
     
     \caption{MAE, average PSD and average PDF graphs for the model predictions on the test set for domain \#7,(2 899 hourly samples from January 2024 to April 2024) for specific, general and zero models, and for baseline obtained from interpolating the predictor grid on the predictand grid using bi-linear interpolation. The ground truth in the PSD and PDF graphs refers to the HR $UV$, i.e., the predictand.}\label{fig:results_domain_III_d7}
\end{figure}

\begin{figure}[ht]
    \centering
     \hspace{2mm}
     \subfloat[RMSE.]{\includegraphics[height=4.6cm,valign=c]{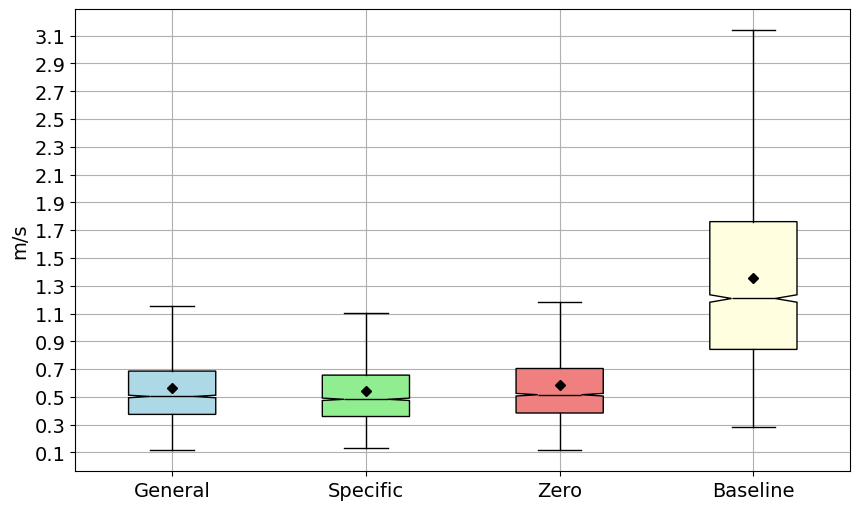}\label{fig:section_III_bp_rmse_d10}}
     \hspace{2mm}
     \subfloat[MAE.]{\includegraphics[height=4.6cm,valign=c]{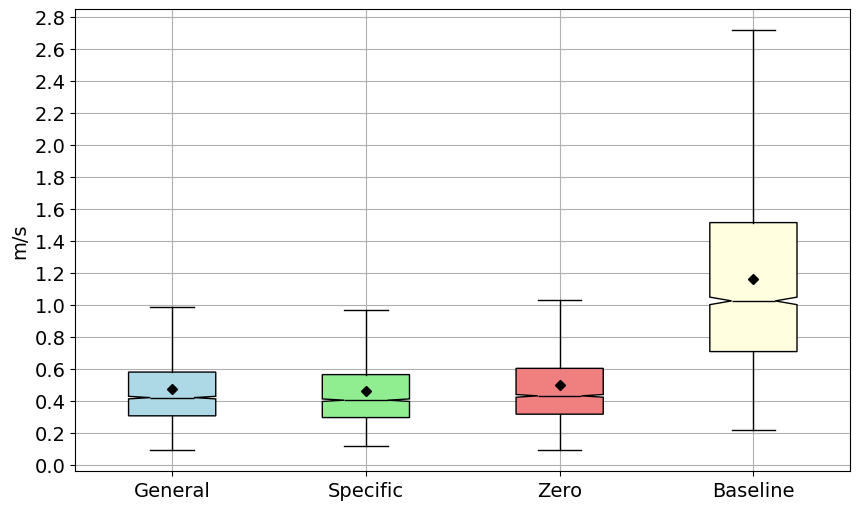}\label{fig:section_III_bp_mae_d10}}
     \hspace{2mm}
     \subfloat[SSIM.]{\includegraphics[height=4.6cm,valign=c]{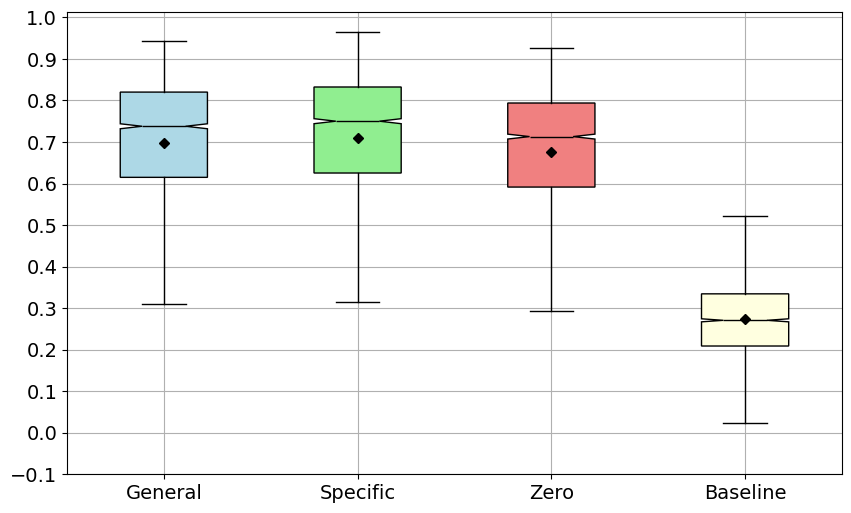}\label{fig:section_III_bp_ssim_d10}}
     
     \caption{RMSE and SSIM metrics box plots for the model predictions on the test set for domain \#8, (2 899 hourly samples from January 2024 to April 2024), for specific, generic, and zero models, and for baseline obtained from interpolating the predictor grid on the predictand grid using bi-linear interpolation. In each graph, the black dot and lines represent respectively the mean and the median, the upper and lower box limits indicate the first (Q1) and third (Q3) quartiles and the whiskers depict the highest (lowest) value within the $1.5 \times$ (Q3-Q1) above Q3 (below Q1).}\label{fig:results_III_d10}
\end{figure}

\begin{figure}[ht]
    \centering
    \subfloat[Pixel-wise MAE (m/s) for the different models and the baseline. The colorscale is capped at the max MAE of the Specific and Zero models. The baseline’s MAE exceeds the maximum value of the colorscale in certain locations, causing saturation. The value at the top of each image is the average over the domain.]{\includegraphics[width=0.99\textwidth,valign=c]{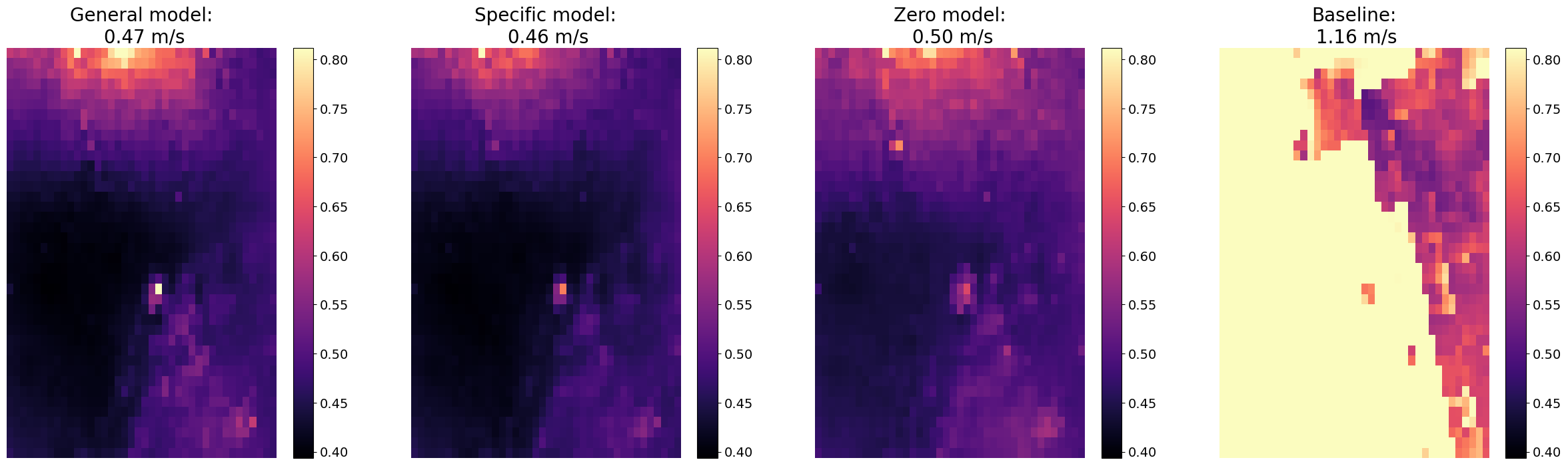}
     \label{fig:section_III_error_d10}}
     \\
     \subfloat[Average PSD.]{\includegraphics[width=0.48\textwidth,valign=c]{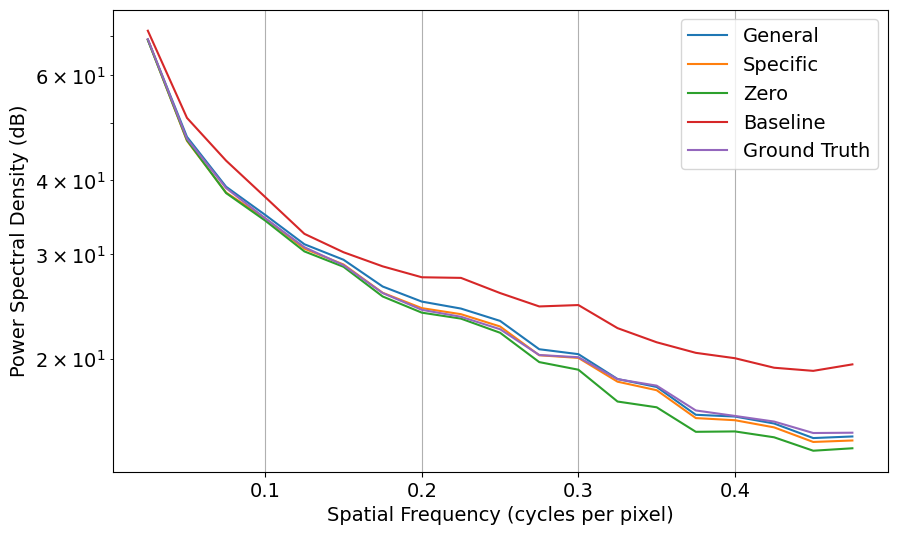}\label{fig:section_III_psd_d10}}
     \hspace{2mm}
     \subfloat[Average PDF. Note that the legend is the same as Figure \ref{fig:section_III_psd_d10}.]{\includegraphics[width=0.48\textwidth,valign=c]{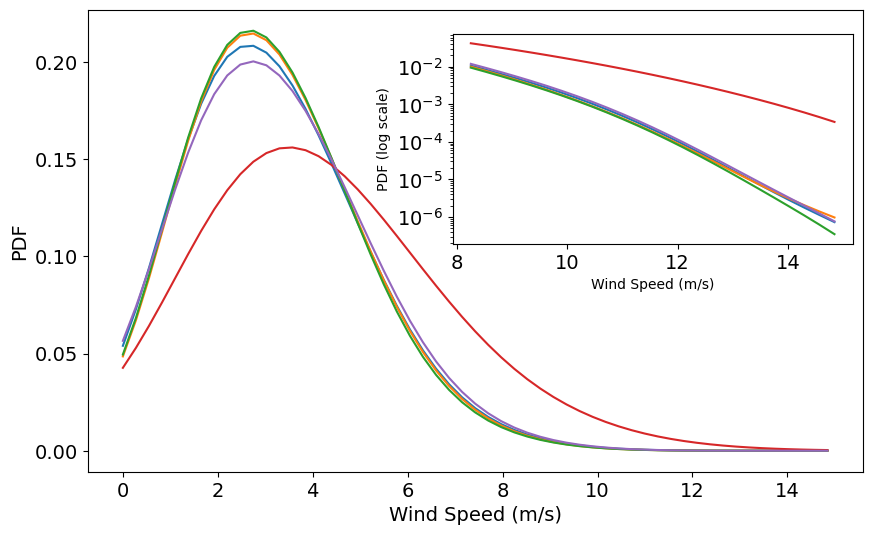}\label{fig:section_III_pdf_d10}}
     
     \caption{MAE, average PSD and average PDF graphs for the model predictions on the test set for domain \#8,(2 899 hourly samples from January 2024 to April 2024) for specific, general and zero models, and for baseline obtained from interpolating the predictor grid on the predictand grid using bi-linear interpolation. The ground truth in the PSD and PDF graphs refers to the HR $UV$, i.e., the predictand.}\label{fig:results_domain_III_d10}
\end{figure}

\begin{figure}[ht]
    \centering
     \hspace{2mm}
     \subfloat[RMSE.]{\includegraphics[height=4.6cm,valign=c]{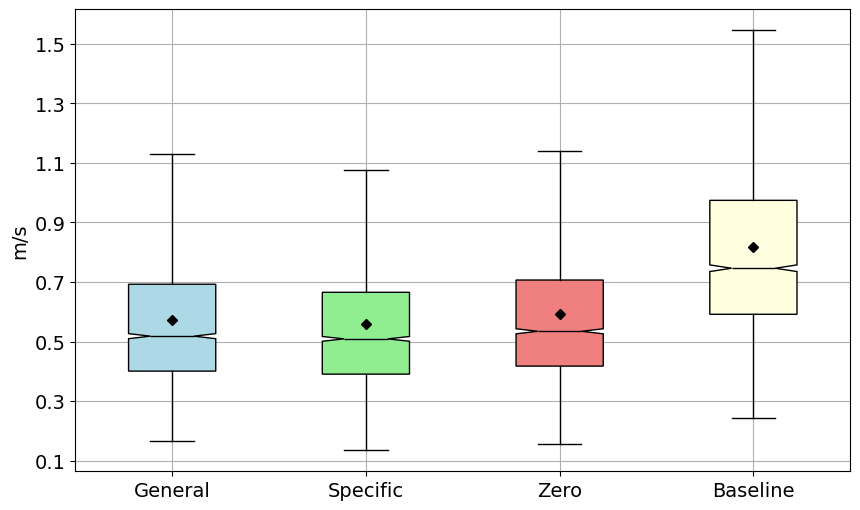}\label{fig:section_III_bp_rmse_d11}}
     \hspace{2mm}
     \subfloat[MAE.]{\includegraphics[height=4.6cm,valign=c]{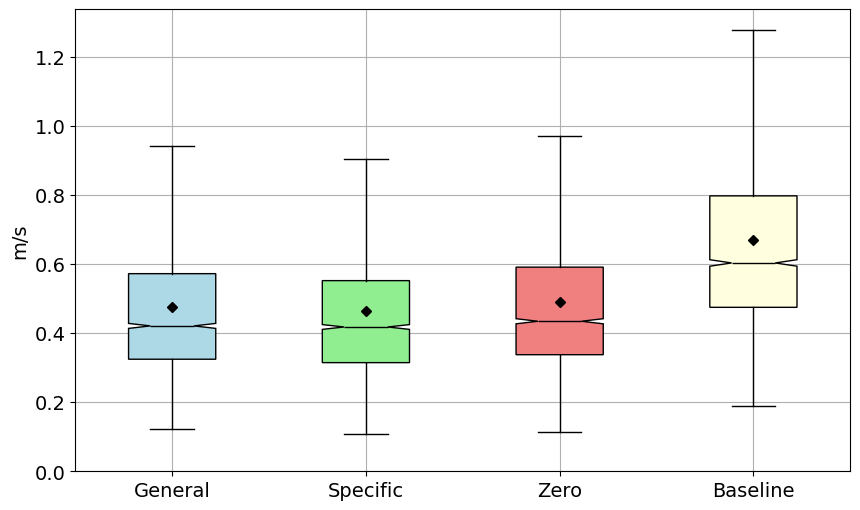}\label{fig:section_III_bp_mae_d11}}
     \hspace{2mm}
     \subfloat[SSIM.]{\includegraphics[height=4.6cm,valign=c]{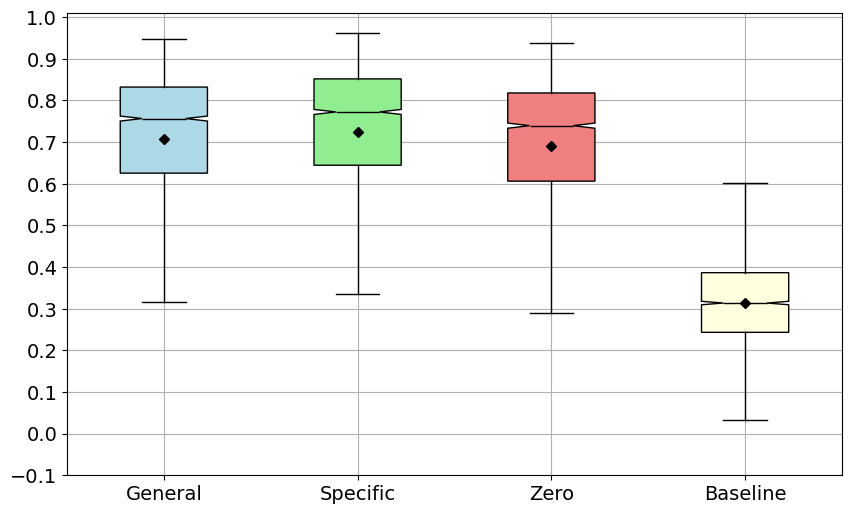}\label{fig:section_III_bp_ssim_d11}}
     
     \caption{RMSE and SSIM metrics box plots for the model predictions on the test set for domain \#9, (2 899 hourly samples from January 2024 to April 2024), for specific, generic, and zero models, and for baseline obtained from interpolating the predictor grid on the predictand grid using bi-linear interpolation. In each graph, the black dot and lines represent respectively the mean and the median, the upper and lower box limits indicate the first (Q1) and third (Q3) quartiles and the whiskers depict the highest (lowest) value within the $1.5 \times$ (Q3-Q1) above Q3 (below Q1).}\label{fig:results_III_d11}
\end{figure}

\begin{figure}[ht]
    \centering
    \subfloat[Pixel-wise MAE (m/s) for the different models and the baseline. The colorscale is capped at the max MAE of the Specific and Zero models. The baseline’s MAE exceeds the maximum value of the colorscale in certain locations, causing saturation. The value at the top of each image is the average over the domain.]{\includegraphics[width=0.99\textwidth,valign=c]{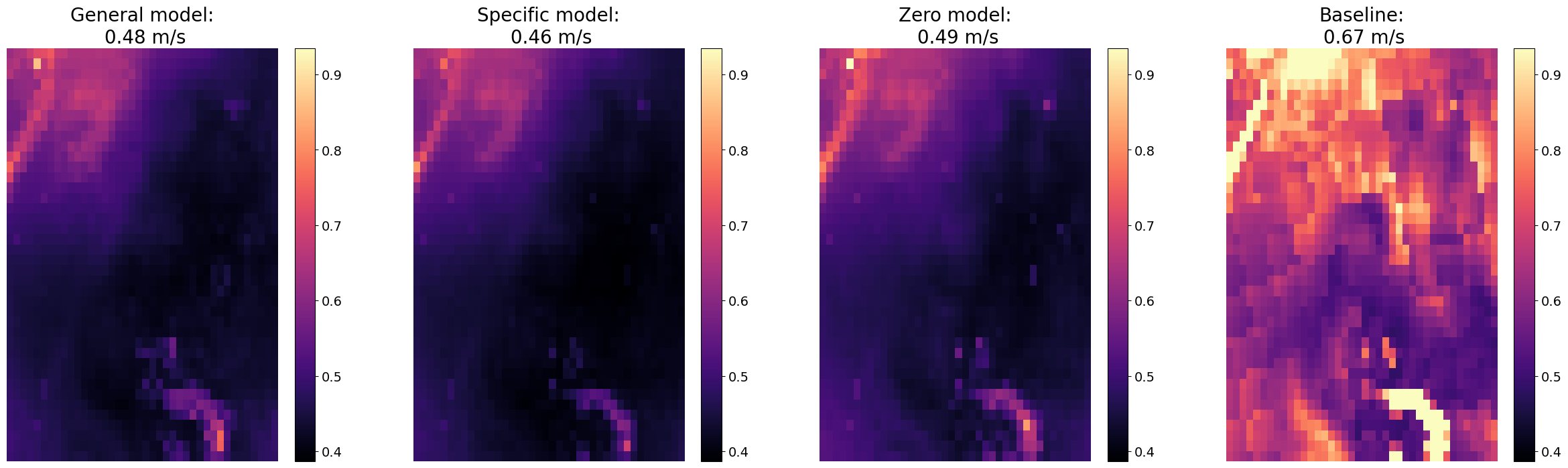}
     \label{fig:section_III_error_d11}}
     \\
     \subfloat[Average PSD.]{\includegraphics[width=0.48\textwidth,valign=c]{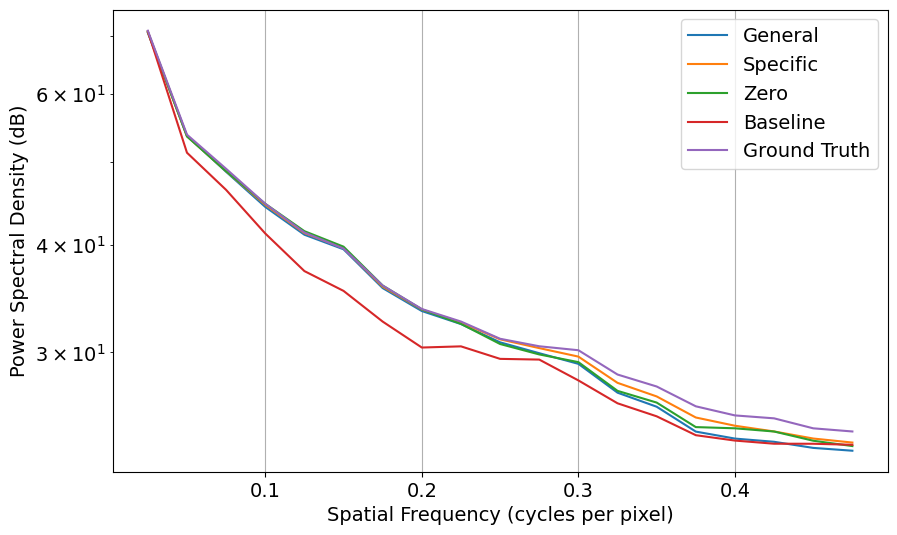}\label{fig:section_III_psd_d11}}
     \hspace{2mm}
     \subfloat[Average PDF. Note that the legend is the same as Figure \ref{fig:section_III_psd_d11}.]{\includegraphics[width=0.48\textwidth,valign=c]{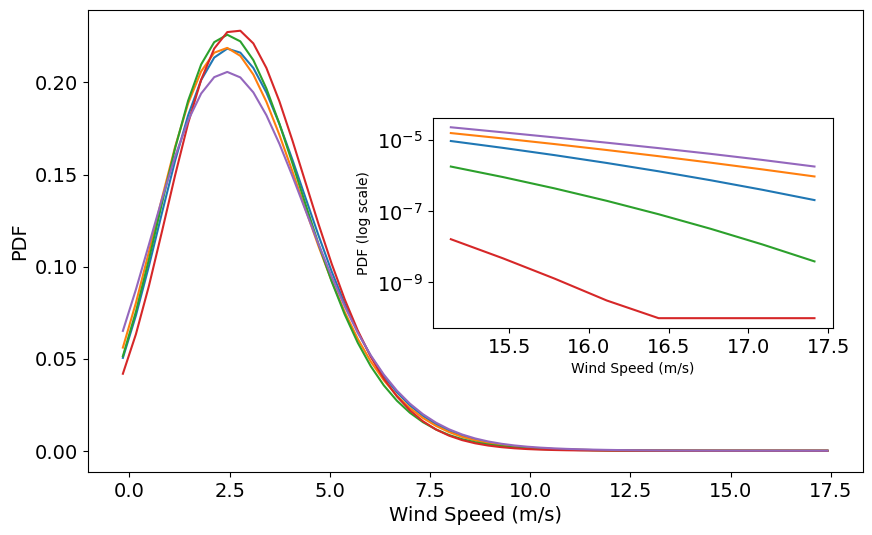}\label{fig:section_III_pdf_d11}}
     
     \caption{MAE, average PSD and average PDF graphs for the model predictions on the test set for domain \#9,(2 899 hourly samples from January 2024 to April 2024) for specific, general and zero models, and for baseline obtained from interpolating the predictor grid on the predictand grid using bi-linear interpolation. The ground truth in the PSD and PDF graphs refers to the HR $UV$, i.e., the predictand.}\label{fig:results_domain_III_d11}
\end{figure}

\begin{figure}[ht]
    \centering
     \hspace{2mm}
     \subfloat[RMSE.]{\includegraphics[height=4.6cm,valign=c]{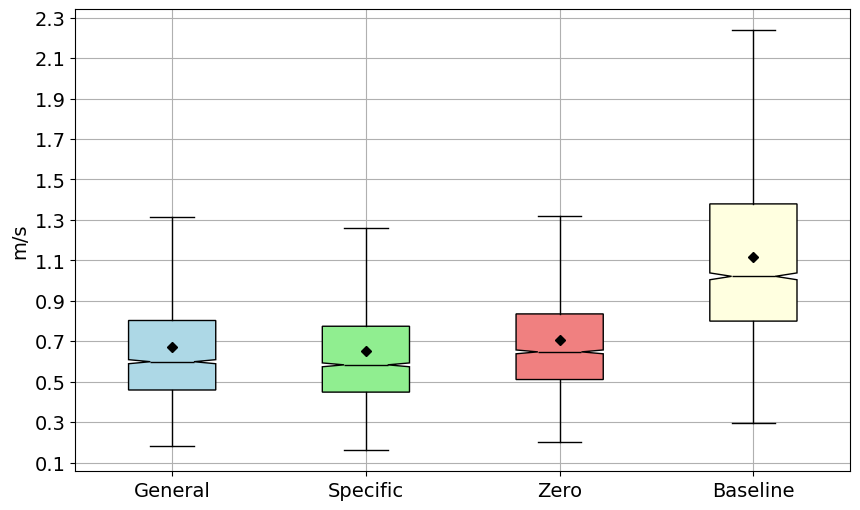}\label{fig:section_III_bp_rmse_d12}}
     \hspace{2mm}
     \subfloat[MAE.]{\includegraphics[height=4.6cm,valign=c]{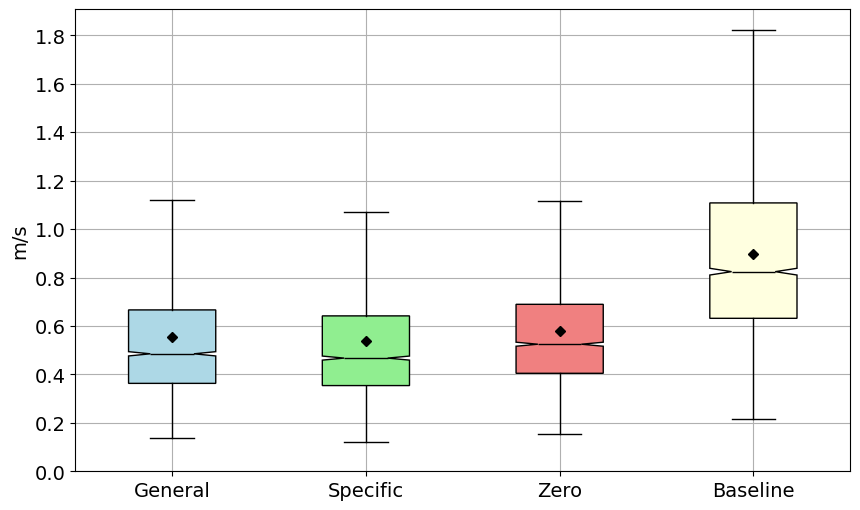}\label{fig:section_III_bp_mae_d12}}
     \hspace{2mm}
     \subfloat[SSIM.]{\includegraphics[height=4.6cm,valign=c]{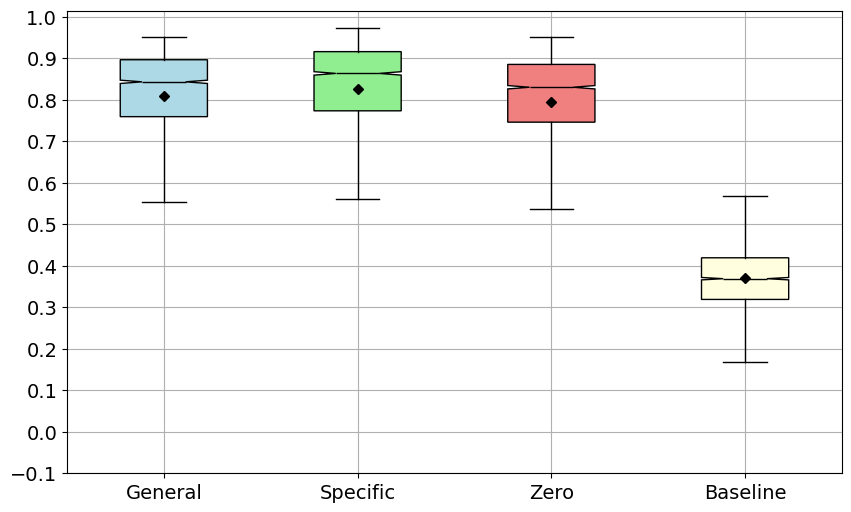}\label{fig:section_III_bp_ssim_d12}}
     
     \caption{RMSE and SSIM metrics box plots for the model predictions on the test set for domain \#10, (2 899 hourly samples from January 2024 to April 2024), for specific, generic, and zero models, and for baseline obtained from interpolating the predictor grid on the predictand grid using bi-linear interpolation. In each graph, the black dot and lines represent respectively the mean and the median, the upper and lower box limits indicate the first (Q1) and third (Q3) quartiles and the whiskers depict the highest (lowest) value within the $1.5 \times$ (Q3-Q1) above Q3 (below Q1).}\label{fig:results_III_d12}
\end{figure}

\begin{figure}[ht]
    \centering
    \subfloat[Pixel-wise MAE (m/s) for the different models and the baseline. The colorscale is capped at the max MAE of the Specific and Zero models. The baseline’s MAE exceeds the maximum value of the colorscale in certain locations, causing saturation. The value at the top of each image is the average over the domain.]{\includegraphics[width=0.99\textwidth,valign=c]{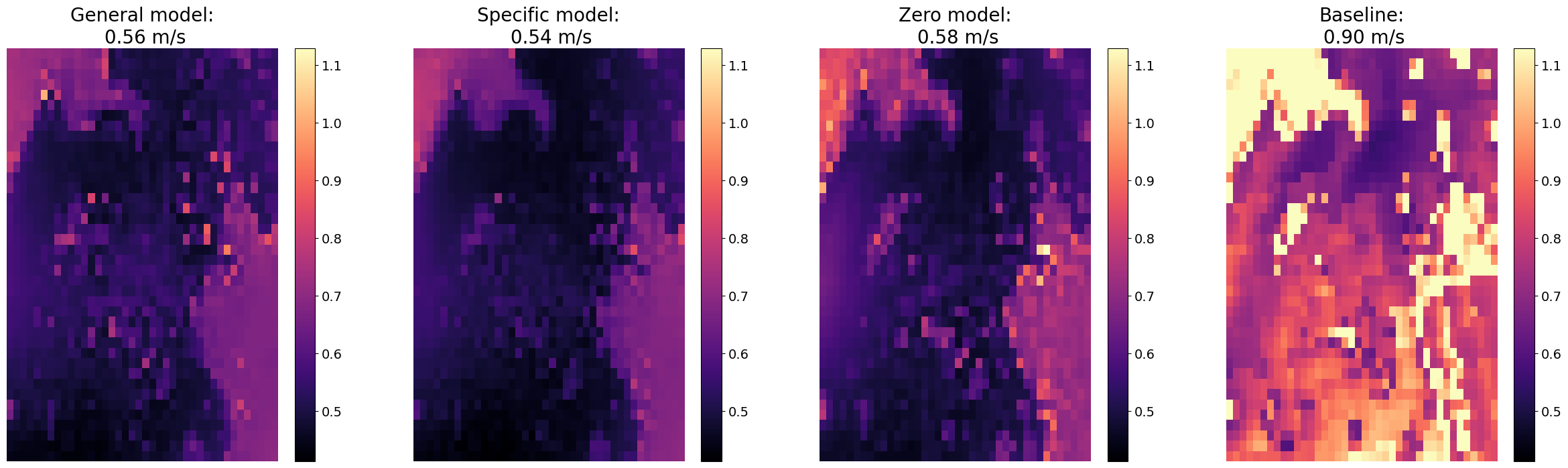}
     \label{fig:section_III_error_d12}}
     \\
     \subfloat[Average PSD.]{\includegraphics[width=0.48\textwidth,valign=c]{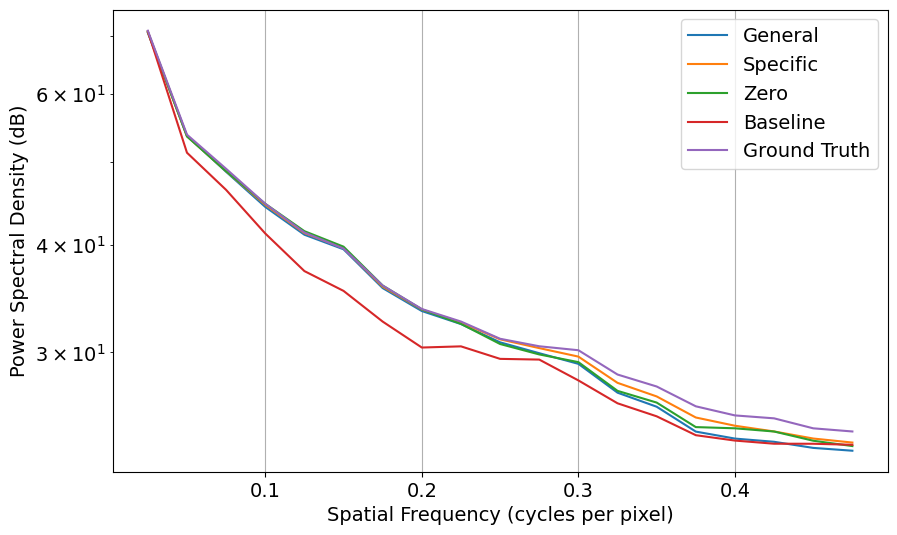}\label{fig:section_III_psd_d12}}
     \hspace{2mm}
     \subfloat[Average PDF. Note that the legend is the same as Figure \ref{fig:section_III_psd_d12}.]{\includegraphics[width=0.48\textwidth,valign=c]{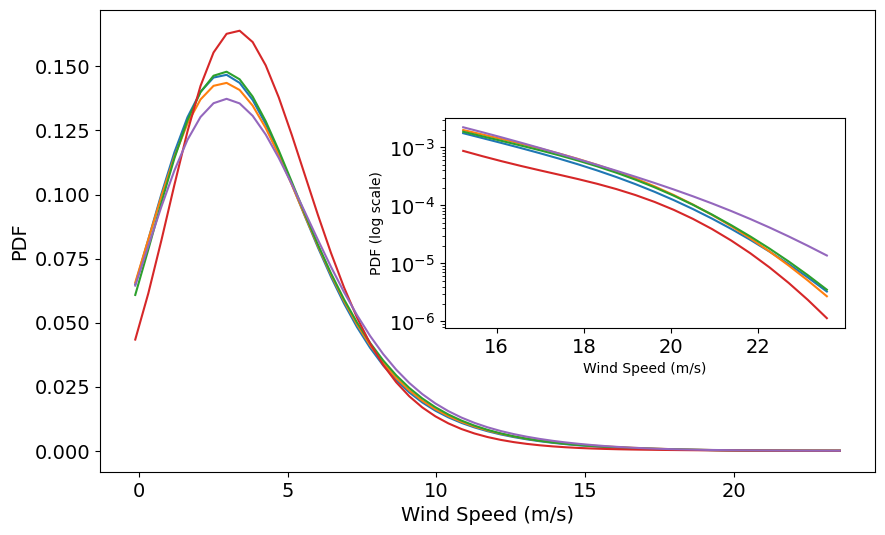}\label{fig:section_III_pdf_d12}}
     
     \caption{MAE, average PSD and average PDF graphs for the model predictions on the test set for domain \#10,(2 899 hourly samples from January 2024 to April 2024) for specific, general and zero models, and for baseline obtained from interpolating the predictor grid on the predictand grid using bi-linear interpolation. The ground truth in the PSD and PDF graphs refers to the HR $UV$, i.e., the predictand.}\label{fig:results_domain_III_d12}
\end{figure}

\begin{figure}[ht]
    \centering
     \hspace{2mm}
     \subfloat[RMSE.]{\includegraphics[height=4.6cm,valign=c]{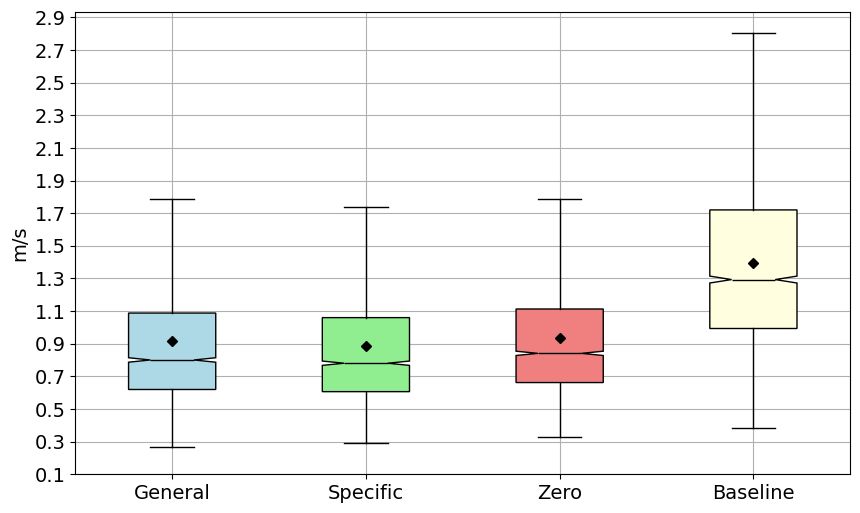}\label{fig:section_III_bp_rmse_d13}}
     \hspace{2mm}
     \subfloat[MAE.]{\includegraphics[height=4.6cm,valign=c]{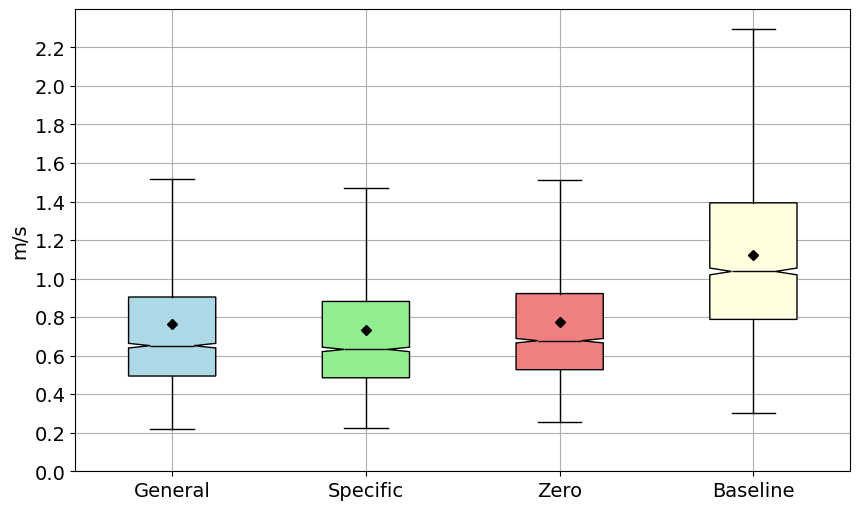}\label{fig:section_III_bp_mae_d13}}
     \hspace{2mm}
     \subfloat[SSIM.]{\includegraphics[height=4.6cm,valign=c]{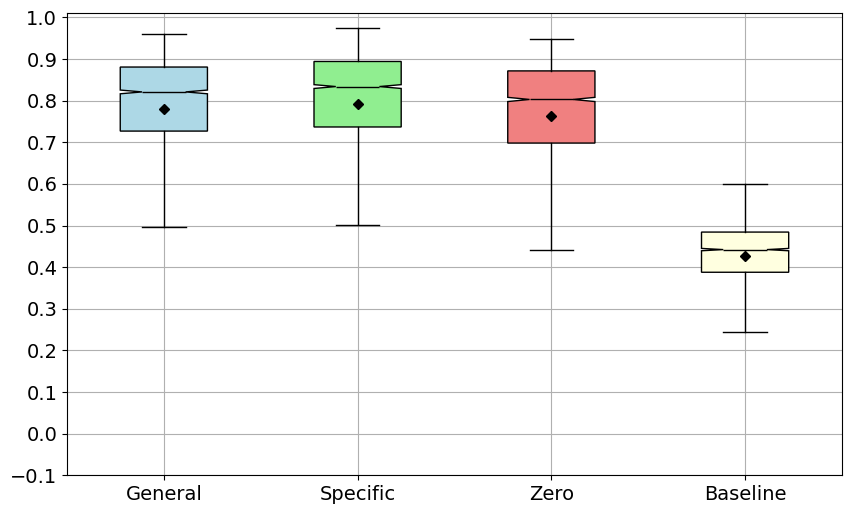}\label{fig:section_III_bp_ssim_d13}}
     
     \caption{RMSE and SSIM metrics box plots for the model predictions on the test set for domain \#11, (2 899 hourly samples from January 2024 to April 2024), for specific, generic, and zero models, and for baseline obtained from interpolating the predictor grid on the predictand grid using bi-linear interpolation. In each graph, the black dot and lines represent respectively the mean and the median, the upper and lower box limits indicate the first (Q1) and third (Q3) quartiles and the whiskers depict the highest (lowest) value within the $1.5 \times$ (Q3-Q1) above Q3 (below Q1).}\label{fig:results_III_d13}
\end{figure}

\begin{figure}[ht]
    \centering
    \subfloat[Pixel-wise MAE (m/s) for the different models and the baseline. The colorscale is capped at the max MAE of the Specific and Zero models. The baseline’s MAE exceeds the maximum value of the colorscale in certain locations, causing saturation. The value at the top of each image is the average over the domain.]{\includegraphics[width=0.99\textwidth,valign=c]{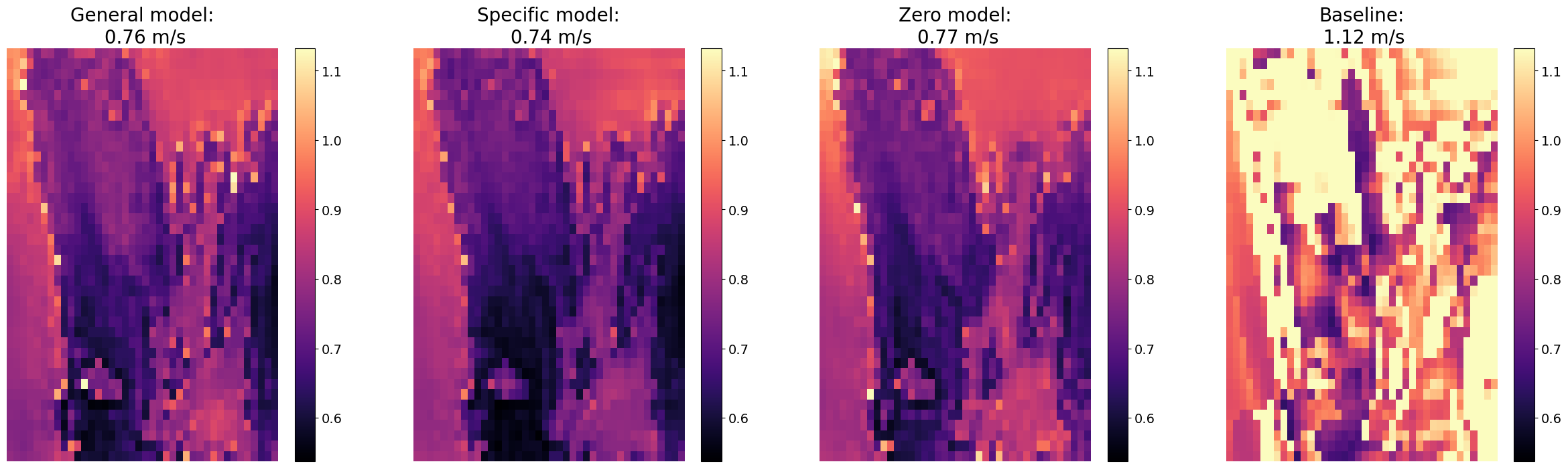}
     \label{fig:section_III_error_d13}}
     \\
     \subfloat[Average PSD.]{\includegraphics[width=0.48\textwidth,valign=c]{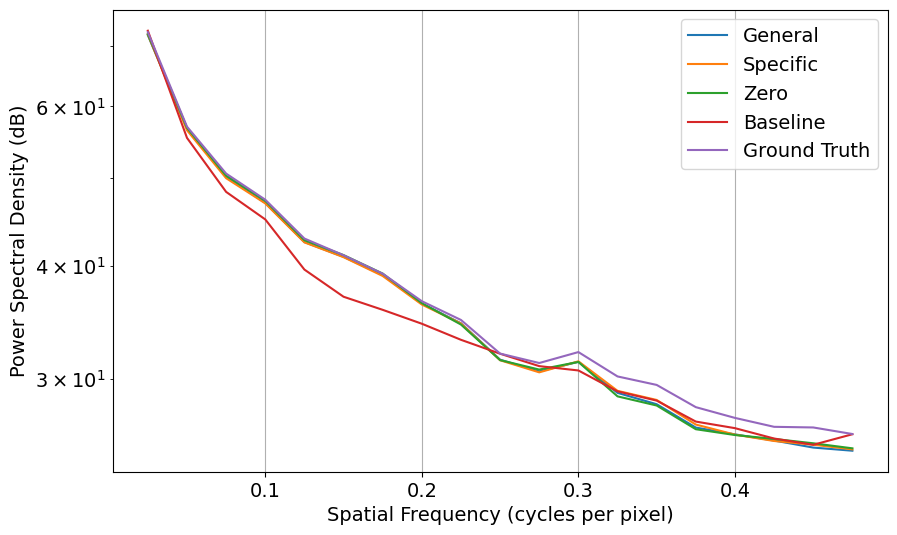}\label{fig:section_III_psd_d13}}
     \hspace{2mm}
     \subfloat[Average PDF. Note that the legend is the same as Figure \ref{fig:section_III_psd_d13}.]{\includegraphics[width=0.48\textwidth,valign=c]{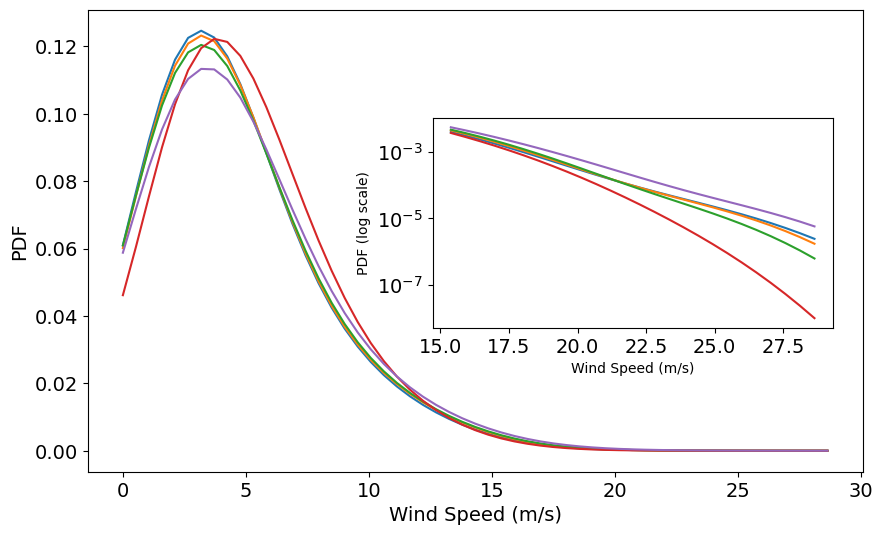}\label{fig:section_III_pdf_d13}}
     
     \caption{MAE, average PSD and average PDF graphs for the model predictions on the test set for domain \#11,(2 899 hourly samples from January 2024 to April 2024) for specific, general and zero models, and for baseline obtained from interpolating the predictor grid on the predictand grid using bi-linear interpolation. The ground truth in the PSD and PDF graphs refers to the HR $UV$, i.e., the predictand.}\label{fig:results_domain_III_d13}
\end{figure}

\begin{figure}[ht]
    \centering
     \hspace{2mm}
     \subfloat[RMSE.]{\includegraphics[height=4.6cm,valign=c]{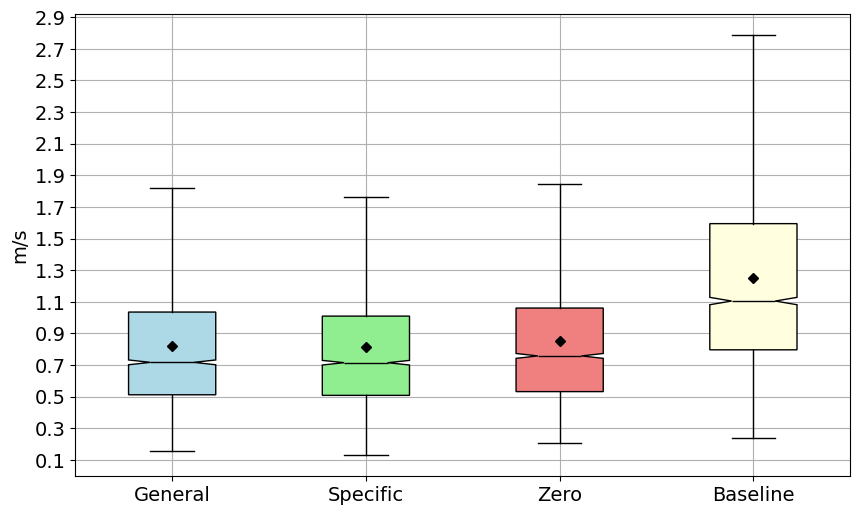}\label{fig:section_III_bp_rmse_d14}}
     \hspace{2mm}
     \subfloat[MAE.]{\includegraphics[height=4.6cm,valign=c]{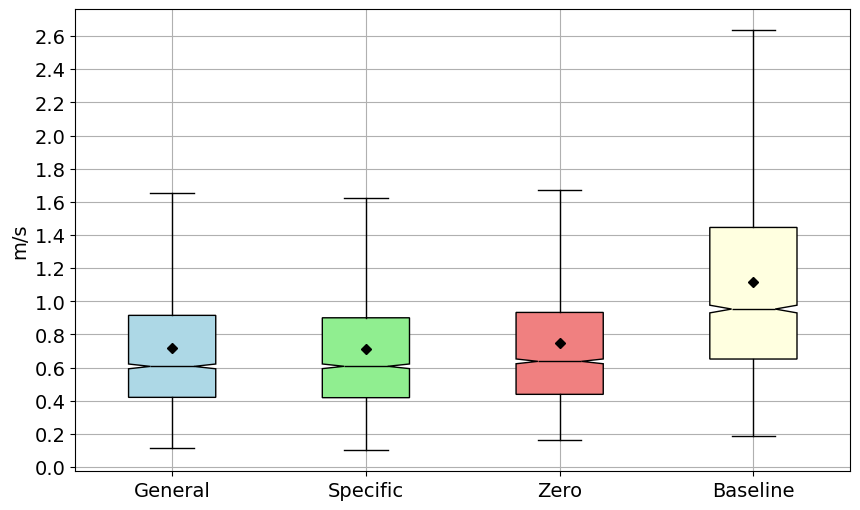}\label{fig:section_III_bp_mae_d14}}
     \hspace{2mm}
     \subfloat[SSIM.]{\includegraphics[height=4.6cm,valign=c]{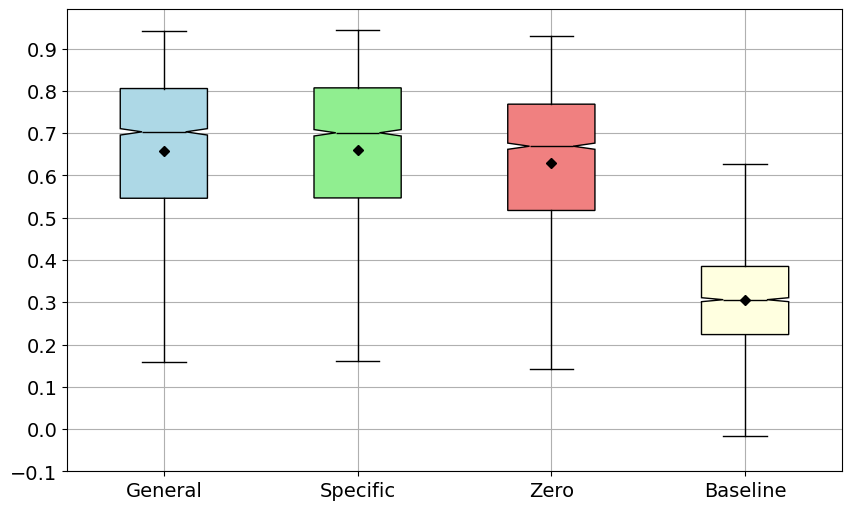}\label{fig:section_III_bp_ssim_d14}}
     
     \caption{RMSE and SSIM metrics box plots for the model predictions on the test set for domain \#12, (2 899 hourly samples from January 2024 to April 2024), for specific, generic, and zero models, and for baseline obtained from interpolating the predictor grid on the predictand grid using bi-linear interpolation. In each graph, the black dot and lines represent respectively the mean and the median, the upper and lower box limits indicate the first (Q1) and third (Q3) quartiles and the whiskers depict the highest (lowest) value within the $1.5 \times$ (Q3-Q1) above Q3 (below Q1).}\label{fig:results_III_d14}
\end{figure}

\begin{figure}[ht]
    \centering
    \subfloat[Pixel-wise MAE (m/s) for the different models and the baseline. The colorscale is capped at the max MAE of the Specific and Zero models. The baseline’s MAE exceeds the maximum value of the colorscale in certain locations, causing saturation. The value at the top of each image is the average over the domain.]{\includegraphics[width=0.99\textwidth,valign=c]{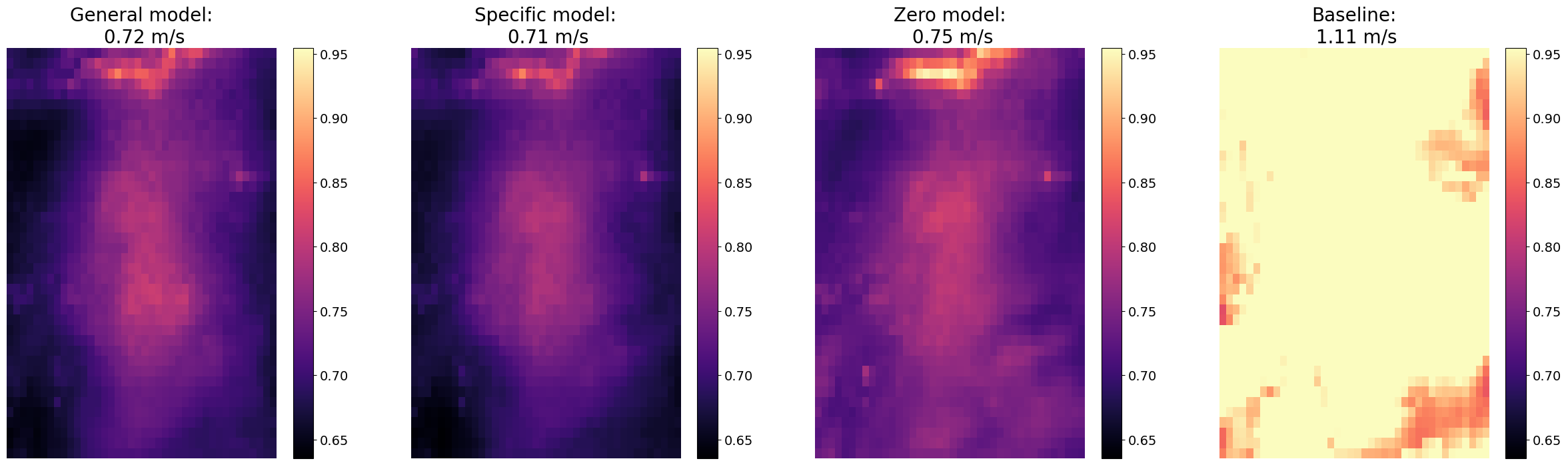}
     \label{fig:section_III_error_d14}}
     \\
     \subfloat[Average PSD.]{\includegraphics[width=0.48\textwidth,valign=c]{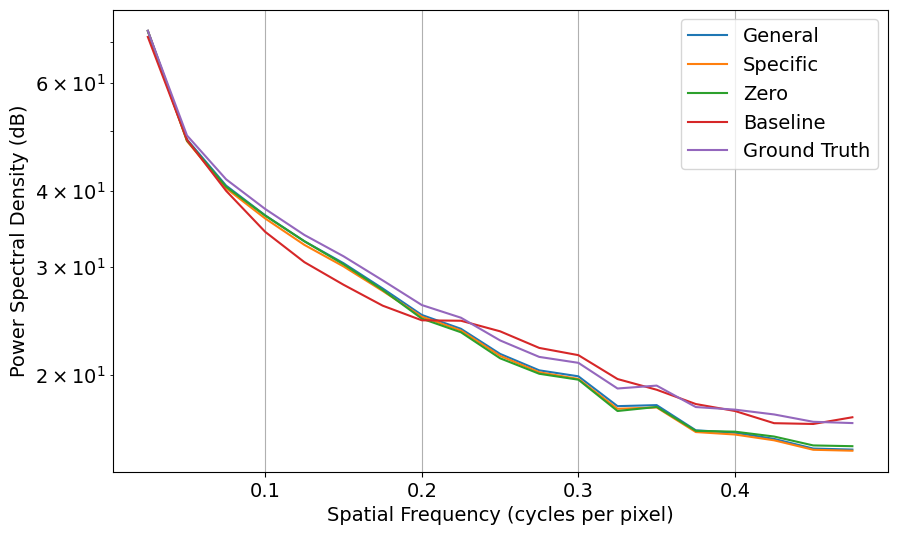}\label{fig:section_III_psd_d14}}
     \hspace{2mm}
     \subfloat[Average PDF. Note that the legend is the same as Figure \ref{fig:section_III_psd_d14}.]{\includegraphics[width=0.48\textwidth,valign=c]{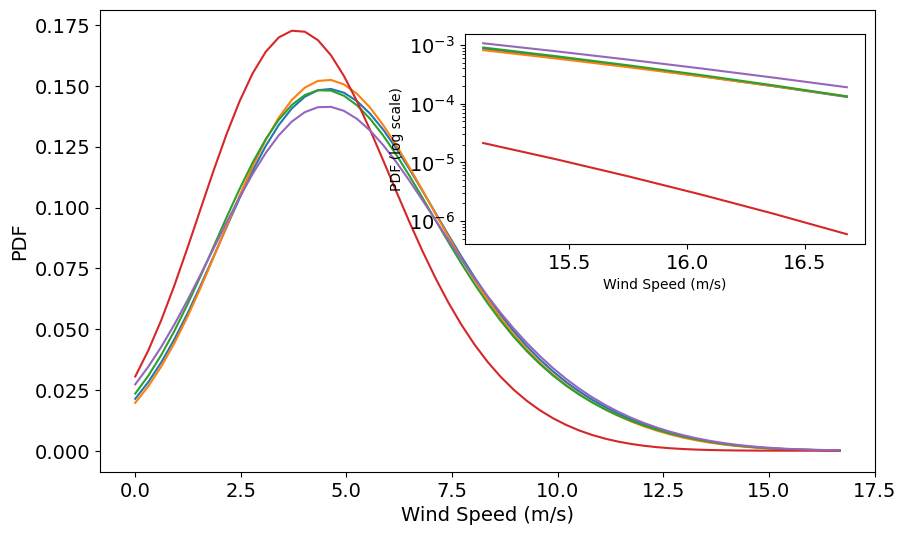}\label{fig:section_III_pdf_d14}}
     
     \caption{MAE, average PSD and average PDF graphs for the model predictions on the test set for domain \#12,(2 899 hourly samples from January 2024 to April 2024) for specific, general and zero models, and for baseline obtained from interpolating the predictor grid on the predictand grid using bi-linear interpolation. The ground truth in the PSD and PDF graphs refers to the HR $UV$, i.e., the predictand.}\label{fig:results_domain_III_d14}
\end{figure}

\begin{figure}[ht]
    \centering
     \hspace{2mm}
     \subfloat[RMSE.]{\includegraphics[height=4.6cm,valign=c]{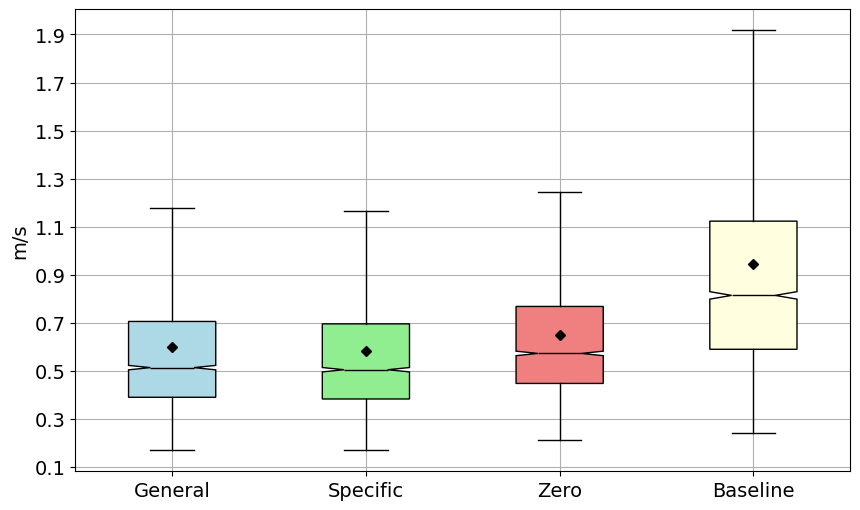}\label{fig:section_III_bp_rmse_d15}}
     \hspace{2mm}
     \subfloat[MAE.]{\includegraphics[height=4.6cm,valign=c]{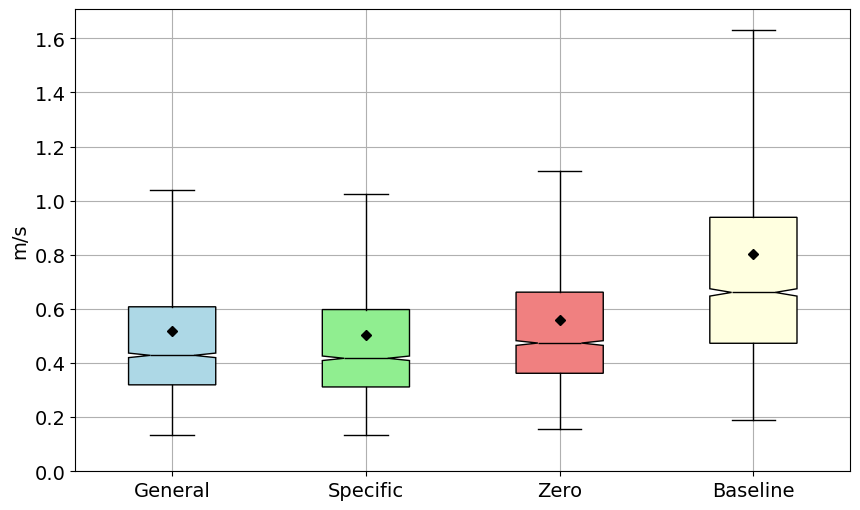}\label{fig:section_III_bp_mae_d15}}
     \hspace{2mm}
     \subfloat[SSIM.]{\includegraphics[height=4.6cm,valign=c]{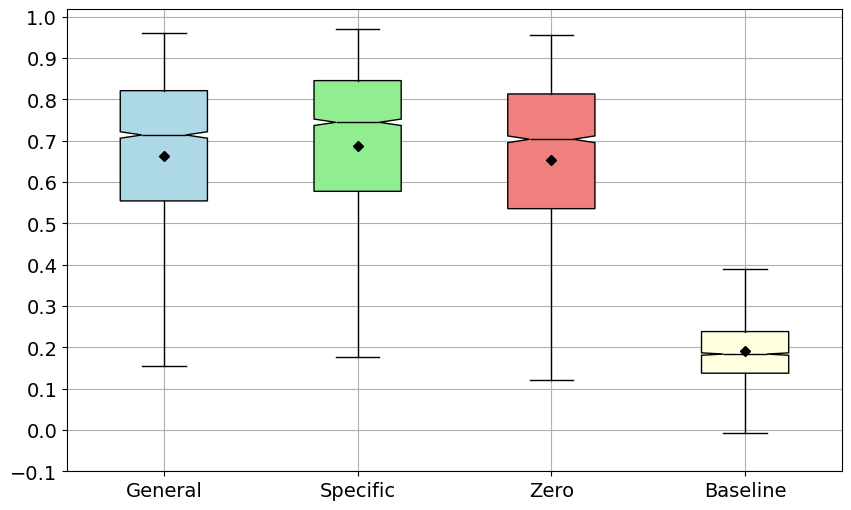}\label{fig:section_III_bp_ssim_d15}}
     
     \caption{RMSE and SSIM metrics box plots for the model predictions on the test set for domain \#13, (2 899 hourly samples from January 2024 to April 2024), for specific, generic, and zero models, and for baseline obtained from interpolating the predictor grid on the predictand grid using bi-linear interpolation. In each graph, the black dot and lines represent respectively the mean and the median, the upper and lower box limits indicate the first (Q1) and third (Q3) quartiles and the whiskers depict the highest (lowest) value within the $1.5 \times$ (Q3-Q1) above Q3 (below Q1).}\label{fig:results_III_d15}
\end{figure}

\begin{figure}[ht]
    \centering
    \subfloat[Pixel-wise MAE (m/s) for the different models and the baseline. The colorscale is capped at the max MAE of the Specific and Zero models. The baseline’s MAE exceeds the maximum value of the colorscale in certain locations, causing saturation. The value at the top of each image is the average over the domain.]{\includegraphics[width=0.99\textwidth,valign=c]{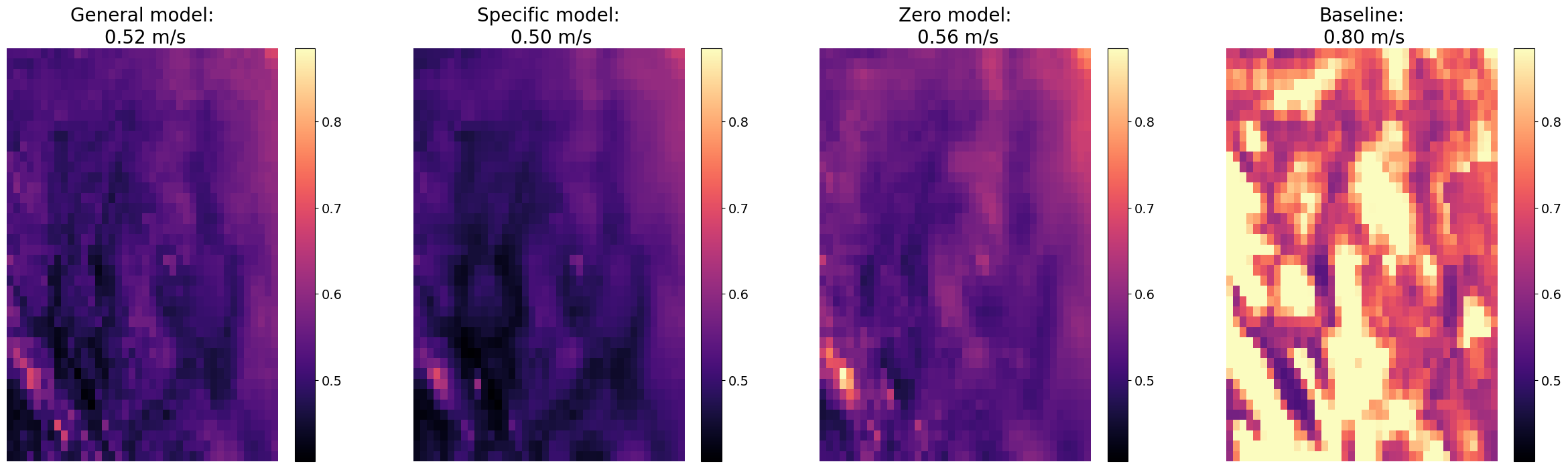}
     \label{fig:section_III_error_d15}}
     \\
     \subfloat[Average PSD.]{\includegraphics[width=0.48\textwidth,valign=c]{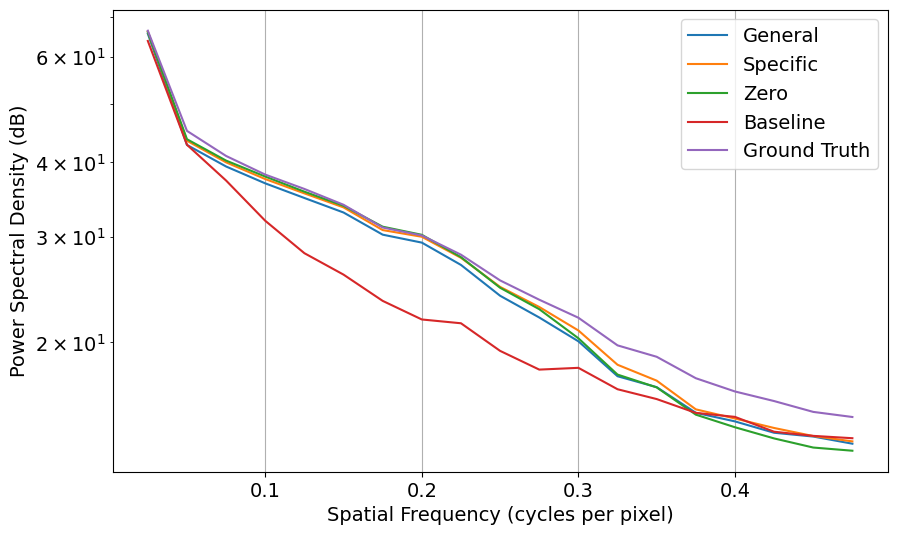}\label{fig:section_III_psd_d15}}
     \hspace{2mm}
     \subfloat[Average PDF. Note that the legend is the same as Figure \ref{fig:section_III_psd_d15}.]{\includegraphics[width=0.48\textwidth,valign=c]{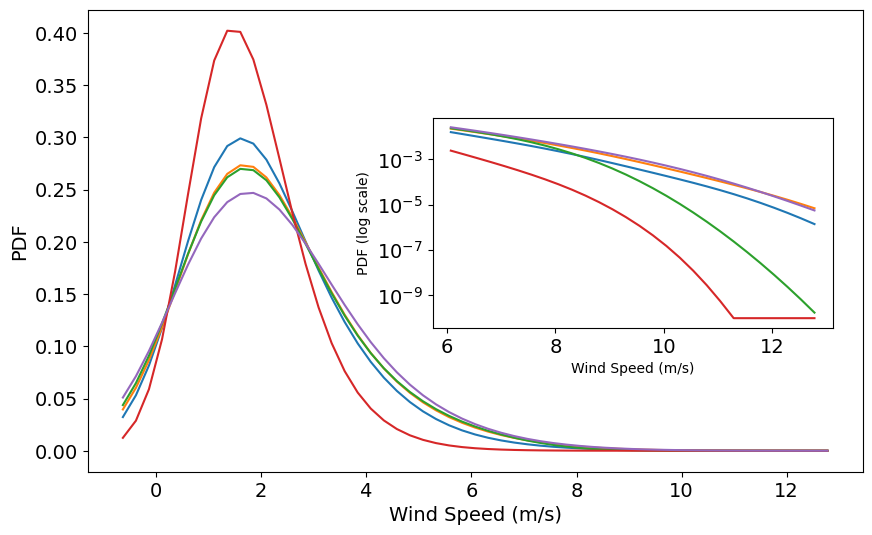}\label{fig:section_III_pdf_d15}}
     
     \caption{MAE, average PSD and average PDF graphs for the model predictions on the test set for domain \#13,(2 899 hourly samples from January 2024 to April 2024) for specific, general and zero models, and for baseline obtained from interpolating the predictor grid on the predictand grid using bi-linear interpolation. The ground truth in the PSD and PDF graphs refers to the HR $UV$, i.e., the predictand.}\label{fig:results_domain_III_d15}
\end{figure}

\end{document}